%% file: jmlr.tex
\documentclass[twoside,11pt]{article}

\usepackage[utf8]{inputenc} % allow utf-8 input
\usepackage[T1]{fontenc}    % use 8-bit T1 fonts
\usepackage{xurl}            % simple URL typesetting
\usepackage{booktabs}       % professional-quality tables
\usepackage{amsfonts}       % blackboard math symbols
\usepackage{nicefrac}       % compact symbols for 1/2, etc.
\usepackage{microtype}      % microtypography
\usepackage{xcolor}         % colors
\usepackage{caption}
\usepackage{subcaption}
\usepackage{amsmath}
\usepackage{mathtools}
\usepackage{multirow}
\usepackage{multicol}
\usepackage{color}
\usepackage{comment}
\usepackage{wrapfig}
\usepackage{float}
\usepackage[preprint]{jmlr2e}

\DeclarePairedDelimiter{\norm}{\lVert}{\rVert}
\newcommand\normx[1]{\left\Vert#1\right\Vert}
\usepackage{lastpage}
\jmlrheading{24}{2023}{1-\pageref{LastPage}}{5/22; Revised
6/23}{7/23}{22-0496}{Sanket Vaibhav Mehta, Darshan Patil, Sarath Chandar and Emma Strubell}

\ShortHeadings{An Empirical Investigation of the Role of Pre-training in Lifelong Learning}{Mehta, Patil, Chandar and Strubell}

\firstpageno{1}

\begin{document}

\title{An Empirical Investigation of the Role of Pre-training in Lifelong Learning}

\author{\name Sanket Vaibhav Mehta \email svmehta@cs.cmu.edu \\
       \addr School of Computer Science\\
       Carnegie Mellon University\\
       Pittsburgh, PA 15213, USA
       \AND
       \name Darshan Patil \email darshan.patil@mila.quebec \\
       \addr Mila - Quebec AI Institute\\
       Université de Montréal \\
       Montreal, QC H3T 1J4, Canada
       \AND
       \name Sarath Chandar \email sarath.chandar@mila.quebec \\
       \addr Mila - Quebec AI Institute\\
       Canada CIFAR AI Chair\\
       École Polytechnique de Montréal\\
       Montreal, QC H3T 1J4, Canada
       \AND
       \name Emma Strubell \email estrubel@cs.cmu.edu \\
       \addr School of Computer Science\\
       Carnegie Mellon University\\
       Pittsburgh, PA 15213, USA}

\editor{Ivan Titov}

\maketitle

\begin{abstract}%   <- trailing '%' for backward compatibility of .sty file
The lifelong learning paradigm in machine learning is an attractive alternative to the more prominent isolated learning scheme not only due to its resemblance to biological learning but also its potential to reduce energy waste by obviating excessive model re-training. A key challenge to this paradigm is the phenomenon of catastrophic forgetting. With the increasing popularity and success of pre-trained models in machine learning, we pose the question: What role does pre-training play in lifelong learning, specifically with respect to catastrophic forgetting? We investigate existing methods in the context of large, pre-trained models and evaluate their performance on a variety of text and image classification tasks, including a large-scale study using a novel data set of 15 diverse NLP tasks. Across all settings, we observe that generic pre-training implicitly alleviates the effects of catastrophic forgetting when learning multiple tasks sequentially compared to randomly initialized models. We then further investigate \emph{why} pre-training alleviates forgetting in this setting. We study this phenomenon by analyzing the loss landscape, finding that pre-trained weights appear to ease forgetting by leading to wider minima. Based on this insight, we propose jointly optimizing for current task loss and loss basin sharpness to explicitly encourage wider basins during sequential fine-tuning. We show that this optimization approach outperforms several state-of-the-art task-sequential continual learning algorithms across multiple settings, occasionally even without retaining a memory that scales in size with the number of tasks.
\end{abstract}


% \newpage
% Introduction
\input{jmlrsections/01_introduction}

% Method
\input{jmlrsections/02_preliminaries}

\input{jmlrsections/03_experimentdesign}

% Experiments
\input{jmlrsections/04_losslandscape}

\input{jmlrsections/05_sam}

% Related Work
\input{jmlrsections/06_relatedwork}

\input{jmlrsections/07_conclusion}

% Acknowledgements should go at the end, before appendices and references
\acks{We thank the anonymous reviewers and Action Editor for their valuable feedback and suggestions, which helped improve the paper. We also thank Janarthanan Rajendran, Sai Krishna Rallabandi, Khyathi Raghavi Chandu, Saujas Vaduguru, and Divyansh Kaushik for reviewing the paper and providing valuable comments. We are also grateful to Mojtaba Faramarzi for helping with ImageNet and CIFAR-100 class hierarchies, and to Hadi Nekoei, and Paul-Aymeric McRae for reviewing our code. We like to acknowledge CMU Workhorse, TIR group, and Compute Canada for providing compute resources for this work. This project is funded in part by DSO National Laboratories. Sarath Chandar is supported by a Canada CIFAR AI Chair and an NSERC Discovery Grant. }

\appendix
\input{jmlrsections/08_appendix}

\bibliography{jmlr}

\end{document}

%% file: jmlrsections/01_introduction.tex
\section{Introduction}
The contemporary machine learning paradigm concentrates on isolated learning \citep{chen2018lifelong} i.e., learning a model from scratch for every new task. 
In contrast, the \textit{lifelong learning} \citep{thrun1995lifelong, thrun1996learning, chen2018lifelong} or \textit{incremental learning} \citep{solomonoff1989system, syed1999incremental, ruping2001incremental} or \textit{never-ending learning} \citep{mitchell2018never} or \textit{continual learning} \citep{parisi2019continual} paradigm 
defines a biologically-inspired learning approach where models learn tasks in sequence, ideally preserving past knowledge and leveraging it to efficiently learn new tasks.
Lifelong learning has the added benefit of avoiding periodical re-training of models from scratch to learn novel tasks or adapt to new data, with the potential to reduce both computational and energy requirements \citep{hazelwood2018hpca,strubell2019energy,schwartz2020green}. 
In the context of modern machine learning where state-of-the-art models are powered by deep neural networks, \textit{catastrophic forgetting} has been identified as a key challenge to implementing successful lifelong learning systems \citep{mccloskey1989catastrophic, french1999catastrophic}. Catastrophic forgetting happens when the model forgets knowledge learned in previous tasks as information relevant to the current task is incorporated. Mitigating or preventing this phenomenon is critical to achieving true lifelong learning.

At the same time, transfer learning has shown impressive results in both computer vision (CV; \citealt{Zhuang2021Survey}) and natural language processing (NLP) applications \citep{howard2018universal,peters2018deep,devlin2019bert}.\footnote{One of the original motivations for transfer learning was as a way to enable lifelong learning, discussed in a NIPS-95 workshop on ``Learning to Learn''  \citep{pan2009survey}.} The modern transfer learning paradigm involves \textit{pre-training} a fixed architecture, like ResNet \citep{he2016deep} or BERT \citep{devlin2019bert}, using copious amounts of data, and then \textit{fine-tuning} the learnt parameters on target tasks. Given the tremendous success of pre-trained models, there has been increased interest in understanding their role in improving generalization \citep{erhan2010does, neyshabur2020being}, speed of convergence \citep{hao2019visualizing}, successful transfer \citep{he2019rethinking, pruksachatkun-etal-2020-intermediate}, and out-of-distribution robustness \citep{hendrycks2020pretrained, tu2020empirical}. Despite these efforts, the role of pre-trained initializations in lifelong learning settings has been under-explored. In contemporary work, it has been shown that pre-trained models can be used as \textit{feature extractors} (i.e., pre-trained weights are frozen) for task-sequential learning \citep{hu2021continual}. Because the pre-trained weights are explicitly frozen in this setting, the model undergoes no catastrophic forgetting. In contrast, \textit{fine-tuning} pre-trained weights update the pre-trained model parameters and are susceptible to severe forgetting. This is typically the most accurate and thus common transfer learning paradigm \citep{peters2019tune}, and the one we consider in this work. To the best of our knowledge, there is no prior work systematically analyzing the role of pre-trained initialization on catastrophic forgetting in lifelong learning scenarios.

\begin{figure}[t!]
  \begin{center}
    \includegraphics[width=0.5\textwidth]{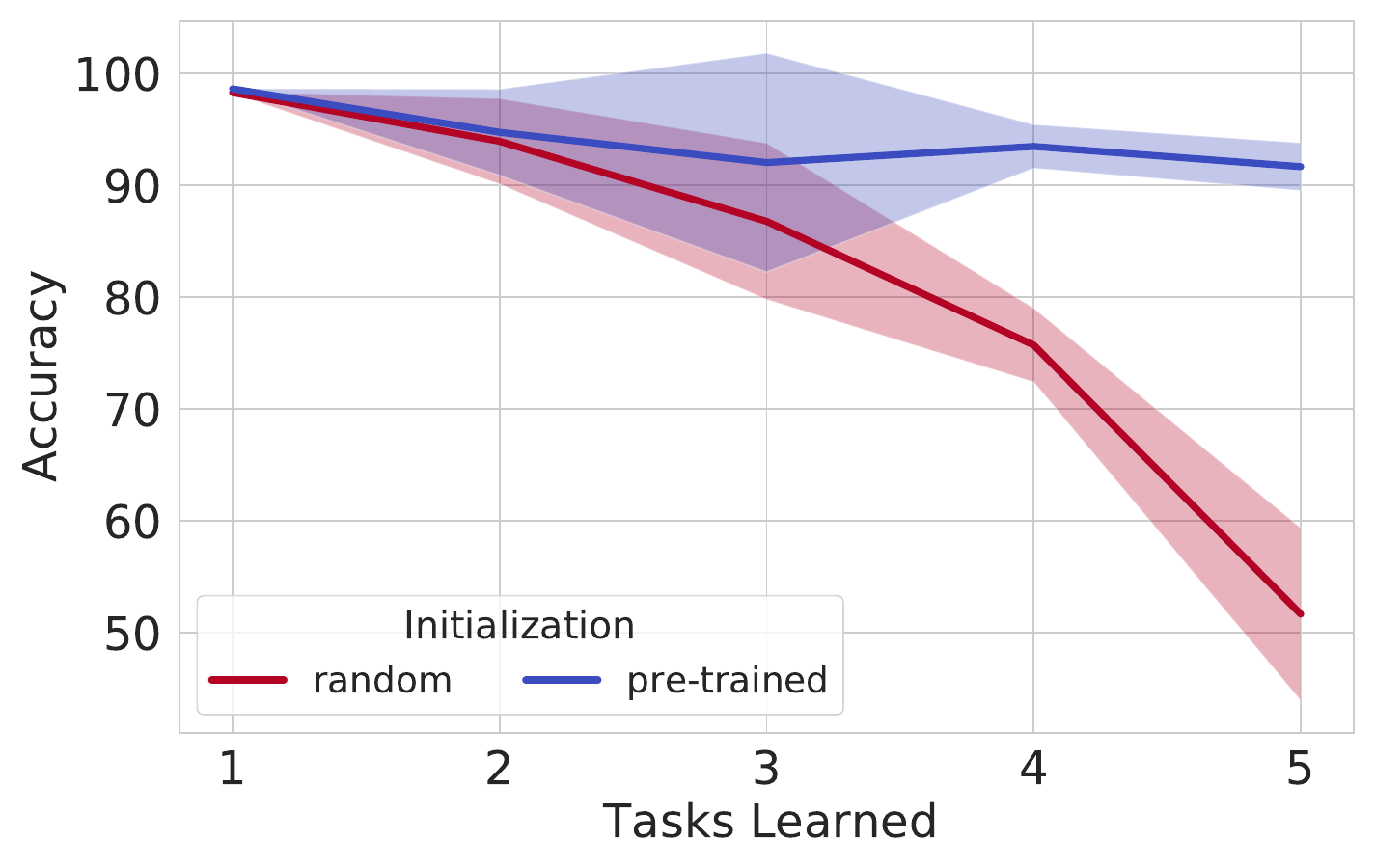}
  \end{center}
  \caption{Pre-trained and randomly initialized DistilBERT on Split YahooQA data set. Performance of the first task visualized over sequential learning of tasks (averaged over $5$ runs). Both models start with approximately equal average task accuracy, but pre-trained initialization leads to significantly less forgetting.}
  \label{fig:motivation}
\end{figure}

Figure \ref{fig:motivation} shows that simply changing the network initialization to generic pre-trained weights can significantly reduce forgetting on the first task when doing sequential training on five tasks.
This observation motivates us to ask---\textit{Does pre-training implicitly alleviate catastrophic forgetting, and if so, why?} To answer this question we conduct a systematic study on existing CV and NLP benchmarks and observe that pre-training indeed leads to less forgetting. We also investigate the effect of the type of pre-trained initialization by analyzing the extent to which different pre-trained Transformer language model variants \citep{sanh2019distilbert, devlin2019bert, liu2019roberta} undergo forgetting, observing that increasing the capacity of the model and diversity of the pre-training corpus play an important role in alleviating forgetting. To further stress-test these models under realistic scenarios, we introduce a data set with 15 diverse NLP tasks and observe a considerable increase in forgetting on this data set. 

We hypothesize that pre-trained weights already have a good inductive bias to implicitly alleviate forgetting. To explain this behavior, we build upon two separate observations---\citet{hao2019visualizing, neyshabur2020being} show that in the context of transfer learning, pre-trained weights lead to a flat loss basin when fine-tuning on a single task.  \citet{mirzadeh2020understanding} argues that the geometric properties of the local minima for each task play a vital role in forgetting, and they suggest modifying the hyper-parameters (learning rate decay, batch size, dropout) to promote flat minima.

To verify the above hypothesis, we analyze the loss landscape of the first task as the model is trained sequentially on subsequent tasks. For pre-trained initializations, we see that minima obtained after training on a sequence of tasks still remain in the relatively low loss basin of the first task when compared with random initialization. Further, linearly interpolating loss across sequentially trained task minima confirms that models initialized with pre-trained weights undergo a more gradual change in loss compared to random initialization. 
%Further, tracking the loss along the linear interpolation between the first task's minima and subsequent ones confirms that models initialized with pre-trained weights undergo a more gradual change in the loss compared to randomly initialized weights. 
These observations hint at the flatness of the minima reached in the case of pre-trained initialized models. To quantify the flatness of the loss landscape, we compute a sharpness metric \citep{keskar2016large} and verify that pre-trained weights indeed lead to flat basins in comparison to random weights while training sequentially. These analyses help us showcase that continual training from pre-trained weights induces wide task minima, therefore, implicitly alleviating forgetting.
% These analyses help us showcase that continual training from pre-trained weights induces wide task minima, which is shown to alleviate forgetting \citep{mirzadeh2020understanding}. 
To further mitigate forgetting, we explicitly optimize for flat loss basins by minimizing the current task loss and the sharpness metric. Concretely, we use the Sharpness-Aware Minimization (SAM) procedure \citep{foret2020sharpness} to seek parameters that lie in the neighborhoods having uniformly low loss values (Section \ref{sec:sam}) and report improved results across many experimental settings.
% To further alleviate forgetting when sequentially fine-tuning the pre-trained weights, we explicitly seek flat basins by minimizing the current task loss value and the sharpness metric. We use the existing Sharpness-Aware Minimization (SAM) procedure \citep{foret2020sharpness} for this purpose and report improved results across many experimental settings. 
Our main contributions can be summarized as follows:

% Our main contributions can be summarized as follows:
% \subsection{Contributions}
% The contributions of this paper can be summarized as follows:
\begin{itemize}
    % \item We observe that initializing models with pre-trained weights results in less forgetting compared to random weights despite achieving higher performance on each task. We bolster this observation with a systematic study validating that this behavior persists across applications (NLP, CV) and three existing approaches: Elastic weight consolidation \citep{kirkpatrick2017overcoming}, A-GEM \citep{chaudhry2018efficient}, and episodic replay \citep{chaudhry2019tiny}. 
    \item We observe that initializing models with pre-trained weights results in less forgetting compared to random weights despite achieving higher performance on each task. We bolster this observation with a systematic study validating that this behavior persists across applications (NLP, CV) and prominent approaches: Elastic weight consolidation \citep{kirkpatrick2017overcoming}, A-GEM \citep{chaudhry2018efficient}, and episodic replay \citep{chaudhry2019tiny} (Section \ref{sec:pretrain_alleviate_forgetting}). 
    % \item %For the pre-training initialization study, 
    % We introduce a new, more challenging benchmark for lifelong learning in NLP consisting of 15 diverse NLP tasks.
    %, which proves more challenging than previous settings for the Transformer models considered in our study.
    % \item To understand the role of varying pre-trained initializations, we analyse a suite of pre-trained Transformer language models and showcase that model capacity and diversity of the pre-training corpus do play a role in alleviating forgetting. We also show that sequential training on diverse tasks is still challenging for models initialized with pre-trained weights by introducing a new, more challenging benchmark for lifelong learning in NLP consisting of 15 diverse NLP tasks.
    \item To understand the role of varying pre-trained initializations, we analyze a suite of pre-trained Transformer language models and showcase that model capacity and diversity of the pre-training corpus do play a role in alleviating forgetting. We also show that sequential training on diverse tasks is still challenging for pre-trained initialized models by introducing a new, more challenging benchmark for lifelong learning in NLP consisting of 15 diverse NLP tasks (Sections \ref{sec:diverse_and_homogeneous},\ref{sec:pretrainingstudy}). 
    \item %We examine the above-mentioned behavior from the loss landscape perspective. 
    We hypothesize and verify empirically that pre-trained models alleviate forgetting as they have an implicit bias towards wider task minima. The effect of these wider minima is that changes in weights from learning subsequent tasks result in a smaller change to the current task loss, which helps reduce forgetting (Section \ref{sec:loss_landscape}). 
    \item We further show that explicitly seeking flat minima during task incremental learning leads to reduced forgetting and achieves state-of-the-art performance on the benchmarks we considered. 
    %results in even less forgetting, resulting in state-of-the-art performance across considered benchmarks (Section \ref{sec:sam}). 
    Our analysis of different initializations, namely task-agnostic meta-learned and supervised pre-trained models explicitly guided towards flat loss regions, highlights the synergistic benefits when combined with the explicit optimization for flatness during sequential fine-tuning (Section \ref{sec:sam}).\footnote{Code is available at---\url{https://github.com/sanketvmehta/lifelong-learning-pretraining-and-sam}}
\end{itemize}

%% file: jmlrsections/02_preliminaries.tex
\section{Preliminaries}
\label{sec:prelim}
In this section, we introduce the notations and outline our problem setup. We also specify our experimental settings, including the baseline methods, benchmarks, and evaluation metrics.
\subsection{Problem Setup: Task Incremental Learning}
We consider a setup where we receive a continuum of data from different tasks sequentially:
$(x_1, y_1, t_1), \cdots, (x_i, y_i, t_i), \cdots$. Each triplet $(x_i, y_i, t_i)$ consists of a task descriptor $t_i \in \mathcal{T}$, input data $x_i \in \mathcal{D}_{t_i}$, target labels $y_i \in \mathcal{Y}_{t_i}$ and satisfies $(x_i, y_i) \stackrel{iid}{\sim} \mathcal{P}_{t_i}(X, Y)$. Following \citep{chaudhry2019tiny}, we consider an explicit task descriptor $t_i$ because the same input $x_i$ can appear in multiple different tasks but with different labels. 
For example, given a product review, we could annotate it with sentiment polarity and grammatical acceptability judgments.
%Following \citet{lopez2017gradient}, we assume that the continuum is locally i.i.d, i.e., each triplet $(x_i, y_i, t_i)$ satisfies $(x_i, y_i) \stackrel{iid}{\sim} \mathcal{P}_{t_i}(X, Y)$. 
Given the observed data, our goal is to learn a predictor $f: \mathcal{X} \times \mathcal{T} \rightarrow \mathcal{Y}$ where we want to evaluate test pairs $(x,t)$ from previously observed tasks (backward transfer) and the current task at any time during task incremental learning of our model \citep{van2019three}. 

\subsection{Benchmarks}
\label{sec:dataset}
We perform extensive experiments on widely adopted task-incremental learning benchmarks \citep{chaudhry2019tiny, ebrahimi2020adversarial, wang2020efficient}  across both CV and NLP domains. These benchmarks help us evaluate our method to be consistent with the literature. Most of the existing works consider Split MNIST, Split CIFAR-10, and Split CIFAR-100 benchmarks, which are homogenous; different tasks in these benchmarks share the same underlying domain. Given the generic nature of the pre-trained initialization, we want to investigate forgetting when subjected to a sequence of diverse tasks. Therefore, we also consider data sets spanning diverse CV and NLP tasks.

\begin{table} 
  \begin{center}
    % \begin{small}
      %\resizebox{\textwidth}{%
      \begin{tabular}{l|rrrr}
        \toprule
        Data set & |Train| & |Dev| & |Test| & |L| \\
        \bottomrule
        MNIST & 51,000  & 9,000   & 10,000  & 10\\
        notMNIST & 15,526  & 2,739  & 459 & 10\\
        Fashion-MNIST & 9,574  & 1,689  & 1,874 & 10\\
        CIFAR10 & 42,500  & 7,500  & 10,000 & 10\\
        SVHN & 62,269 & 10,988  & 26,032 & 10\\
        \bottomrule
      \end{tabular}
      %}
    % \end{small}
  \end{center}
  \caption[caption]{5-dataset-CV statistics. |Train|, |Dev|, |Test| denotes the number of examples in the train, dev, and test splits respectively. |L| denotes the number of classes for each task.}
  \label{tab:5data_tasks}
\end{table}

\subsubsection{CV Benchmarks}
% \textbf{CV benchmarks.} 
Below, we provide further details on the data sets we utilized for our CV experiments: 5-dataset-CV (diverse) and Split CIFAR-50/ Split CIFAR-100 (homogenous).

\begin{itemize}
    \item \textit{5-dataset-CV} consists of five diverse 10-way image classification tasks: CIFAR-10 \citep{krizhevsky2009learning}, MNIST \citep{lecun1998mnist}, Fashion-MNIST \citep{xiao2017fashion}, SVHN \citep{Netzer2011}, and notMNIST \citep{notmnist}. It is one of the largest data sets used for lifelong learning experiments \citep{ebrahimi2020adversarial} with overall $180.9$k train examples (see Table~\ref{tab:5data_tasks} for task-specific statistics).
    \item \textit{Split CIFAR-50} takes the first 50 classes of the CIFAR-100 image classification data set \citep{krizhevsky2009learning} and randomly splits them into five homogenous 10-way classification tasks. Each task contains $5,000/1,000$ (train/test) examples. We built this data set as a homogenous counterpart to 5-dataset-CV by mimicking its task structure (10 classes/task) and the number of tasks. Further, we note that Split CIFAR-50 (10 classes/ task) is more challenging than Split MNIST/ CIFAR-10 (2 classes/ task) because of the more number of classes per task. Therefore, in this work, we prefer Split CIFAR-50 over MNIST/CIFAR-10 for our experimentation. 
    \item \textit{Split CIFAR-100} splits the CIFAR-100 data set into 20 disjoint 5-way classification tasks, with each task containing $2,500/500$ (train/test) examples. Due to the large number of tasks in this data set, it is regarded as one of the most challenging and realistic CV benchmarks for lifelong learning \citep{chaudhry2018efficient}.
\end{itemize}

\begin{table*}  
  \begin{center}
    \begin{small}
      %\resizebox{\textwidth}{%
      \begin{tabular}{lll|rrrrl}
        \toprule
        Task                     & Data set/ & Domain(s)/         & |Train| & |Dev| & |Test| & |L| & Metrics     \\
                                 & Corpus   & Text source(s)     &         &       &        &     &             \\
        \bottomrule
        Linguistic               & CoLA     & Journal articles   & 7,695   & 856   & 1,043  & 2   & Matthews    \\
        Acceptability            &          & \& books           &         &       &        &     & correlation \\
        \midrule
        Boolean          & BoolQ    & Google queries,    & 8,483   & 944   & 3,270  & 2   & Acc.        \\
        Question                &          & Wikipedia &         &       &        &     &             \\
        Answering                &          & passages &         &       &        &     &             \\
        \midrule
        Sentiment        & SST-2    & Movie      & 9,971   & 873   & 872    & 2   & Acc.        \\
        Analysis       &     & reviews      &    &    &     &   &         \\
        \midrule
        Paraphrase      & QQP      & Quora    & 10,794  & 4,044 & 4,043  & 2   & Acc. \& F1  \\
        Detection     &       & questions    &   &  &   &    &   \\
        \midrule
        Q \& A     & YahooQA  & Yahoo!      & 13,950  & 4,998 & 4,998  & 10  & Acc.        \\
        Categorization    &   & Answers     &   &  &   &   &         \\
        \midrule
        Review Rating            & Yelp     & Business  & 12,920  & 3,999 & 3,998  & 5   & Acc.        \\
        Prediction               &          &   reviews                 &         &       &        &     &             \\
        \midrule
        Event Factuality        & Decomp   & FactBank                  & 10,176  & 4,034 & 3,934  & 2   & Acc.        \\
        \midrule
        Argument Aspect         & AAC      & Web      & 10,893  & 2,025 & 4,980  & 3   & Acc. \& F1  \\
        Detection                &          & documents         &         &       &        &     &             \\
        \midrule
        Explicit Discourse      & DISCONN8 & Penn Discourse    & 9,647   & 1,020 & 868    & 8   & Acc. \& F1  \\
        Marker Prediction        &          & TreeBank                   &         &       &        &     &             \\
        \midrule
        Question & QNLI     & Wikipedia          & 9,927   & 5,464 & 5,463  & 2   & Acc.        \\
        Answering NLI &          &                    &         &       &        &     &             \\
        \midrule
        Binary Sentence         & RocBSO   & Roc story,         & 10,000  & 2,400 & 2,400  & 2   & Acc.        \\
        Order Prediction         &          & corpus             &         &       &        &     &             \\
        \midrule
        Natural Language        & MNLI     & speech, fiction,   & 11,636  & 4,816 & 4,815  & 3   & Acc.        \\
        Inference                &          & govt. reports      &         &       &        &     &             \\
        \midrule
        Multi-choice & SciTAIL  & Science exams      & 11,145  & 1,305 & 1,304  & 2   & Acc.        \\
        Science QA &          &                    &         &       &        &     &             \\
        \midrule
        Implicit Discourse      & PDTB2L1  &   Penn Discourse                 & 13,046  & 1,183 & 1,046  & 4   & Acc. \& F1  \\
        Relation  &          &   TreeBank                 &         &       &        &     &             \\
        Classification  &          &               &         &       &        &     &             \\
        \midrule
        Emotion                 & Emotion  & Twitter            & 9,600   & 2,000 & 2,000  & 6   & Acc. \& F1  \\
        Detection                &          &                    &         &       &        &     &             \\
        \bottomrule
      \end{tabular}
      %}
    \end{small}
  \end{center}
  \caption[caption]{15-dataset-NLP: Task/Data set description and statistics. All tasks are either a single sentence or sentence pair classification. |Train|, |Dev|, |Test| denotes the number of examples in the train, dev, and test splits respectively. |L| denotes the number of classes for each task.}
  \label{tab:summary_tasks}
\end{table*}

\subsubsection{NLP Benchmarks}
% \textbf{NLP benchmarks.} 
Below, we provide further details on the data sets we utilized for our NLP experiments: Split YahooQA (homogenous) and 5-dataset-NLP (diverse). 
\begin{itemize}
    \item \textit{Split YahooQA} consists of five homogenous 2-way classification tasks and is built from a 10-way topic classification data set (YahooQA; \citealp{zhang2015character}) by randomly splitting topics into different tasks. Each task includes around $279$k/$12$k (train/test) examples.
    \item \textit{5-dataset-NLP} consists of text classification data sets \citep{zhang2015character} from 5 diverse domains: AGNews, Yelp, Amazon, DBPedia, and YahooQA. Following \citet{wang2020efficient}, we have $115$k/$7.6$k (train/test) examples per task.
\end{itemize}

\subsubsection{15-Dataset-NLP Benchmark}
One of the objectives of our work is to study the role of different pre-trained initializations in lifelong learning. To enable this study, we introduce \textit{15-dataset-NLP}, a novel suite of diverse tasks for lifelong learning. It consists of fifteen text classification tasks covering a broad range of domains and data sources. Although there exists a setup with 4 tasks spanning 5 data sets, 5-dataset-NLP \citep{de2019episodic}, we show that our introduced benchmark proves more challenging (see Table \ref{tab:pretrainingstudy} and Section \ref{sec:pretrainingstudy}) than the previous setup for the Transformer models (e.g., DistilBERT, BERT, RoBERTa) considered in our study. 

The 15-dataset-NLP benchmark consists of single-sentence or sentence pair classification tasks. We design our benchmark from existing tasks such that (1) the overall data set includes various domains, (2) different tasks are (dis)similar to each other, thereby, facilitating both transfer and interference phenomena. All tasks under consideration differ in data set size (from 8.5k-400k), so for our experiments, we only use between 8.5-14k training examples from each task. Lifelong learning from highly imbalanced data is an interesting problem, and we feel that our introduced benchmark can be used to investigate this problem as well. % As we gather data for all tasks from publicly available sources, for some tasks we do not have access to hidden test examples. As our data is gathered from publicly available sources, for some tasks we do not have access to hidden test examples. 
In such cases, we consider dev examples as test split and sample examples from train split for validation %\footnote{We plan to release sampled example indices for replicability of our results}.  
We describe the tasks below and Table~\ref{tab:summary_tasks} details the evaluation metrics and train/dev/test split sizes for each task.

\begin{enumerate}
    \item Linguistic acceptability aims at identifying whether the given sequence of words is a grammatical sentence. The Corpus of Linguistic Acceptability (\textit{CoLA}; \citealp{warstadt2019neural}) consists of English sentences annotated with their grammatical judgments. The data spans multiple domains, specifically books, and journal articles.    
    
    \item Boolean QA is a reading comprehension task of answering yes/no questions for a given passage. The Boolean Questions (\textit{BoolQ}; \citealp{clark2019boolq}) data set consists of short passages with yes/no questions about the passage. The questions are sourced from anonymous Google users and paired up with passages from Wikipedia articles.
    
    \item Sentiment analysis is a binary classification task of identifying the polarity (positive/negative sentiment) of a given text. The Stanford Sentiment Treebank (\textit{SST-2}; \citealp{socher2013recursive}) corpus consists of sentences from Rotten Tomatoes movie reviews annotated with their sentiment.
    
    \item Paraphrase detection aims at identifying whether two sentences are semantically equivalent. The Quora Question Pairs (\textit{QQP}) corpus constitutes of question pairs from \textit{Quora}\footnote{https://www.quora.com/share/First-Quora-Dataset-Release-Question-Pairs} website annotated for semantic equivalence of question pairs.
    
    \item Q\&A categorization is a topic classification task of categorizing question and answer text pairs into existing topics. The Yahoo! Answers Comprehensive Questions and Answers  (\textit{YahooQA}; \citealp{zhang2015character}) corpus contains data corresponding to the ten largest categories from Yahoo! Webscope program.
    
    \item Review rating prediction is a five-way classification task of predicting the number of stars the user has given in a review given the corresponding text. The \textit{Yelp} \citep{zhang2015character} data set contains business reviews obtained from the Yelp Dataset Challenge (2015).
    
    \item Event factuality prediction is the task of determining whether an event described in the text occurred. The factuality annotations from the \textit{Decomp} corpus are recast into an NLI structure and we use the modified data set from Diverse NLI Collection \citep{poliak2018collecting}. 
    
    \item Argument aspect mining is concerned with the automatic recognition and interpretation of arguments (assessing the stance, source, and supportability for a given topic). The Argument Aspect Corpus (\textit{AAC}; \citealp{stab2018cross}) has over 25,000 arguments spanning eight topics annotated with three labels (no argument, supporting argument, opposing argument). \citet{stab2018cross} collected the data from web documents representing a range of genres and text types, including blogs, editorials, forums, and encyclopedia articles.
    
    \item The explicit discourse marker prediction task aims at classifying the discourse markers between sentences. Specifically, words like 'and', 'but', 'because', 'if', 'when', 'also', 'while', and 'as' mark the conceptual relationship between sentences (\textit{DISCONN8}) and are considered as labels for this task as discussed in \citep{prasad2019penn, kim2020implicit}. We use examples from the Penn Discourse Treebank 3.0 marked for explicit discourse relationship for our experimentation.
    
    \item Question-answering NLI (\textit{QNLI}) is a task adapted from the SQuAD by converting it into the sentence pair classification task  \citep{wang2018glue}. QNLI is a binary classification task of detecting whether the context sentence contains the answer to the question.
    
    \item Binary Sentence Ordering (BSO) is a binary classification task to determine the order of two sentences. This task is similar to pre-training objectives considered in recent works. We use Roc Stories (\textit{RocBSO}; \citealp{mostafazadeh-etal-2016-corpus}) corpus for constructing the data set for this task.
    
    \item Natural language inference (NLI) is a three-way classification task of predicting whether the premise entails the hypothesis (entailment), contradicts the hypothesis (contradiction), or neither (neutral). The Multi-Genre Natural Language Inference (\textit{MNLI}; \citealp{williams2018broad}) corpus consists of sentence pairs from different sources (transcribed speech, fiction, and government report) annotated for textual entailment. 
    
    \item Multi-choice QA is a reading comprehension task wherein given a passage and question, models need to pick up the right option out of provided ones. \citet{khot2018scitail} cast the multiple-choice science exam questions into an NLI structure to convert them to the binary classification task. We use the \textit{SciTAIL} \citep{khot2018scitail} data set released by them for our experimentation.
    
    \item Implicit discourse relation classification is a common task of identifying discourse relations between two text spans or arguments. The Penn Discourse Treebank 3.0 (\textit{PDTB3L1}; \citealp{prasad2019penn, kim2020implicit}) contains a hierarchical annotation scheme (top-level senses, fine-grained level-2 senses) and we use top-level senses comprising of four labels (expansion, comparison, contingency, temporal) for our experimentation.
    
    \item Emotion detection is a classification task of detecting the emotions from a given text snippet. We use \textit{Emotion} data set \citep{saravia-etal-2018-carer} which contains Twitter messages with six emotions: anger, fear, joy, love, sadness, and surprise.
\end{enumerate}

\subsection{Task sequences}
\label{sec:task_sequences}
One of the desiderata of a lifelong learning method is to be robust to different task sequences as task ordering is unknown beforehand. Hence, we run all of our experiments with different random task sequences and report average performance. Below, we list our task sequences.

\begin{itemize}
    \item For Split CIFAR-50 and Split CIFAR-100, the task sequences were generated by randomly sampling classes without replacement for each task, similar to \citet{chaudhry2019tiny}. Thus, the sequences were different for every random seed, but since we ran each method with the same 5 seeds, each method was trained and tested on the same 5 sequences. 
    \item For Split YahooQA, we created 5 tasks by using disjoint groups of consecutive classes (e.g. $\{0, 1\}, \{2, 3\} \dots$). These tasks were then randomly ordered for each task sequence, and each method was trained and tested using the same 5 random sequences.
    \item For 5-dataset-CV, we randomly select the following data set orders:
        \begin{enumerate}
            \item [Seq1.] SVHN$\rightarrow$notMNIST$\rightarrow$Fashion-MNIST$\rightarrow$CIFAR-10$\rightarrow$MNIST
            \item [Seq2.] SVHN$\rightarrow$MNIST$\rightarrow$notMNIST$\rightarrow$Fashion-MNIST$\rightarrow$CIFAR-10
            \item [Seq3.] CIFAR-10$\rightarrow$SVHN$\rightarrow$notMNIST$\rightarrow$Fashion-MNIST$\rightarrow$MNIST
            \item [Seq4.] notMNIST$\rightarrow$Fashion-MNIST$\rightarrow$CIFAR-10$\rightarrow$SVHN$\rightarrow$MNIST
            \item [Seq5.] CIFAR-10$\rightarrow$MNIST$\rightarrow$notMNIST$\rightarrow$SVHN$\rightarrow$Fashion-MNIST
        \end{enumerate}
    \item For 5-dataset-NLP, we randomly select the following data set orders (the first 4 are consistent with \citealp{de2019episodic}): 
        \begin{enumerate}
            \item [Seq1.] Yelp$\rightarrow$AGNews$\rightarrow$DBPedia$\rightarrow$Amazon$\rightarrow$YahooQA 
            \item [Seq2.] DBPedia$\rightarrow$YahooQA$\rightarrow$AGNews$\rightarrow$Amazon$\rightarrow$Yelp
            \item [Seq3.] Yelp$\rightarrow$YahooQA$\rightarrow$Amazon$\rightarrow$DBpedia$\rightarrow$AGNews
            \item [Seq4.] AGNews$\rightarrow$Yelp$\rightarrow$Amazon$\rightarrow$YahooQA$\rightarrow$DBpedia 
            \item [Seq5.] YahooQA$\rightarrow$Yelp$\rightarrow$DBPedia$\rightarrow$AGNews$\rightarrow$Amazon
        \end{enumerate}
    \item For 15-dataset-NLP, we randomly select and use the following $5$ data set orders: 
        \begin{enumerate}
          \item [Seq1.] Decomp$\rightarrow$BoolQ$\rightarrow$AAC$\rightarrow$Yelp$\rightarrow$DISCONN8$\rightarrow$SST-2$\rightarrow$QQP$\rightarrow$YahooQA$\rightarrow$QNLI$\\
          \rightarrow$RocBSO$\rightarrow$MNLI$\rightarrow$SciTAIL$\rightarrow$CoLA$\rightarrow$PDTB2L1$\rightarrow$Emotion
          \item [Seq2.] CoLA$\rightarrow$QQP$\rightarrow$MNLI$\rightarrow$QNLI$\rightarrow$Emotion$\rightarrow$SST-2$\rightarrow$BoolQ$\rightarrow$Decomp$\rightarrow$AAC\\$\rightarrow$SciTAIL$
          \rightarrow$RocBSO$\rightarrow$Yelp$\rightarrow$PDTB2L1$\rightarrow$YahooQA$\rightarrow$DISCONN8
          \item [Seq3.] SciTAIL$\rightarrow$BoolQ$\rightarrow$SST-2$\rightarrow$AAC$\rightarrow$DISCONN8$\rightarrow$YahooQA$\rightarrow$QNLI$\rightarrow$RocBSO\\$\rightarrow$PDTB2L1$
          \rightarrow$Emotion$\rightarrow$Decomp$\rightarrow$MNLI$\rightarrow$QQP$\rightarrow$CoLA$\rightarrow$Yelp
          \item [Seq4.] DISCONN8$\rightarrow$QNLI$\rightarrow$CoLA$\rightarrow$YahooQA$\rightarrow$AAC$\rightarrow$SciTAIL$\rightarrow$PDTB2L1$\rightarrow$Emotion$\\
          \rightarrow$Decomp$\rightarrow$RocBSO$\rightarrow$QQP$\rightarrow$Yelp$\rightarrow$MNLI$\rightarrow$BoolQ$\rightarrow$SST-2
          \item [Seq5.] Emotion$\rightarrow$SST-2$\rightarrow$RocBSO$\rightarrow$YahooQA$\rightarrow$AAC$\rightarrow$MNLI$\rightarrow$CoLA$\rightarrow$DISCONN8\\$\rightarrow$QQP$\rightarrow$QNLI$\rightarrow$Decomp$\rightarrow$PDTB2L1$\rightarrow$SciTAIL$\rightarrow$Yelp$\rightarrow$BoolQ
        \end{enumerate}
\end{itemize}

\subsection{Evaluation}
Let $S_{t,\tau}$ denote the accuracy on task $\tau$ after training on task $t$. After model finishes training on the task $t$, we compute the \textit{average accuracy} ($A_t$), \textit{forgetting} ($F_t$) and \textit{learning accuracy} ($LA_t$) metrics as proposed by \cite{lopez2017gradient, riemer2018learning}. $F_{t}$ (also referred to as backward transfer) measures the influence of learning task $t$ on the performance of all previously seen tasks $\tau, (1 \leq \tau < t)$. As the model learns multiple tasks in the sequence, we hope that knowledge acquired during lifelong learning aids the learning of new tasks (forward transfer). $LA_t$ measures the learning capability when the model sees the new task $t$ (indirectly measuring the forward transfer). Say we learn the $t^\text{th}$ task, then $A_t$, $F_t$ and $LA_t$ are defined as follows 
\begin{align}
    \centering
    A_t = \frac{1}{t} \sum_{\tau=1}^{t} S_{t,\tau} , \qquad F_t = \frac{1}{t-1} \sum_{\tau=1}^{t-1} \max_{\tau' \in \{1,\cdots, t-1\}} (S_{\tau',\tau} - S_{t,\tau}), \qquad LA_t = \frac{1}{t} \sum_{\tau=1}^{t} S_{\tau,\tau}. \label{eq:metrics}
\end{align}

\subsection{Methods}
\label{sec:baseline_methods}
We compare our approach with state-of-the-art methods for task-incremental lifelong learning \citep{chaudhry2019tiny, mirzadeh2020linear}. 
\begin{itemize}
\item \textit{Finetune (FT)}: The model is sequentially fine-tuned on each task without additional learning constraints.
\item \textit{Elastic weight consolidation} (\textit{EWC}; \citealp{kirkpatrick2017overcoming}): A regularization-based approach that tries to mitigate forgetting by restricting learning to parameters important to previously learned tasks, as measured by the Fisher information matrix.
\item \textit{Averaged Gradient Episodic Memory} (\textit{A-GEM}; \citealp{chaudhry2018efficient}): A data-based regularization approach that augments the base model with an episodic memory module that retains examples from the previously seen tasks, and during training, uses these stored examples to enforce a constraint on the gradients, ensuring that the model does not forget previously learned tasks. 
\item \textit{Episodic replay} (\textit{ER}; \citealp{chaudhry2019tiny}): This approach involves the use of a replay buffer to store and replay past experiences during training. This enables the model to revisit and learn from previously seen examples, mitigating catastrophic forgetting and enhancing its ability to retain knowledge across multiple tasks. Following \citet{chaudhry2019tiny}, we retain one example per task per class and randomly select examples for storage. \citet{prabhu2020gdumb, hussain2021towards} show that the straightforward ER method outperforms all of the previous methods under realistic task-incremental learning settings, and therefore, we compare our approach mainly with ER.
\item \textit{Stable SGD} \citep{mirzadeh2020understanding}: This method alters the training process by adjusting hyperparameters such as learning rate, batch size, learning rate decay, and dropout regularization (see Appendix~\ref{sec:impldetails} for hyperparameter sweep). The goal is to introduce inherent noise in the stochastic gradients, resulting in convergence to wide regions within the loss landscape, which in turn leads to reduced forgetting during continual learning.
\item \textit{Mode Connectivity SGD} (\textit{MC-SGD}; \citealp{mirzadeh2020linear}): This approach restricts the minima of continual learning within a region of low loss by all previous minima. MC-SGD can be viewed as a combination of both regularization and replay-based methods in continual learning as it uses a small replay buffer to approximate a low-loss path for previous tasks.
\end{itemize}

%% file: jmlrsections/03_experimentdesign.tex
\section{Does pre-training implicitly alleviate forgetting?}
\label{sec:experiment}
Having defined the formal problem setup, evaluation metrics, and methods for alleviating the forgetting phenomenon, in this section, we conduct experiments to tease apart the role of pre-training for lifelong learning. We are interested in answering the following questions---\textit{(Q1) How much does pre-training help in alleviating forgetting? (Q2) Do pre-trained models undergo similar forgetting on diverse and homogeneous tasks? (Q3) How do different pre-trained initializations affect forgetting?} 

\textbf{Experimental design.} To answer these questions convincingly, we conduct experiments on the above-discussed CV and NLP benchmarks. We utilize the $\text{DistilBERT}_{\text{base}}$ \citep{sanh2019distilbert} architecture for text classification and the $\text{ResNet-18}$ \citep{he2016deep} architecture for image classification. To isolate the effect of pre-training, we consider two variants for each of these architectures: pre-trained models (\textit{DistilBERT-PT}/\textit{ResNet-18-PT}) and randomly initialized models (\textit{DistilBERT-R}/\textit{ResNet-18-R}). For our study, we need to ensure that there are as few confounding factors as possible. Therefore, we keep all other hyperparameters the same and vary only the initialization (for more details refer to Appendix~\ref{sec:impldetails}). To measure the severity of forgetting, ideally, we want sufficient training samples so that both a pre-trained model or a randomly initialized model (of the same capacity) achieves similar learning accuracy on each task. 
To control for this behavior we either select a large training corpus whenever available (e.g., 279k examples/task for Split YahooQA) or run our experiments for multiple epochs (5 epochs for CV benchmarks). 

\textbf{ImageNet pre-training corpus.} For a fair comparison between pre-trained and randomly initialized models, we explicitly control for and remove the overlap between pre-training and downstream tasks. Publicly available ResNet models are pre-trained on ImageNet that overlaps with CIFAR-100 in terms of class labels. Therefore, we make sure that the subset of the ImageNet corpus we use does not have any visually and semantically overlapping classes with the CIFAR-100 data set. We use the publicly available \citep{abdelsalam2021iirc} two-level class hierarchies for ImageNet, where semantically and visually similar labels are grouped under one super-category. We iterate over all CIFAR-100 labels and drop the complete super-category from ImageNet corresponding to each of these labels. For example, CIFAR-100 contains a \textit{castle} class and we have a \textit{building} super-category in ImageNet that contains \textit{castle, palace, monastery, church, etc.}. We remove all building-related labels from our pre-training data set. In total, we remove 267 classes and pre-train the \textit{ResNet-18-PT} model on the remaining subset of the ImageNet data set.

\subsection{How much does pre-training help in alleviating forgetting?}
\label{sec:pretrain_alleviate_forgetting}
\begin{figure}[t!]
    \centering
    \begin{subfigure}{.33\textwidth}
      \centering
      \includegraphics[width=\textwidth]{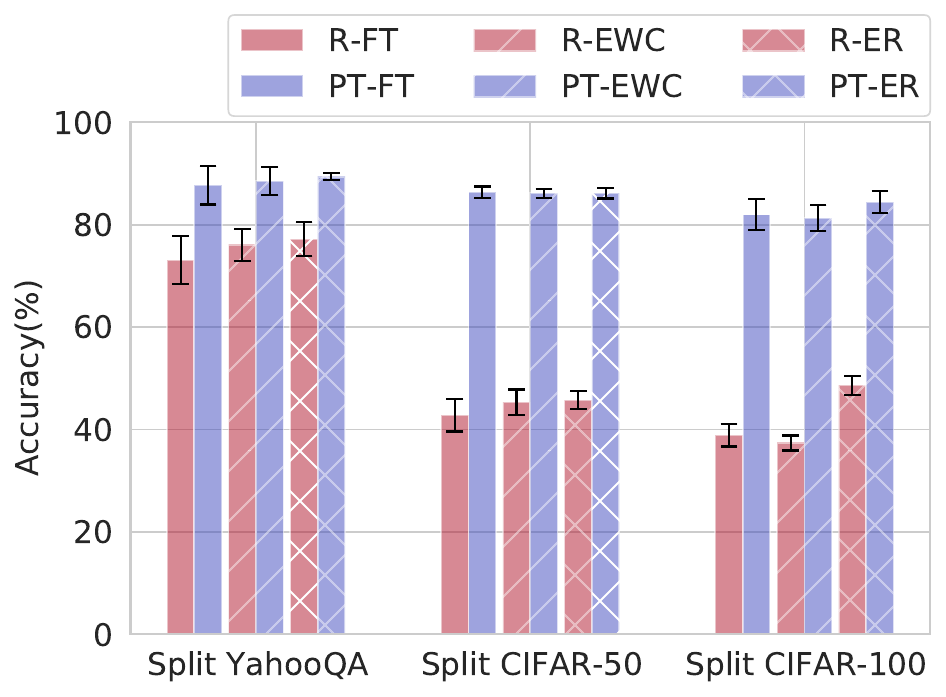}
      \caption{Accuracy ($\uparrow$)}
      \label{fig:homogenous_acc}
    \end{subfigure}\hspace{\fill}%
    \begin{subfigure}{.33\textwidth}
      \centering
      \includegraphics[width=\textwidth]{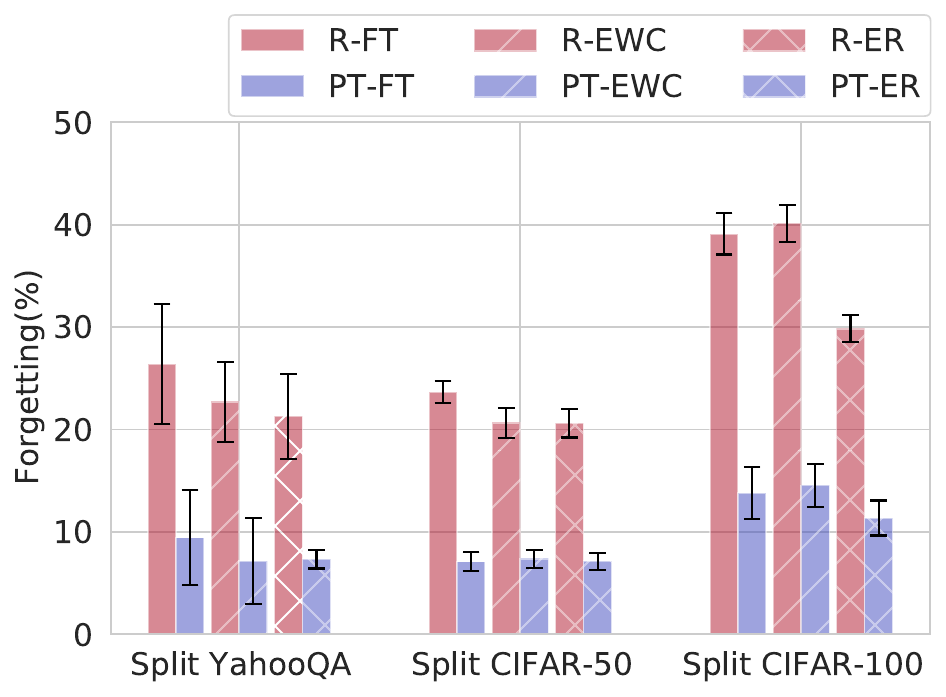}
      \caption{Forgetting ($\downarrow$)}
      \label{fig:homogenous_forget}
    \end{subfigure}\hspace{\fill}%
    \begin{subfigure}{.33\textwidth}
      \centering
      \includegraphics[width=\textwidth]{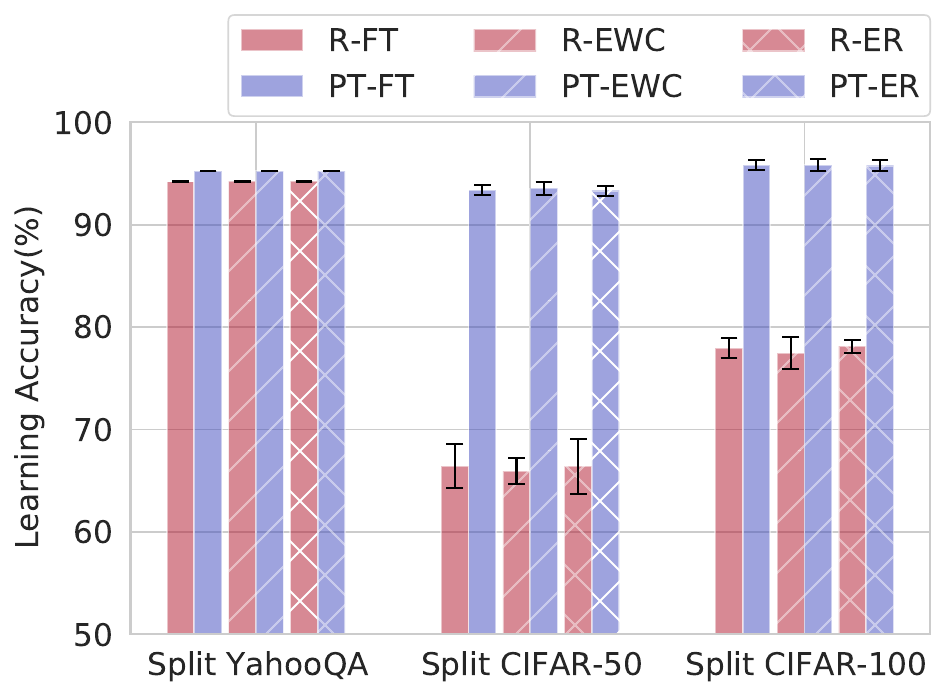}
      \caption{Learning Accuracy ($\uparrow$)}
      \label{fig:homogenous_learnacc}
    \end{subfigure}\hspace{\fill}%
    \caption{Comparing performance on homogenous tasks (Split YahooQA/ CIFAR-50/ CIFAR-100) across initialization (R: random, PT: pre-trained) and methods (FT: finetune, EWC: elastic weight consolidation, ER: episodic replay) after training on the last task. $\uparrow$ indicates higher is better, $\downarrow$ indicates lower is better. All metrics are averaged over 5 random task sequences. We observe that pre-trained models undergo significantly less forgetting in comparison to randomly initialized models.}
    \label{fig:homogenous_tasks}
\end{figure}

\begin{figure}
    \centering
    \begin{subfigure}{.33\textwidth}
      \centering
      \includegraphics[width=\textwidth]{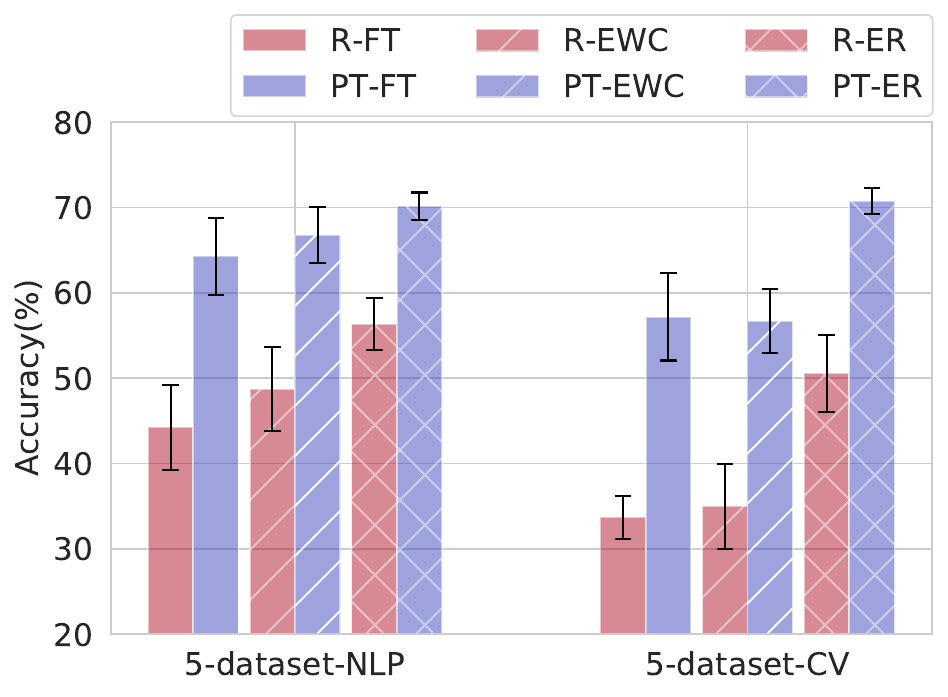}
      \caption{Accuracy ($\uparrow$)}
      \label{fig:diverse_acc}
    \end{subfigure}\hspace{\fill}%
    \begin{subfigure}{.33\textwidth}
      \centering
      \includegraphics[width=\textwidth]{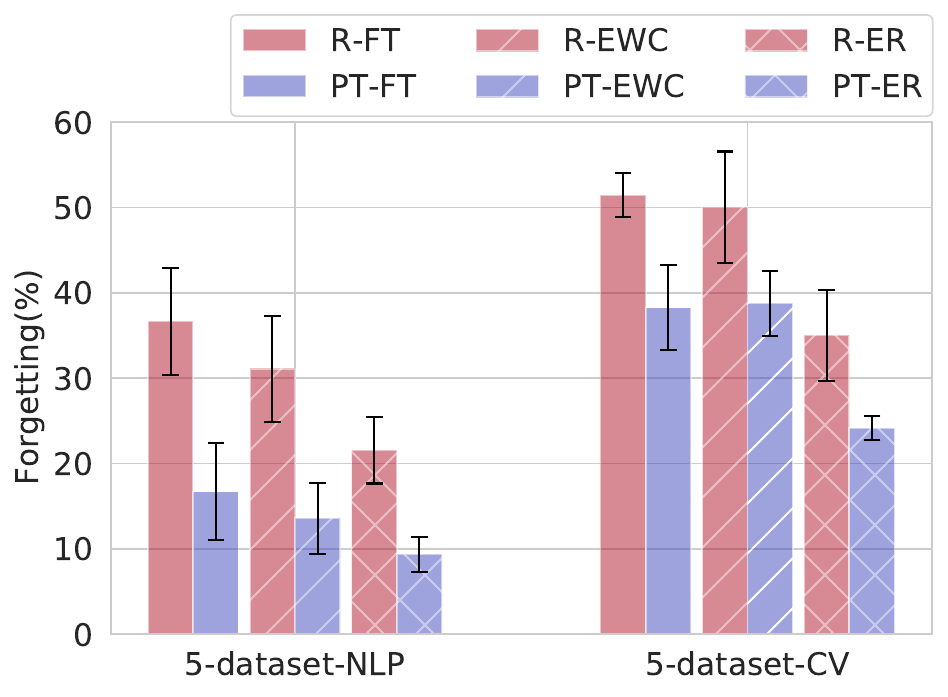}
      \caption{Forgetting ($\downarrow$)}
      \label{fig:diverse_forget}
    \end{subfigure}\hspace{\fill}%
    \begin{subfigure}{.33\textwidth}
      \centering
      \includegraphics[width=\textwidth]{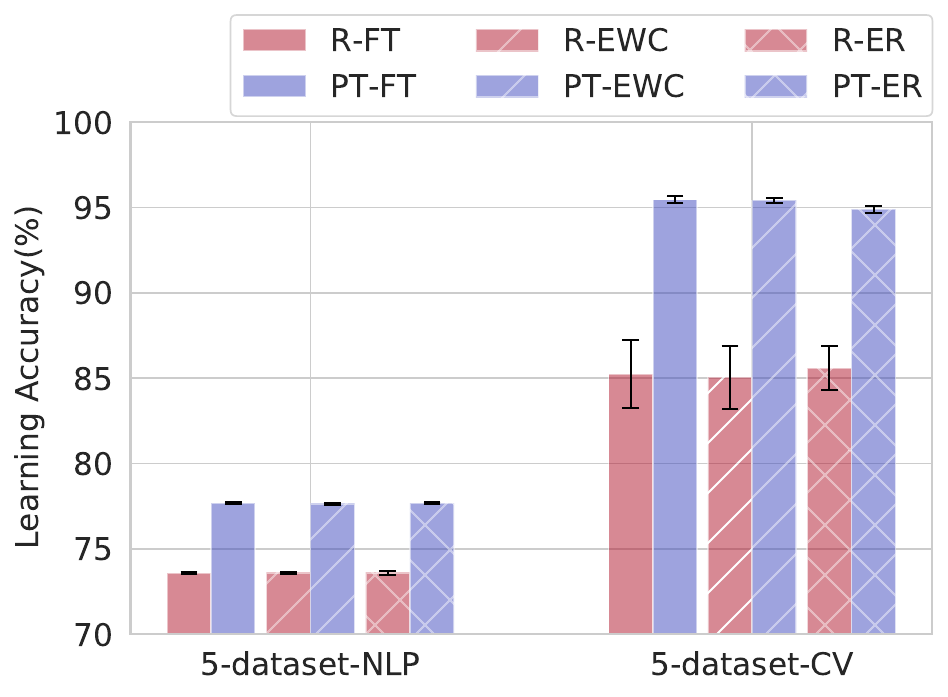}
      \caption{Learning Accuracy ($\uparrow$)}
      \label{fig:diverse_learnacc}
    \end{subfigure}\hspace{\fill}%
    \caption{Comparing performance on diverse tasks (5-dataset-NLP/ CV) across initialization (R: random, PT: pre-trained) and methods (FT: finetune, EWC: elastic weight consolidation, ER: episodic replay) after training on the last task. $\uparrow$ indicates higher is better, $\downarrow$ indicates lower is better. All metrics are averaged over 5 task sequences. In comparison to homogenous tasks, we observe that pre-trained models are more susceptible to forgetting when exposed to a diverse sequence of tasks.}
    \label{fig:diverse_tasks}
\end{figure}

From Figures~\ref{fig:homogenous_forget} and \ref{fig:diverse_forget} (and Table~\ref{tab:summary}), we see that pre-trained models (ResNet-18-PT, DistilBERT-PT) undergo significantly less forgetting in comparison to models with random initialization (ResNet-18-R, DistilBERT-R). This trend holds across all three methods --- FT, EWC, and ER. For text data sets (Split YahooQA, 5-dataset-NLP), we see that both models have comparable learning accuracy (see Figures~\ref{fig:homogenous_learnacc} and \ref{fig:diverse_learnacc}) and significantly less forgetting for DistilBERT-PT. This can be completely attributed to the pre-trained initialization.  
On 5-dataset-CV, ResNet-18-PT undergoes less forgetting ($38.3$) when compared to ResNet-18-R ($51.5$) (see Table~\ref{tab:summary}). Specifically, despite task accuracy starting at a higher base for ResNet-18-PT, \textit{the absolute forgetting value is still lower compared to ResNet-18-R models}. Additionally, this effect also holds when considering a sequentially finetuned pre-trained model (with no additional mechanism to alleviate forgetting) to a randomly initialized model trained with lifelong learning methods. For example, on 5-dataset-NLP, sequentially finetuning DistilBERT-PT undergoes less forgetting ($16.7$) compared to the competitive ER method ($21.6$) when applied to DistillBERT-R. This raises an interesting research direction---\textit{explicitly focusing on learning generic initialization for future tasks apart from just concentrating on the forgetting aspect of lifelong learning}. 

\subsection{Do pre-trained models undergo similar forgetting on diverse and homogeneous tasks?}
\label{sec:diverse_and_homogeneous}

From Figure~\ref{fig:homogenous_forget}, we see that ResNet-18-PT does not undergo a significant amount of forgetting when sequentially fine-tuned on Split CIFAR-50, and Split CIFAR-100 (homogenous tasks). On Split CIFAR-50, forgetting is around $7\%$ absolute points. Surprisingly, the competitive ER method also undergoes a similar amount of forgetting, thereby raising a question about the applicability of these data sets when studying forgetting in the context of the pre-trained models. It may be possible to manually cluster tasks based upon semantic closeness, rendering severe interference to make these benchmarks more challenging \citep{ramasesh2020anatomy}. We leave an analysis of pre-trained models on such variants to future work. 
Given the generic nature of the pre-trained initialization, we ask---\textit{what happens when we train the model sequentially on diverse tasks?} To answer this question, we conduct experiments on 5-dataset-CV and 5-dataset-NLP. From Figures~\ref{fig:homogenous_forget}, \ref{fig:diverse_forget} (and Table~\ref{tab:summary}), \textit{we empirically observe that pre-trained models are more susceptible to forgetting when exposed to diverse tasks in comparison to homogenous tasks}. Particularly, DistilBERT-PT/ResNet-18-PT undergoes a $16.73/38.28\%$ absolute points drop in accuracy when trained on 5-dataset-NLP/5-dataset-CV (see Table~\ref{tab:summary} for exact values). Figures~\ref{fig:homogenous_acc}, \ref{fig:diverse_acc} report average accuracy after training on the last task. We report task-specific results for 5-dataset-NLP/5-dataset-CV in Appendix \ref{sec:taskspecificres}.

\begin{figure}
    \centering
    \begin{subfigure}{.33\textwidth}
      \centering
      \includegraphics[width=\textwidth]{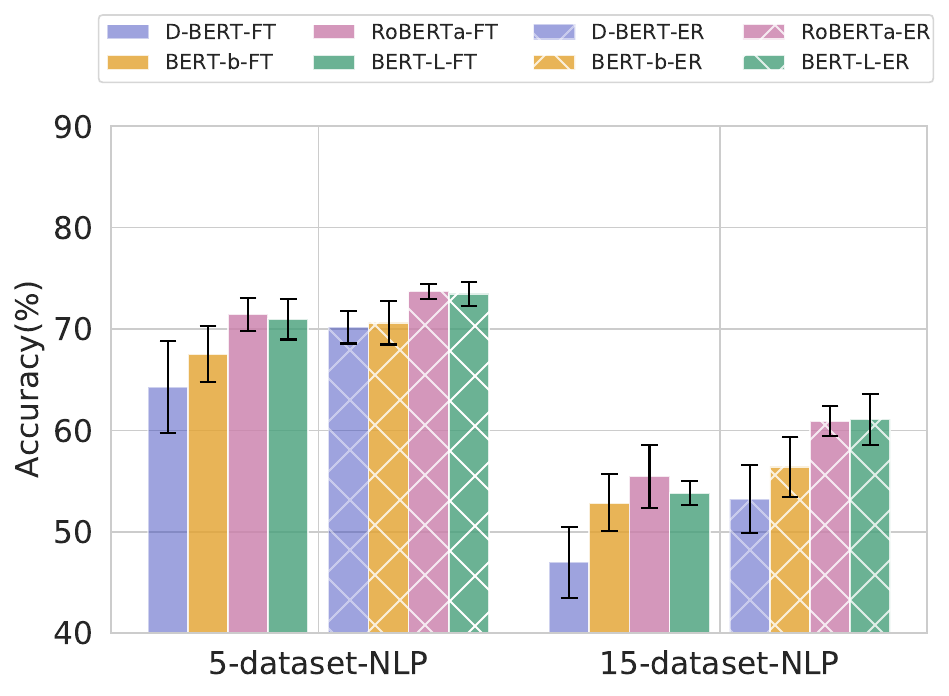}
      \caption{Accuracy ($\uparrow$)}
      \label{fig:pretrain_acc}
    \end{subfigure}\hspace{\fill}%
    \begin{subfigure}{.33\textwidth}
      \centering
      \includegraphics[width=\textwidth]{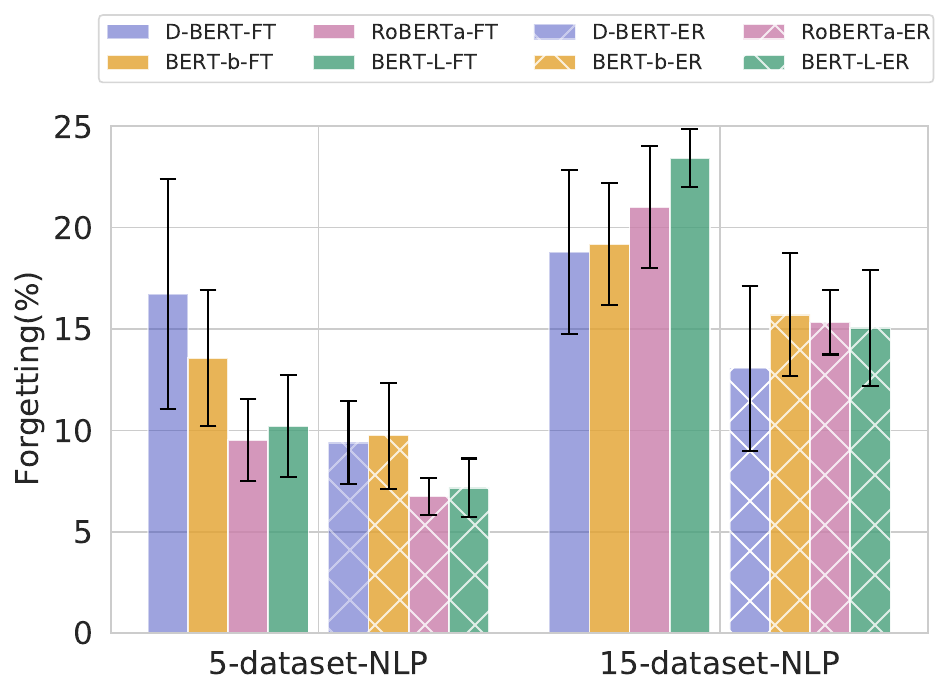}
      \caption{Forgetting ($\downarrow$)}
      \label{fig:pretrain_forget}
    \end{subfigure}\hspace{\fill}%
    \begin{subfigure}{.33\textwidth}
      \centering
      \includegraphics[width=\textwidth]{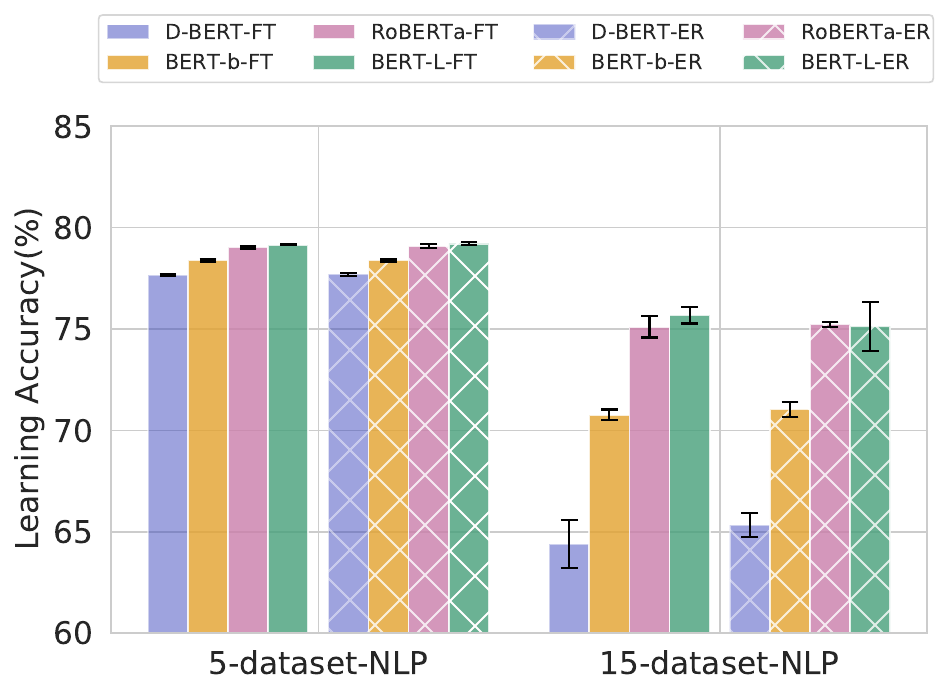}
      \caption{Learning Accuracy ($\uparrow$)}
      \label{fig:pretrain_learnacc}
    \end{subfigure}\hspace{\fill}%
    \caption{Comparing performance on diverse tasks (5-dataset-NLP/ 15-dataset-NLP) across different pre-trained Transformer models (D-BERT: DistilBERT, BERT-b: BERT-base, RoBERTa: RoBERTa-base, BERT-L: BERT-Large) and methods (FT: finetune, ER: episodic replay) after training on the last task. $\uparrow$ indicates higher is better, $\downarrow$ indicates lower is better. All metrics are averaged over 5 random task sequences. Overall, we observe that larger models and models pre-trained on diverse corpora (RoBERTa-base) undergo less forgetting on both $5$ and $15$ diverse tasks.}
    \label{fig:pretrain_initialization}
\end{figure}

\subsection{How do different pre-trained initialization affect forgetting?}
\label{sec:pretrainingstudy}

To examine the impact of varying pre-trained initialization on forgetting, we evaluate different pre-trained Transformer models, DistilBERT \citep{sanh2019distilbert}, BERT \citep{devlin2019bert}, RoBERTa \citep{liu2019roberta}, on text classification tasks. 
From the previous subsection, we observe that pre-trained models are relatively more susceptible to forgetting on LL of diverse tasks. In response, we conduct a thorough investigation on the 5-dataset-NLP. From Figure~\ref{fig:pretrain_initialization} (and Table~\ref{tab:pretrainingstudy}), we observe that when keeping the pre-training corpora the same and increasing the capacity of the model -- DistilBERT (66M), BERT-base (110M), and BERT-large (336M) -- we observe that larger models undergo less forgetting on sequential finetuning of diverse tasks. Further, to understand the impact of the diversity of the pre-training corpora, we compare BERT-base (110M) with RoBERTa-base (125M). We observe that the RoBERTa-base model performs far superior to BERT-base, thus hinting at the necessity of diverse pre-training corpora to alleviate forgetting implicitly. To stress-test these models, we experiment with the 15-dataset-NLP. We observe that by increasing the number of tasks in the sequence, pre-trained models undergo severe forgetting. Surprisingly, the RoBERTa-base model outperforms BERT-Large despite having many fewer parameters. Empirically, we infer that \textit{diversity of pre-training corpora plays a vital role in easing forgetting during lifelong learning of diverse tasks.}

%% file: jmlrsections/04_losslandscape.tex
\section{Exploring the Loss Landscape}
\label{sec:loss_landscape}
To better understand how pre-training reduces forgetting, we perform experiments analyzing where models are situated in the loss landscape after training on each task.
We denote model parameters after training on task $k$ as $w_k$.
If we define forgetting as the increase in loss for a given task during training (instead of a decrease in accuracy), \citet{mirzadeh2020understanding} show that the forgetting can actually be bounded by
\begin{equation}
    % L_1(w_2^*) - L_1(w_1^*) \approx \frac{1}{2}\Delta w^{\top} \nabla^2 L_1(w_1^*)\Delta w \leq \frac{1}{2} \lambda_1^{max}\lVert{\Delta w}\rVert^2
    \label{eq:forgetting_bound}
    L_1(w_2) - L_1(w_1) \approx \frac{1}{2}\Delta w^{\top} \nabla^2 L_1(w_1)\Delta w \leq \frac{1}{2} \lambda_1^{max}\lVert{\Delta w}\rVert^2,
\end{equation}
where $L_1(w)$ represents the loss on task 1 with parameters $w$, $\Delta w = w_2 - w_1$, and $\lambda_1^{max}$ is the largest eigenvalue of $\nabla^2 L_1(w_1)$. The magnitude of the eigenvalues of $L_1(w)$ can be used to characterize the curvature of the loss function \citep{keskar2016large}, and thus $\lambda_1^{max}$ can be thought of as a proxy for the flatness of the loss function (lower is flatter). From Equation~\ref{eq:forgetting_bound}, we can see that the flatter the minima, the less forgetting occurs in the model.

We hypothesize that the one explanation of improvements from pre-training shown in the previous section might be because pre-training leads to a more favorable loss landscape. 
Specifically, pre-training results in wider/flatter minima for each task. The effect of these wider minima is that the change in weights from learning on future tasks results in a gradual change of the current task loss, thereby reducing forgetting. 
We verify this idea in two parts. First, we use loss contours and then linearly interpolate between sequential minima to show that the flat loss basins lead to smaller changes in the loss. Next, we compute a sharpness metric to show that pre-training indeed leads to flat loss basins. All models analyzed in this section are sequentially trained using the finetune method.

\begin{figure}
    \centering
    \begin{subfigure}{.245\textwidth}
      \centering
      \includegraphics[width=\textwidth]{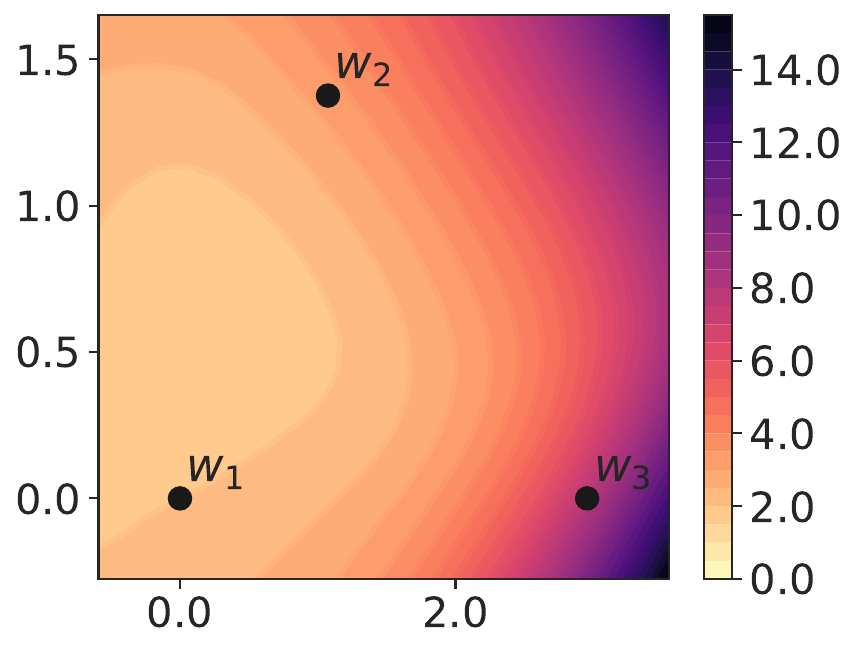}
      \subcaption{5-dataset-CV (R)}
      \label{fig:5data_no_contour}
    \end{subfigure}\hspace{\fill}%
    \begin{subfigure}{.245\textwidth}
      \centering
      \includegraphics[width=\textwidth]{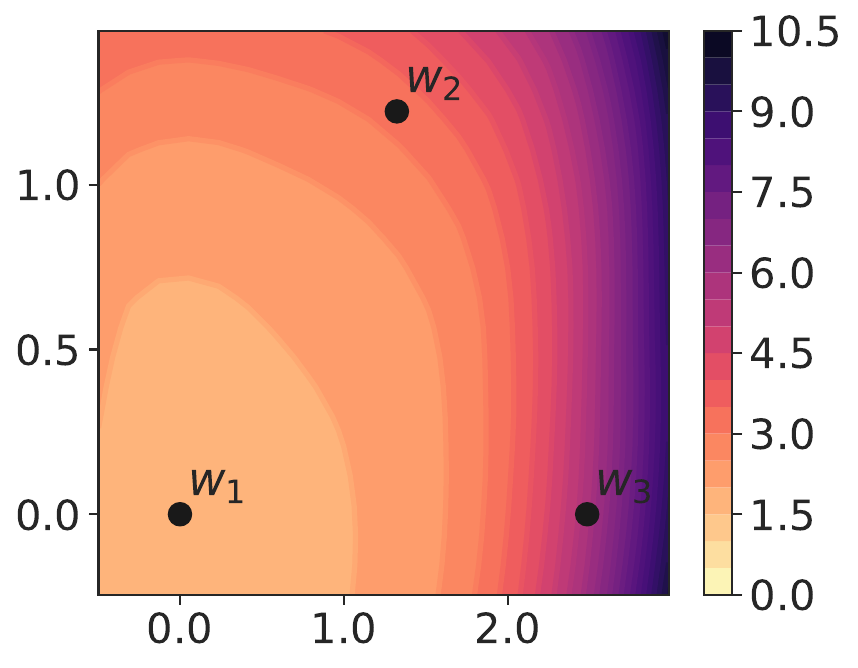}
      \caption{Split CIFAR-50 (R)}
      \label{fig:c50_no_contour}
    \end{subfigure}\hspace{\fill}%
    \begin{subfigure}{.245\textwidth}
      \centering
      \includegraphics[width=\textwidth]{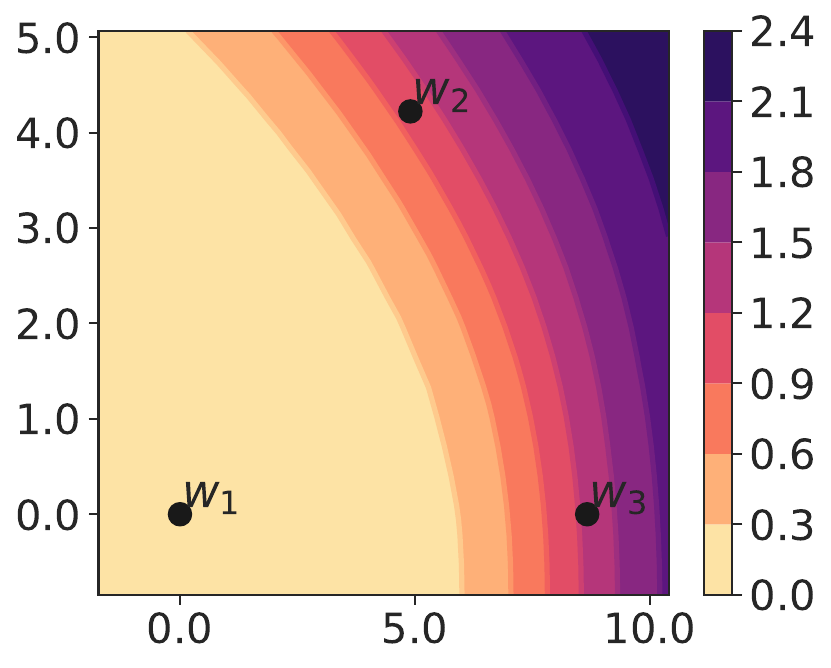}
      \caption{5-dataset-NLP (R)}
      \label{fig:5data_nlp_no_contour}
    \end{subfigure}\hspace{\fill}%
    \begin{subfigure}{.245\textwidth}
      \centering
      \includegraphics[width=\textwidth]{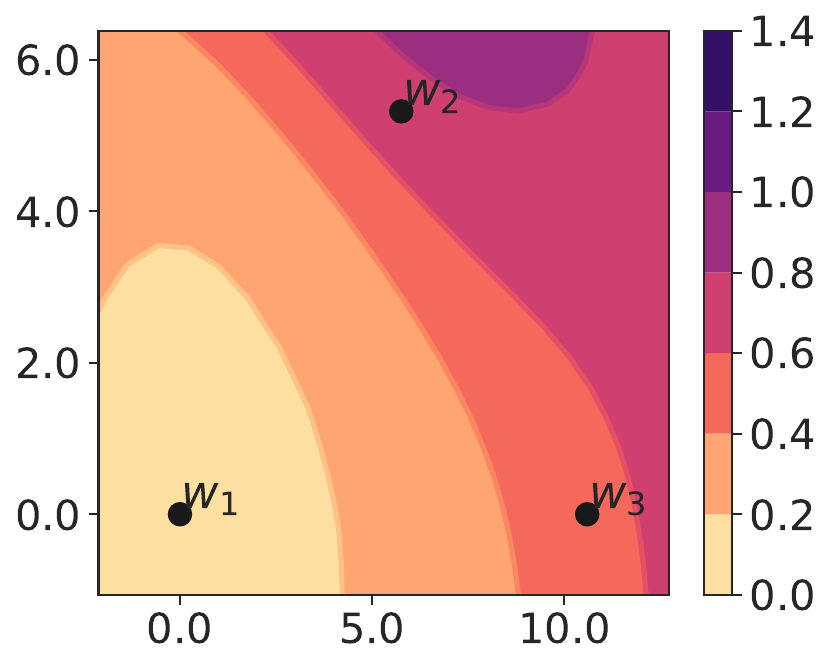}
      \caption{Split YahooQA (R)}
      \label{fig:yahooqa_no_contour}
    \end{subfigure}\hspace{\fill}%
    \bigskip
    \begin{subfigure}{.245\textwidth}
      \centering
      \includegraphics[width=\textwidth]{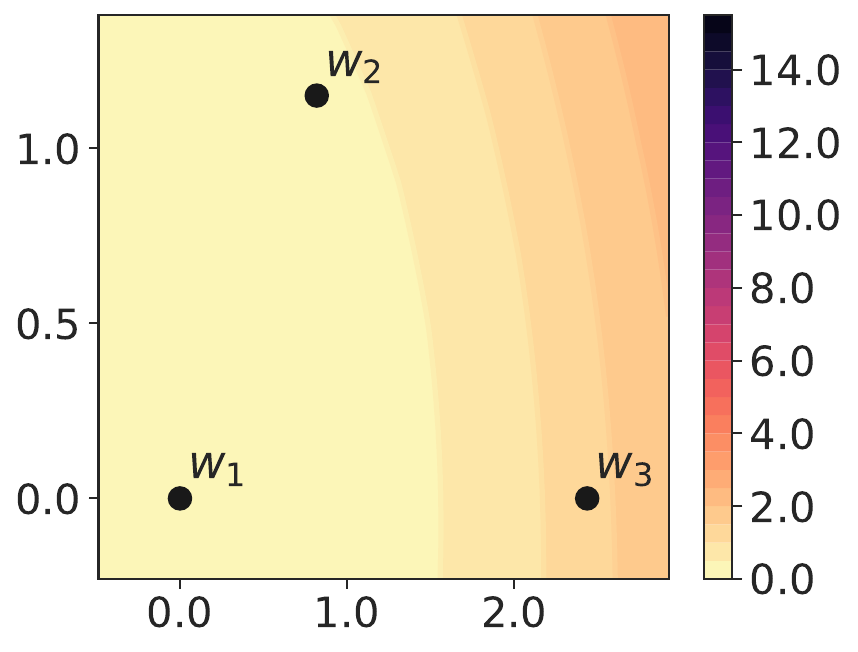}
      \caption{ 5-dataset-CV (PT)}
      \label{fig:5data_pt_contour}
    \end{subfigure}\hspace{\fill}
    \begin{subfigure}{.245\textwidth}
      \centering
      \includegraphics[width=\textwidth]{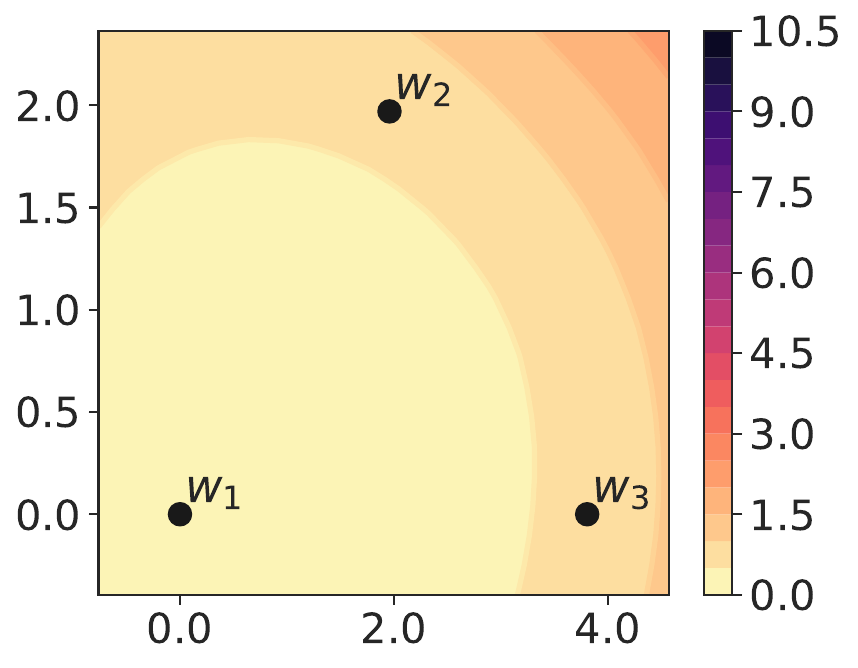}
      \caption{Split CIFAR-50 (PT)}
      \label{fig:c50_pt_contour}
    \end{subfigure}\hspace{\fill}%
    \begin{subfigure}{.245\textwidth}
      \centering
      \includegraphics[width=\textwidth]{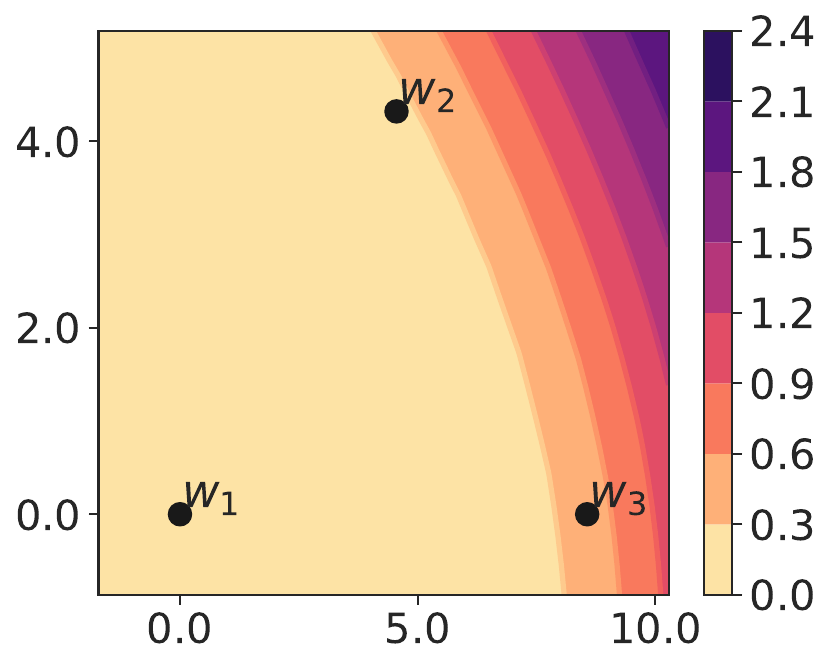}
      \caption{5-dataset-NLP (PT)}
      \label{fig:5data_nlp_pt_contour}
    \end{subfigure}\hspace{\fill}
    \begin{subfigure}{.245\textwidth}
      \centering
      \includegraphics[width=\textwidth]{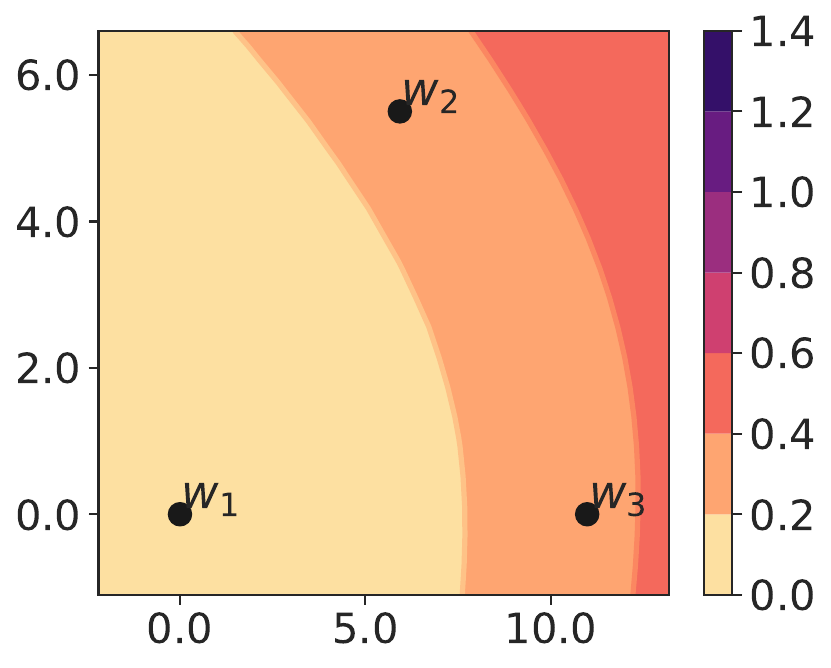}
      \caption{Split YahooQA (PT)}
      \label{fig:yahooqa_pt_contour}
    \end{subfigure}\hspace{\fill}%
    \caption{Loss contours for task 1 where w$_1$, w$_2$, and w$_3$ are minima obtained after sequential training on tasks 1, 2, and 3, respectively. The top row visualizes loss contours for randomly initialized models (R) and the bottom row visualizes loss contours for pre-trained models (PT).}    
    \label{fig:contours}
\end{figure}

\subsection{Loss Contour}
To better understand the changes in task loss during continual training on a sequence of tasks, we utilize loss contours, which involve linearly interpolating between the continual learning minima. In order to construct a 2D loss contour, we require three points to define two basis vectors. Specifically, we train on tasks 1, 2, and 3 sequentially, resulting in minima represented by w$_1$, w$_2$, and w$_3$, respectively.
We designate w$_1$ as our reference point (0, 0) and calculate $\vec{u}$=w$_3 - $w$_1$ as one basis vector (representing the x-axis in our plots). Additionally, we compute an orthogonal projection $\vec{v}$ of w$_2 - $w$_1$ onto $\vec{u}$, which serves as the second basis vector (representing the y-axis in our plots). Consequently, for any coordinate $(x, y)$ within the 2D loss contour, we derive the corresponding model parameters as w$(x, y)$ = w$_1$ + $x$.u + $y$.v and compute the validation loss for the task under consideration. In this setup, the distance between any two points on the loss contour reflects the Euclidean distance between the corresponding model parameters.

In Figure~\ref{fig:contours}, we visualize loss contours for the first task across different data set-specific task sequences. For every contour, we plot minima (w$_1$, w$_2$, w$_3$) of the model after continual training on three tasks (T$_1$, T$_2$, T$_3$). As the model is trained continuously on a sequence of tasks, the pre-training initialized model remains largely at the same loss level (for task 1) as compared to the randomly initialized model, despite drifting a comparable distance away from the original model (w$_1$).
For example, in the loss contour for the pre-trained model on 5-dataset-CV (Figure~\ref{fig:5data_pt_contour}), we observe that the model after training on task 2 (w$_2$) remains at the same loss level as after training on task 1 (w$_1$) and slightly higher loss level after training on task 3 (w$_3$). For the randomly initialized model (Figure~\ref{fig:5data_no_contour}), the Euclidean distance between the model parameter vectors is approximately the same as for the pre-trained model, but the differences in task 1 loss levels are significantly higher. We visualize more instances of loss contours over different task sequences in Appendix \ref{sec:loss_landscape_appendix}. In summary, \textit{we observe that pre-trained models consistently lead to wider loss basins across different data sets (NLP and CV domains), model architectures (ResNet and Transformer), and task sequences (5 random orderings)}.

\begin{figure}
    \centering
    % \begin{subfigure}[b]{.64\textwidth}
        \begin{subfigure}{.36\textwidth}
          \centering
          \includegraphics[width=\textwidth]{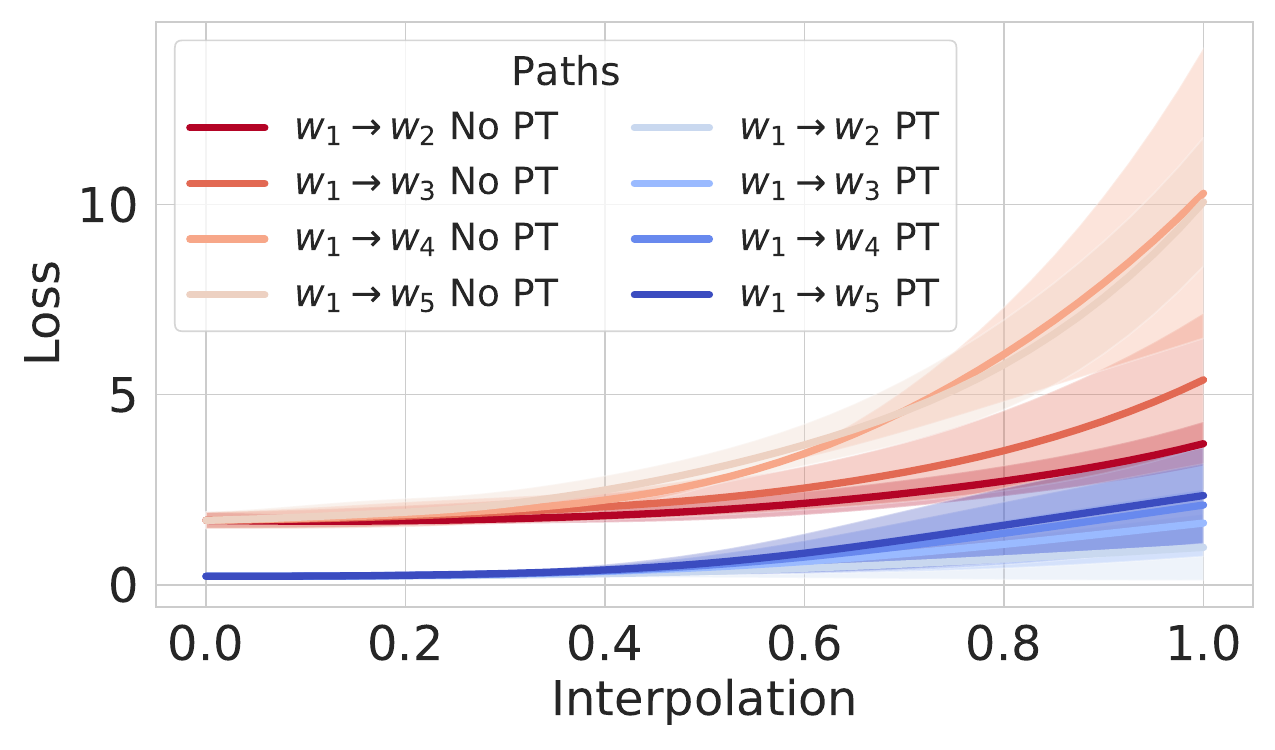}
          \caption{5-dataset-CV (task 1)}
          \label{fig:5data_lmc}
        \end{subfigure}
        \begin{subfigure}{.36\textwidth}
          \centering
          \includegraphics[width=\textwidth]{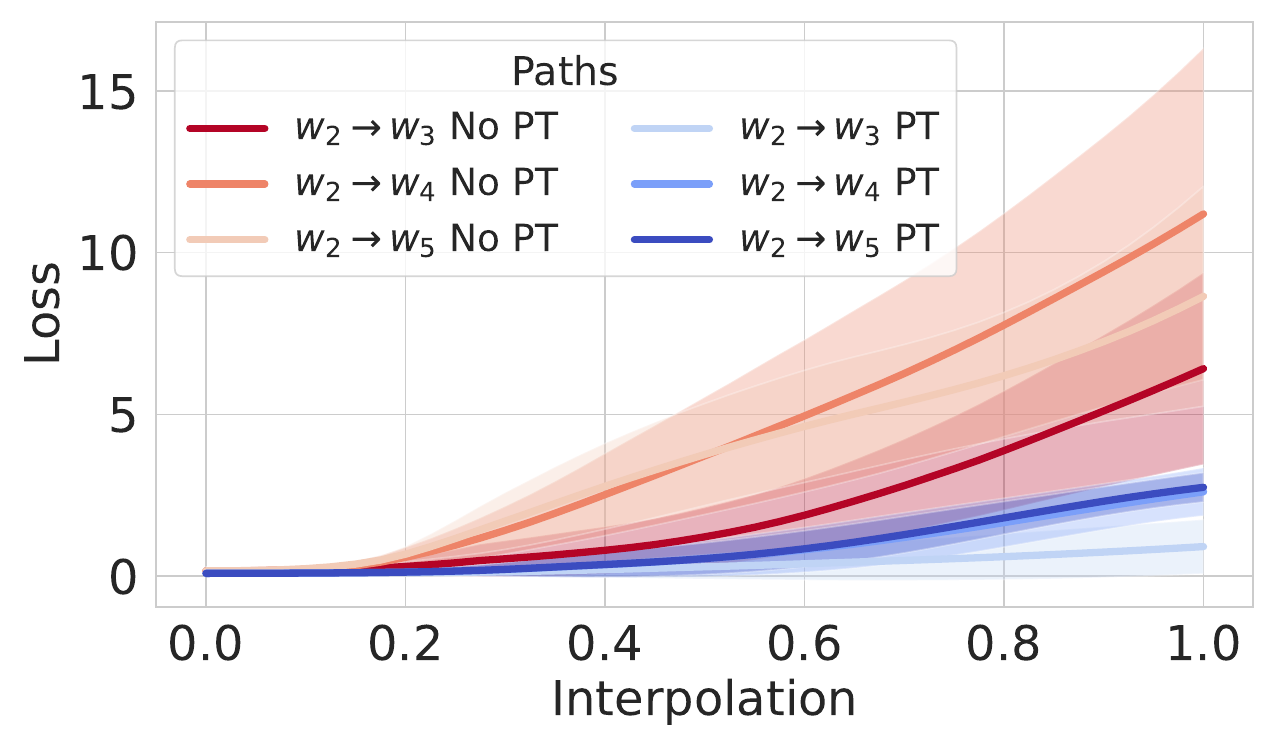}
          \caption{5-dataset-CV (task 2)}
          \label{fig:5data_5epoch_lmcw2}
        \end{subfigure}
        \bigskip
        % \hspace{\fill}%
        \begin{subfigure}{.36\textwidth}
        % \addtocounter{subfigure}{-1}
        % \renewcommand\thesubfigure{\alph{subfigure}2}
          \centering
          \includegraphics[width=\textwidth]{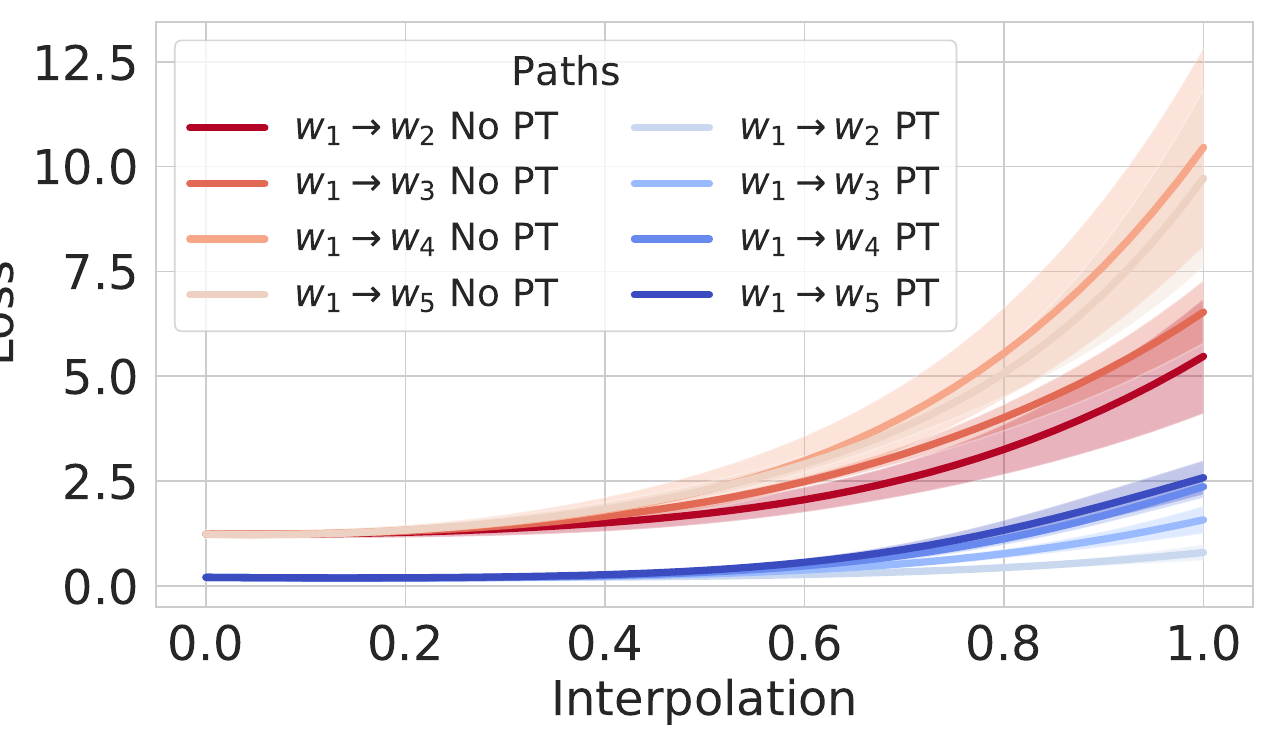}
          \caption{Split CIFAR-50 (task 1)}
          \label{fig:c50_lmc}
        \end{subfigure}
        \begin{subfigure}{.36\textwidth}
        %\addtocounter{subfigure}{-1}
        %\renewcommand\thesubfigure{\alph{subfigure}2}
          \centering
          \includegraphics[width=\textwidth]{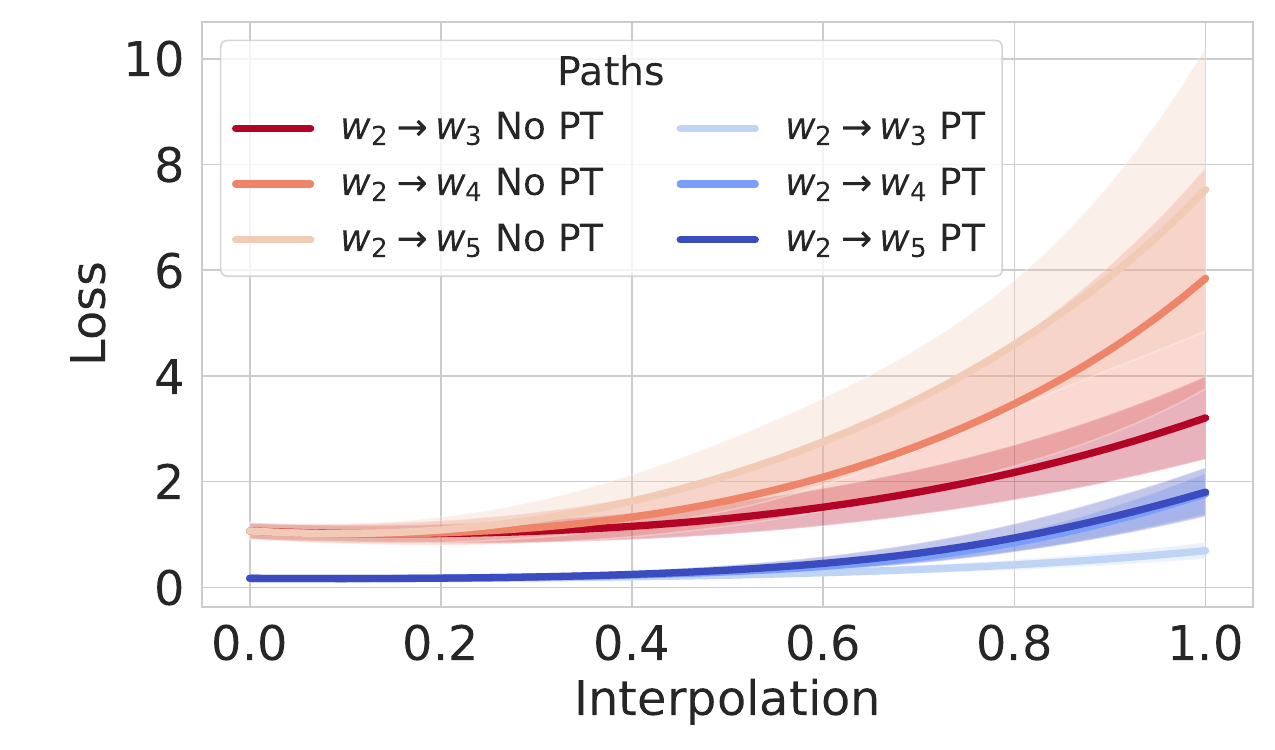}
          \caption{Split CIFAR-50 (task 2)}
          \label{fig:c50_5epoch_lmcw2}
        \end{subfigure}
        \bigskip
        % \hspace{\fill}%
        % \bigskip
        % \hspace{\fill}%
        \begin{subfigure}{.36\textwidth}
        % \addtocounter{subfigure}{-1}
        % \renewcommand\thesubfigure{\alph{subfigure}3}
          \centering
          \includegraphics[width=\textwidth]{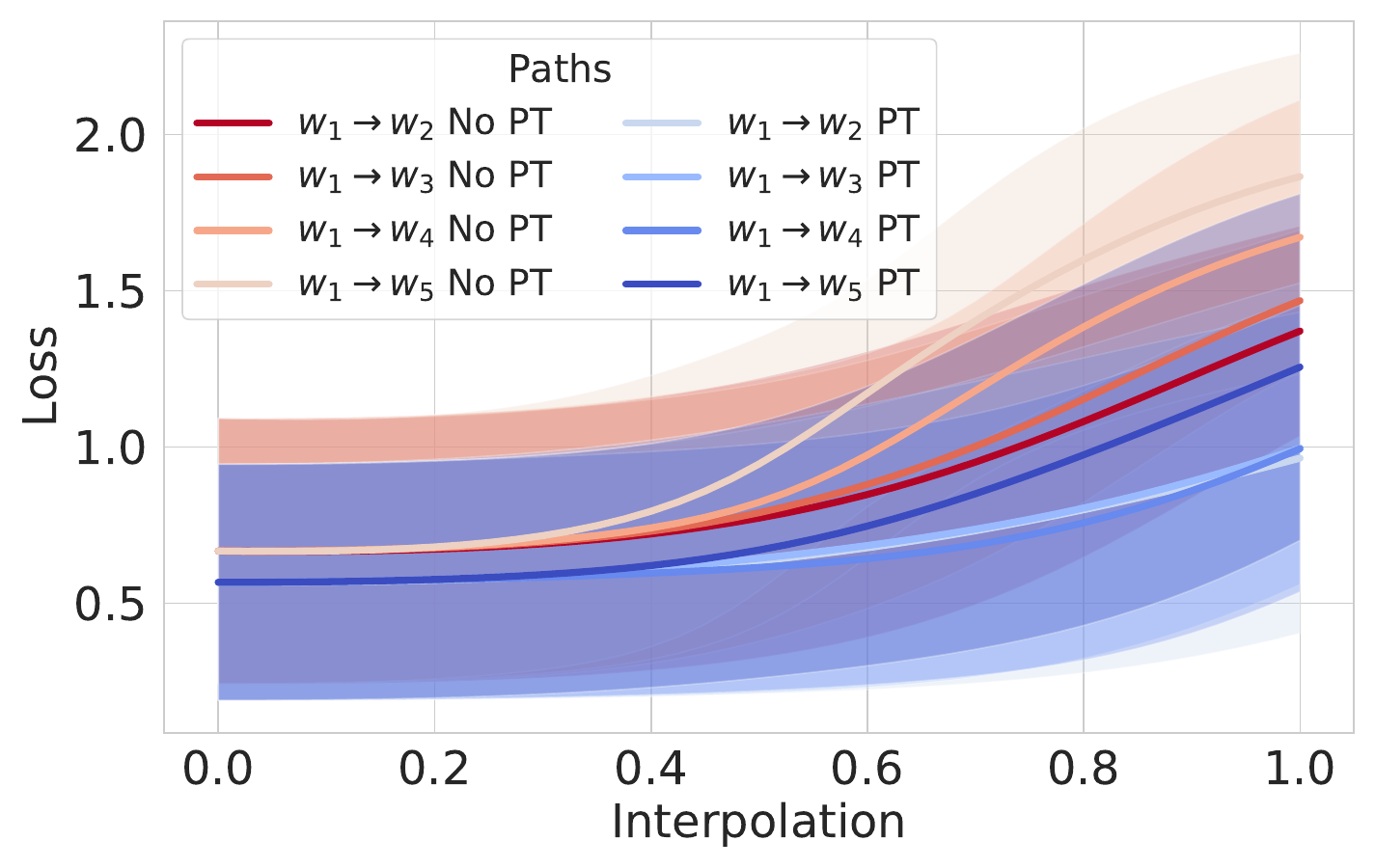}
          \caption{5-dataset-NLP (task 1)}
          \label{fig:5data_nlp_lmc}
        \end{subfigure}
        \begin{subfigure}{.36\textwidth}
        % \addtocounter{subfigure}{-1}
        % \renewcommand\thesubfigure{\alph{subfigure}3}
          \centering
          \includegraphics[width=\textwidth]{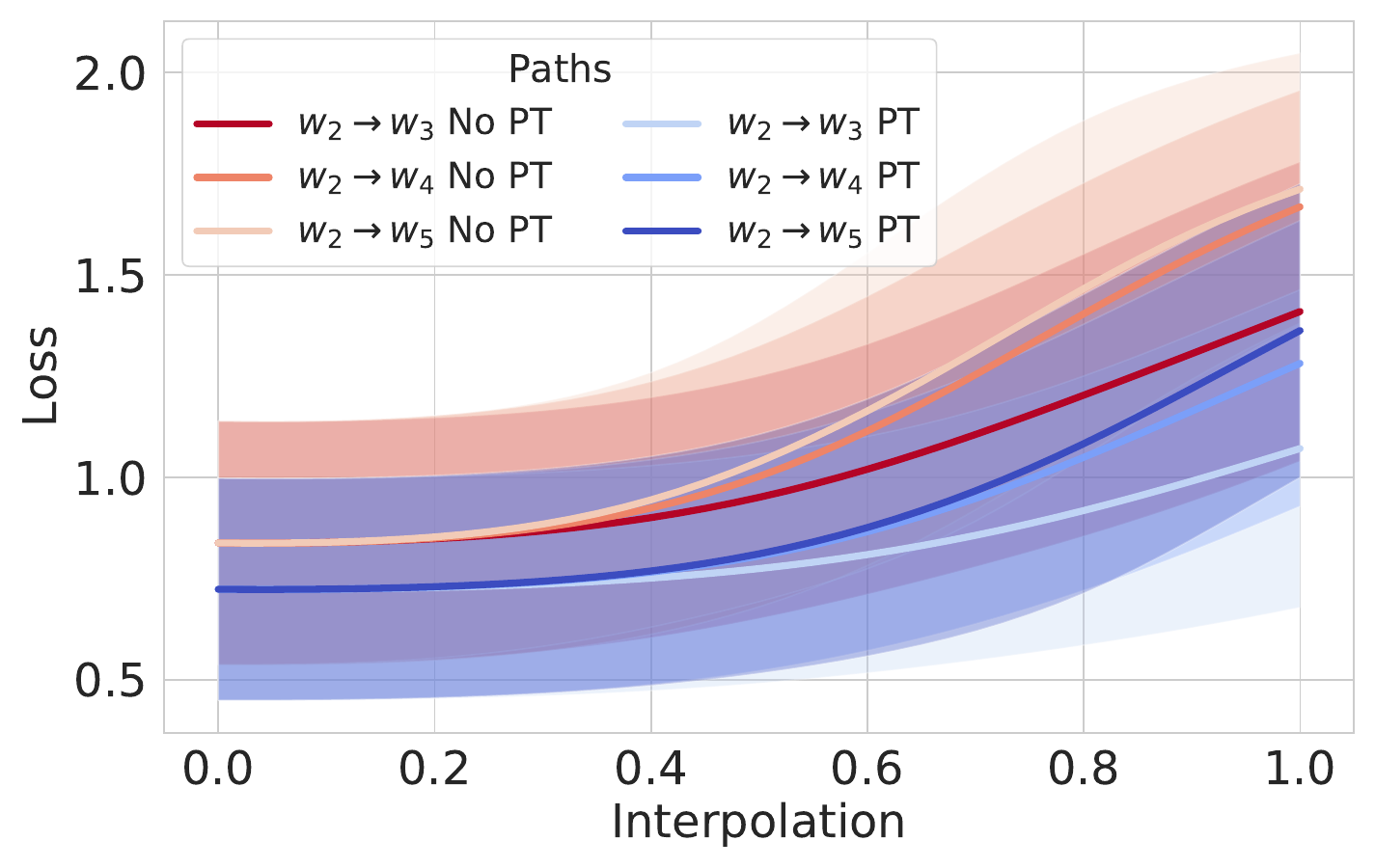}
          \caption{5-dataset-NLP (task 2)}
          \label{fig:5data_nlp_lmcw2}
        \end{subfigure}
        \bigskip
        % \hspace{\fill}%
        \begin{subfigure}{.36\textwidth}
        % \addtocounter{subfigure}{-1}
        % \renewcommand\thesubfigure{\alph{subfigure}4}
          \centering
          \includegraphics[width=\textwidth]{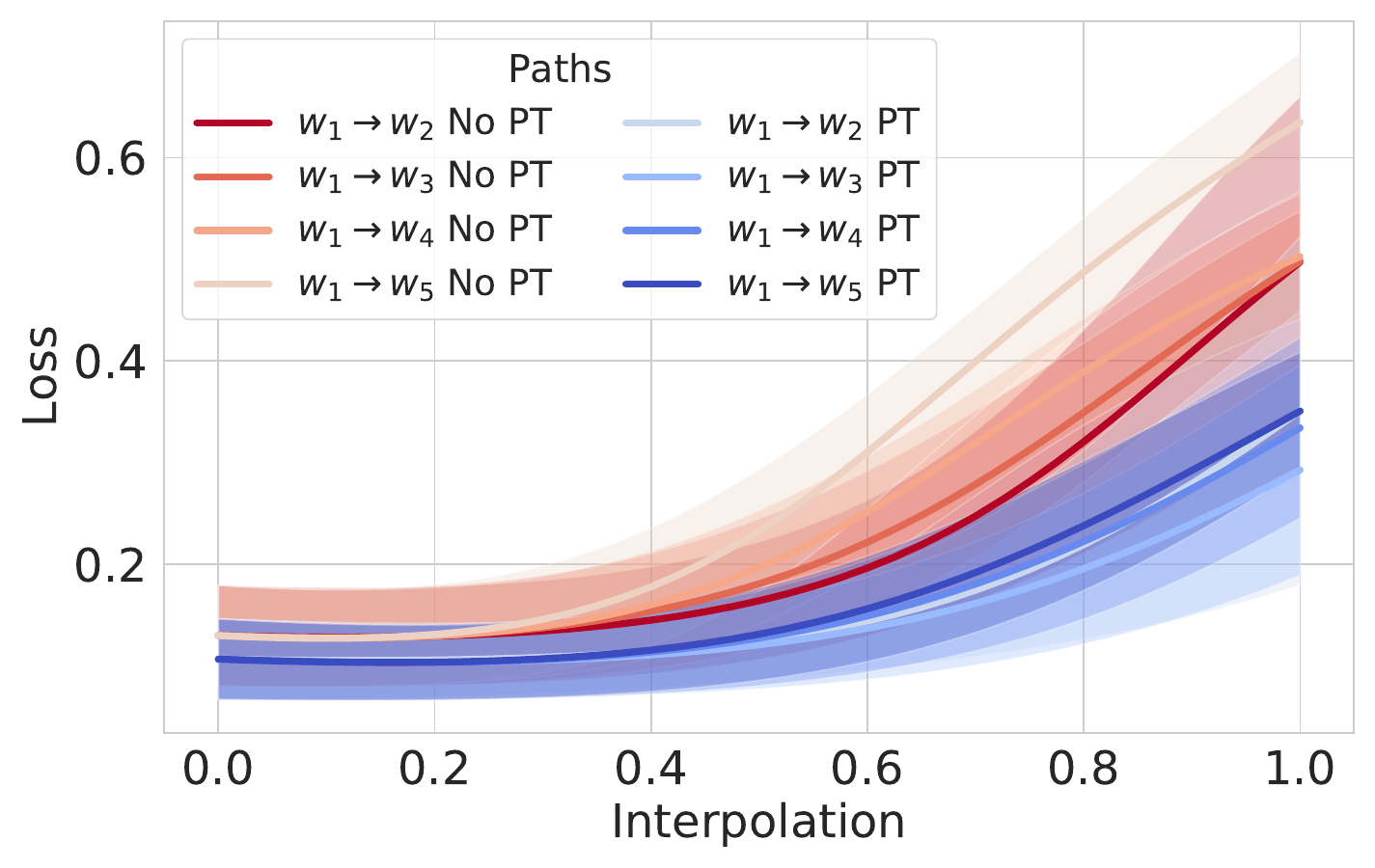}
          \caption{Split YahooQA (task 1)}
          \label{fig:yqa_lmc}
        \end{subfigure}
        % \hspace{\fill}%
        \begin{subfigure}{.36\textwidth}
        % \addtocounter{subfigure}{-1}
        % \renewcommand\thesubfigure{\alph{subfigure}4}
          \centering
          \includegraphics[width=\textwidth]{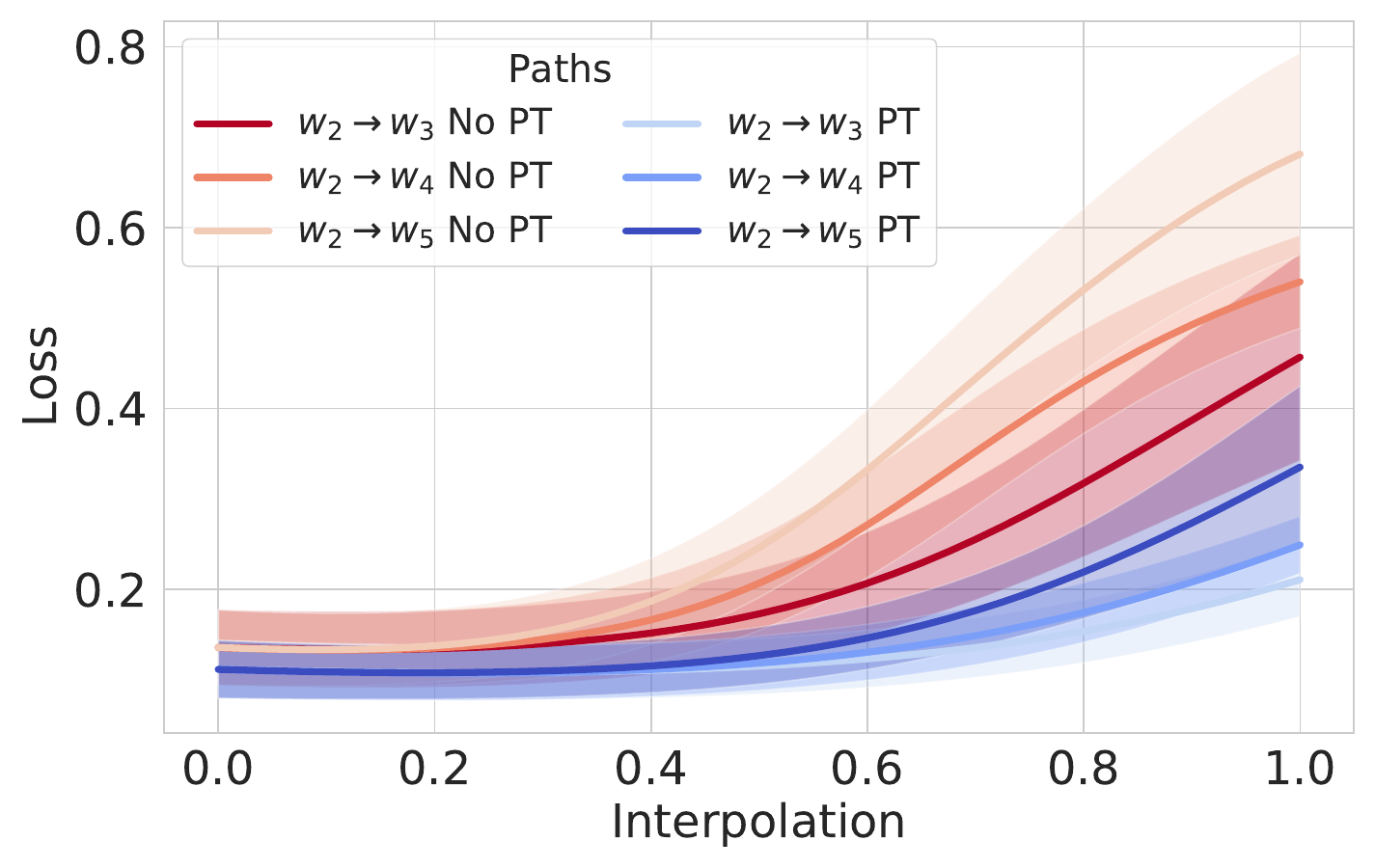}
          \caption{Split YahooQA (task 2)}
          \label{fig:yqa_lmcw2}
        \end{subfigure}
        % \hspace{\fill}
        % \addtocounter{subfigure}{-1}        
        % \end{subfigure}
    \caption{Linear model interpolation plots for different data sets. The plots for pre-training initialized (PT) models are shown in hues of blue, and the randomly initialized (No PT) models are shown in hues of red. We linearly interpolate between the task 1/ task 2 minimum (w$_1$/w$_2$) to the subsequent task minimum (w$_i \rightarrow$ w$_j$, j > i), tracking the loss in the process. In general, the loss landscape is flatter along these paths for pre-trained initialized models compared to randomly initialized models.}
    \label{fig:lmc}
\end{figure}

\subsection{Linear Model Interpolation}
\label{sec:lmi}

Ideally, to ease forgetting during sequential training of tasks, loss on previous tasks should undergo minimal change. This desideratum would be satisfied if a previous task minimum lands in a flat loss region and subsequent task minima also remained in that flat loss region. To probe this behavior, we linearly interpolate between w$_1$ (w$_2$) and subsequent task minima, tracking the (validation) loss on task 1 (task 2). This probe can be interpreted as viewing a slice of the loss contours in Figure \ref{fig:contours} along the linear trajectory connecting w$_1$ (w$_2$) to a subsequent minimum. In Figure~\ref{fig:lmc}, we visualize the results from linear interpolation with the first row for task 1 and the second row for task 2. The plots for pre-training initialized models are shown in hues of blue, and the randomly initialized models are shown in hues of red. These plots illustrate that the pre-training initialized models experience a much more gradual increase in loss compared to the randomly initialized models, even when interpolating to minima after training on several tasks. Moreover, these results hold for task 2 as well, thereby reinforcing that pre-training initialized models lead to flat minima even for subsequent tasks in a sequence. 

\begin{table}      
    %   \small
      \centering
    %   \small
      % \begin{small}
      \begin{tabular}{lrr|rr}
            \toprule
                                    %  \multirow{2}{*}{Dataset}&  
            % \cmidrule(r){2-4}\cmidrule(r){4-5}
                                     & w/o PT & w/ PT       & w/o PT  & w/ PT    \\
            \midrule
            & \multicolumn{2}{c|}{$\epsilon = 5\times 10^{-4}$} & \multicolumn{2}{c}{$\epsilon = 10^{-3}$}                                         \\
            5-dataset-CV & $2.1_{0.6}$ & $0.1_{0.0}$ & $5.7_{1.6}$ & $0.2_{0.0}$ \\
            Split CIFAR-50 & $2.3_{0.7}$ & $0.2_{0.1}$ & $6.1_{1.5}$ & $0.4_{0.1}$\\
            Split CIFAR-100 & $2.3_{0.6}$ & $0.1_{0.0}$ & $5.9_{1.3}$ & $0.2_{0.1}$ \\
            \midrule
            & \multicolumn{2}{c|}{$\epsilon = 5\times 10^{-5}$} & \multicolumn{2}{c}{$\epsilon = 10^{-4}$}                                         \\
            5-dataset-NLP & $32.7_{1.2}$ & $28.3_{1.2}$ & $213.6_{11.5}$ & $129.0_{10.5}$ \\
            Split YahooQA & $10.4_{0.4}$ & $8.8_{0.4}$ & $53.2_{7.0}$ & $43.0_{4.2}$ \\
            \bottomrule
      \end{tabular}
      % \end{small}
      \caption[caption]{Average sharpness value (lower value corresponds to flat loss basin) of task minima. ResNet-18-PT/DistilBERT-PT (w/ PT) has lower average sharpness in comparison to ResNet-18-R/DistilBERT-R (w/o PT). Pre-training reduces the sharpness of minima for each task in training by an order of magnitude.}
      \label{tab:sharpness}
\end{table}

\subsection{Sharpness}
\label{sec:sharpness}

As discussed earlier, the flatness of the minima can be estimated based upon the magnitude of eigenvalues of $\nabla^2L(w)$. However, computing these eigenvalues is computationally expensive. Therefore, \cite{keskar2016large} introduces a sensitivity measure, termed \textit{sharpness metric}, as a computationally feasible alternative to computing eigenvalues. The sharpness metric estimates the flatness by computing the maximum value of the loss function $L$ in a constrained neighborhood around the minima. Given that the maximization process can be inaccurate, \cite{keskar2016large} suggests performing maximization in a random subspace $\mathbb{R}^p$ of the entire parameter space $\mathbb{R}^n$, specified by a projection matrix $A\in \mathbb{R}^{n\times p}$. For our experiments, we randomly sample our matrix $A$ and set $p=100$ as in \citet{keskar2016large}. The neighborhood maximization region ($C_\epsilon$) is defined as
\begin{equation}
\label{eq:sharpness_bounds}
C_\epsilon = \{z \in \mathbb{R}^p: -\epsilon(|(A^+w)_i| + 1) \leq z_i \leq \epsilon(|(A^+w)_i| + 1); \forall i \in \{1\dots p\}\},
\end{equation}
where $A^+$ is the pseudo inverse of $A$, $w$ is the parameter vector and $\epsilon$ is a hyperparameter controlling the size of the neighborhood. Formally, the sharpness metric is defined as follows
\begin{equation}
\label{eq:sharpness_value}
\phi_{w,L} \coloneqq \frac{(\max_{z\in C_\epsilon} L(w + Az)) - L(w)}{1 + L(w)} \times 100,
\end{equation}
where $L(w)$ denotes the loss function with parameters $w$. According to \citet{keskar2016large}, the sharpness metric is closely related to the spectrum $\nabla^2L(w)$, therefore acts as a proxy measure for $\lambda_1^{max}$ in Equation~\ref{eq:forgetting_bound}. 
After training on each task, we evaluate the minimum reached by the model for its sharpness (alternatively flatness). We average the sharpness values across all tasks in a given task sequence and report the mean and standard deviation across 5 random task orderings. In Appendix~\ref{sec:impldetails}, we provide implementation details about the sharpness metric.

In Table \ref{tab:sharpness}, we report sharpness values for ResNet-18 on 5-dataset-CV, Split CIFAR-50/CIFAR-100 for $\epsilon \in \{5e^{-4}, 1e^{-3}\}$ and DistilBERT on 5-dataset-NLP and Split YahooQA data sets for $\epsilon \in \{5e^{-5}, 1e^{-4}\}$. We see that for all data sets, \textit{the average sharpness value for the pre-trained initialized models is significantly lower than for the randomly initialized models, validating the relative flatness of the task minima in the case of pre-trained models}. 

%% file: jmlrsections/05_sam.tex
\section{Lifelong Learning with Sharpness Aware Minimization (SAM)}
\label{sec:sam}

In the previous section, we looked at the role of initialization in alleviating forgetting. Specifically, pre-trained initializations favor flat loss basins and implicitly mitigate forgetting to some extent. On the other hand, \citet{mirzadeh2020understanding} suggests modifying the training regime by varying learning rate decay, batch size, and dropout regularization such that inherent noise in the stochastic gradients leads to flat basins in the loss landscape. However, the procedure for tuning these hyperparameters is ill-defined for lifelong learning, thereby rendering their strategy less helpful. Furthermore, the suggested hyperparameter sweep is expensive (e.g., $48$ separate runs just for one data set) and does not transfer across different architectures and data sets. Motivated by these shortcomings, we pose a question---\textit{What if we modify the optimization dynamics by explicitly seeking flat loss basins during lifelong learning of the model?} Alternatively, what if we jointly optimize the sharpness metric and the current task loss? Towards this objective, we employ the Sharpness-Aware Minimization (SAM) procedure \citep{foret2020sharpness} that seeks parameters that lie in the neighborhoods having uniformly low loss values by jointly minimizing the task loss value and sharpness metric. SAM defines the sharpness of the loss function $L$ at parameters $w$ as
\begin{equation}
\max_{||\epsilon||_2 \leq \rho} L(w+\epsilon) - L(w), \label{eq:sam_sharpness_metric}
\end{equation}
where maximization region is an $\ell^p$ ball with radius $\rho$ for $p=2$ in Equation 
\ref{eq:sam_sharpness_metric}. SAM problem can be defined in terms of the following minimax optimization
\begin{equation}
    \min_w \max_{||\epsilon||_2 \leq \rho} L(w+\epsilon) + \lambda ||w||_2^2. \label{eq:sam_minimax}
\end{equation}
The gradient of the result of the maximization problem can be approximated as 
\begin{equation}
    \nabla_w \max_{||\epsilon||_2 \leq \rho} L(w+\epsilon) \approx \nabla_w L(w) \Big\rvert_{w+\hat{\epsilon}(w)} + \frac{d\hat{\epsilon}(w)}{dw} \nabla_w L(w) \Big\rvert_{w+\hat{\epsilon}(w)}, 
\end{equation}
where 
\begin{equation}
    \hat{\epsilon}(w) = \rho \operatorname{sign}\big(\nabla_w L(w)\big)\Bigg(\frac{\big\lvert\nabla_w L(w)\big\rvert}{\norm{\nabla_w L(w)}_2}\Bigg). 
\end{equation}
To make the optimization simpler, the second-order term in the gradient is dropped, leaving us with 
\begin{equation}
    \nabla_w \max_{||\epsilon||_2 \leq \rho} L(w+\epsilon) \approx \nabla_w L(w) \Big\rvert_{w+\hat{\epsilon}(w)}.
\end{equation}
For the complete derivation of this gradient, we defer readers to \citet{foret2020sharpness}.

\begin{table}%[ht] 
  \centering
  \begin{scriptsize}
    \begin{tabular}{lrrrrrr}
      \toprule
               & \multicolumn{3}{c}{w/o PT (ResNet-18-R/DistilBERT-R)} & \multicolumn{3}{c}{w/ PT (ResNet-18-PT/DistilBERT-PT)}                                                                                 \\
      \cmidrule(r){2-4}\cmidrule(r){5-7}
               & Accuracy($\uparrow$) & Forgetting($\downarrow$) & LearnAcc($\uparrow$) & Accuracy($\uparrow$) & Forgetting($\downarrow$) & LearnAcc($\uparrow$) \\
      \midrule
      \multicolumn{5}{l}{\textbf{Split YahooQA}}                                                                                                                \\
      FT & $73.1_{4.7}$ & $26.4_{5.9}$ & $94.2_{0.0}$ & $87.7_{3.7}$ & $9.5_{4.7}$ & $95.2_{0.0}$ \\
      FT w/ SAM & $73.5_{4.0}$  & $25.9_{5.0}$ & $94.2_{0.0}$ & $88.5_{2.8}$ & $8.4_{3.5}$  & $95.2_{0.0}$ \\
      EWC    & $76.1_{3.1}$  & $22.7_{3.9}$ & $94.2_{0.0}$ & $89.5_{3.4}$ & $7.1_{4.2}$  & $95.2_{0.0}$ \\
      ER    & $77.2_{3.3}$  & $21.3_{4.2}$ & $94.2_{0.0}$ & $\mathbf{89.4_{0.7}}$ & $\mathbf{7.3_{0.9}}$ & $95.2_{0.0}$ \\
      ER w/ SAM    & $\mathbf{77.5_{1.4}}$  & $\mathbf{20.9_{1.8}}$ & $94.2_{0.0}$ & $89.0_{0.7}$ & $7.8_{0.9}$ & $95.2_{0.0}$ \\
      \midrule
      \multicolumn{5}{l}{\textbf{5-dataset-NLP}}                                                                                                                \\
      FT & $44.3_{5.0}$  & $36.7_{6.3}$ & $73.6_{0.1}$ & $64.3_{4.5}$ & $16.7_{5.7}$ & $77.7_{0.1}$ \\
      FT w/ SAM    & $46.0_{5.0}$  & $34.3_{6.3}$ & $73.4_{0.1}$ & $66.4_{2.8}$ & $13.9_{3.5}$  & $77.6_{0.1}$ \\
      EWC      & $48.7_{4.9}$ & $31.1_{6.2}$ & $73.6_{0.0}$ & $66.8_{3.3}$ & $13.6_{4.2}$  & $77.6_{0.1}$ \\
      ER       & $\mathbf{56.3_{3.1}}$ & $\mathbf{21.6_{3.9}}$ & $73.6_{0.1}$ & $70.2_{1.6}$ & $9.4_{2.0}$  & $77.7_{0.1}$ \\
      ER w/ SAM    & $56.3_{3.9}$  & $21.5_{5.0}$ & $73.4_{0.1}$ & $\mathbf{71.1_{1.2}}$ & $\mathbf{8.1_{1.5}}$  & $77.5_{0.0}$ \\
      \midrule
      \midrule
      \multicolumn{5}{l}{\textbf{Split CIFAR-50}}                                                                                                               \\
      FT & $42.8_{3.1}$ & $23.7_{1.1}$ & $66.4_{2.1}$ & $86.3_{1.2}$ & $7.1_{0.9}$ & $93.4_{0.5}$ \\
      FT w/ NSGD & $46.4_{2.3}$ & $24.7_{1.6}$ & $71.1_{0.9}$ & $86.5_{0.9}$ &	$7.4_{0.7}$ & $93.5_{0.3}$ \\
      FT w/ SAM & $50.3_{2.2}$  & $15.0_{2.1}$   & $65.3_{1.2}$ & $\mathbf{90.5_{1.1}}$ & $\mathbf{4.2_{1.0}}$ & $94.7_{0.4}$ \\
      Stable SGD  & $46.0_{2.3}$ & $12.1_{0.4}$ & $58.1_{2.5}$ & $84.1_{1.9}$ & $5.2_{1.6}$ & $89.2_{0.7}$ \\
      EWC      & $45.3_{2.5}$ & $20.7_{1.5}$ & $65.9_{1.3}$ & $86.2_{0.9}$ & $7.4_{0.9}$ & $93.5_{0.7}$ \\      
      A-GEM      & $47.3_{2.7}$ & $21.1_{2.0}$ & $68.4_{0.7}$ & $87.3_{1.0}$ & $6.2_{0.6}$ & $93.4_{0.4}$ \\
      ER      & $45.8_{1.8}$ & $20.6_{1.4}$ & $66.4_{2.7}$ & $86.2_{1.1}$ & $7.1_{0.8}$ & $93.3_{0.5}$ \\
      ER w/ SAM    & $50.8_{0.5}$  & $16.9_{0.9}$   & $67.6_{0.7}$ & $88.4_{1.3}$ & $6.0_{1.1}$ & $94.4_{0.3}$ \\
    %   MC-SGD &  &  &  & $64.5_{2.7}$ & $25.9_{2.9}$ & $90.2_{0.3}$ \\
      MC-SGD & $59.0_{2.3}$ & $5.4_{1.3}$ & $63.7_{1.8}$ & $86.5_{0.9}$ & $4.1_{0.5}$ & $90.4_{0.6}$ \\
      MC-SGD w/ SAM & $\mathbf{59.1_{2.9}}$ & $\mathbf{5.2_{1.7}}$ & $63.9_{2.1}$ & $87.9_{0.7}$ & $3.8_{0.2}$ & $91.7_{0.7}$  \\
      % Stable SGD  & $46.0_{2.3}$ & $12.1_{0.4}$ & $58.1_{2.5}$ & $84.1_{1.9}$ & $5.2_{1.6}$ & $89.2_{0.7}$ \\
      \midrule
      \multicolumn{5}{l}{\textbf{5-dataset-CV}}                                                                                                                   \\
      FT    & $33.7_{2.5}$ & $51.5_{2.6}$ & $85.2_{2.0}$ & $57.2_{5.1}$ & $38.3_{5.0}$ & $95.5_{0.2}$ \\
      FT w/ NSGD & $35.8_{3.7}$ & $52.4_{3.3}$	& $88.2_{1.3}$ & $56.6_{5.1}$ & $39.0_{4.9}$	& $95.6_{0.3}$ \\
      FT w/ SAM  & $47.6_{3.8}$  & $40.6_{4.0}$   & $88.2_{1.3}$ & $70.4_{4.4}$ & $25.6_{4.4}$ & $96.0_{0.1}$ \\
      Stable SGD      & $50.2_{7.0}$ & $40.3_{7.8}$ & $90.5_{1.0}$ & $71.3_{2.7}$ & $20.5_{2.5}$ & $91.9_{0.8}$ \\
      EWC    & $35.0_{4.9}$ & $50.1_{6.5}$  & $85.1_{1.9}$ & $56.7_{3.8}$  & $38.8_{3.8}$  & $95.4_{0.2}$ \\
      A-GEM    & $46.1_{6.8}$  & $39.5_{7.1}$  & $85.2_{2.5}$ & $72.0_{2.3}$ & $23.0_{2.3}$   & $95.0_{0.2}$ \\
      ER    & $50.6_{4.5}$ & $35.0_{5.4}$ & $85.6_{1.3}$ & $70.7_{1.5}$  & $24.2_{1.4}$  & $94.9_{0.2}$ \\
      ER w/ SAM    & $60.3_{3.9}$  & $27.3_{4.1}$  & $87.6_{1.3}$ & $77.4_{3.9}$ & $18.2_{3.9}$ & $95.6_{0.2}$ \\
      % {\color{blue}MC-SGD} & $56.2_{6.7}$ & $36.4_{6.3}$ & $92.5_{0.5}$ & $60.5_{2.8}$ & $34.5_{3.1}$ & $95.0_{0.4}$ \\
      % {\color{blue}MC-SGD w/ SAM} & $57.7_{7.4}$ & $34.7_{7.2}$ & $92.4_{0.2}$ & $63.7_{2.2}$ & $31.8_{2.3}$ & $95.5_{0.2}$ \\
      MC-SGD & $71.3_{5.9}$ & $21.3_{5.7}$ & $92.6_{0.3}$ & $81.9_{2.6}$ & $13.3_{2.6}$ & $95.2_{0.3}$ \\
      MC-SGD w/ SAM & $\mathbf{72.7_{7.8}}$ & $\mathbf{19.9_{7.6}}$ & $92.6_{0.4}$ & $\mathbf{87.1_{1.6}}$ & $\mathbf{8.5_{1.7}}$ & $95.5_{0.2}$ \\
      % Stable SGD      & $50.2_{7.0}$ & $40.3_{7.8}$ & $90.5_{1.0}$ & $71.3_{2.7}$ & $20.5_{2.5}$ & $91.9_{0.8}$ \\
      \midrule
      \multicolumn{5}{l}{\textbf{Split CIFAR-100}}    \\
      FT & $38.9_{2.2}$ & $39.1_{2.0}$ & $78.0_{1.0}$ & $82.0_{3.0}$ & $13.8_{2.6}$  & $95.8_{0.5}$ \\
      FT w/ NSGD & $40.2_{3.5}$ &	$41.7_{3.2}$ &	$81.9_{0.8}$ & $79.8_{3.4}$ & $16.2_{3.1}$ & $96.0_{0.6}$\\
      FT w/ SAM   & $48.9_{4.8}$  & $28.5_{5.0}$  & $77.5_{0.9}$ & $88.3_{1.7}$ & $8.6_{1.3}$ & $96.9_{0.6}$ \\
      Stable SGD  & $52.9_{1.7}$ & $21.0_{2.0}$ & $73.8_{1.5}$ & $86.6_{2.2}$ & $5.5_{1.5}$ & $91.8_{0.7}$ \\
      EWC    & $37.4_{1.5}$ & $40.1_{1.8}$ & $77.5_{1.5}$ & $81.3_{2.5}$  & $14.5_{2.1}$   & $95.8_{0.6}$ \\
      A-GEM    & $46.8_{3.5}$ & $32.0_{3.8}$ & $78.8_{1.0}$ & $84.0_{1.6}$ & $11.7_{1.0}$  & $95.7_{0.7}$ \\
      ER    & $48.6_{1.9}$ & $29.8_{1.3}$ & $78.1_{0.7}$ & $84.4_{2.2}$ & $11.4_{1.7}$  & $95.8_{0.5}$ \\
      ER w/ SAM    & $60.5_{0.5}$  & $20.9_{0.7}$  & $81.4_{0.7}$ & $88.4_{0.7}$ & $8.6_{0.2}$ & $96.7_{0.5}$ \\
    %   MC-SGD    & $55.8_{3.1}$ & $22.1_{2.0}$ & $77.7_{1.2}$ & $86.7_{1.0}$ & $6.5_{0.5}$  & $93.1_{0.6}$ \\
      % {\color{blue}MC-SGD}$^*$    & $62.3_{1.3}$ & $13.4_{1.2}$ & $75.5_{0.9}$ & $86.5_{1.2}$ & $6.5_{0.7}$  & $92.9_{0.7}$ \\
      % {\color{blue}MC-SGD w/ SAM}   & $\mathbf{64.5_{0.9}}$ & $\mathbf{10.9_{0.6}}$ & $75.2_{0.6}$ & $88.1_{1.1}$ & $6.2_{0.8}$  & $94.2_{0.6}$ \\
      MC-SGD    & $62.2_{2.4}$ & $13.3_{1.8}$ & $75.2_{0.7}$ & $84.7_{2.4}$ & $8.5_{1.8}$  & $93.0_{0.7}$ \\
      MC-SGD w/ SAM   & $\mathbf{65.1_{1.1}}$ & $\mathbf{10.4_{1.1}}$ & $75.4_{1.0}$ & $\mathbf{89.0_{1.7}}$ & $\mathbf{5.3_{1.1}}$  & $94.2_{0.6}$ \\
      % Stable SGD  & $52.9_{1.7}$ & $21.0_{2.0}$ & $73.8_{1.5}$ & $86.6_{2.2}$ & $5.5_{1.5}$ & $91.8_{0.7}$ \\
      \bottomrule
    \end{tabular}
  \end{scriptsize}
   \caption[caption]{Comparing performance in terms of accuracy($\%$), forgetting($\%$), and learning accuracy ($\%$) across methods after training on the last task (all metrics are averaged over 5 random task sequences). $\uparrow$ indicates higher is better, $\downarrow$ indicates lower is better. Augmenting the FT baseline with SAM results in performance competitive with state-of-the-art methods, and augmenting the ER or MC-SGD method with SAM often outperforms state-of-the-art methods demonstrating SAM as a valuable addition to current lifelong learning methods.}
  \label{tab:summary}
\end{table}

\begin{table}[ht]    
  \centering
  \begin{scriptsize}
  \begin{tabular}{lrrrrrr}
    \toprule
                         & \multicolumn{3}{c}{5-dataset-NLP} & \multicolumn{3}{c}{15-dataset-NLP}                                       \\
    \cmidrule(r){2-4}\cmidrule(r){5-7}
                         & Accuracy($\uparrow$) & Forgetting($\downarrow$) & LearnAcc($\uparrow$) & Accuracy($\uparrow$)  & Forgetting($\downarrow$) & LearnAcc($\uparrow$)\\
    \midrule
    \multicolumn{3}{l}{\textbf{DistilBERT}} \\
    FT            & $64.3_{4.5}$ & $16.7_{5.7}$ & $77.7_{0.1}$ & $47.0_{3.5}$ & $18.8_{4.0}$ & $64.4_{1.2}$  \\
    FT w/ SAM        & $66.4_{2.8}$ & $13.9_{3.5}$ & $77.6_{0.1}$ & $47.5_{3.1}$ & $16.5_{3.8}$ & $62.5_{0.8}$  \\
    ER          & $70.2_{1.6}$ & $9.4_{2.0}$ & $77.7_{0.1}$ & $53.2_{3.4}$ & $13.1_{4.1}$ & $65.3_{0.6}$  \\
    ER w/ SAM   & $\mathbf{71.1_{1.2}}$ & $\mathbf{8.1_{1.5}}$ & $77.5_{0.0}$ & $\mathbf{53.5_{2.0}}$ & $\mathbf{11.0_{3.1}}$ & $63.1_{1.0}$  \\
    \midrule
    \multicolumn{3}{l}{{\textbf{T5-Small}}} \\
    FT            & $65.9_{2.8}$ & $12.6_{3.7}$ & $76.0_{0.2}$ & $44.9_{3.2}$ & $21.9_{3.0}$ & $65.2_{0.6}$  \\
    FT w/ SAM        & $66.4_{2.3}$ & $11.9_{3.0}$ & $75.9_{0.3}$ & $46.6_{3.0}$ & $19.3_{3.0}$ & $64.4_{0.7}$  \\
    ER          & $69.4_{1.0}$ & $8.2_{1.2}$ & $75.9_{0.2}$ & $48.1_{1.3}$ & $18.9_{1.0}$ & $65.5_{0.8}$  \\
    ER w/ SAM   & $\mathbf{69.9_{0.2}}$ & $\mathbf{7.5_{0.1}}$ & $75.9_{0.2}$ & $\mathbf{48.6_{1.6}}$ & $\mathbf{17.4_{1.2}}$ & $64.5_{0.7}$ \\
    \midrule
    \multicolumn{3}{l}{\textbf{BERT-base}} \\
    FT &      $67.6_{2.8}$ & $13.6_{3.4}$ & $78.4_{0.1}$ & $52.9_{2.8}$ & $19.2_{3.0}$ & $70.8_{0.3}$ \\
    FT w/ SAM   & $70.8_{2.1}$ & $9.5_{2.5}$ & $78.4_{0.1}$ & $55.1_{2.6}$ & $16.4_{3.2}$ & $70.4_{0.7}$  \\
    ER          & $70.6_{2.1}$ & $9.7_{2.6}$ & $78.4_{0.1}$ & $56.4_{2.9}$ & $15.7_{3.0}$ & $71.0_{0.4}$  \\
    ER w/ SAM   & $\mathbf{73.0_{1.5}}$ & $\mathbf{6.9_{1.9}}$ & $78.5_{0.1}$ & $\mathbf{57.8_{1.9}}$ & $\mathbf{13.7_{1.9}}$ & $70.5_{0.5}$  \\
    \midrule
    \multicolumn{3}{l}{\textbf{RoBERTa-base}} \\
    FT & $71.4_{1.7}$ & $9.5_{2.0}$ & $79.0_{0.1}$ & $55.5_{3.1}$ & $21.0_{3.0}$ & $75.1_{0.5}$ \\
    FT w/ SAM & $72.6_{1.2}$ & $7.8_{1.5}$ & $78.8_{0.0}$ & $57.8_{1.7}$ & $15.4_{1.9}$ & $72.1_{1.4}$  \\
    ER  & $73.7_{0.8}$ & $6.7_{0.9}$ & $79.1_{0.1}$ & $60.9_{1.4}$ & $15.3_{1.6}$ & $75.2_{0.1}$  \\
    ER w/ SAM & $\mathbf{74.3_{0.6}}$ & $\mathbf{5.7_{0.8}}$ & $78.9_{0.0}$ & $\mathbf{62.1_{1.5}}$ & $\mathbf{12.2_{2.1}}$ & $73.3_{0.8}$  \\
    \midrule
    \multicolumn{3}{l}{\textbf{BERT-Large}} \\
    FT & $71.0_{2.0}$ & $10.2_{2.5}$ & $79.2_{0.0}$ & $53.8_{1.2}$ & $23.4_{1.4}$ & $75.7_{0.4}$ \\
    FT w/ SAM & $73.7_{1.3}$ & $6.9_{1.6}$ & $79.2_{0.0}$ & $58.7_{3.4}$ & $17.1_{4.3}$ & $74.6_{2.2}$ \\
    ER       & $73.5_{1.2}$ & $7.2_{1.4}$ & $79.2_{0.1}$ & $61.1_{2.6}$ & $15.1_{2.9}$ & $75.1_{1.2}$  \\
    ER w/ SAM & $\mathbf{74.6_{0.6}}$ & $\mathbf{5.7_{0.8}}$ & $79.2_{0.1}$ & $\mathbf{61.7_{1.3}}$ & $\mathbf{13.7_{2.8}}$ & $74.5_{1.5}$  \\
    \bottomrule
  \end{tabular}
  \end{scriptsize}
  \caption[caption]{Comparing performance in terms of average accuracy (\%), forgetting (\%), and learning accuracy (\%) across pre-trained Transformers after continual learning the last task. $\uparrow$ indicates higher is better, $\downarrow$ indicates lower is better. All metrics are averaged over 5 random task sequences. Overall, we observe that models pre-trained on diverse corpora (RoBERTa-base) undergo less forgetting on both $5$ and $15$ diverse tasks. Furthermore, augmenting the FT and ER methods with SAM often outperforms state-of-the-art methods.}
  \label{tab:pretrainingstudy}
\end{table}

In Tables \ref{tab:summary} and \ref{tab:pretrainingstudy}, we report the results with the discussed SAM procedure. We see that SAM results in a consistent improvement in performance over non-SAM counterparts. Simply augmenting SAM with the finetune method (FT w/ SAM) results in a competitive baseline, sometimes outperforming state-of-the-art baselines like ER \citep{chaudhry2019tiny}, Stable SGD \citep{mirzadeh2020understanding} and MC-SGD \citep{mirzadeh2020linear} (see Table~\ref{tab:summary}, Split CIFAR-50 w/PT ResNet-18-PT). Note SAM requires minimal hyper-parameter tuning (we set $\rho=0.05$ to the default value for all our CV experiments). Since SAM is just a modification to the optimization procedure, we propose augmenting it with existing state-of-the-art continual learning methods like ER \citep{chaudhry2019tiny} and MC-SGD \citep{mirzadeh2020linear}. 

From Table~\ref{tab:summary}, MC-SGD w/ SAM outperforms all existing baselines in terms of overall accuracy and forgetting across all considered CV benchmarks (Split CIFAR-50, Split CIFAR-100, and 5-dataset-CV) when evaluated in the context of random as well as pre-trained initialized models. 
From Tables \ref{tab:summary} and \ref{tab:pretrainingstudy}, ER w/ SAM results in a method that outperforms all existing baselines in terms of overall accuracy and forgetting across all NLP benchmarks. Furthermore, in Table~\ref{tab:pretrainingstudy}, we present the results obtained with T5-Small (v1.1) \citep{raffel2019exploring} on the 5-dataset-NLP and 15-dataset-NLP benchmarks. Consistent with encoder-only models such as DistilBERT, BERT, and RoBERTa, we observe that FT w/ SAM and ER w/ SAM outperform their non-SAM counterparts on the encoder-decoder architecture. This finding highlights the broad applicability of SAM across different model architectures, including both encoder-only and encoder-decoder setups.
To summarize, \textit{SAM serves as a valuable addition to current continual learning methods and can be seamlessly incorporated to enhance overall performance.}

\begin{figure}%[ht!]
    \centering
    \begin{subfigure}{.245\textwidth}
      \centering
      \includegraphics[width=\textwidth]{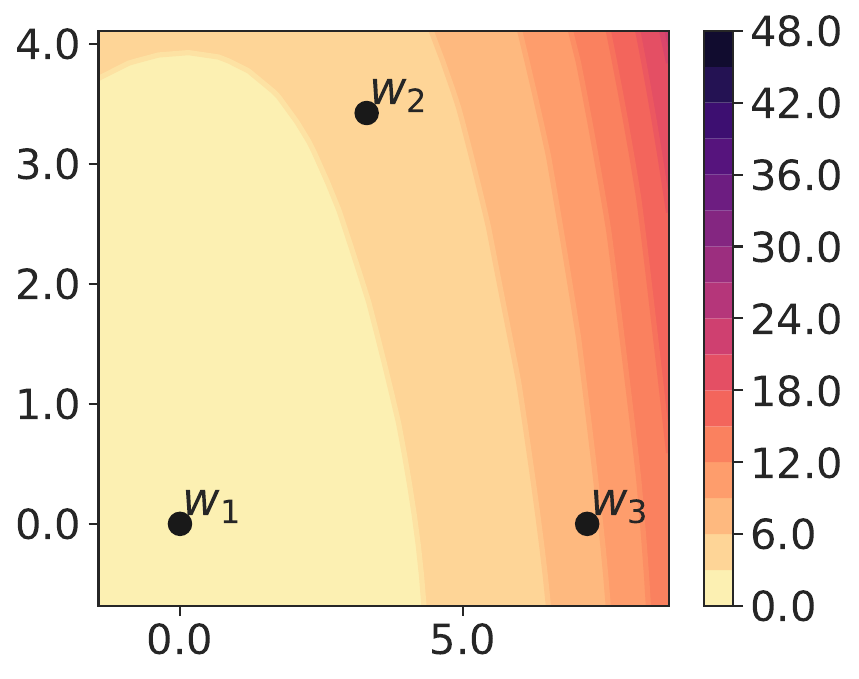}
      \caption{FT (T1)}
      \label{fig:c50_seq2_no_pt_sgd_contour_w1}
    \end{subfigure}\hspace{\fill}%
    \begin{subfigure}{.245\textwidth}
      \centering
      \includegraphics[width=\textwidth]{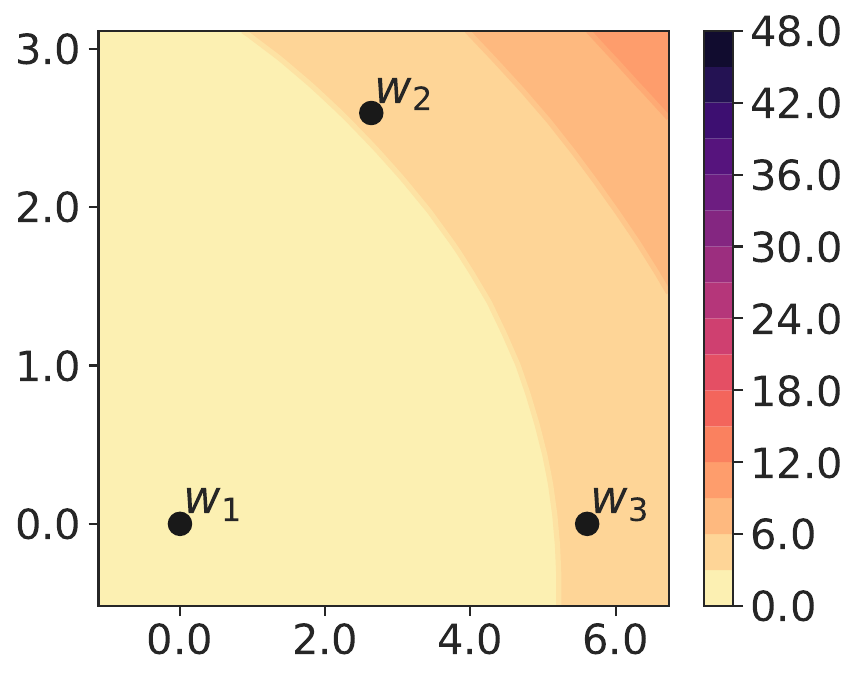}
      \caption{FT w/ SAM (T1)}
      \label{fig:c50_seq2_no_pt_sam_contour_w1}
    \end{subfigure}\hspace{\fill}%
    \begin{subfigure}{.245\textwidth}
      \centering
      \includegraphics[width=\textwidth]{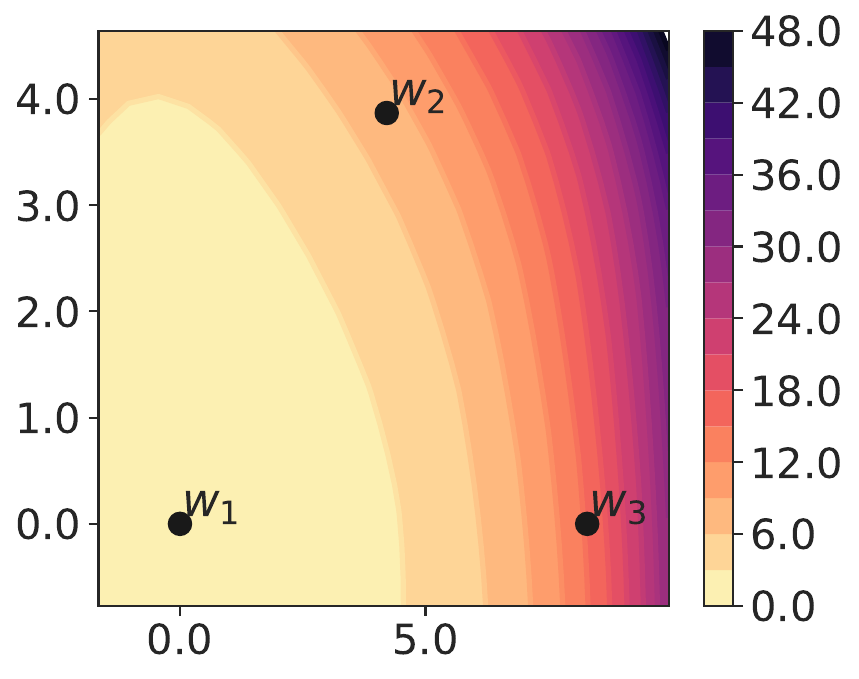}
      \caption{ER (T1)}
      \label{fig:c50_seq2_no_pt_er_contour_w1}
    \end{subfigure}\hspace{\fill}%
    \begin{subfigure}{.245\textwidth}
      \centering
      \includegraphics[width=\textwidth]{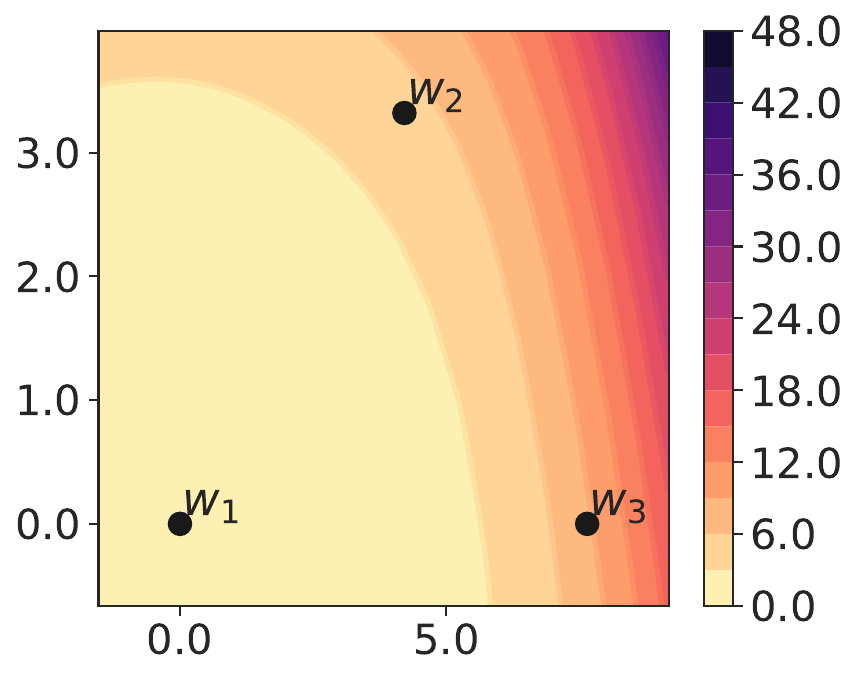}
      \caption{ER w/ SAM (T1)}
      \label{fig:c50_seq2_no_pt_ersam_contour_w1}
    \end{subfigure}\hspace{\fill}%
    \bigskip
    \begin{subfigure}{.245\textwidth}
      \centering
      \includegraphics[width=\textwidth]{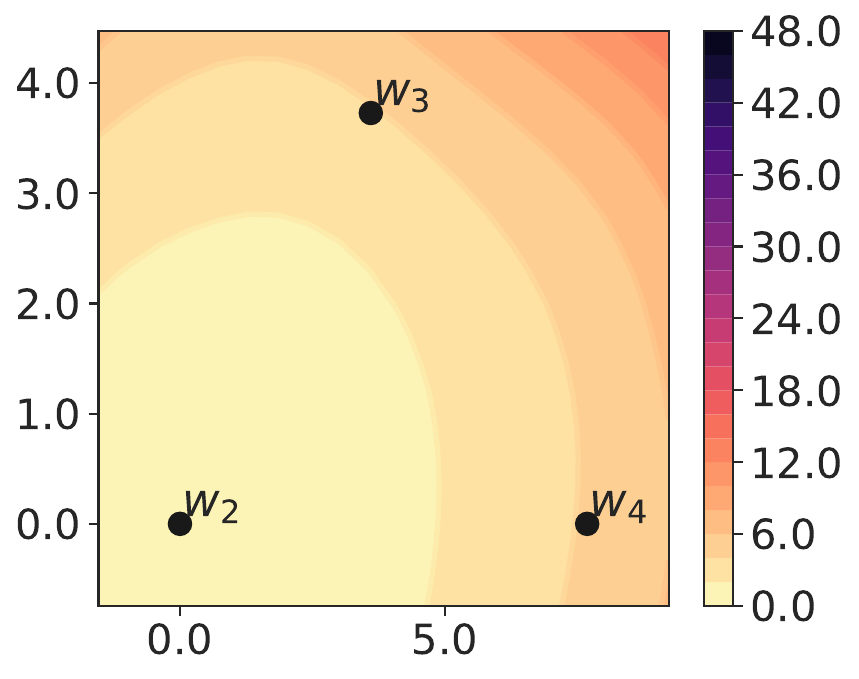}
      \caption{FT (T2)}
      \label{fig:c50_seq2_no_pt_sgd_contour_w2}
    \end{subfigure}\hspace{\fill}%
    \begin{subfigure}{.245\textwidth}
      \centering
      \includegraphics[width=\textwidth]{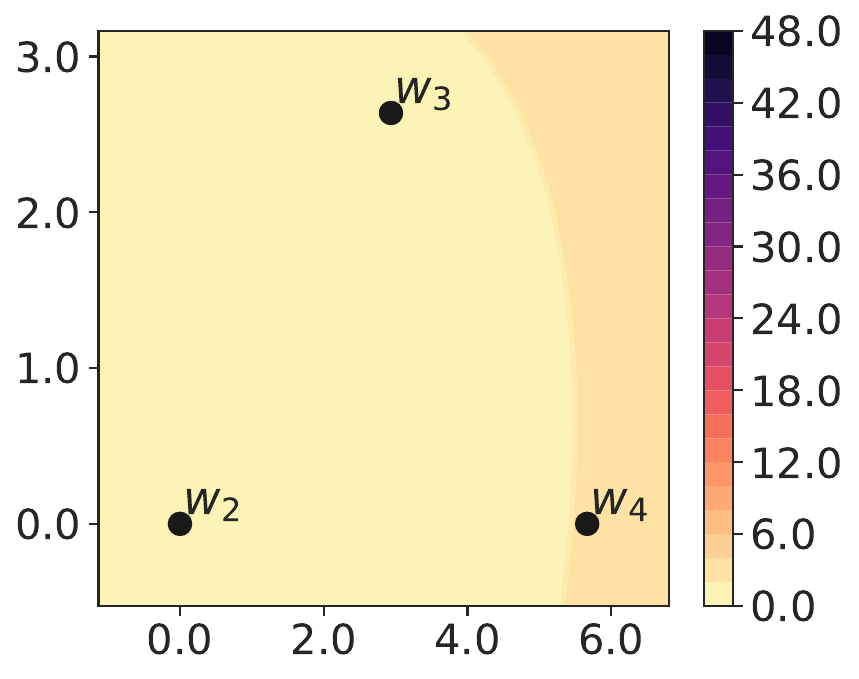}
      \caption{FT w/ SAM (T2)}
      \label{fig:c50_seq2_no_pt_sam_contour_w2}
    \end{subfigure}\hspace{\fill}%
    \begin{subfigure}{.245\textwidth}
      \centering
      \includegraphics[width=\textwidth]{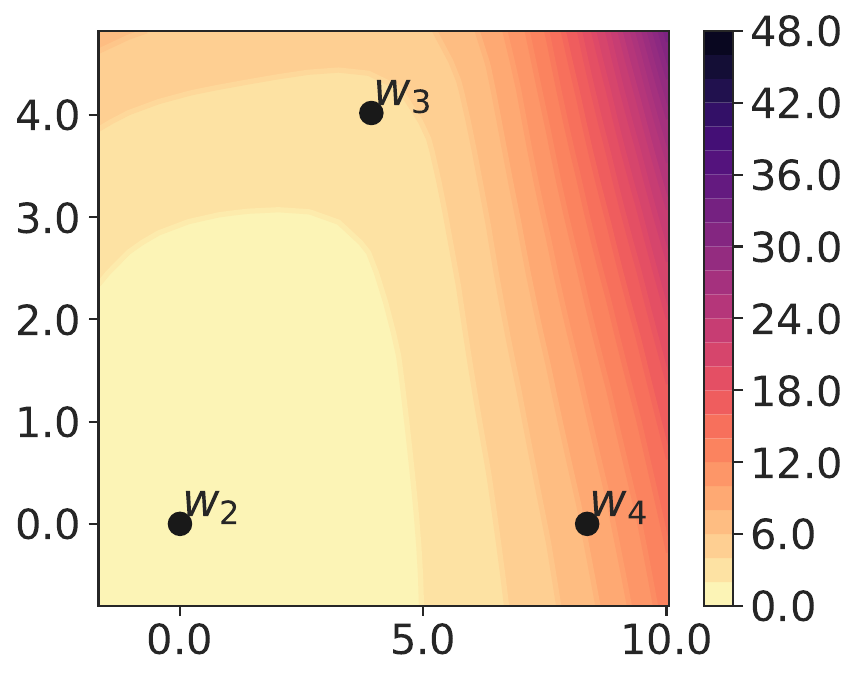}
      \caption{ER (T2)}
      \label{fig:c50_seq2_no_pt_er_contour_w2}
    \end{subfigure}\hspace{\fill}%
    \begin{subfigure}{.245\textwidth}
      \centering
      \includegraphics[width=\textwidth]{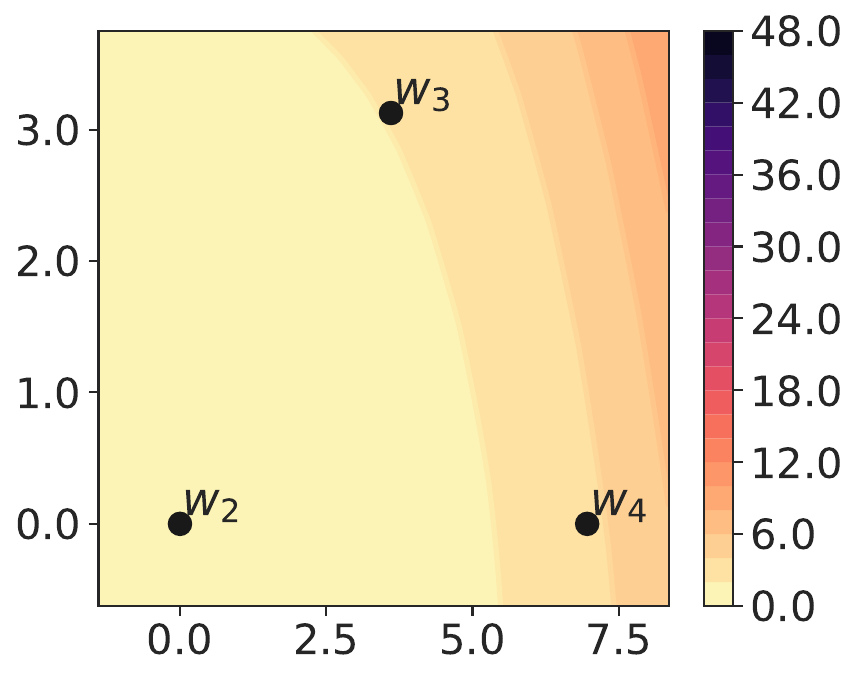}
      \caption{ER w/ SAM (T2)}
      \label{fig:c50_seq2_no_pt_ersam_contour_w2}
    \end{subfigure}\hspace{\fill}%
    \caption{Loss contours for task 1 (T1) and task 2 (T2) of Split CIFAR-50. The top row visualizes loss contours for task 1 where w$_1$, w$_2$, and w$_3$ are minima obtained after sequential training on tasks 1, 2, and 3, respectively. Similarly, the bottom row visualizes loss contours for task 2 after sequential training on tasks 2, 3, and 4. All of the above models start with random weights. SAM (FT w/ SAM, ER w/ SAM) leads to wide task minima compared to finetune (FT) and ER methods.}
    \label{fig:contours_sam_er_c50}
\end{figure}

\subsection{Loss Contours and Sharpness with SAM}

\begin{figure}
    \centering
    \begin{subfigure}{.245\textwidth}
      \centering
      \includegraphics[width=\textwidth]{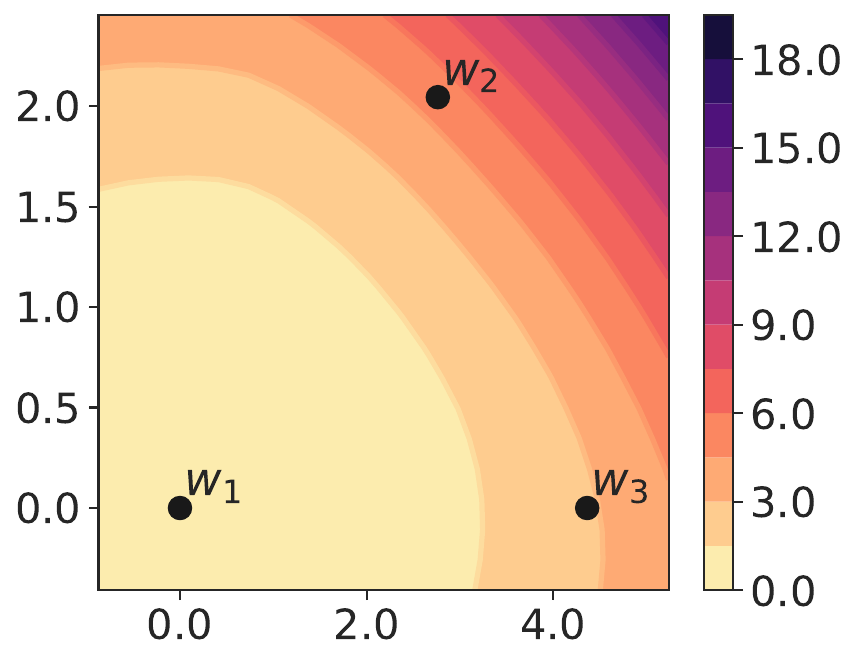}
      \caption{FT (T1)}
      \label{fig:5data_seq2_no_pt_sgd_contour_w1}
    \end{subfigure}\hspace{\fill}%
    \begin{subfigure}{.245\textwidth}
      \centering
      \includegraphics[width=\textwidth]{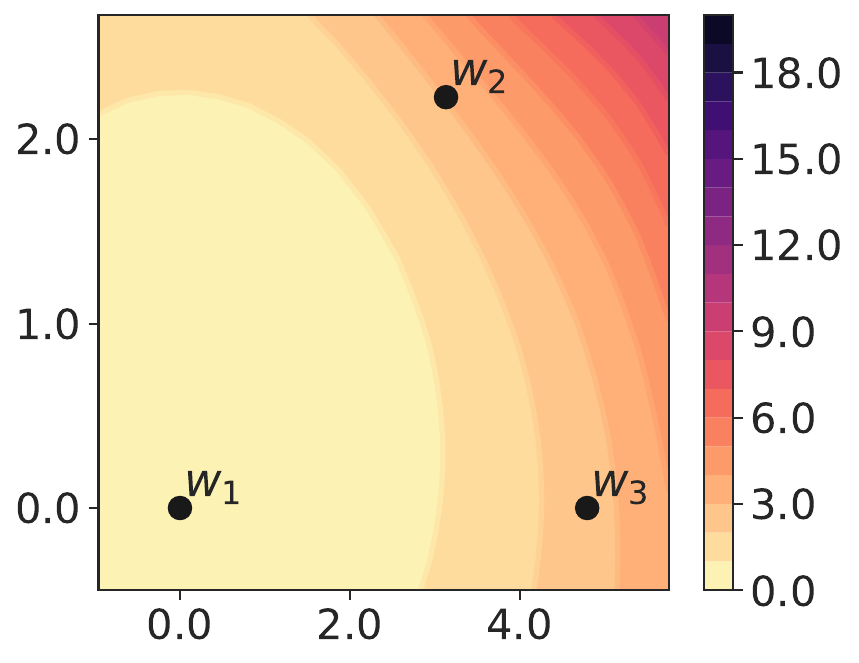}
      \caption{FT w/ SAM (T1)}
      \label{fig:5data_seq2_no_pt_sam_contour_w1}
    \end{subfigure}\hspace{\fill}%
    \begin{subfigure}{.245\textwidth}
      \centering
      \includegraphics[width=\textwidth]{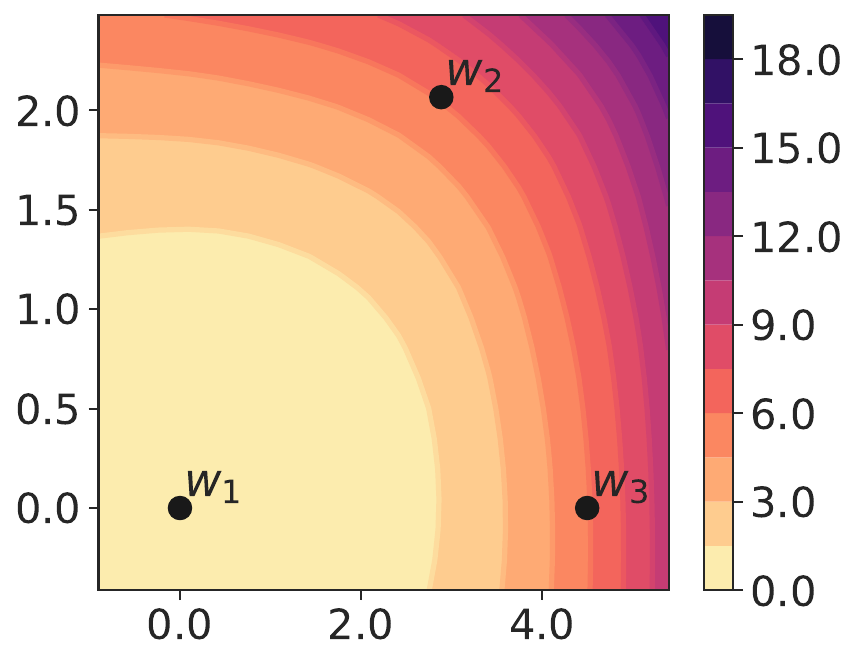}
      \caption{ER (T1)}
      \label{fig:5data_seq2_no_pt_er_contour_w1}
    \end{subfigure}\hspace{\fill}%
    \begin{subfigure}{.245\textwidth}
      \centering
      \includegraphics[width=\textwidth]{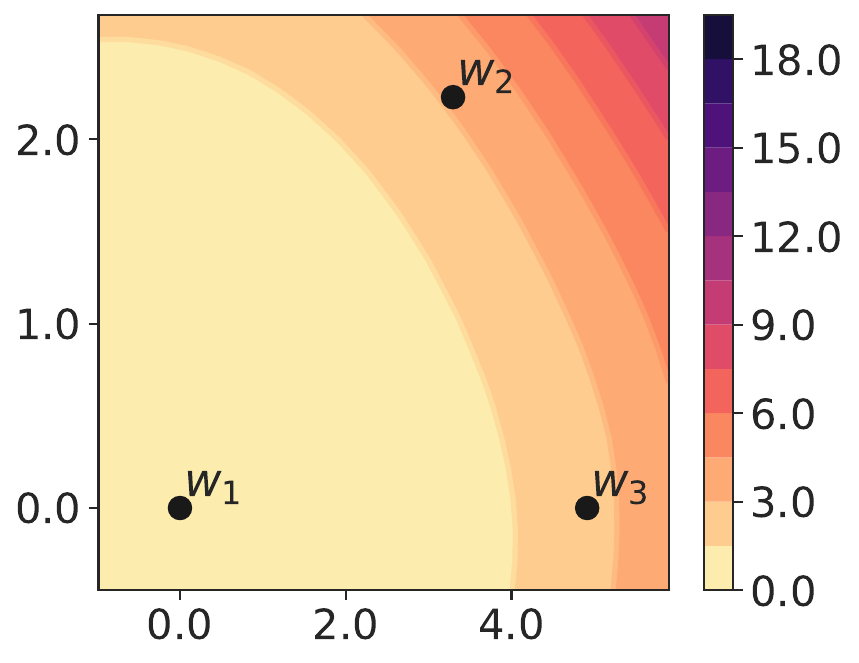}
      \caption{ER w/ SAM (T1)}
      \label{fig:5data_seq2_no_pt_ersam_contour_w1}
    \end{subfigure}\hspace{\fill}%
    \bigskip
    \begin{subfigure}{.245\textwidth}
      \centering
      \includegraphics[width=\textwidth]{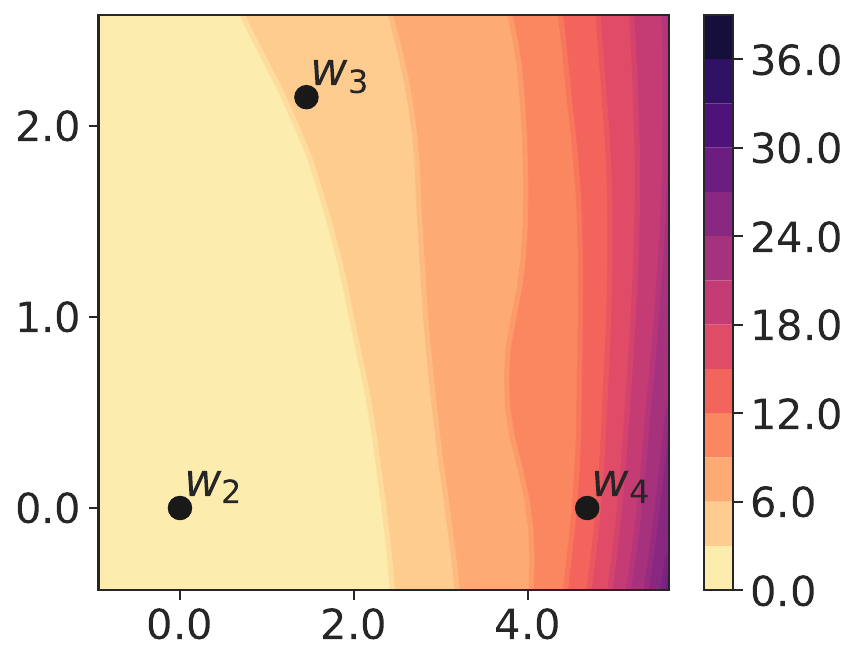}
      \caption{FT (T2)}
      \label{fig:5data_seq2_no_pt_sgd_contour_w2}
    \end{subfigure}\hspace{\fill}%
    \begin{subfigure}{.245\textwidth}
      \centering
      \includegraphics[width=\textwidth]{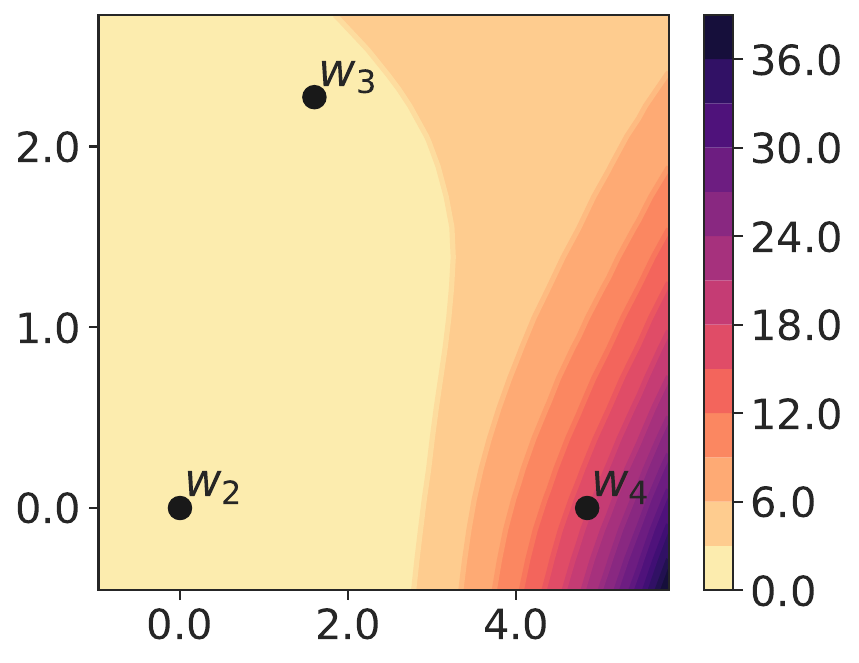}
      \caption{FT w/ SAM (T2)}
      \label{fig:5data_seq2_no_pt_sam_contour_w2}
    \end{subfigure}\hspace{\fill}%
    \begin{subfigure}{.245\textwidth}
      \centering
      \includegraphics[width=\textwidth]{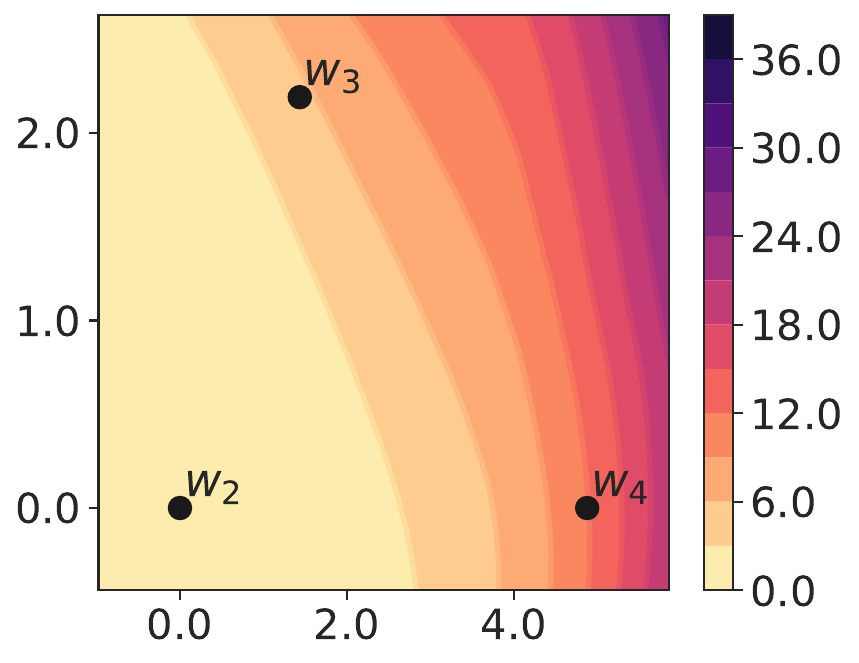}
      \caption{ER (T2)}
      \label{fig:5data_seq2_no_pt_er_contour_w2}
    \end{subfigure}\hspace{\fill}%
    \begin{subfigure}{.245\textwidth}
      \centering
      \includegraphics[width=\textwidth]{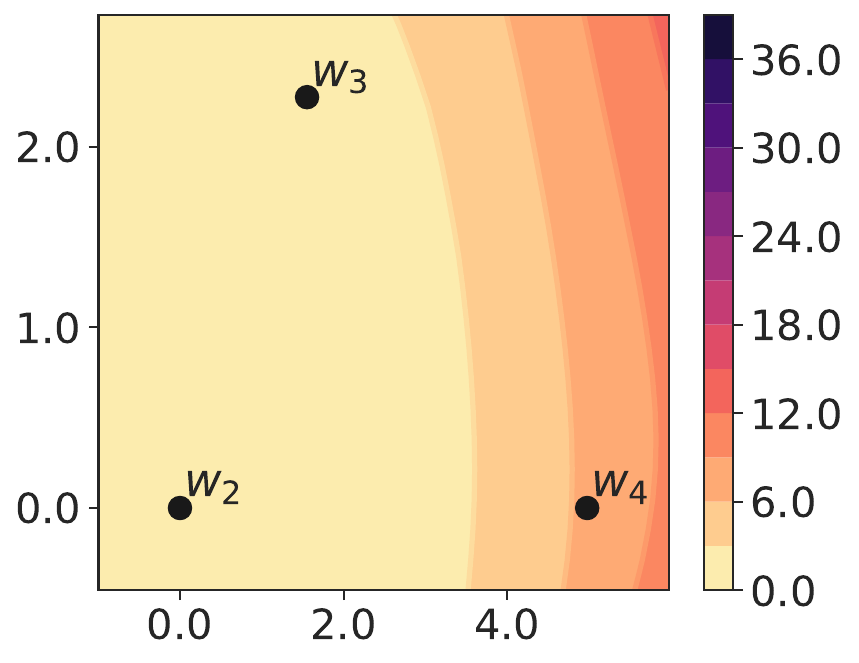}
      \caption{ER w/ SAM (T2)}
      \label{fig:5data_seq2_no_pt_ersam_contour_w2}
    \end{subfigure}\hspace{\fill}%
    \caption{Loss contours for SVHN (T1) and MNIST (T2) of 5-dataset-CV. The top row visualizes loss contours for SVHN where w$_1$, w$_2$, and w$_3$ are minima obtained after sequential training on SVHN, MNIST, and nonMNIST, respectively. Similarly, the bottom row visualizes loss contours for MNIST where w$_2$, w$_3$, and w$_4$ are minima obtained after sequential training on tasks MNIST, nonMNIST, and Fashion-MNIST. All of the above models start with random weights. SAM (FT w/ SAM, ER w/ SAM) leads to wide task minima compared to finetune (FT) and ER methods.}
    \label{fig:contours_sam_er_5data}
\end{figure}

In order to understand the effectiveness of the SAM, we visualize the loss contours and compute the sharpness metric (Equation~\ref{eq:sharpness_value}). We plot loss contours for task 1/ task 2 of Split CIFAR-50 (Figure~\ref{fig:contours_sam_er_c50}) and 5-dataset-CV (Figure~\ref{fig:contours_sam_er_5data}), under continual training from randomly initialized weights, and compare them across four different methods: FT, FT w/ SAM, ER, and ER w/ SAM. We show that SAM (FT w/ SAM and ER w/ SAM) leads to wide task minima (task 1/ task 2) across both data sets as compared to FT and ER methods. Moreover, from Table \ref{tab:summary} for ResNet-18-R (w/o PT) initialization, we see that for Split CIFAR-50, FT w/ SAM (15.0), ER w/ SAM (16.9), and MC-SGD w/ SAM (5.2) undergoes lesser forgetting than FT (23.7), ER (20.6), and MC-SGD (5.4) methods, respectively. These results convincingly demonstrate the effectiveness of SAM when used with vanilla FT, ER, and/ or MC-SGD methods. Similarly, we see that for 5-dataset-CV, MC-SGD w/ SAM (19.9) undergoes lesser forgetting than ER w/ SAM (27.3) and FT w/ SAM (40.6), which in turn significantly improves over FT (51.5). 
Next, we compare the loss contours between FT and ER methods and do not notice any stark difference in terms of flatness. However, in the presence of SAM, qualitatively we see that ER w/ SAM (Figures~\ref{fig:5data_seq2_no_pt_ersam_contour_w1}, \ref{fig:5data_seq2_no_pt_ersam_contour_w2}) leads to a flat loss basin in comparison to FT w/ SAM (Figures~\ref{fig:5data_seq2_no_pt_sam_contour_w1}, \ref{fig:5data_seq2_no_pt_sam_contour_w2}). 

We compute the sharpness metric for FT and FT w/ SAM methods. In Table~\ref{tab:sharpness_sam} we report sharpness metrics for 5-dataset-CV and Split CIFAR-50. We see that the SAM significantly reduces the sharpness in the case of randomly initialized models. Concretely, on the 5-dataset-CV, we see that the sharpness value (for $\epsilon=5$x$10^{-4}$ ) decreases from 2.1 (FT) to 0.7 (FT w/ SAM). Similarly, on the Split CIFAR-50, we see a drop from 2.3 (FT) to 0.7 (FT w/ SAM). These results validate that \textit{SAM indeed leads to flat minima, therefore, explaining the superior performance (in terms of average accuracy and forgetting) of SAM optimization procedure over baseline}.

\begin{table}[t!]      
    %   \small
      \centering
    %   \small
      % \begin{small}
      \begin{tabular}{llrrrr}
            \toprule
                                     \multirow{2}{*}{Data set}& \multirow{2}{*}{Method} & \multicolumn{2}{c}{$\epsilon = 5\times 10^{-4}$} & \multicolumn{2}{c}{$\epsilon = 10^{-3}$}                                         \\
            \cmidrule(r){3-4}\cmidrule(r){5-6}
                                     & & w/o PT                                 & w/ PT                           & w/o PT  & w/ PT    \\
            \midrule

            % 5-dataset (1 epoch)      & $3.16 \pm 1.40$ & $0.10 \pm 0.02$ & $8.13 \pm 3.26$ & $0.25 \pm 0.04$ \\
            \multirow{2}{*}{5-dataset-CV} & FT & $2.1_{0.6}$ & $0.1_{0.0}$ & $5.7_{1.6}$ & $0.2_{0.0}$ \\
            % & {\color{blue}FT + N-SGD} \\
                                          & FT w/ SAM & $0.7_{0.2}$ & $0.1_{0.0}$ & $1.8_{0.4}$ & $0.3_{0.0}$ \\
            \midrule
            \multirow{2}{*}{Split CIFAR-50} & FT & $2.3_{0.7}$ & $0.2_{0.1}$ & $6.1_{1.5}$ & $0.4_{0.1}$\\
            % & {\color{blue}FT + N-SGD} \\
            & FT w/ SAM & $0.7_{0.1}$ & $0.2_{0.0}$ & $2.0_{0.3}$ & $0.6_{0.0}$\\
            % Split-CIFAR50 (1 epoch)  & $1.51 \pm 0.28$ & $0.21 \pm 0.05$ & $4.03 \pm 0.65$ & $0.51 \pm 0.11$ \\
            % Split-CIFAR50 (5 epoch)  & $2.25 \pm 0.61$ & $0.13 \pm 0.04$ & $6.13 \pm 1.29$ & $0.33 \pm 0.08$ \\
            % Split-CIFAR100 (1 epoch) & $1.64 \pm 0.31$ & $0.15 \pm 0.04$ & $4.45 \pm 0.80$ & $0.35 \pm 0.09$ \\
            % Split-CIFAR100 (5 epoch) & $2.26 \pm 0.58$ & $0.10 \pm 0.02$ & $5.89 \pm 1.33$ & $0.24 \pm 0.06$ \\
            \bottomrule
      \end{tabular}
      % \end{small}
      \caption[caption]{Average sharpness (lower is flatter) of minima across tasks in a 100-dimensional random subspace. SAM significantly decreases the sharpness metric in comparison to Finetune (FT) method in the case of randomly initialized models (w/o PT).}
      \label{tab:sharpness_sam}
\end{table}

\subsection{Analyzing the influence of pre-training task minima curvature on forgetting}
\label{sec:pt_task_minima_curvature}% during fine-tuning}
In the prior sections, we demonstrate that pre-trained initializations alleviate forgetting in lifelong learning scenarios by guiding optimization towards flat minima in the loss basin for sequentially trained tasks. Furthermore, we show that explicitly optimizing for the flatness of the loss basin using the sharpness-aware minimization (SAM) procedure leads to an additional reduction in forgetting. It is important to note that our discussion so far has mainly focused on the flatness of the loss basin near the fine-tuned task minima. However, we now inquire about the impact of the loss basin's flatness (or sharpness) near the pre-training task minima on forgetting during lifelong learning. Specifically, we actively push a pre-trained model toward a region with low (or high) curvature. While such a model would benefit from learning structure from pre-training data, it would reside in flat (or sharp) regions of the loss basin with respect to the pre-training task. The question we pose is---\textit{What role does the curvature of the pre-training task minima play in lifelong learning, particularly in relation to forgetting in the fine-tuned task?}

\textbf{Experimental design.} To answer the above question, we conduct a controlled experiment with SVHN as our pre-training task and analyze the forgetting in MNIST (and its subsets) when learning homogeneous as well as diverse tasks in a sequence. We pre-train two separate models on SVHN, ensuring that one model converges to a flat minimum using the SAM procedure, while the other model converges to a sharper minimum using the Nudged-SGD procedure (NSGD; \citet{jastrzbski2018on}). It is important to ensure that both models exhibit comparable generalization performance on SVHN. Subsequently, we initialize these models and perform sequential training on four diverse task sequences, starting with MNIST as our fine-tuning task. The task sequences are as follows: MNIST $\rightarrow$ SVHN, MNIST $\rightarrow$ notMNIST, MNIST $\rightarrow$ Fashion-MNIST, and MNIST $\rightarrow$ CIFAR10. 
To create homogeneous tasks in the Split MNIST data set, we randomly select two digits for each task. Considering that MNIST comprises 10 digits, we then create a random sequence of five tasks. By utilizing different random seeds, we generate different tasks and consequently obtain a variety of task orderings. In total, we generate 25 distinct task sequences for the Split MNIST data set.
By examining the degree of forgetting observed in the case of (Split) MNIST, we can gain insights into the influence of the flatness (or sharpness) of the pre-training task minimum on the subsequent forgetting phenomenon. 
To ensure the broad applicability of our findings, we further conduct experiments using MNIST as the pre-training task and SVHN as the initial task for fine-tuning over four diverse task sequences.

\textbf{Nudged-SGD (NSGD).}  \citet{jastrzbski2018on} investigate the SGD dynamics in relation to the sharpest directions of the loss basin and demonstrate that although SGD updates align closely with these directions, they generally fail to minimize the loss when solely constrained to these directions. In order to enhance both the convergence speed and the generalization of the resulting model, \citet{jastrzbski2018on} introduces a variant of SGD, called Nudged-SGD (NSGD). NSGD aims to reduce the alignment between the SGD update direction and the sharpest directions. Specifically, NSGD is implemented as follows: instead of the standard SGD update $\Delta w(t) = - \eta \mathbf{g}(t)$, NSGD employs a reduced learning rate, $\eta' = \gamma\eta$, along the top $K$ eigenvectors. Meanwhile, NSGD continues to follow the standard SGD update in other directions. By setting $\gamma < 1.0$, NSGD diminishes the updates along the top $K$ eigenvectors. As a result, training speed improves while converging to sharper and better generalizing minima in comparison to vanilla SGD. In our experimental setup, we utilize a randomly initialized ResNet-18 model. Following \citet{jastrzbski2018on}, we set the parameters $\gamma$ to 0.01 and $K$ to 20. Additionally, we employ 100 training examples to compute the top $K$ eigenvectors using pytorch-hessian-eigenthings \citep{hessian-eigenthings} for every 100 training updates. Lastly, we set tolerance (relative accuracy for eigenvalues) to $1e^{-5}$ for determining the convergence of the Lanczos algorithm.

To begin, we conduct supervised pre-training of ResNet18 models using two different optimization procedures: SAM and NSGD, with the SVHN data set. Utilizing SAM, we achieve a validation accuracy of $92.0 (\pm 0.4)$ and a maximum eigenvalue $\lambda_1^{max}$ of $338.5 (\pm 91.5)$ (averaged over five runs). On the other hand, employing NSGD yields a validation accuracy of $92.4 (\pm 0.3)$ and a maximum eigenvalue $\lambda_1^{max}$ of $1416.8 (\pm 411.2)$ (also averaged over five runs). It is evident from these results that the NSGD leads to convergence towards sharper minima (as indicated by higher $\lambda_1^{max}$ values) and better generalization (as reflected in higher accuracy) compared to SAM and vanilla SGD. With vanilla SGD, we obtain a validation accuracy of $91.5 (\pm 0.6)$ and a maximum eigenvalue $\lambda_1^{max}$ of $1241.3 (\pm 236.5)$. Consequently, we have successfully obtained pre-trained models with varying levels of flatness or sharpness concerning the pre-training SVHN task, achieved through the SAM (\textit{Init:Flat}) and NSGD (\textit{Init:Sharp}) procedures.

\begin{figure}[t!]
    \centering
    % \captionsetup{font=footnotesize}
    \begin{subfigure}{.49\textwidth}
      \centering
      \includegraphics[width=\textwidth]{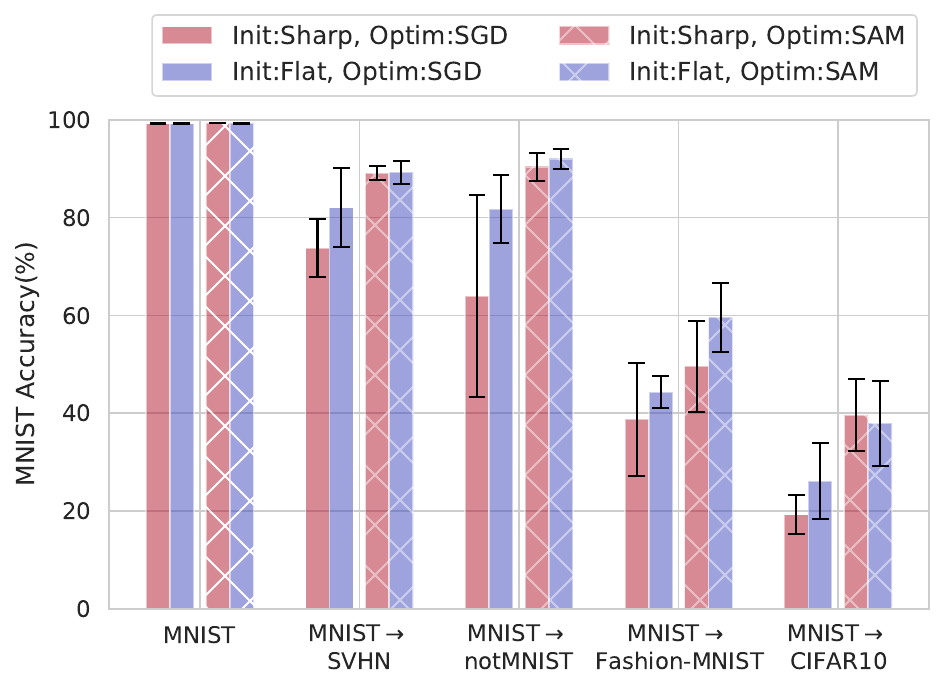}
      \caption{MNIST Accuracy ($\uparrow$)}
      \label{fig:mnist_nsgdinit_acc}
    \end{subfigure}\hspace{\fill}%
    \begin{subfigure}{.49\textwidth}
      \centering
      \includegraphics[width=\textwidth]{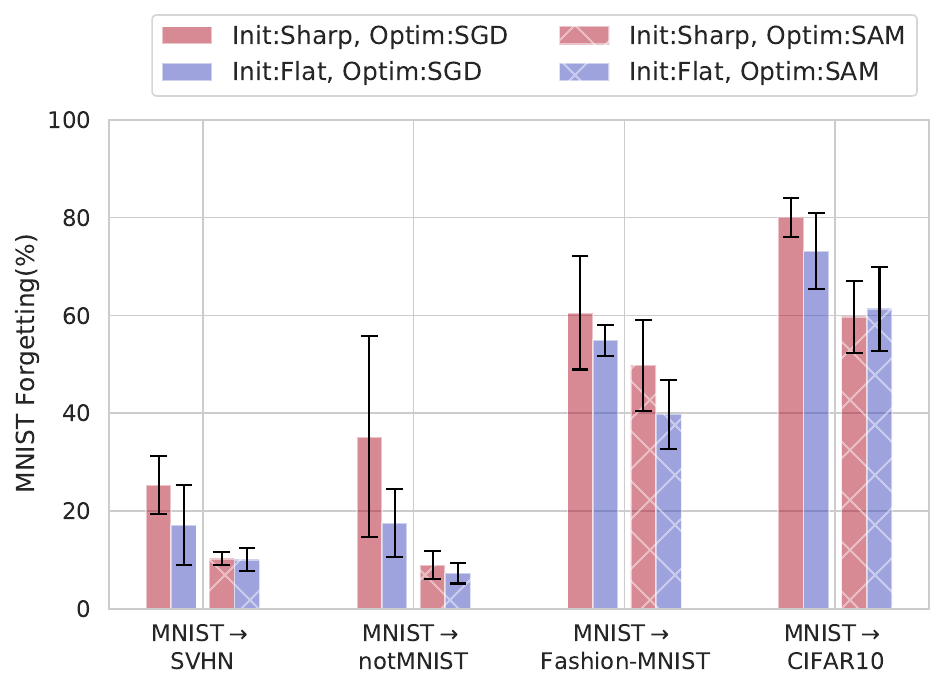}
      \caption{MNIST Forgetting ($\downarrow$)}
      \label{fig:mnist_nsgdinit_forget}
    \end{subfigure}\hspace{\fill}%
    \bigskip
    \begin{subfigure}{.49\textwidth}
      \centering
      \includegraphics[width=\textwidth]{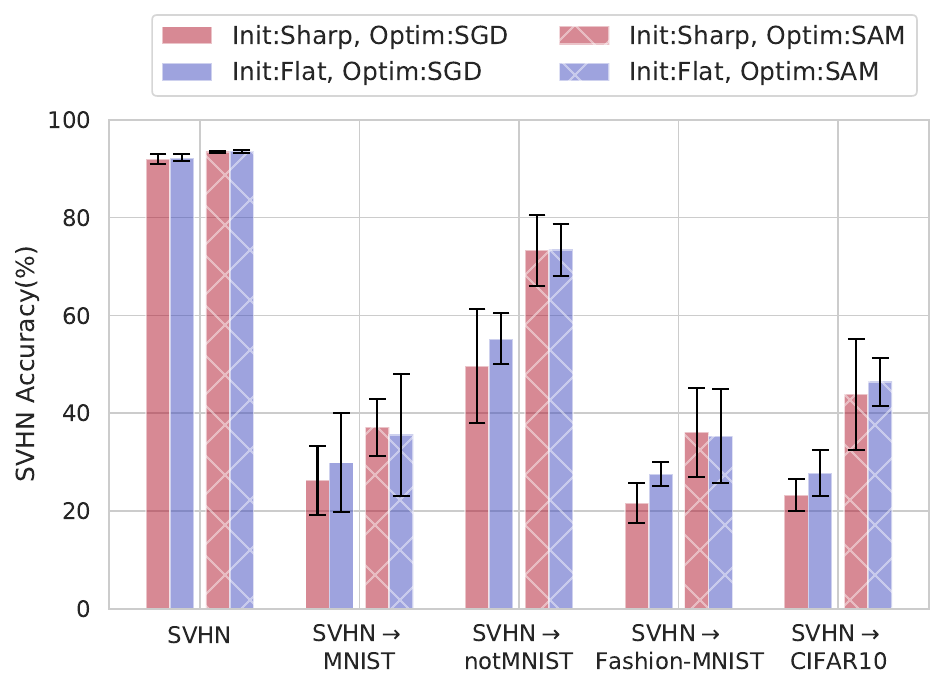}
      \caption{SVHN Accuracy ($\uparrow$)}
      \label{fig:svhn_nsgdinit_acc}
    \end{subfigure}\hspace{\fill}
    \begin{subfigure}{.49\textwidth}
      \centering
      \includegraphics[width=\textwidth]{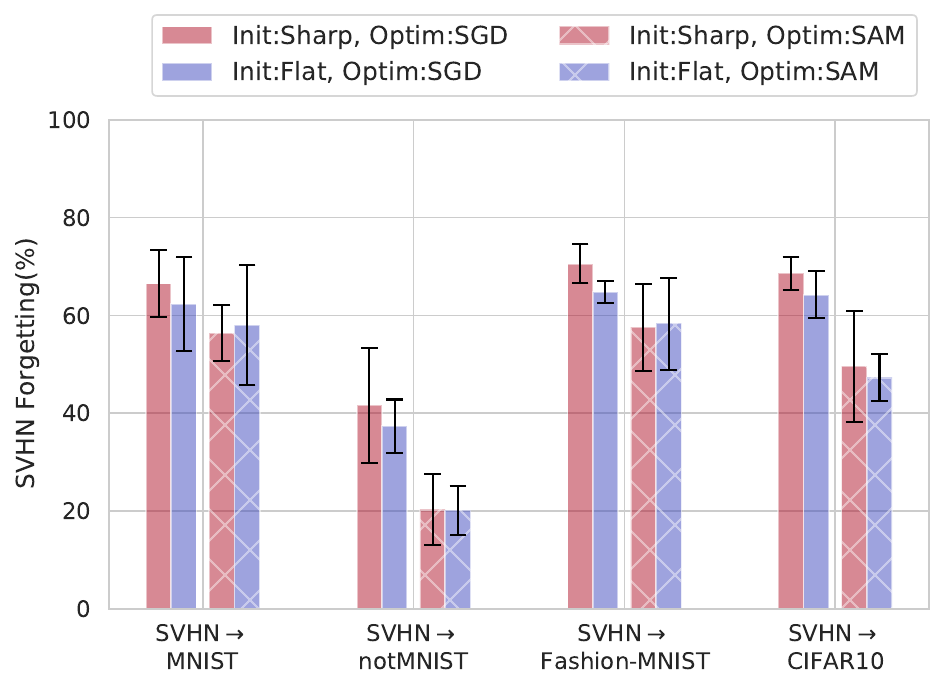}
      \caption{SVHN Forgetting ($\downarrow$)}
      \label{fig:svhn_nsgdinit_forget}
    \end{subfigure}\hspace{\fill}%
    \caption{Comparing performance of the first task (MNIST in the top row, SVHN in the bottom row) after sequential training on the second task across different supervised pre-training initializations (Init:Sharp, Init:Flat) and optimization procedures (Optim:SGD, Optim:SAM). $\uparrow$ indicates higher performance, $\downarrow$ indicates lower performance. All metrics are averaged over 5 runs. Pre-trained models converged to flat minima with respect to the pre-training task (Init:Flat, Optim:SGD) exhibit reduced forgetting with SGD in comparison to sharp minima(Init:Sharp, Optim:SGD). Notably, explicitly promoting flatness (Optim:SAM) for the fine-tuning task yields an even greater reduction in forgetting.}    
    \label{fig:nsgd}
\end{figure}

\begin{figure}[ht]
    \centering
    \begin{subfigure}{.245\textwidth}
      \centering
      \captionsetup{justification=centering}
      \includegraphics[width=\textwidth]{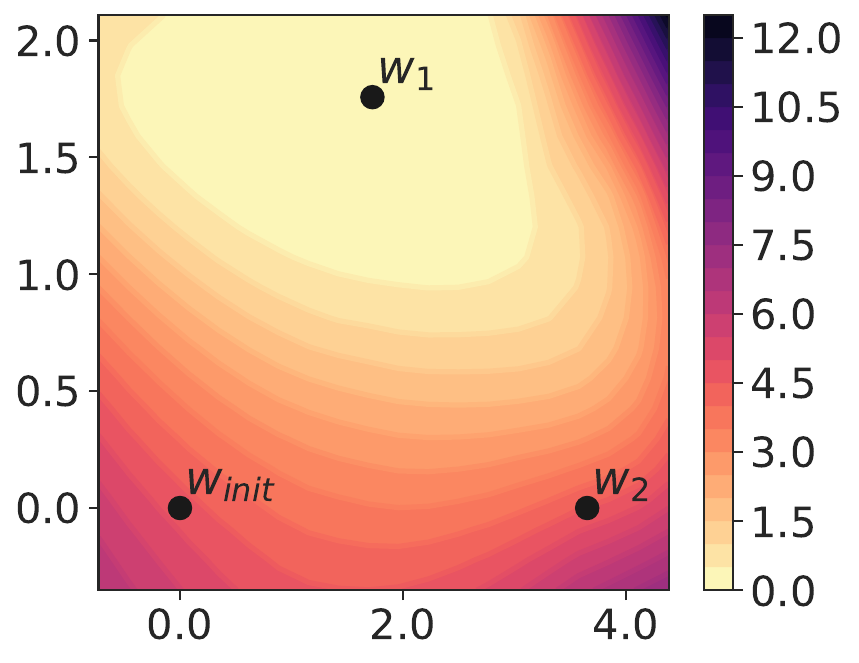}
      \caption{Init:Sharp, \\ Optim:SGD}
      \label{fig:mnist_not_mnist_sharp_sgd}
    \end{subfigure}\hspace{\fill}%
    \begin{subfigure}{.245\textwidth}
      \centering
      \captionsetup{justification=centering}
      \includegraphics[width=\textwidth]{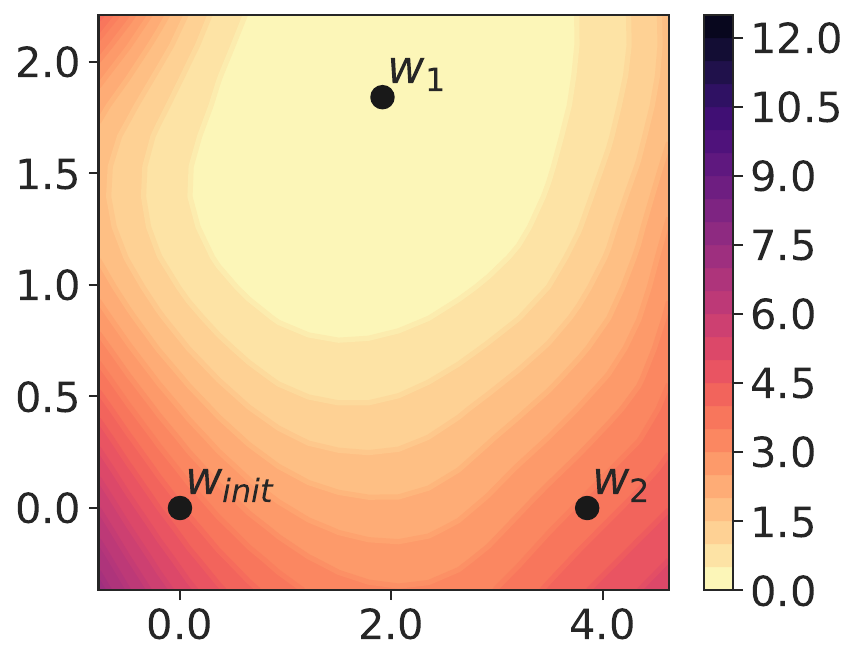}
      \caption{Init:Flat, \\ Optim:SGD}
      \label{fig:mnist_not_mnist_flat_sgd}
    \end{subfigure}\hspace{\fill}%
    \begin{subfigure}{.245\textwidth}
      \centering
      \captionsetup{justification=centering}
      \includegraphics[width=\textwidth]{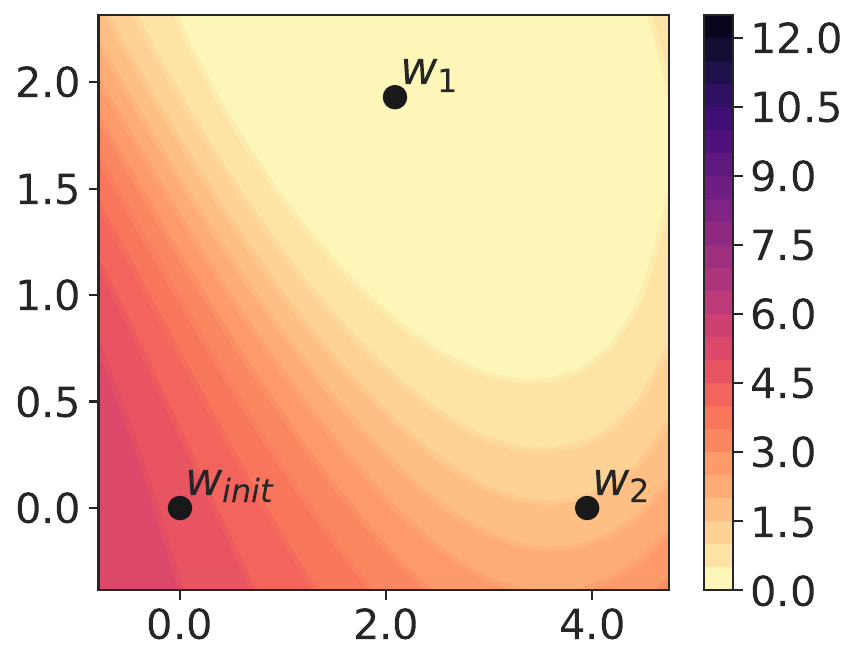}
      \caption{Init:Sharp, \\ Optim:SAM}
      \label{fig:mnist_not_mnist_sharp_sam}
    \end{subfigure}\hspace{\fill}%
    \begin{subfigure}{.245\textwidth}
      \centering
      \captionsetup{justification=centering}
      \includegraphics[width=\textwidth]{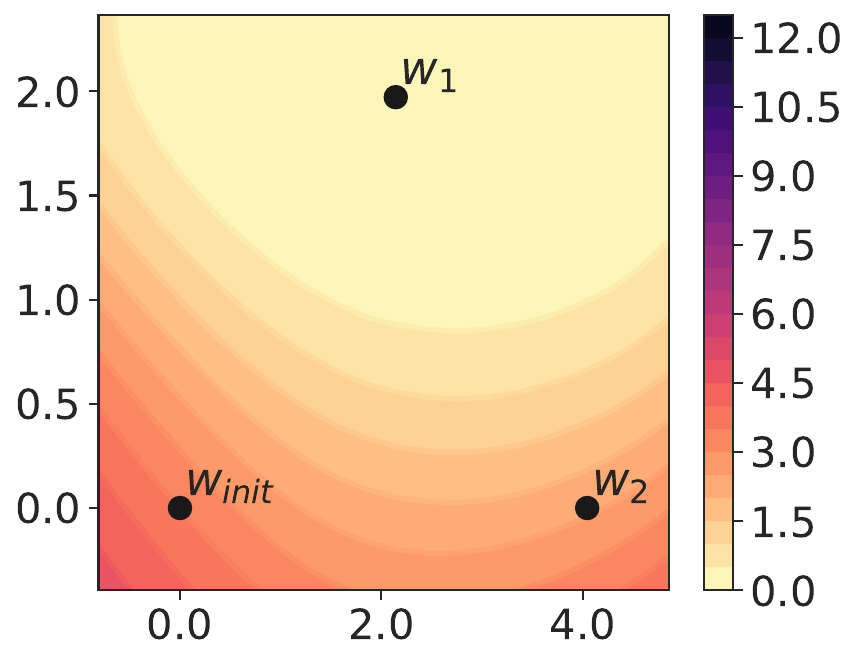}
      \caption{Init:Flat, \\ Optim:SAM}
      \label{fig:mnist_not_mnist_flat_sam}
    \end{subfigure}\hspace{\fill}%    
    \caption{Loss contours are shown for MNIST, with w$_{\text{init}}$, w$_1$, and w$_2$ representing the minima obtained after supervised pre-training on SVHN, followed by sequential training on MNIST and nonMNIST, respectively. The models are initialized either with a sharp pre-trained model (Init:Sharp) or a flat pre-trained model (Init:Flat). (a), (b) starting with a flat pre-trained model results in a flatter loss basin for MNIST during sequential fine-tuning. (c), (d) explicitly optimizing for flat MNIST minima using SAM (Optim:SAM) leads to even wider task minima compared to vanilla SGD.}
    \label{fig:contours_nsgd}
\end{figure}

\textbf{Discussion.} In order to examine the influence of pre-training task minima curvature on forgetting during fine-tuning, we proceed to sequentially fine-tune the aforementioned pre-trained models (Init:Sharp or Init:Flat) on MNIST, followed by one of four tasks: SVHN, notMNIST, Fashion-MNIST, and CIFAR10. Additionally, to investigate the interplay between pre-trained minima curvature and optimization dynamics, we conduct sequential fine-tuning using either the vanilla SGD optimizer (\textit{Optim:SGD}) or the SAM procedure (\textit{Optim:SAM}). Figures~\ref{fig:mnist_nsgdinit_acc} and \ref{fig:mnist_nsgdinit_forget} illustrate the accuracy and forgetting values for the MNIST task, respectively.
From Figure~\ref{fig:mnist_nsgdinit_forget}, it is evident that when fine-tuning with vanilla SGD (Optim:SGD), pre-trained models with flat minima (Init:Flat; shown in blue) consistently exhibit lower levels of forgetting compared to models with sharp minima (Init:Sharp; shown in red) across various task sequences. However, the advantage provided by the flat pre-trained models appears to diminish when employing the SAM optimization procedure (see Init:Sharp, Optim:SAM and Init:Flat, Optim:SAM). This finding highlights that the flatness of the MNIST task minima plays a more significant role in reducing forgetting for MNIST compared to the initialization flatness with respect to the pre-training task (SVHN in this case) minima.
Similar observations are reported in Figures~\ref{fig:svhn_nsgdinit_acc} and \ref{fig:svhn_nsgdinit_forget} when MNIST is used as the pre-training task, and forgetting is analyzed for the SVHN task during continual learning. 

In Figure~\ref{fig:contours_nsgd}, we provide a comparison of the loss contours for MNIST using sharp pre-trained initialization (Figure~\ref{fig:mnist_not_mnist_sharp_sgd}) and flat pre-trained initialization (Figure~\ref{fig:mnist_not_mnist_flat_sgd}). Upon visual inspection, we observe that the flat pre-trained initialization results in a wider loss basin for the MNIST minima (w$_1$), thereby, explaining lesser forgetting of MNIST when continually training on notMNIST in the case of pre-trained initialized models (see Figure~\ref{fig:mnist_nsgdinit_forget}; Init:Flat, Optim:SGD). Furthermore, when SAM is applied, Optim:SAM (Figures~\ref{fig:mnist_not_mnist_sharp_sam}, \ref{fig:mnist_not_mnist_flat_sam}) yields an even wider loss basin compared to Optim:SGD (Figures~\ref{fig:mnist_not_mnist_sharp_sgd}, \ref{fig:mnist_not_mnist_flat_sgd}).

In Table~\ref{tab:nsgd_metainit_study}, we present a comparison of average accuracy and forgetting on Split MNIST, focusing on sharp and flat supervised SVHN pre-trained initializations. Consistent with experiments involving diverse tasks (refer to Figure~\ref{fig:mnist_nsgdinit_forget}), we find that Init:Flat, Optim:SGD ($4.3$) exhibits lower forgetting compared to Init:Sharp, Optim:SGD ($7.6$) in homogeneous tasks. This finding reinforces the notion that actively promoting flatness during pre-training, in addition to learning structure from abundant data, is beneficial for reducing forgetting in sequentially fine-tuned tasks. Moreover, initializing with a sharp pre-trained model and explicitly optimizing for flatness using SAM, as seen in Init:Sharp, Optim:SAM ($3.7$), yields an even greater reduction in forgetting compared to flat pre-trained initialization with vanilla SGD, i.e., Init:Flat, Optim:SGD ($4.3$), aligning with our previous observations (refer to Figure~\ref{fig:nsgd}). However, we observe synergistic advantages when utilizing flat pre-trained models in conjunction with the SAM optimization procedure, resulting in minimal forgetting. Specifically, the combination of Init:Flat and Optim:SAM yields a forgetting value of $2.0$, showcasing the complementary benefits of these approaches.

To summarize, \textit{initiating fine-tuning with pre-trained models that have converged to flat minima with respect to the pre-training task helps mitigate forgetting. However, explicitly promoting flatness with respect to the fine-tuning task leads to a more pronounced reduction in forgetting.}

\begin{table}[t!]    
  \centering
  \begin{scriptsize}
  \begin{tabular}{lrrrrrr}
    \toprule
                         & \multicolumn{3}{c}{Optim:SGD} & \multicolumn{3}{c}{Optim:SAM}                                       \\
    \cmidrule(r){2-4}\cmidrule(r){5-7}
                         & Accuracy($\uparrow$) & Forgetting($\downarrow$) & LearnAcc($\uparrow$) & Accuracy($\uparrow$) & Forgetting($\downarrow$) & LearnAcc($\uparrow$) \\
    \midrule
    \multicolumn{3}{l}{\textbf{Task-agnostic}} \\
    Init:Random      & $80.9_{9.7}$ & $17.6_{9.0}$ & $96.9_{5.9}$ & $91.9_{5.3}$ & $7.7_{5.2}$ & $99.6_{0.3}$ \\
    Init:Meta        & $89.5_{9.1}$ & $9.8_{8.0}$ & $98.8_{3.2}$ & $\mathbf{92.3_{6.4}}$ & $\mathbf{7.5_{6.3}}$ & $99.6_{0.3}$ \\
    \midrule
    \multicolumn{3}{l}{\textbf{Supervised pre-training (SVHN)}} \\
    Init:Sharp       & $91.9_{8.1}$ & $7.6_{7.6}$ & $99.2_{1.8}$ & $96.1_{4.7}$ & $3.7_{4.6}$ & $99.8_{0.2}$ \\
    Init:Flat        & $95.2_{4.8}$ & $4.3_{4.7}$ & $99.1_{2.1}$ & $\mathbf{97.8_{2.2}}$ & $\mathbf{2.0_{2.1}}$ & $99.7_{0.4}$ \\    
    \bottomrule
  \end{tabular}
  \end{scriptsize}
  \caption[caption]{Comparing the performance of Split MNIST in terms of average accuracy(\%), forgetting(\%), and learning accuracy(\%), we analyze the impact of different initializations: random (Init:Random), task-agnostic meta-learned (Init:Meta), supervised pre-trained with explicit optimization for sharp minima (Init:Sharp), and supervised pre-trained with explicit optimization for flat minima (Init:Flat). We also consider different optimization approaches: vanilla SGD (Optim:SGD) and explicit optimization for flatness (Optim:SAM). $\uparrow$ indicates higher performance, while $\downarrow$ symbolizes lower performance. All metrics are averaged over 25 random task sequences. Our observations indicate that with Optim:SGD, MetaInit significantly reduces forgetting compared to random initialization, suggesting the benefits of initializing in flatter regions with minimal susceptibility to second-order effects. Also, pushing supervised pre-trained models towards flatter regions contributes to reduced forgetting. However, these gains diminish across all initialization schemes when explicitly optimizing for flatness (Optim:SAM) during lifelong learning. Nevertheless, we observe synergistic advantages when utilizing flat pre-trained or meta-initialized models in conjunction with the SAM procedure, resulting in minimal forgetting.}
  \label{tab:nsgd_metainit_study}
\end{table}

\subsection{Analyzing the influence of task-agnostic favorable initializations on forgetting}

Initialization is widely recognized as a crucial factor in a model's learning dynamics and overall performance \citep{glorot2010understanding, lecun2015deep, mishkin2015all}. Various initialization schemes have been developed for specific network architectures, such as fully-connected networks \citep{pennington2017resurrecting}, residual networks \citep{he2015delving}, convolutional networks \citep{xiao2018dynamical}, recurrent networks \citep{chen2018dynamical}, and attention-based networks \citep{huang2020improving}. However, these schemes are often architecture-specific and do not readily transfer to different or novel architectures. To overcome this limitation, \citet{dauphin2019metainit} propose \textit{MetaInit}, an automated initialization search approach using task-agnostic meta-learning. On the other hand, pre-training has also been shown to yield favorable initializations in terms of optimization \citep{erhan2009difficulty, hao2019visualizing, neyshabur2020being}, making it a data-driven method for finding effective initialization schemes. In the previous sections, we observe that pre-trained initializations reduce forgetting during sequential fine-tuning. As these observations pertain to pre-trained initializations, this section aims to directly investigate the contribution of favorable initializations by removing the pre-training aspect. Specifically, we pose the question---\textit{Does good initializations from MetaInit, shown to facilitate gradient descent by starting in locally linear (or wider) regions with minimal second-order effects, also mitigate forgetting when compared to a random initialization in a sharper region?}

\textbf{Experimental Design.} 
To address the above question, we perform controlled experiments to investigate the phenomenon of forgetting in the MNIST data set, both in the context of homogeneous and diverse task sequences. For this purpose, we employ the MetaInit algorithm \citep{dauphin2019metainit} to obtain a task-agnostic favorable initialization, which is independent of the specific task at hand. By comparing models initialized with the MetaInit approach to randomly initialized models, we aim to assess the impact of task-agnostic favorable initialization on forgetting. We examine the degree of forgetting specifically in the MNIST data set while sequentially learning four task sequences: MNIST $\rightarrow$ SVHN, MNIST $\rightarrow$ notMNIST, MNIST $\rightarrow$ Fashion-MNIST, and MNIST $\rightarrow$ CIFAR10, where MNIST serves as the initial task. Additionally, similar to the methodology described in Section~\ref{sec:pt_task_minima_curvature}, we conduct experiments involving homogeneous tasks from the Split MNIST data set.

\begin{figure}[t!]
    \centering
    % \captionsetup{font=footnotesize}
    \begin{subfigure}{.49\textwidth}
      \centering
      \includegraphics[width=\textwidth]{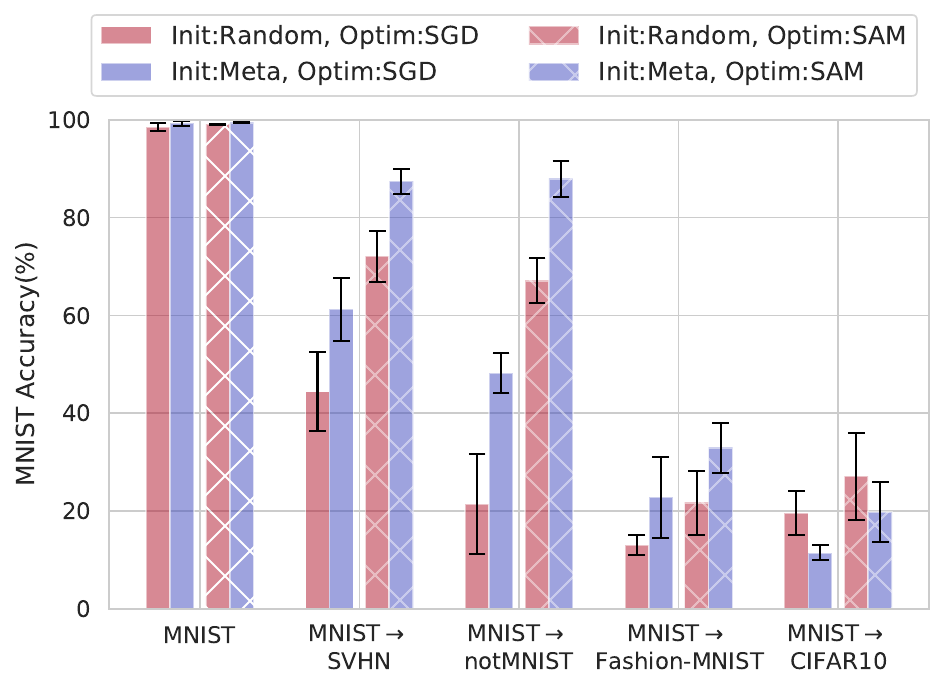}
      \caption{MNIST Accuracy ($\uparrow$)}
      \label{fig:mnist_metainit_acc}
    \end{subfigure}\hspace{\fill}%
    \begin{subfigure}{.49\textwidth}
      \centering
      \includegraphics[width=\textwidth]{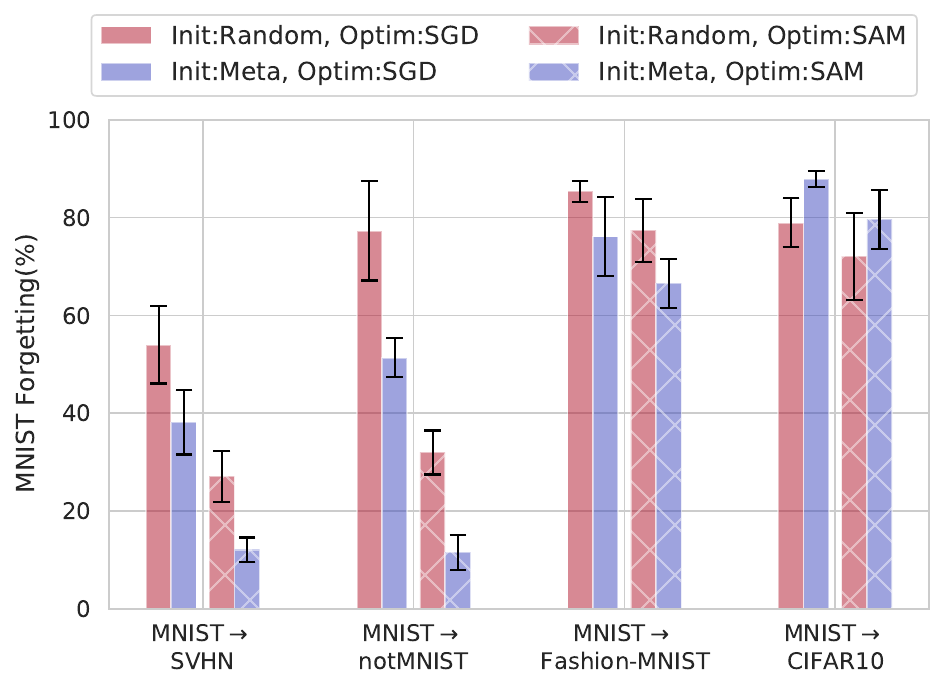}
      \caption{MNIST Forgetting ($\downarrow$)}
      \label{fig:mnist_metainit_forget}
    \end{subfigure}\hspace{\fill}%    
    \bigskip
    \begin{subfigure}{.49\textwidth}
      \centering
      \includegraphics[width=\textwidth]{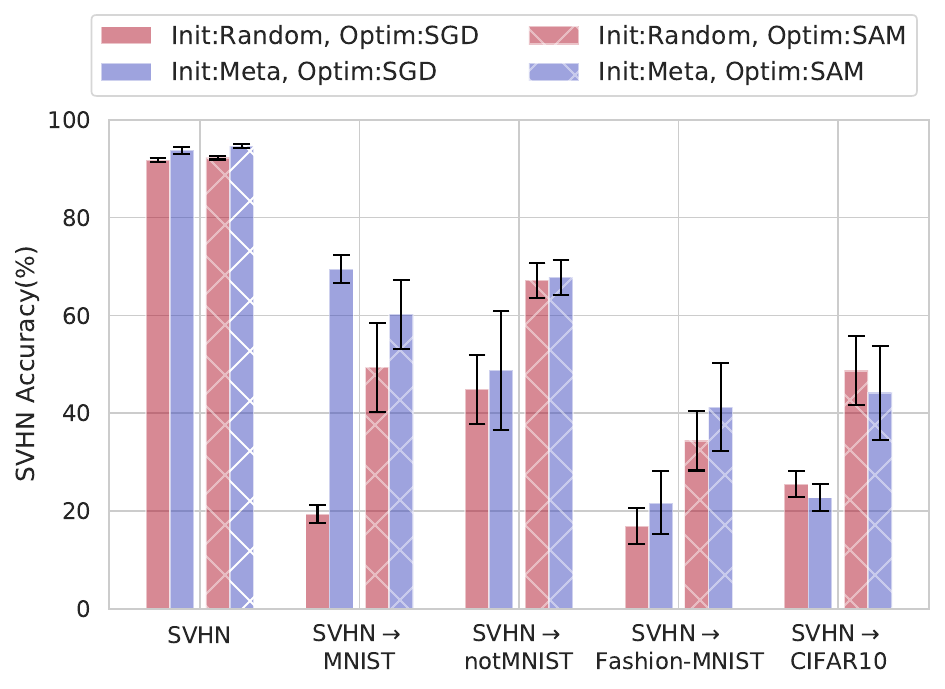}
      \caption{SVHN Accuracy ($\uparrow$)}
      \label{fig:svhn_metainit_acc}
    \end{subfigure}\hspace{\fill}
    \begin{subfigure}{.49\textwidth}
      \centering
      \includegraphics[width=\textwidth]{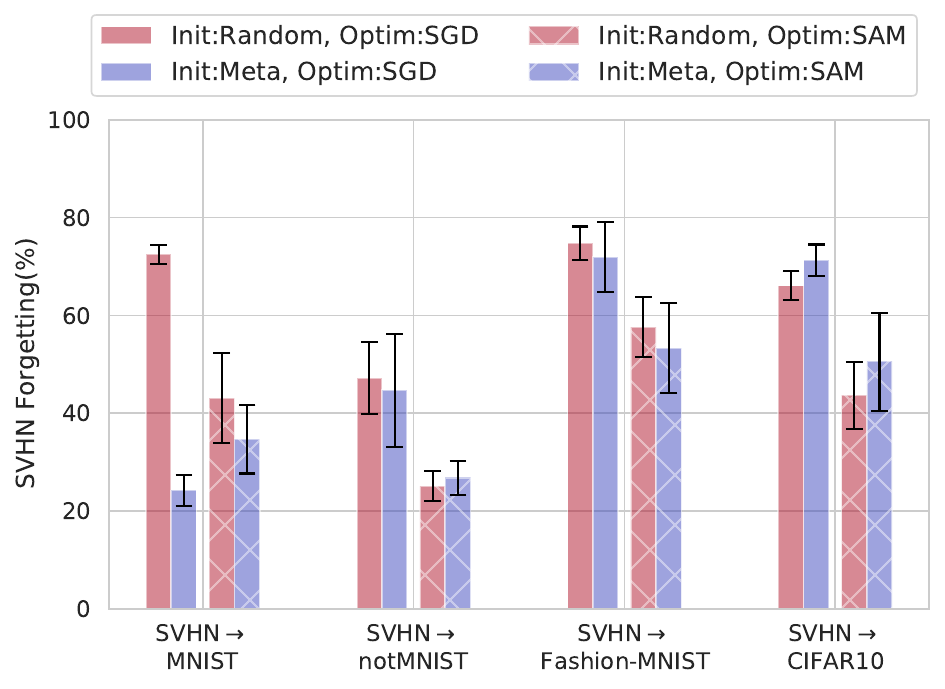}
      \caption{SVHN Forgetting ($\downarrow$)}
      \label{fig:svhn_metainit_forget}
    \end{subfigure}\hspace{\fill}%    
    \caption{Comparing the performance of the first task (\textit{MNIST} in the top row, \textit{SVHN} in the bottom row) after sequential training on the second task, we examine the impact of random and task-agnostic MetaInit initialization strategy (Init:Random, Init:Meta) and optimization procedures (Optim:SGD, Optim:SAM). $\uparrow$ indicates higher performance, while $\downarrow$ symbolizes lower performance. All metrics are averaged over 5 runs. The results show that task-agnostic MetaInit models (Init:Meta, Optim:SGD) exhibit reduced forgetting with SGD compared to random initialization (Init:Random, Optim:SGD). Similarly to Figure~\ref{fig:nsgd}, explicitly promoting flatness (Optim:SAM) for the sequential task leads to an even greater reduction in forgetting.}    
    \label{fig:metainit}
\end{figure}

\textbf{MetaInit.}  \citep{dauphin2019metainit} demonstrate that effective initializations exhibit characteristics that aid gradient descent by beginning in regions with minimal susceptibility to second-order effects. To quantify the impact of curvature (or second-order effects) around an initial choice of parameters, they introduce a quantity known as the gradient quotient (GQ). GQ measures the change in the gradient of a function following a single gradient descent step. Mathematically, the GQ is defined as follows
\begin{equation}
    \text{GQ}(L, w) = \frac{1}{N} \normx{\frac{\mathbf{g}(w) - \mathbf{H}(w)\mathbf{g}(w)}{\mathbf{g}(w) + \epsilon} - 1}_1 \approx \frac{1}{N} \normx{\frac{\mathbf{g}(w - \mathbf{g}(w))}{\mathbf{g}(w) + \epsilon} - 1}_1, \label{eq:grad_quotient}
\end{equation}
where $w \in \mathbb{R}^N$ are network parameters, $L$ is an empirical loss computed over the batch of the examples, $\mathbf{g}(w) = \nabla L(w)$ is the gradient, $\mathbf{H}(w) = \nabla^2L(w)$ denotes Hessian matrix, $\epsilon=\epsilon_0(2_{\mathbf{g}(w) \geq 0} - 1)$ with $\epsilon_0$ as a small constant and $||.||_1$ is the L1 vector norm. 
Alternatively, the GQ can be interpreted as the relative change in the gradient per parameter after a single step of gradient descent. As a result, parameters that cause a rapid change in the gradient exhibit large gradient quotients, while an optimal GQ of 0 is achieved when the loss function $L(w)$ behaves almost linearly, indicating a negligible Hessian matrix $\mathbf{H}(w) \approx 0$. Having established this metric to assess initialization quality, \citet{dauphin2019metainit} present MetaInit, a task-agnostic meta-learning algorithm aimed at obtaining a good initialization from suboptimal ones. The meta-objective of MetaInit is defined as follows
\begin{equation}
    \text{MetaInit}(L, w^*) = \arg \min_w \text{GQ}(L, w). \label{eq:meta_init}
\end{equation}

Following the methodology proposed by \citet{dauphin2019metainit}, we optimize the aforementioned meta-objective using random input data ($\mathbf{x} \sim \mathcal{N}(0, 1)$). It can be argued that solving for the meta-objective resembles a form of pre-training; however, no task-specific data is utilized in the learning process, thereby characterizing it as a task-agnostic initialization. Moreover, consistent with previous studies \citep{glorot2010understanding, dauphin2019metainit}, we solely adjust the norms of the initial weight matrices. We begin with five ResNet-18 models that are randomly initialized (\textit{Init:Random}), with a GQ averaging $5476.2 (\pm 2804.3)$. Through the MetaInit procedure (\textit{Init:Meta}), we attain a final GQ value of $0.9 (\pm 0.0)$.

\textbf{Discussion.} 
To investigate the impact of task-agnostic favorable initializations on forgetting, we sequentially train starting from the aforementioned initialization (Init:Random or Init:Meta) on MNIST, followed by one of four diverse tasks: SVHN, notMNIST, Fashion-MNIST, and CIFAR10. Furthermore, to explore the relationship between meta-learned initializations and optimization dynamics, we use either vanilla SGD (\textit{Optim:SGD}) or the SAM procedure (\textit{Optim:SAM}). The accuracy and forgetting values for the MNIST task are visualized in Figures~\ref{fig:mnist_metainit_acc} and \ref{fig:mnist_metainit_forget}, respectively. Figure~\ref{fig:mnist_metainit_forget} demonstrates that when optimized with vanilla SGD (Optim:SGD), meta-initialized models (Init:Meta; shown in blue) exhibit lower forgetting levels compared to randomly initialized models with high gradient quotient (Init:Random; shown in red) across various task sequences, except for CIFAR10, which differs significantly from MNIST. However, the advantage of meta-initialized models appears to diminish when employing the SAM optimization procedure (see Init:Random, Optim:SAM and Init:Meta, Optim:SAM), highlighting the more significant role of MNIST task minima flatness in reducing forgetting compared to the curvature of task-agnostic initialization.
Similar observations are reported in Figures~\ref{fig:svhn_metainit_acc}, \ref{fig:svhn_metainit_forget} when starting from SVHN task followed by one of MNIST, notMNIST, Fashion-MNIST, and CIFAR10 task during lifelong learning.

\begin{figure}[t]
    \centering
    \begin{subfigure}{.245\textwidth}
      \centering
      \captionsetup{justification=centering}
      \includegraphics[width=\textwidth]{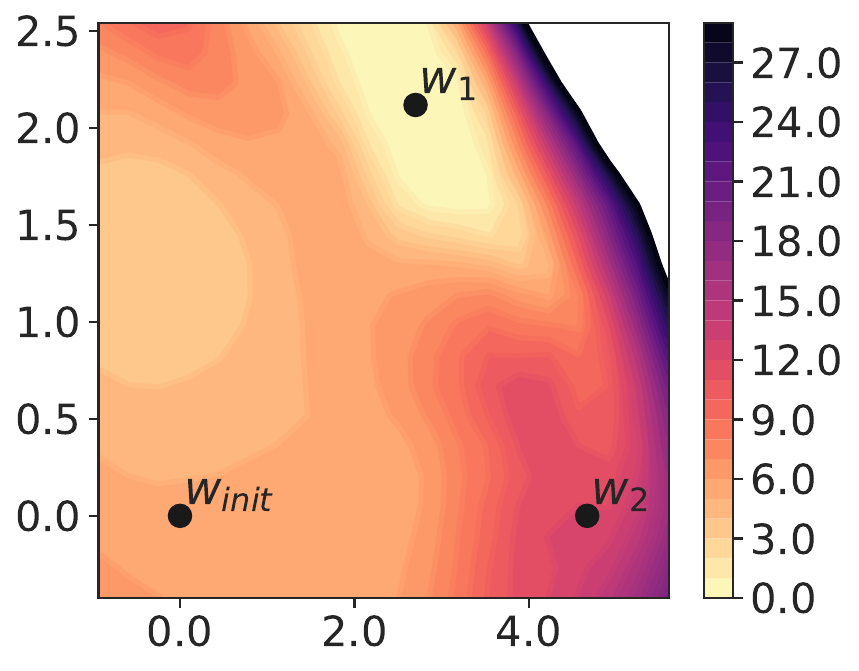}
      \caption{Init:Random, \\   Optim:SGD}
      \label{fig:mnist_not_mnist_random_sgd}
    \end{subfigure}\hspace{\fill}%
    \begin{subfigure}{.245\textwidth}
      \centering
      \captionsetup{justification=centering}
      \includegraphics[width=\textwidth]{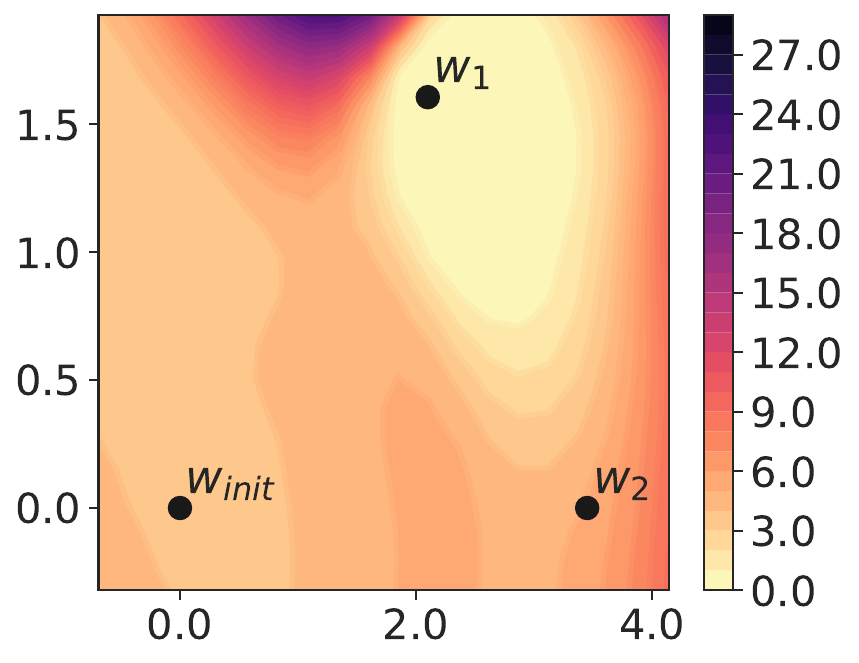}
      \caption{Init:Meta, \\ Optim:SGD}
      \label{fig:mnist_not_mnist_meta_sgd}
    \end{subfigure}\hspace{\fill}%
    \begin{subfigure}{.245\textwidth}
      \centering
      \captionsetup{justification=centering}
      \includegraphics[width=\textwidth]{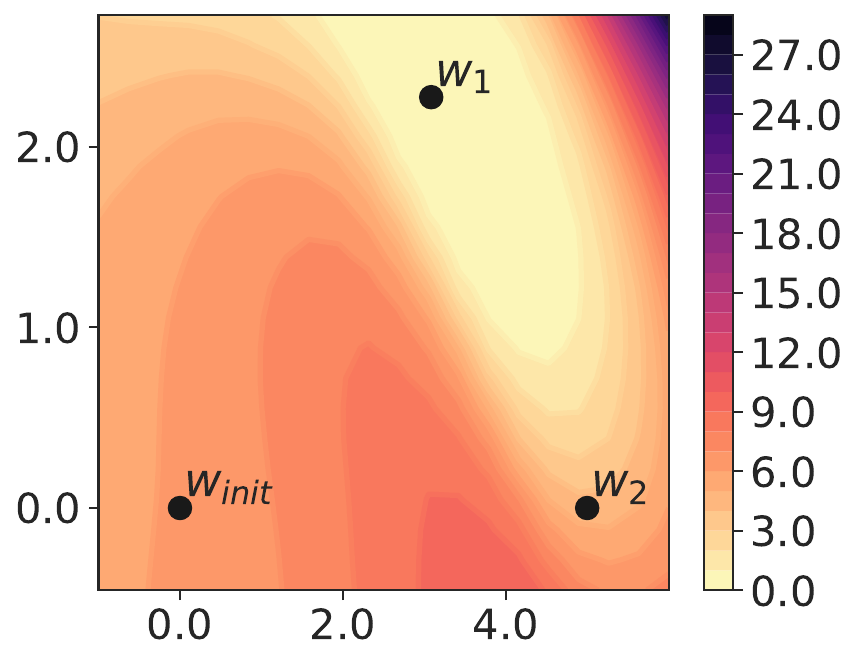}
      \caption{Init:Random, \\ Optim:SAM}
      \label{fig:mnist_not_mnist_random_sam}
    \end{subfigure}\hspace{\fill}%
    \begin{subfigure}{.245\textwidth}
      \centering
      \captionsetup{justification=centering}
      \includegraphics[width=\textwidth]{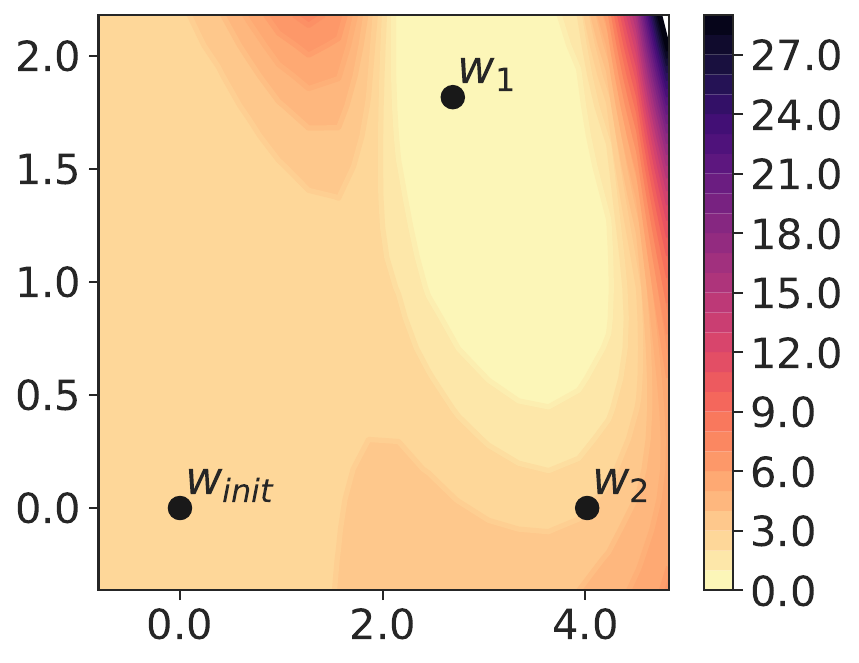}
      \caption{Init:Meta, \\ Optim:SAM}
      \label{fig:mnist_not_mnist_meta_sam}
    \end{subfigure}\hspace{\fill}%    
    \caption{Loss contours are shown for MNIST, with w$_{\text{init}}$, w$_1$, and w$_2$ representing either random or task-agnostic initialization, followed by sequential training on MNIST and nonMNIST, respectively. The models are initialized either with a random strategy (Init:Random) or a MetaInit strategy (Init:Meta). (a), (b) starting with a MetaInit initialization results in a flatter loss basin for MNIST during sequential fine-tuning. (c), (d) explicitly optimizing for flat MNIST minima using SAM (Optim:SAM) leads to even wider task minima compared to vanilla SGD.}
    \label{fig:contours_metainit}
\end{figure}

In Figure~\ref{fig:contours_metainit}, we provide a comparison of the loss contours for MNIST using random initialization (Figure~\ref{fig:mnist_not_mnist_random_sgd}) and task-agnostic MetaInit initialization (Figure~\ref{fig:mnist_not_mnist_meta_sgd}). Upon visual inspection, we observe that the MetaInit initialization results in a wider loss basin for the MNIST minima (w$_1$), thereby, explaining lesser forgetting of MNIST when continually training on notMNIST in the case of MetaInit initialized models (see Figure~\ref{fig:mnist_metainit_forget}; Init:Meta, Optim:SGD). Furthermore, when SAM is applied, Optim:SAM (Figures~\ref{fig:mnist_not_mnist_random_sam}, \ref{fig:mnist_not_mnist_meta_sam}) yields an even wider loss basin compared to Optim:SGD (Figures~\ref{fig:mnist_not_mnist_random_sgd}, \ref{fig:mnist_not_mnist_meta_sgd}), explaining superior results with Optim:SAM (refer to Figures~\ref{fig:mnist_metainit_acc}, \ref{fig:mnist_metainit_forget}).

Table~\ref{tab:nsgd_metainit_study} presents a comparative analysis of average accuracy and forgetting on the Split MNIST data set, focusing on random initialization and task-agnostic MetaInit initialization. Consistent with our experiments involving diverse tasks (see Figure~\ref{fig:mnist_metainit_forget}), we observe that Init:Meta, Optim:SGD ($9.8$) exhibits significantly lower levels of forgetting compared to Init:Random, Optim:SGD ($17.6$) in the context of homogeneous tasks. This finding supports the notion that initializing model parameters with regions that are locally linear (or wider) and less affected by second-order effects can be advantageous for mitigating forgetting in lifelong learning. Furthermore, we find that random initialization and explicit optimization for flatness using SAM, as demonstrated by Init:Random, Optim:SAM ($7.7$), result in an even greater reduction in forgetting compared to MetaInit initialization with vanilla SGD, i.e., Init:Meta, Optim:SGD ($9.8$), which aligns with our previous observations (see Figure~\ref{fig:metainit}).

To summarize, \textit{initializing models with lower gradient quotients (achieved through task-agnostic meta-learning) in regions that are less susceptible to second-order effects help reduce forgetting during lifelong learning. However, the reduction in forgetting is more significant when explicitly promoting flatness in relation to the specific sequential learning task.}

%% file: jmlrsections/06_relatedwork.tex
\section{Related Work}
\label{sec:related:app}

In this section, we establish the connections to significant related works in the field.

\textbf{Transfer learning} from generic pre-trained models has sparked significant advancements in machine learning \citep{Zhuang2021Survey}. Initially emerging in CV with the introduction of the ImageNet data set \citep{deng2009imagenet}, the practice of transfer learning has also undergone its own ``ImageNet revolution'' in NLP. Notably, large models pre-trained on self-supervised tasks have exhibited remarkable performance across various language-related tasks \citep{peters2018deep, howard2018universal, radford2018improving, devlin2019bert, liu2019roberta, raffel2019exploring}. The NIPS-95 workshop on ``Learning to Learn'' \citep{pan2009survey} initially motivated transfer learning as a means to facilitate lifelong learning. Our work revisits it in light of recent progress in transfer learning.

\textbf{Lifelong learning} approaches focus on the idea of mitigating the forgetting phenomenon and can be grouped into four categories: (1) \textit{regularization-based} approaches either augment the loss function with extra penalty terms preventing important parameters learned on previous tasks from significantly deviating while training on the new task \citep{kirkpatrick2017overcoming, zenke2017continual, chaudhry2018riemannian, aljundi2018memory} and/ or enforce distillation-based penalty \citep{li2017learning, dhar2019learning}; (2) \textit{memory-based} approaches that augment the model with episodic memory for sparse experience replay of previous task examples, either during training \citep{lopez2017gradient, chaudhry2018efficient, chaudhry2019tiny, guo2020improved} and/ or during inference \citep{rebuffi2017icarl, de2019episodic, wang2020efficient}; some approaches learn generative model to simulate replay buffers \citep{shin2017continual, sun2020lamal}; (3) \textit{optimization-based} approaches that either maintain a space of gradient directions for previous tasks and projects the gradients of a new task in a direction orthogonal to that space, ensuring less disruption of older tasks \citep{farajtabar2020orthogonal} or modify training regimes by specific hyperparameter configurations yielding wider minima and reducing the forgetting \citep{mirzadeh2020understanding}; and (4) \textit{architecture-based} approaches that dynamically expand the network based upon new tasks \citep{rusu2016progressive, aljundi2017expert, yoon2018lifelong, sodhani2020toward}, or iteratively learn mask \citep{mallya2018piggyback, serra2018overcoming} or prunes networks \citep{mallya2018packnet} for new tasks. By formulation, these approaches do not undergo forgetting and hence we consider regularization and episodic memory-based approaches for our analysis.

\textbf{Meta-learning} involves the development of models capable of learning over time and has been employed in various studies focusing on lifelong learning \citep{riemer2018learning, pmlr-v97-finn19a, javed2019meta, wang2020efficient, gupta2020look}. Notably, \citet{Caccia2020OnlineFA} propose a two-phase continual learning scenario, where the initial phase entails pre-training utilizing MAML \citep{finn2017model}, followed by continual deployment with task revisiting. They emphasize that deploying agents without any pre-training in lifelong learning scenarios would be impractical, a sentiment shared by other studies \citep{lomonaco2019rehearsal}. While some of these works employ pre-trained initializations, the comprehensive examination of the impact of pre-training in lifelong learning remains largely unexplored and is the focus of our work.

\textbf{Optimization and loss landscape} works have explored the relationship between pre-training and wider optima in single-task generalization \citep{hao2019visualizing, neyshabur2020being}, as well as the effects of larger batch sizes on sharper minima and poorer generalization in single-task learning \citep{keskar2016large}. Additionally, \citet{mirzadeh2020linear} compare minima resulting from multitask learning and continual learning, establishing that minima from continual learning are linearly connected to optimal sequential multitask minima but not to each other, leading to forgetting. 
While these studies examine the connection between pre-training and flat minima in single-task scenarios or the relationship between flat minima and model generalization, we extend this research by investigating whether the benefits of pre-training persist during sequential training on multiple tasks. We explore the effects of pre-training on loss landscapes throughout lifelong learning and validate a hypothesis that elucidates the role of pre-training in lifelong learning.

%% file: jmlrsections/07_conclusion.tex
% \section{Conclusion and Future Work}
\section{Discussion}
In this paper, we investigate the role of pre-training in lifelong learning, examining various benchmarks and modalities. Our findings reveal that models with pre-trained initializations exhibit significantly reduced forgetting compared to models with random initializations. Despite pre-trained models starting with higher task performance, they undergo lesser forgetting. This observation holds even when comparing a sequentially fine-tuned pre-trained model, without additional regularization to mitigate forgetting, to a randomly initialized model trained with state-of-the-art lifelong learning methods. A key insight is that lifelong learning methods should explore the learning of generic initializations for future tasks rather than solely focusing on alleviating the forgetting of previous tasks. Furthermore, our analysis relating to different pre-trained models indicates that while increased model capacity provides benefits up to a certain point, the quality of pre-trained representations becomes more crucial when considering longer and more diverse task sequences.

To explain the above phenomenon, we conduct several analyses of the loss landscapes throughout continual training for models initialized randomly and with pre-training. Our findings reveal that the minima attained by the pre-trained models after each task exhibit significantly flatter characteristics compared to those obtained by randomly initialized models. Consequently, even if pre-trained models deviate from the original flat task minima, the task loss does not experience a significant increase, thereby undergoing less forgetting.
Furthermore, explicitly seeking flat basins using the SAM procedure yields even lower forgetting than existing methods. Integrating SAM into the baselines surpasses state-of-the-art techniques, underscoring its valuable contribution to advancing lifelong learning methods. Lastly, our analysis of various initializations, including task-agnostic meta-learned and supervised pre-trained models explicitly guided towards flat loss regions, showcases the synergistic behavior that arises when combined with the SAM procedure during sequential fine-tuning.

%% file: jmlrsections/08_appendix.tex
\section{Implementation Details}
\label{sec:impldetails}

\subsection{CV Experiments} For all vision experiments, we use the full ResNet-18 \cite{he2016deep} architecture, with the final linear layer replaced (the number of outputs corresponds to the total number of classes in all given tasks). During inference, only the subset of outputs corresponding to the given task is considered. In accordance with \citep{he2015delving}, for random initialization strategy, we initialize weight tensors using the Kaiming normal distribution, while setting batch normalization weights to 1 and biases to 0. 
All images are resized to $224\times224$, and normalized with $\mu=(0.485, 0.456, 0.406)$ and $\sigma=(0.229, 0.224, 0.225)$. We used an SGD optimizer with the learning rate set to $.01$ for all methods (we did a hyperparameter search for both pre-trained and randomly initialized models and found the learning rate $0.01$ resulted in a good learning accuracy for both pre-trained and randomly initialized models). The batch size was set to $10$ for the Split CIFAR-50 and Split CIFAR-100 experiments and $64$ for the 5-dataset-CV experiments. The memory per class for ER was set to 1, and the $\lambda$ parameter for EWC was also set to 1. For Stable SGD, we performed a hyperparameter sweep over the parameters specified in the original paper, namely: 
\begin{itemize}
    \item initial learning rate: [.25 (Split CIFAR-100-R, Split CIFAR-50-R, 5-dataset-CV-R), .1, .01 (Split CIFAR-100-PT, Split CIFAR-50-PT), .001 (5-dataset-CV-PT)]
    \item learning rate decay: [0.9 (Split CIFAR-50-R, 5-dataset-CV-R, Split CIFAR-100-PT), 0.85 (Split CIFAR-100-R, Split CIFAR-50-PT), 0.8 (5-dataset-CV-PT)]
    \item batch size: [10 (all), 64]
    \item dropout: [0.5 (5-dataset-R), 0.25 (Split CIFAR-100-R, Split CIFAR-50-R, Split CIFAR-100-PT, Split CIFAR-50-PT, 5-dataset-CV-PT)]
\end{itemize}
For Mode Connectivity SGD, we adopted the hyperparameters specified in the original paper by \citet{mirzadeh2020linear}. To achieve continual learning minima, in experiments starting with the random initialization, we used an initial learning rate of 0.1, a learning rate decay of 0.8, a momentum of 0.8, a dropout of 0.1, a batch size of 10 for Split CIFAR-50 and Split CIFAR-100, and a batch size of 64 for the 5-dataset-CV. Conversely, in experiments starting with pre-trained initialization, we used an initial learning rate of 0.01, no learning rate decay, no momentum, no dropout, and a batch size of 64 for the 5-dataset-CV and 10 for Split CIFAR-10 and Split CIFAR-100. To obtain the linear mode connectivity minima, we employed 10 line samples, a learning rate of 0.01, a momentum of 0.8, a batch size of 64, a number of epochs set to 5, and initialized the position for the minima at 0.5.

\subsection{NLP Experiments} 
For most of the text classification experiments, we use the Transformer architecture-based text encoder, DistilBERT-base \citep{sanh2019distilbert} to encode our input. In a single-sentence text classification task, $x_t$ is an input sentence to be classified. In a sentence-pair classification task, the concatenation of $x_t^{1}$ and $x_t^{2}$ sentences separated by a $[SEP]$ symbol is considered as an input $x_t$. DistilBERT produces a contextual representation of each token in $x_t$ including a special beginning of the sentence token symbol $[CLS]$. We use the representation of the $[CLS]$ symbol from the model as features for a linear task classifier. We have a separate classifier for each task. We mainly set hyper-parameters to default implementation from HuggingFace.\footnote{https://github.com/huggingface/transformers} For random initialization, we initialize weight tensors using the normal distribution $\mathcal{N} \sim (0, 0.02)$, biases to 0, layer normalization weights to 1 and biases to 0. We use Adam as our optimizer, set dropout $0.1$, the base learning rate to $2e^{-5}$, batch size to $32$, and the maximum total input sequence length after tokenization to $128$. For EWC, we set the regularization strength $\lambda$ to $100$ (as this ended up with comparable LA across other methods), and for ER, following \citep{chaudhry2019tiny}, the memory per class per task is set to $1$. For SAM, we set $\rho=0.02$ for all models (random as well as pre-trained) on 5-dataset-NLP and 15-dataset-NLP. For SplitYahooQA we set $\rho=0.001$.
For our experiments involving encoder-decoder architecture, we utilize the pre-trained T5-Small v1.1 checkpoint from HuggingFace.\footnote{https://huggingface.co/google/t5-v1\_1-small} Since encoder-only models have a separate classification head, we also incorporate a separate classification head in the T5 decoder. We employ the Adam optimizer, with a dropout rate of 0.1. The base learning rate is set to $3e^{-4}$, the batch size is set to 32, and the maximum total input sequence length is set to 128. In the case of SAM, we set the hyperparameter $\rho$ to 0.05 for both the 5-dataset-NLP and 15-dataset-NLP experiments.

\subsection{Sharpness metric} 
The matrix $A \in \mathbb{R}^{n\times p}$ used for projecting the parameters onto a subspace is randomly sampled and then normalized row-wise. 
%Since this matrix is very large, the computation of the bounds from Equation \ref{eq:sharpness_bounds} does not actually calculate the pseudo inverse $A^+$, as this is very memory intensive and unstable. 
Since this matrix is very large, the computation of the pseudo-inverse $A^+$ (required for calculating the bounds in Equation~\ref{eq:sharpness_bounds}) is very memory intensive and unstable.
Instead, we directly calculate $A^+w$ by finding the least squares solution to $Ab=w$. To find the maximum referenced in Equation~\ref{eq:sharpness_value}, we use the L-BFGS-B algorithm.\footnote{We used the implementation provided by scipy at \url{https://docs.scipy.org/doc/scipy/reference/optimize.minimize-lbfgsb.html}} We set the maximum number of iterations for the algorithm to 10, and to speed up computation, we directly provide the gradients along with the loss to the algorithm, instead of using the default 2-point finite difference gradient estimation. 

For ResNet-18 ($n=11M$), we set $p=100$. However, for DistilBERT ($n=66M$) when we set $p=100$, we notice extremely small values for the sharpness metric. With the increase in the number of parameters, $n$, we should ideally increase random subspace projection dimension $p$. Setting larger $p(>100)$ values for DistilBERT, however, leads to memory issues relating to allocating space for $A$ and computing the bounds (even with the more efficient method discussed above). So instead of evaluating the sharpness metric in a random manifold, we perform the maximization in the entire space $\mathbb{R}^{n}$ (basically setting $A=I_n$). According to \citet{keskar2016large}, when $\epsilon$ is small enough and $A=I_n$, the sharpness metric in Equation~\ref{eq:sharpness_value} relates to the largest eigenvalue of $\nabla^2L(w)$.

\section{Task-specific results}
\label{sec:taskspecificres}

\begin{figure}%[ht]
    \centering
    \bigskip
    \begin{subfigure}{.32\textwidth}
      \centering
      \includegraphics[width=0.8\textwidth]{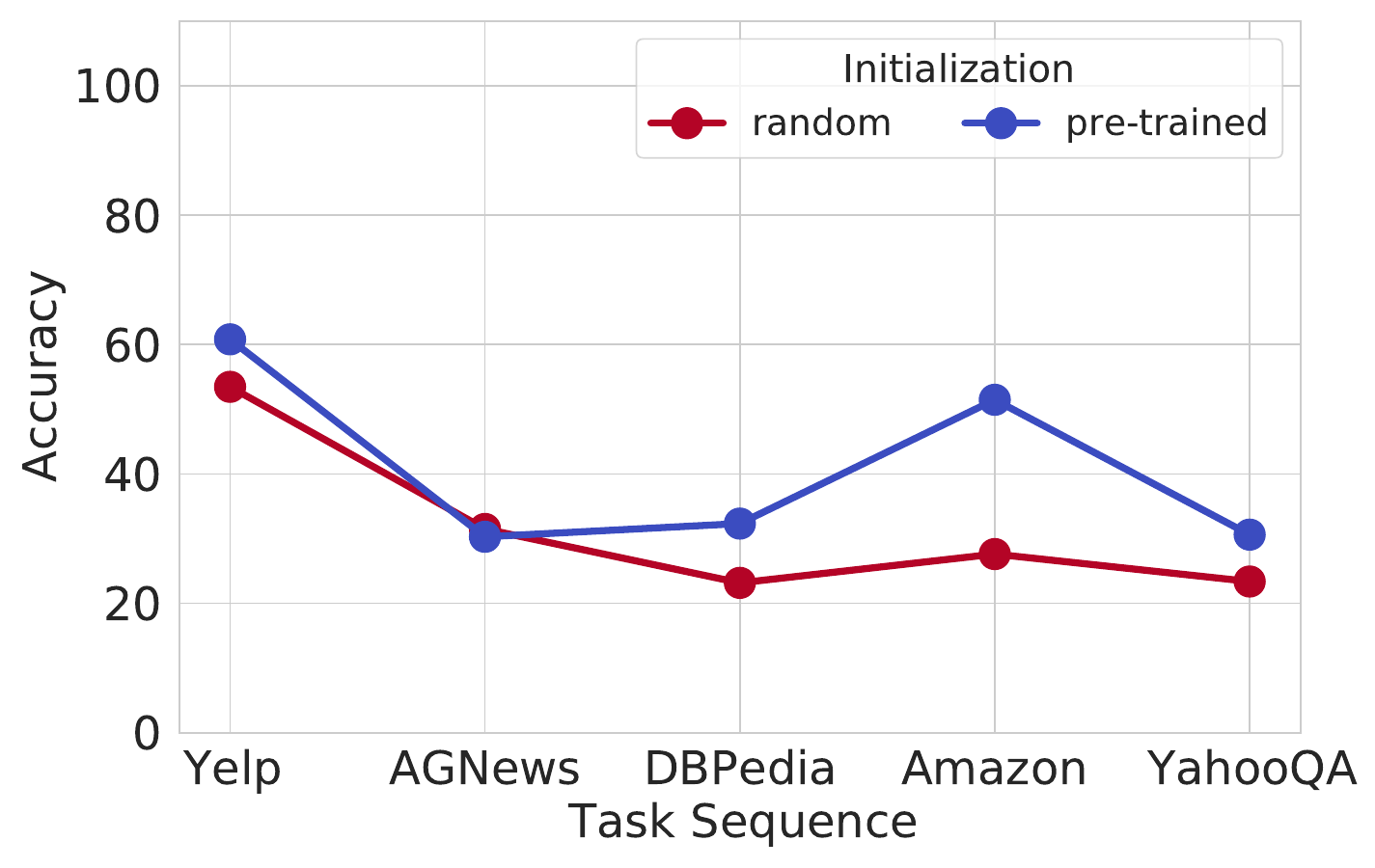}
      \caption{Yelp (Seq1)}
      \label{fig:5datanlp_seq1_task1}
    \end{subfigure}\hspace{\fill}%
    \begin{subfigure}{.32\textwidth}
      \centering
      \includegraphics[width=0.8\textwidth]{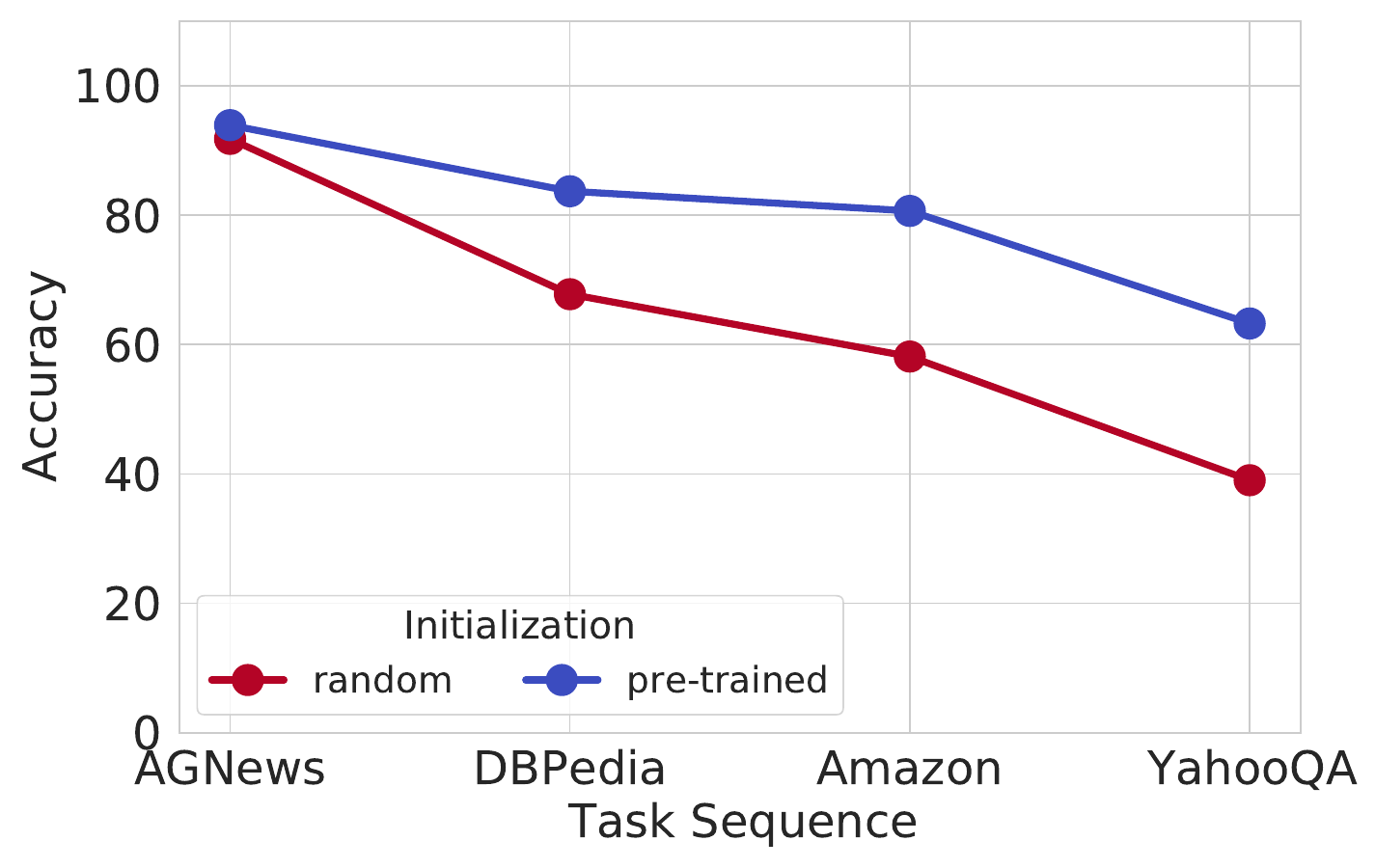}
      \caption{AGNews (Seq1)}
      \label{fig:5datanlp_seq1_task2}
    \end{subfigure}\hspace{\fill}%
    \begin{subfigure}{.32\textwidth}
      \centering
      \includegraphics[width=0.8\textwidth]{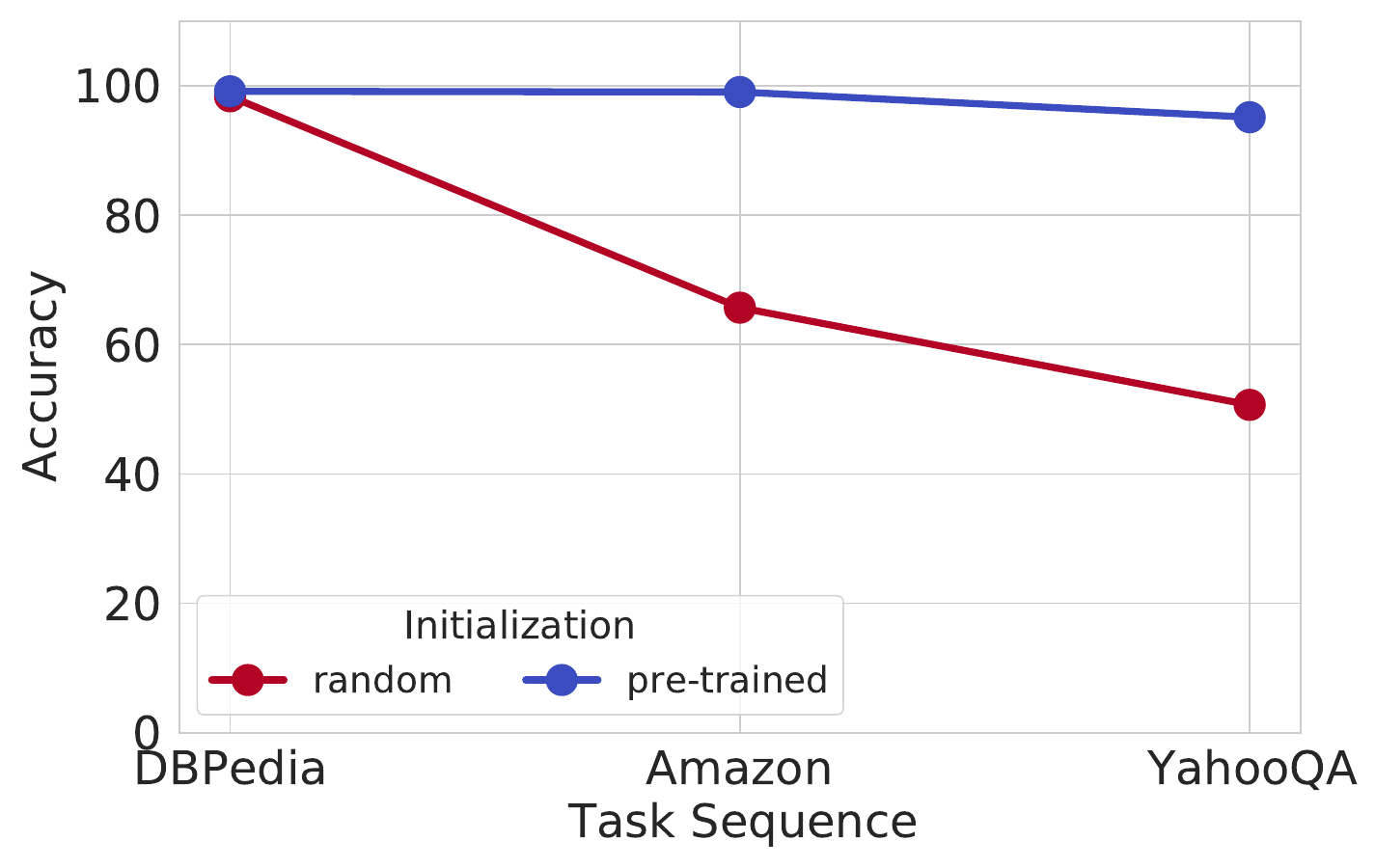}
      \caption{DBPedia (Seq1)}
      \label{fig:5datanlp_seq1_task3}
    \end{subfigure}\hspace{\fill}%
    \bigskip
    \begin{subfigure}{.32\textwidth}
      \centering
      \includegraphics[width=0.8\textwidth]{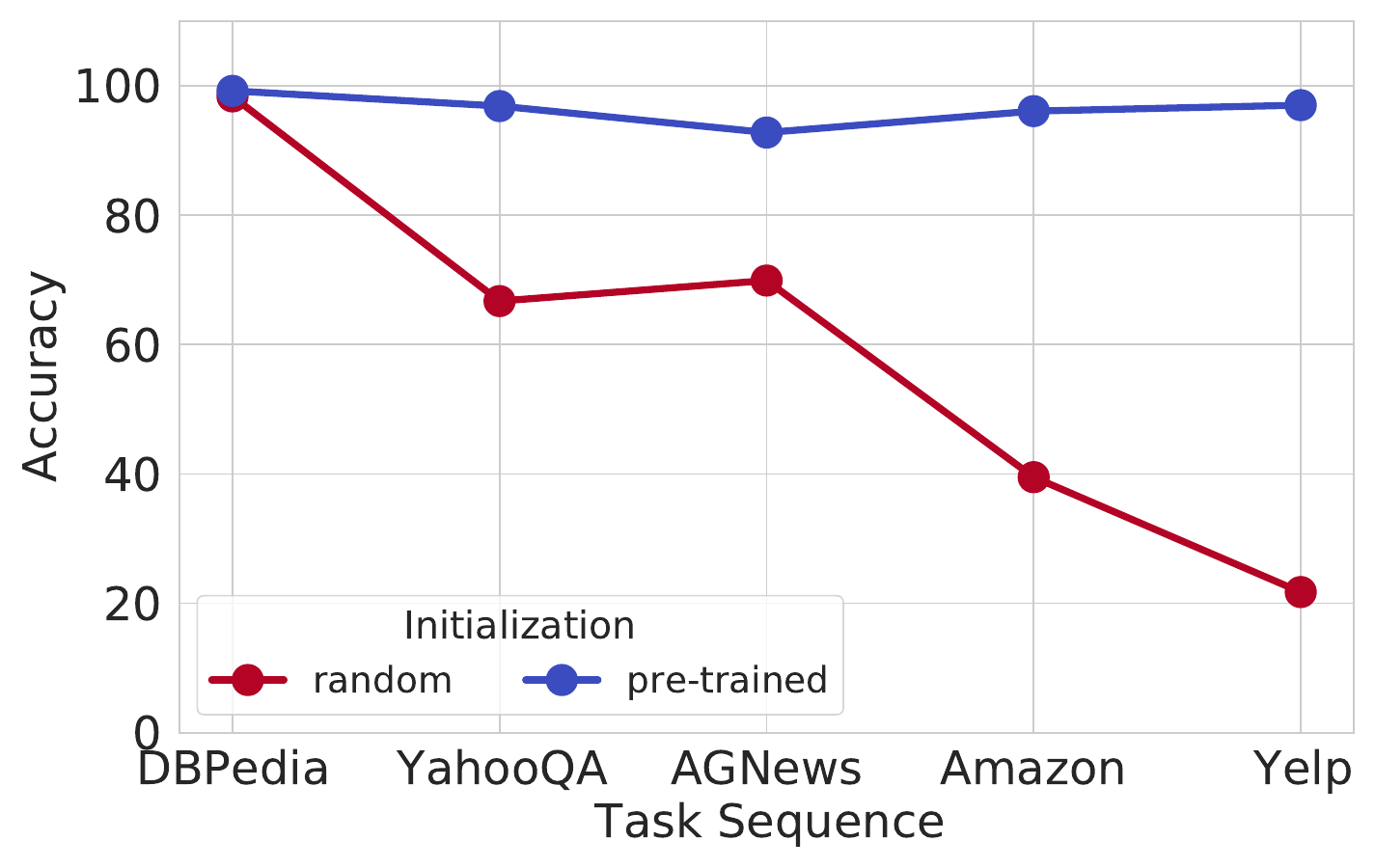}
      \caption{DBPedia (Seq2)}
      \label{fig:5datanlp_seq2_task1}
    \end{subfigure}\hspace{\fill}%
    \begin{subfigure}{.32\textwidth}
      \centering
      \includegraphics[width=0.8\textwidth]{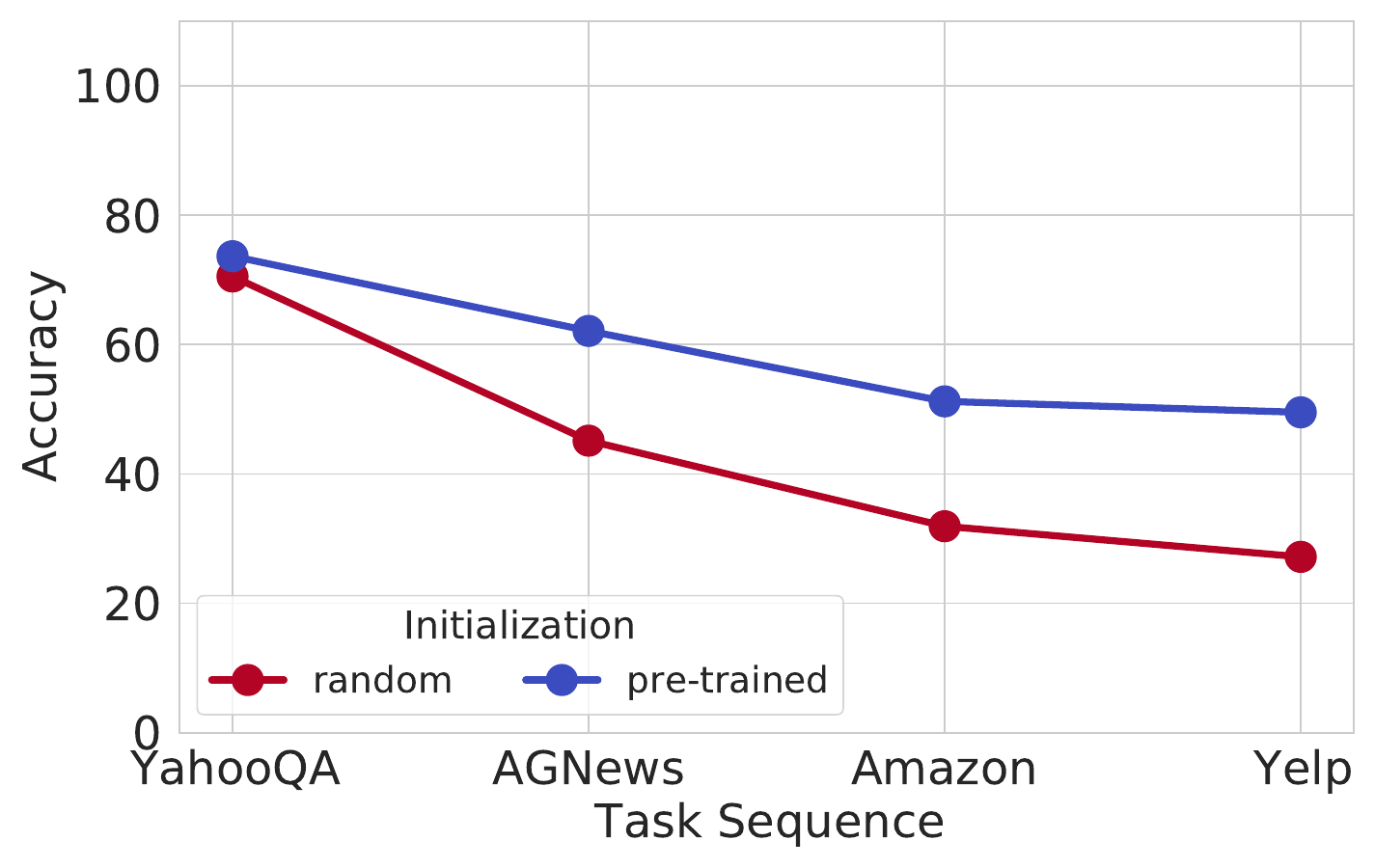}
      \caption{YahooQA (Seq2)}
      \label{fig:5datanlp_seq2_task2}
    \end{subfigure}\hspace{\fill}%
    \begin{subfigure}{.32\textwidth}
      \centering
      \includegraphics[width=0.8\textwidth]{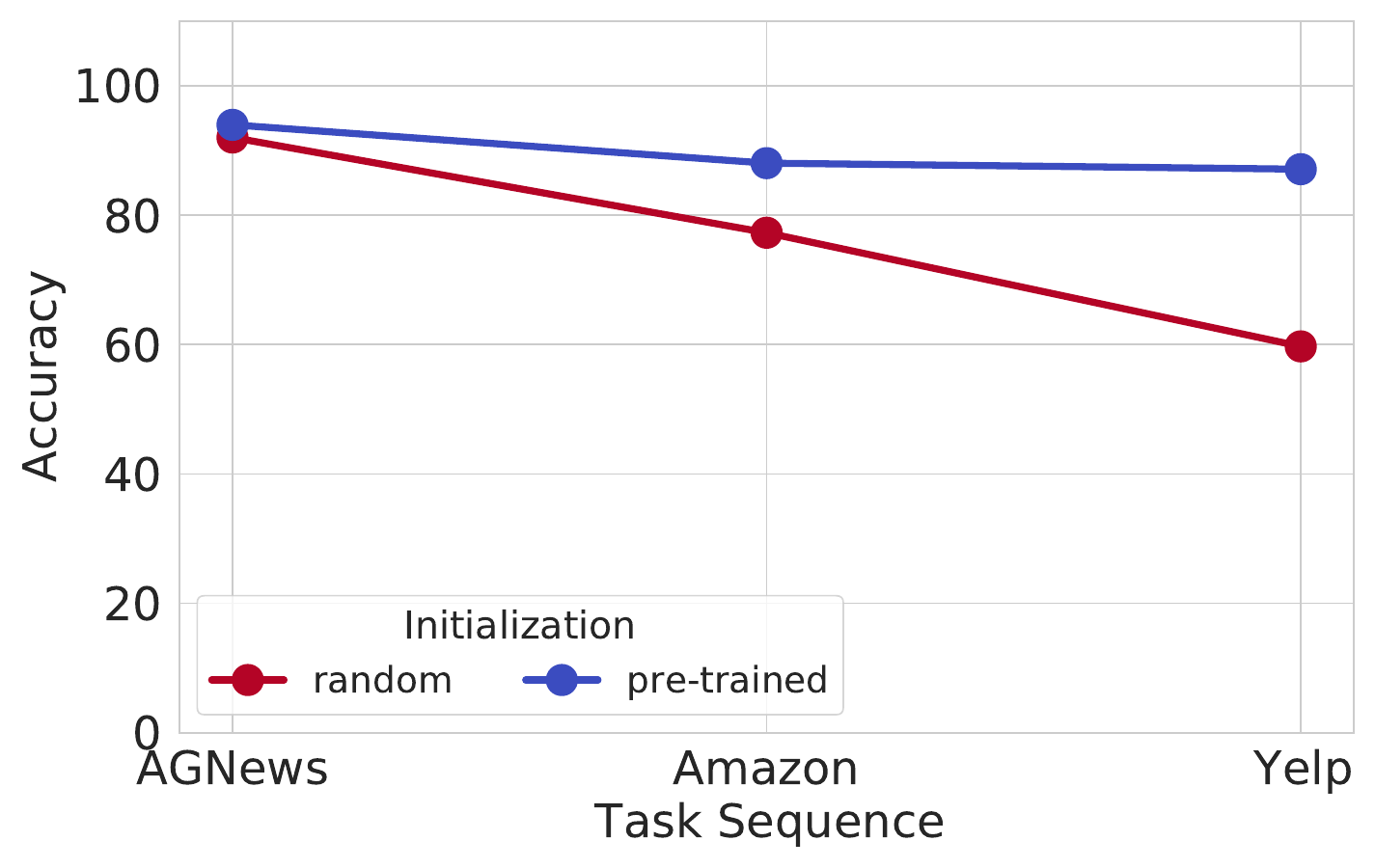}
      \caption{AGNews (Seq2)}
      \label{fig:5datanlp_seq2_task3}
    \end{subfigure}\hspace{\fill}%
    \bigskip
    \begin{subfigure}{.32\textwidth}
      \centering
      \includegraphics[width=0.8\textwidth]{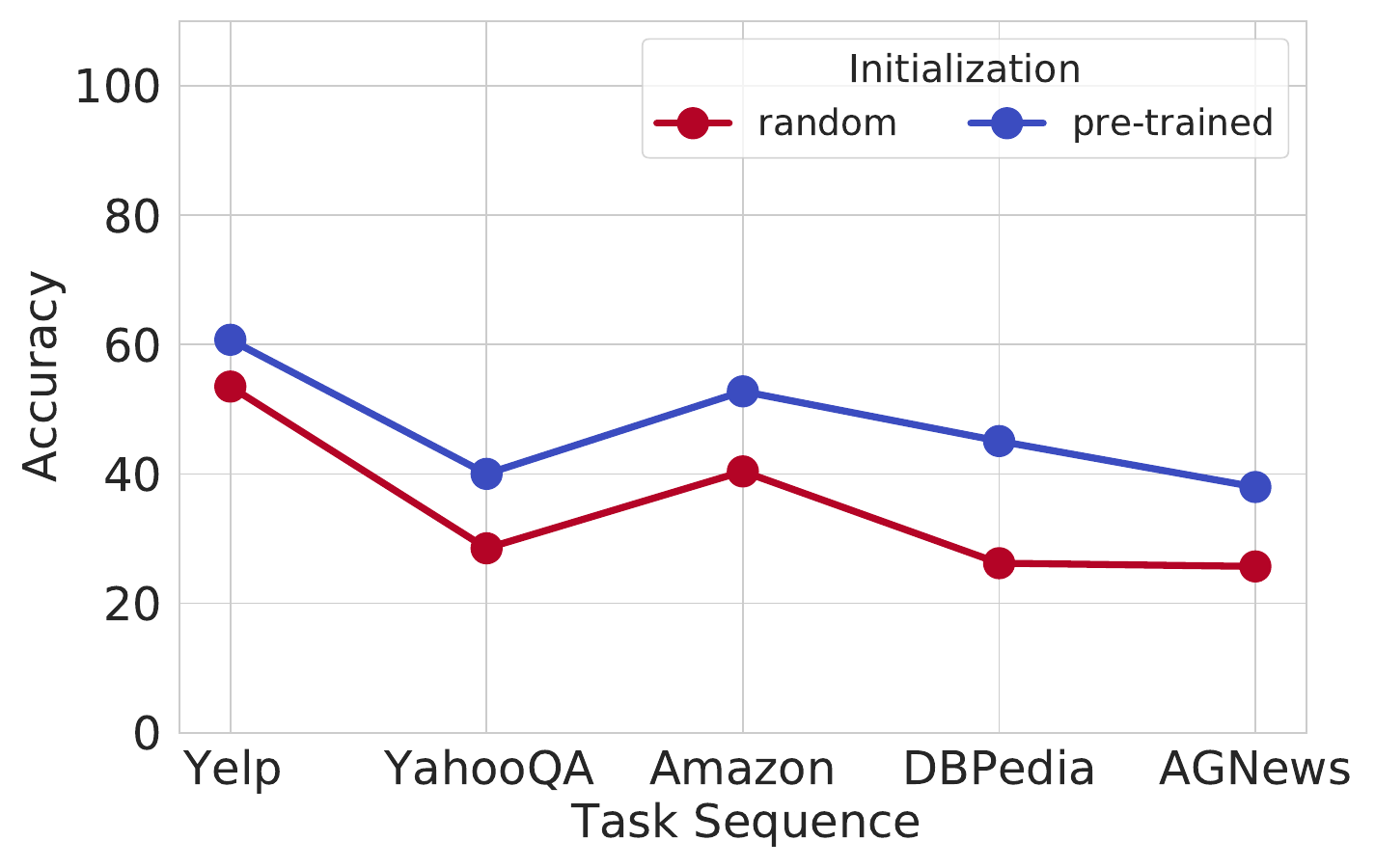}
      \caption{Yelp (Seq3)}
      \label{fig:5datanlp_seq3_task1}
    \end{subfigure}\hspace{\fill}%
    \begin{subfigure}{.32\textwidth}
      \centering
      \includegraphics[width=0.8\textwidth]{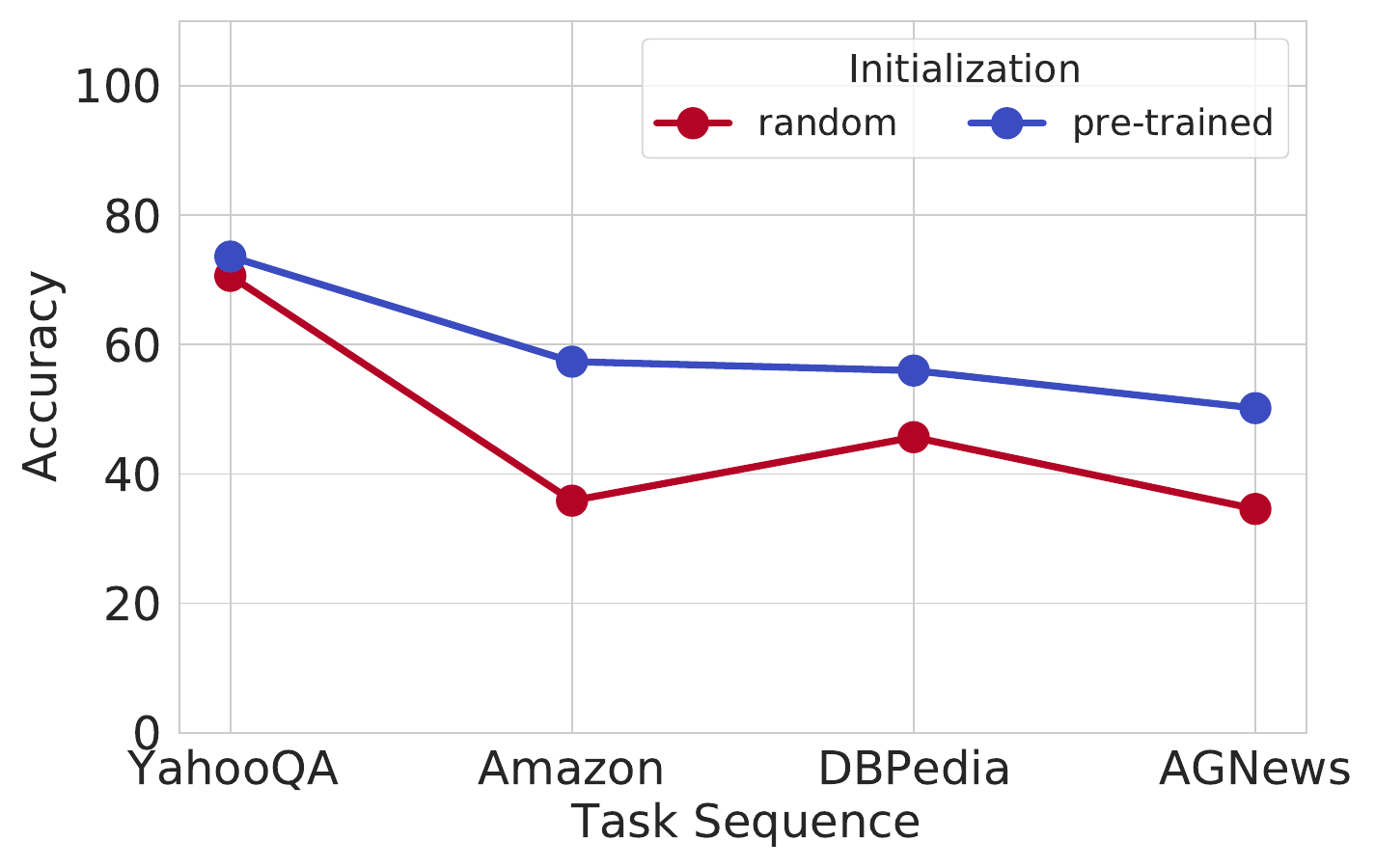}
      \caption{YahooQA (Seq3)}
      \label{fig:5datanlp_seq3_task2}
    \end{subfigure}\hspace{\fill}%
    \begin{subfigure}{.32\textwidth}
      \centering
      \includegraphics[width=0.8\textwidth]{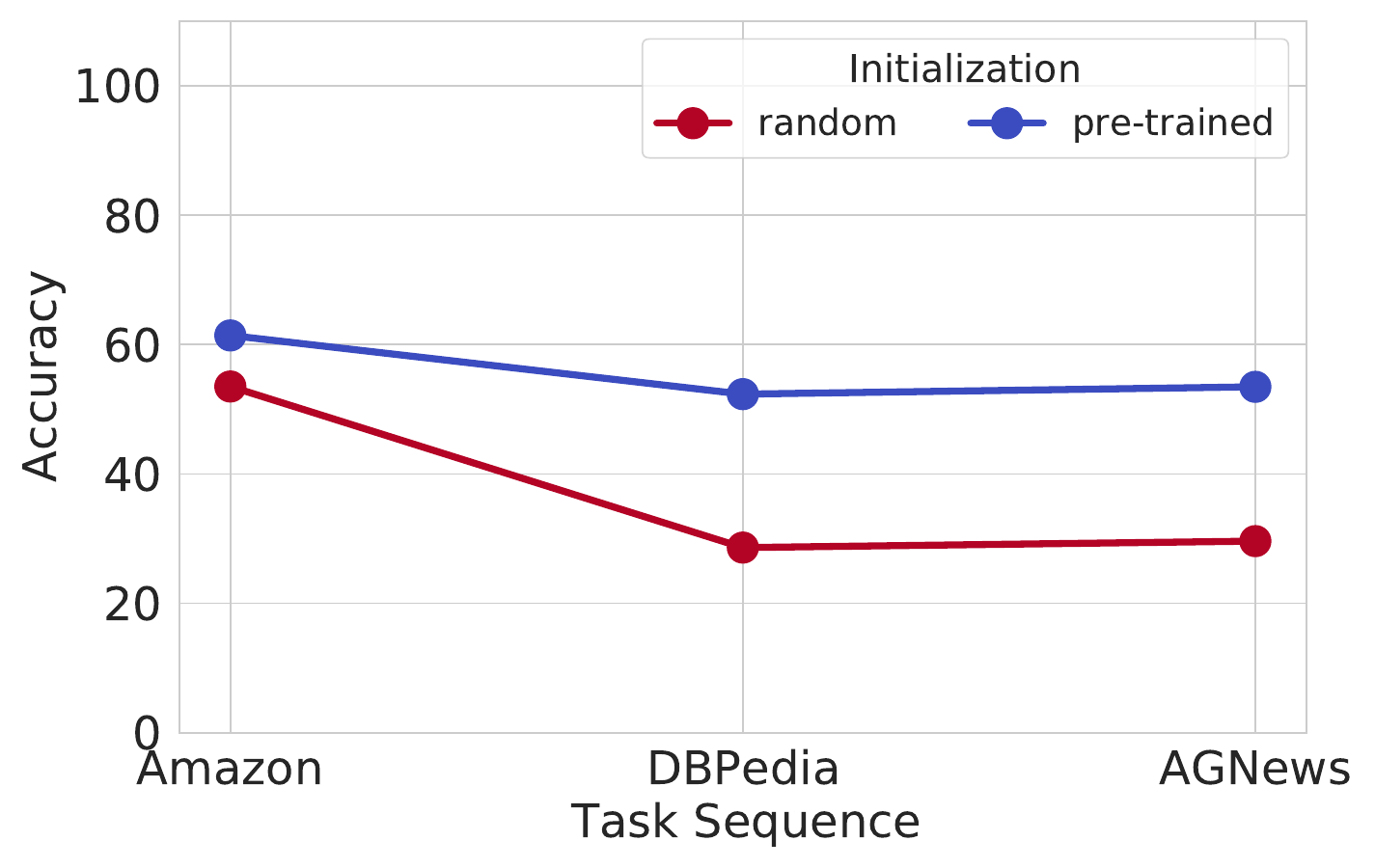}
      \caption{Amazon (Seq3)}
      \label{fig:5datanlp_seq3_task3}
    \end{subfigure}\hspace{\fill}%
    \bigskip
    \begin{subfigure}{.32\textwidth}
      \centering
      \includegraphics[width=0.8\textwidth]{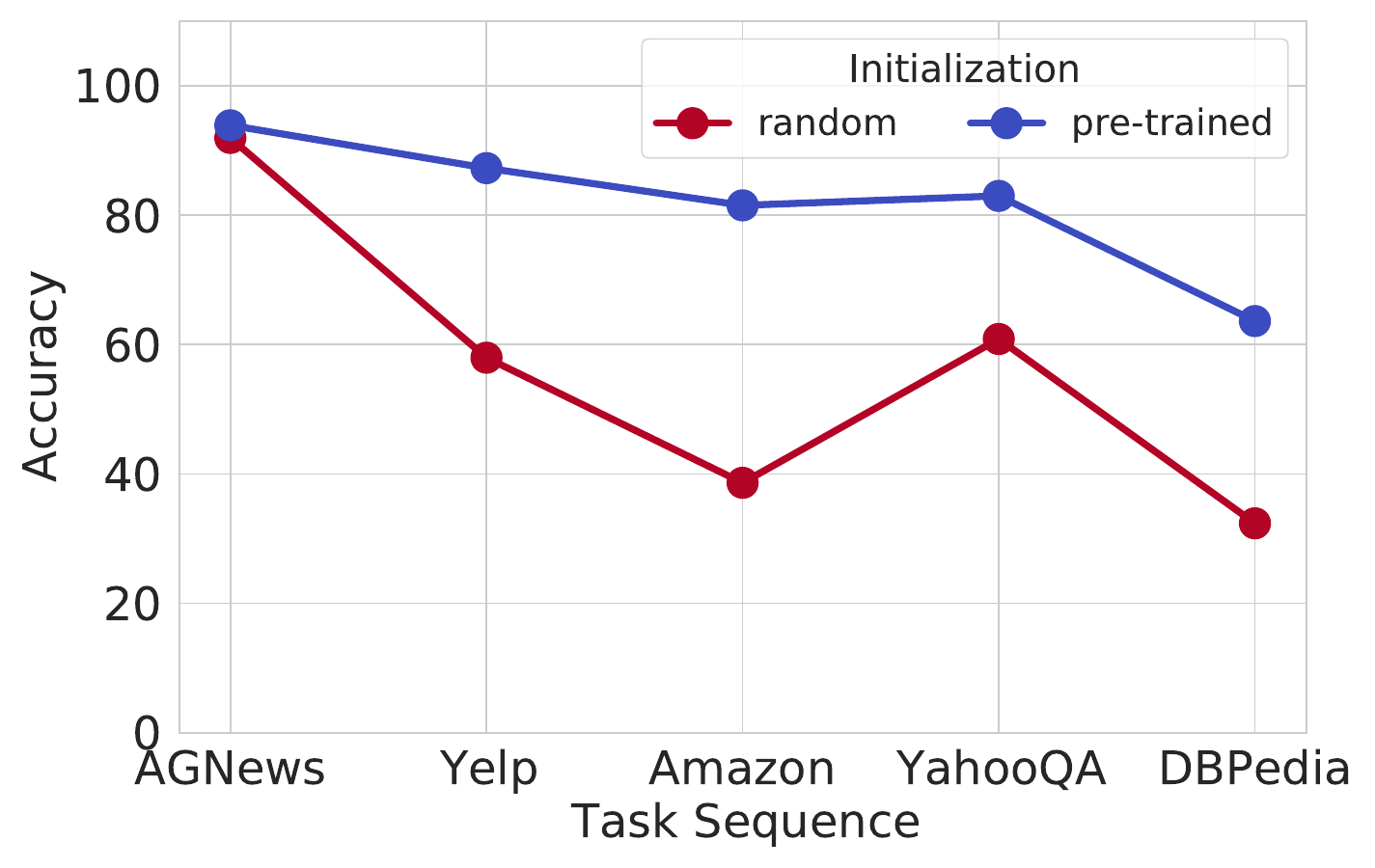}
      \caption{AGNews (Seq4)}
      \label{fig:5datanlp_seq4_task1}
    \end{subfigure}\hspace{\fill}%
    \begin{subfigure}{.32\textwidth}
      \centering
      \includegraphics[width=0.8\textwidth]{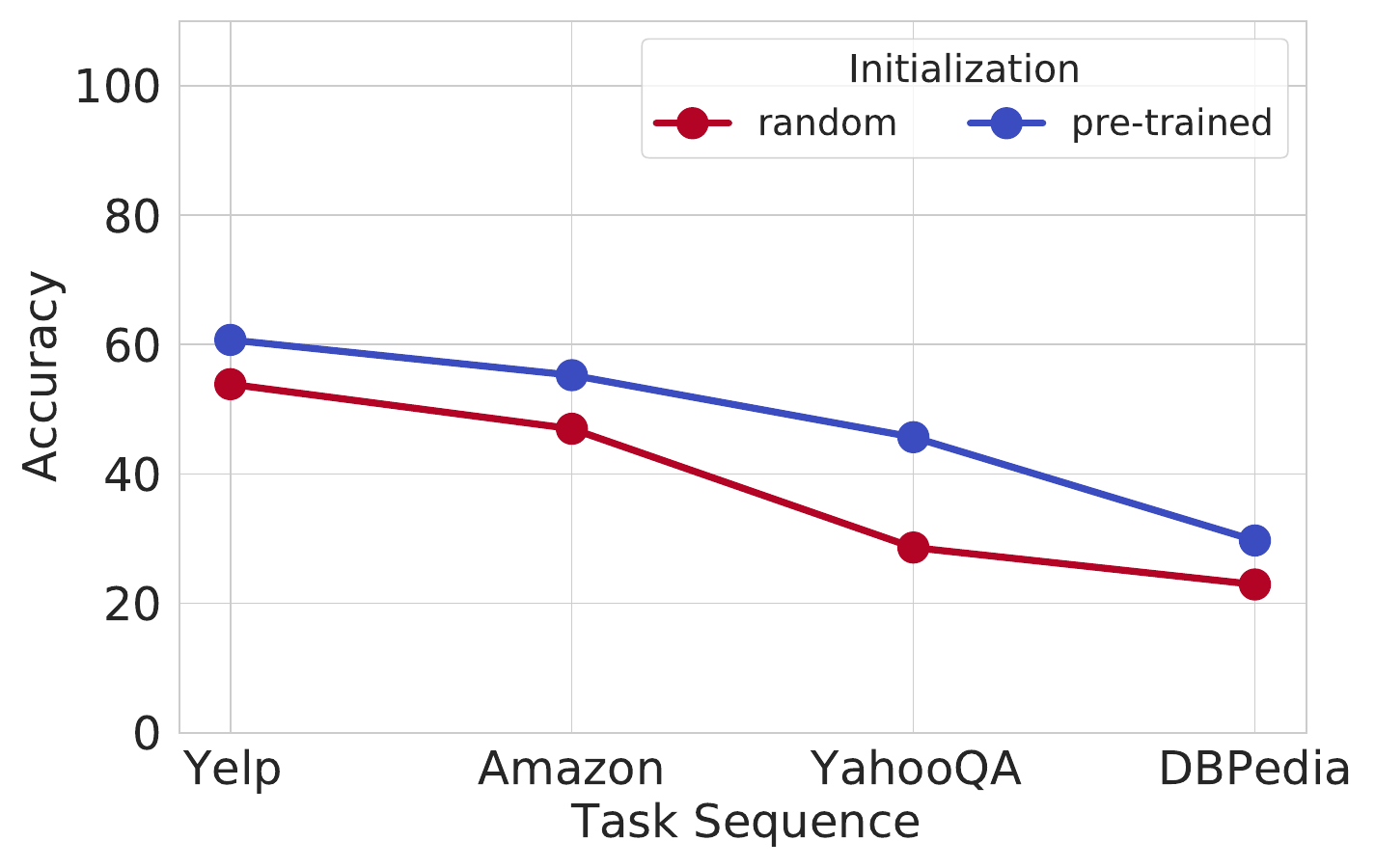}
      \caption{Yelp (Seq4)}
      \label{fig:5datanlp_seq4_task2}
    \end{subfigure}\hspace{\fill}%
    \begin{subfigure}{.32\textwidth}
      \centering
      \includegraphics[width=0.8\textwidth]{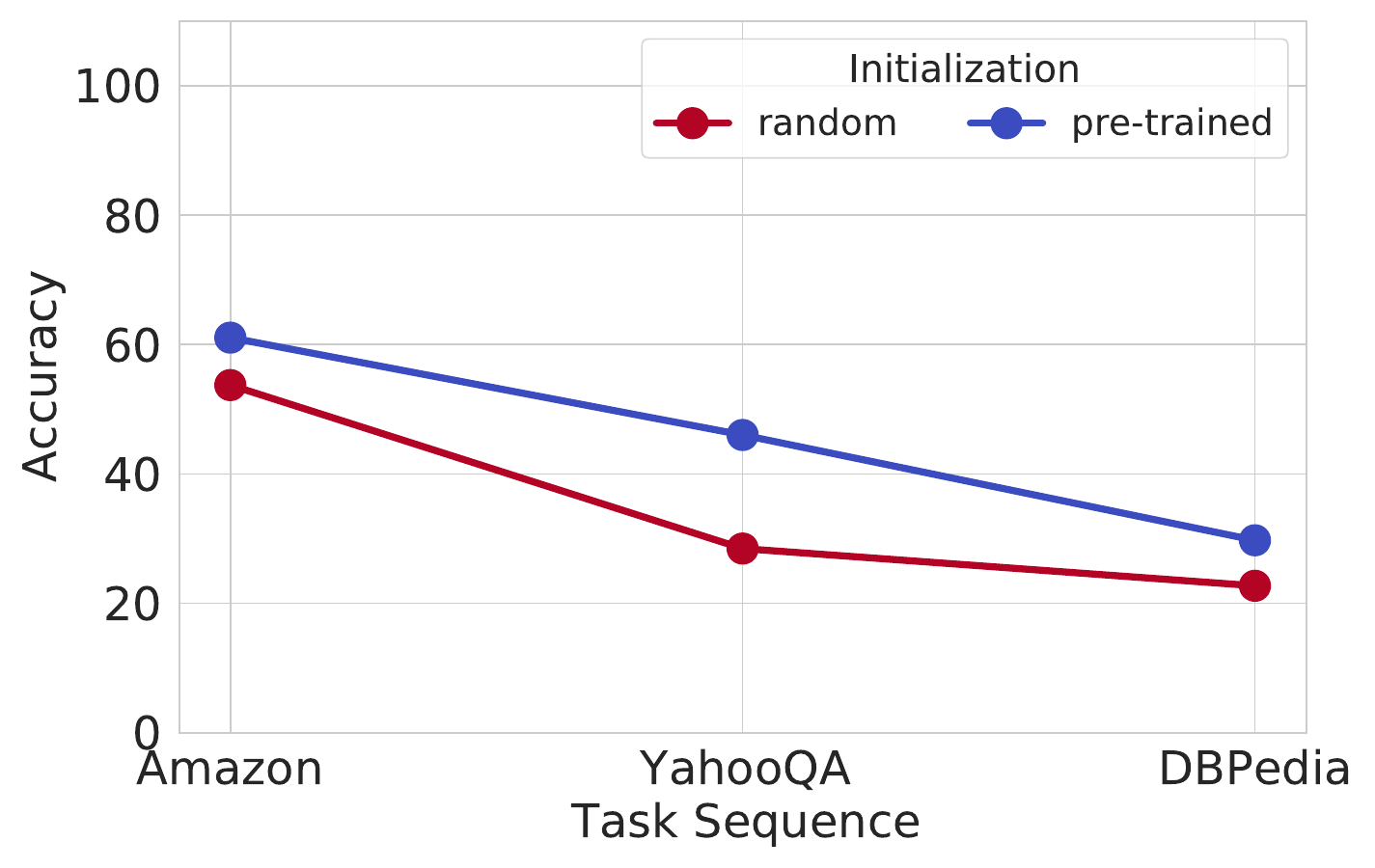}
      \caption{Amazon (Seq4)}
      \label{fig:5datanlp_seq4_task3}
    \end{subfigure}\hspace{\fill}%
    \bigskip
    \begin{subfigure}{.32\textwidth}
      \centering
      \includegraphics[width=0.8\textwidth]{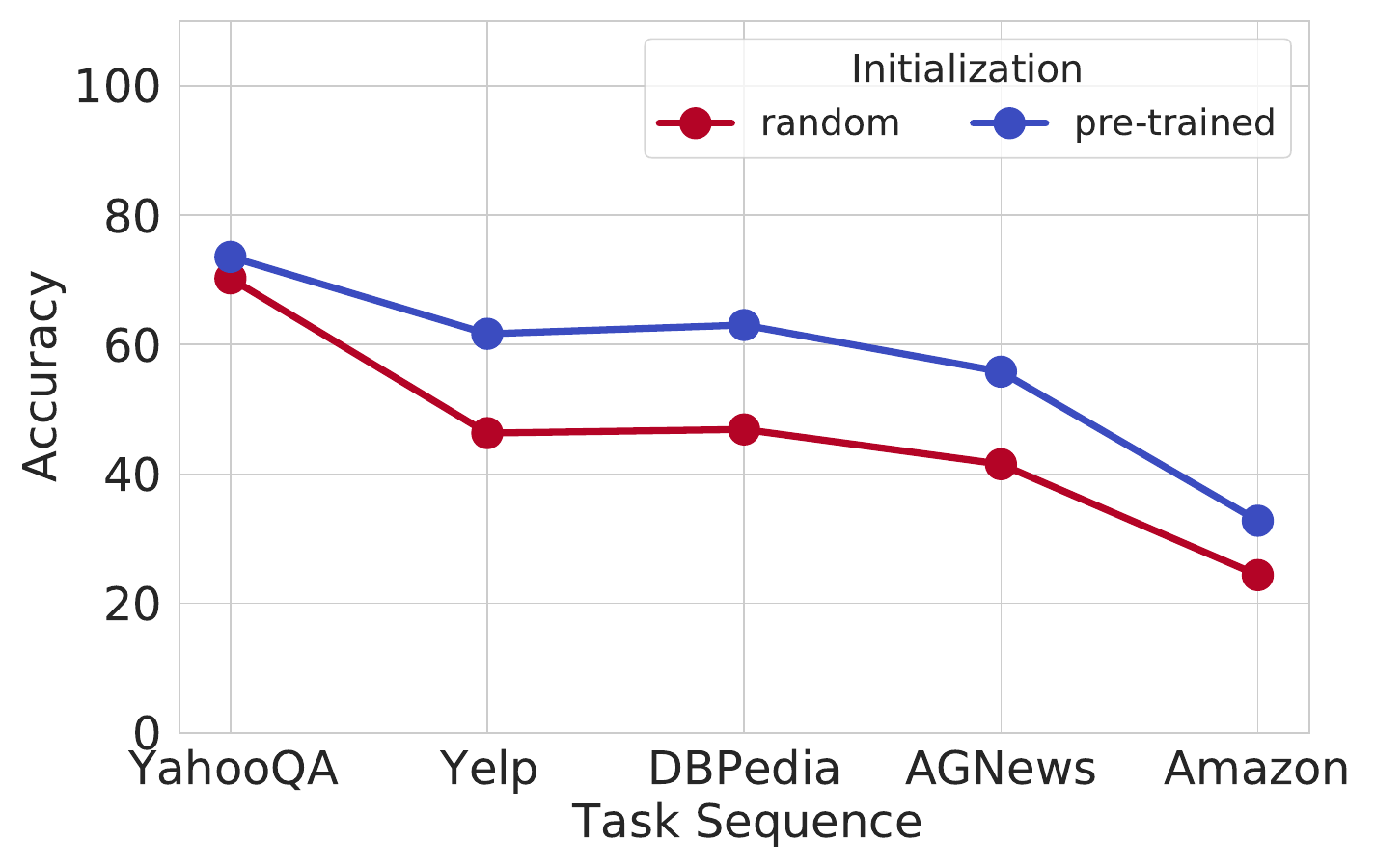}
      \caption{YahooQA (Seq5)}
      \label{fig:5datanlp_seq5_task1}
    \end{subfigure}\hspace{\fill}%
    \begin{subfigure}{.32\textwidth}
      \centering
      \includegraphics[width=0.8\textwidth]{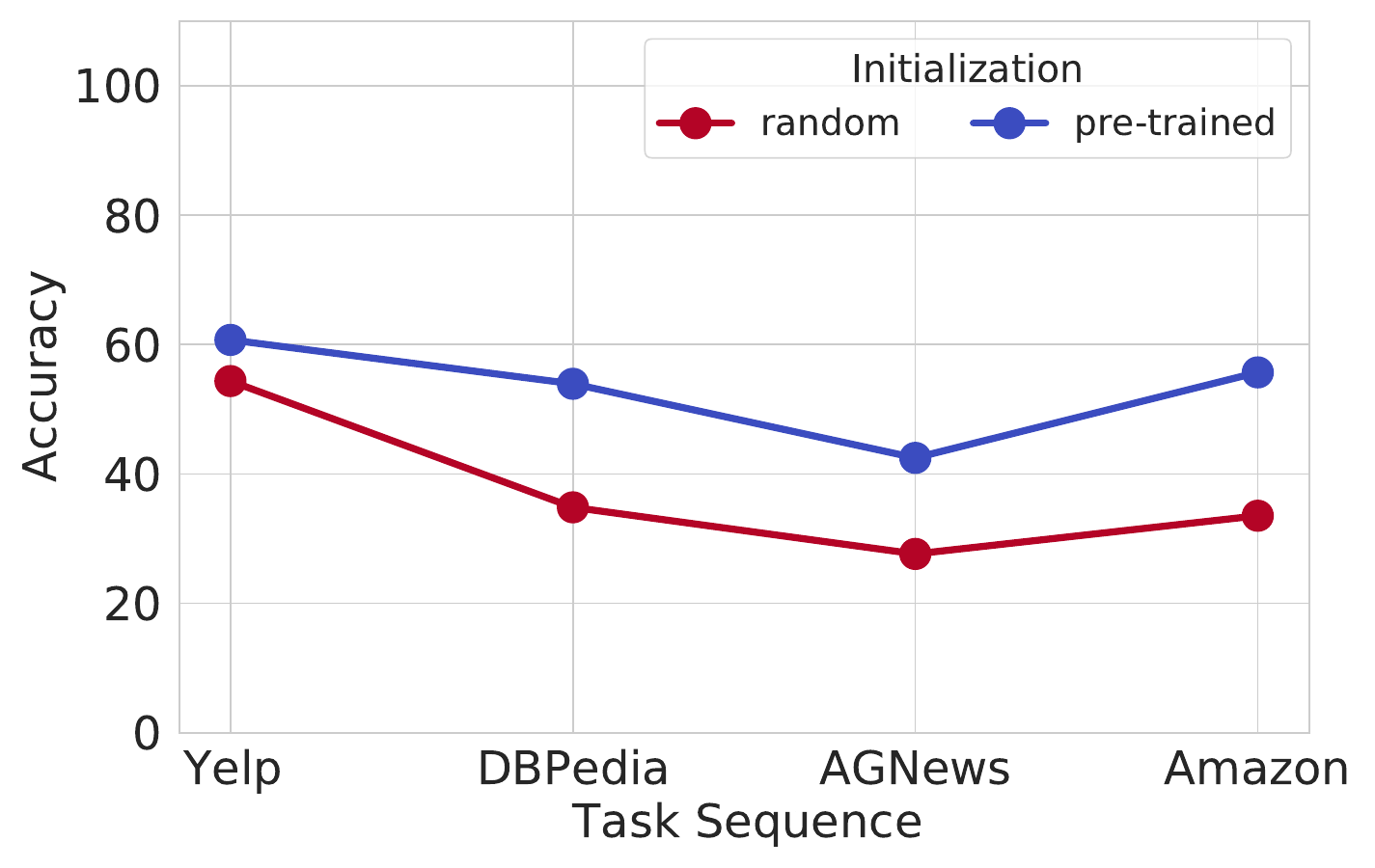}
      \caption{Yelp (Seq5)}
      \label{fig:5datanlp_seq5_task2}
    \end{subfigure}\hspace{\fill}%
    \begin{subfigure}{.32\textwidth}
      \centering
      \includegraphics[width=0.8\textwidth]{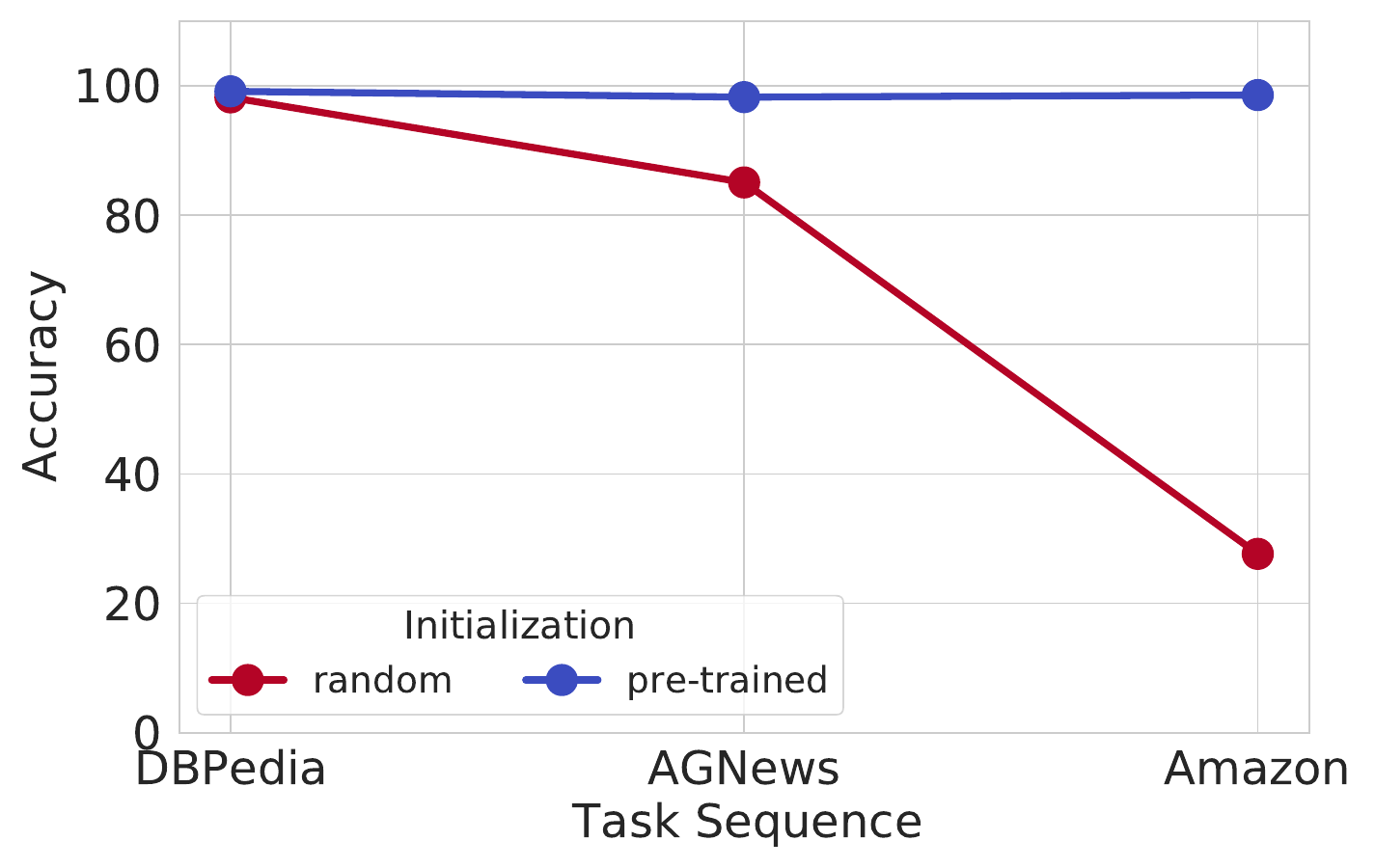}
      \caption{DBPedia (Seq5)}
      \label{fig:5datanlp_seq5_task3}
    \end{subfigure}\hspace{\fill}%
    \caption{Evolution of task accuracy during sequential training on 5-dataset-NLP. We compare the performance of pre-trained and randomly initialized models, for the first three tasks in a sequence, across five different random task orderings (Seq1, Seq2, Seq3, Seq4, Seq5). We see that both models start with approximately equal task accuracy, but pre-trained initialized models undergo lesser forgetting than randomly initialized models.}
    \label{fig:taskwiseplots_5datanlp}
\end{figure}

\begin{figure}%[ht]
    \centering
    \bigskip
    \begin{subfigure}{.32\textwidth}
      \centering
      \includegraphics[width=0.8\textwidth]{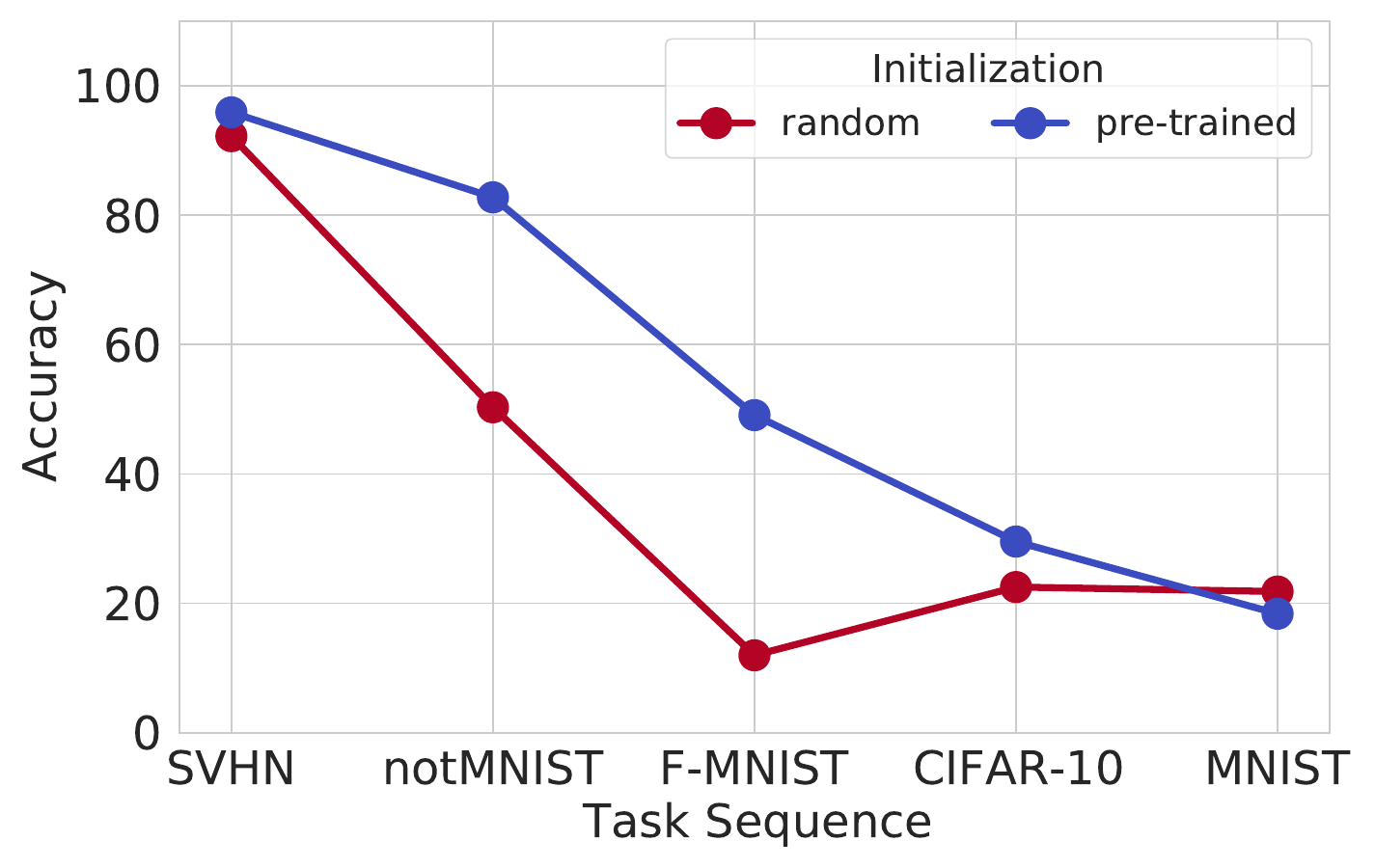}
      \caption{SVHN (Seq1)}
      \label{fig:5data_seq1_task1}
    \end{subfigure}\hspace{\fill}%
    \begin{subfigure}{.32\textwidth}
      \centering
      \includegraphics[width=0.8\textwidth]{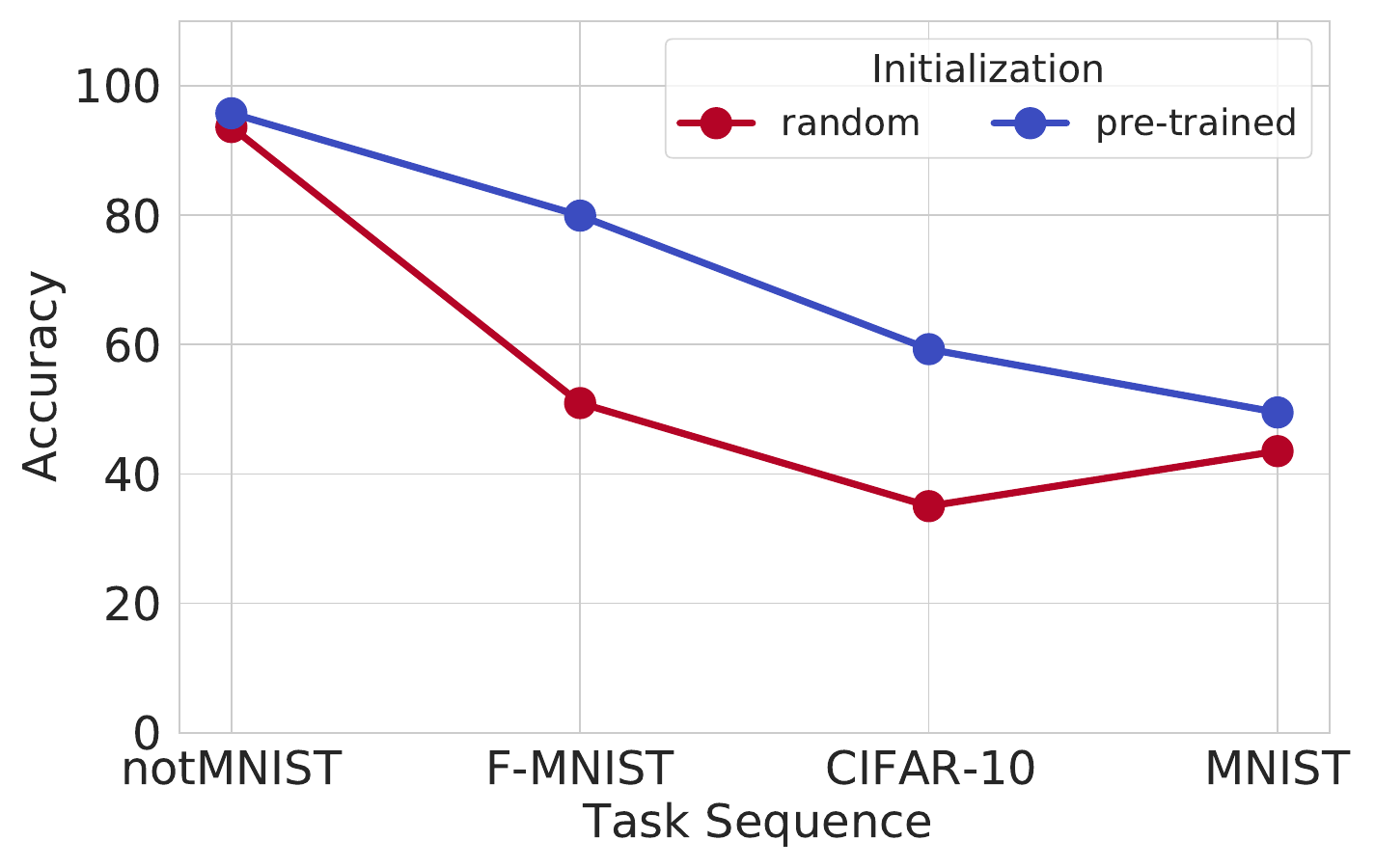}
      \caption{notMNIST (Seq1)}
      \label{fig:5data_seq1_task2}
    \end{subfigure}\hspace{\fill}%
    \begin{subfigure}{.32\textwidth}
      \centering
      \includegraphics[width=0.8\textwidth]{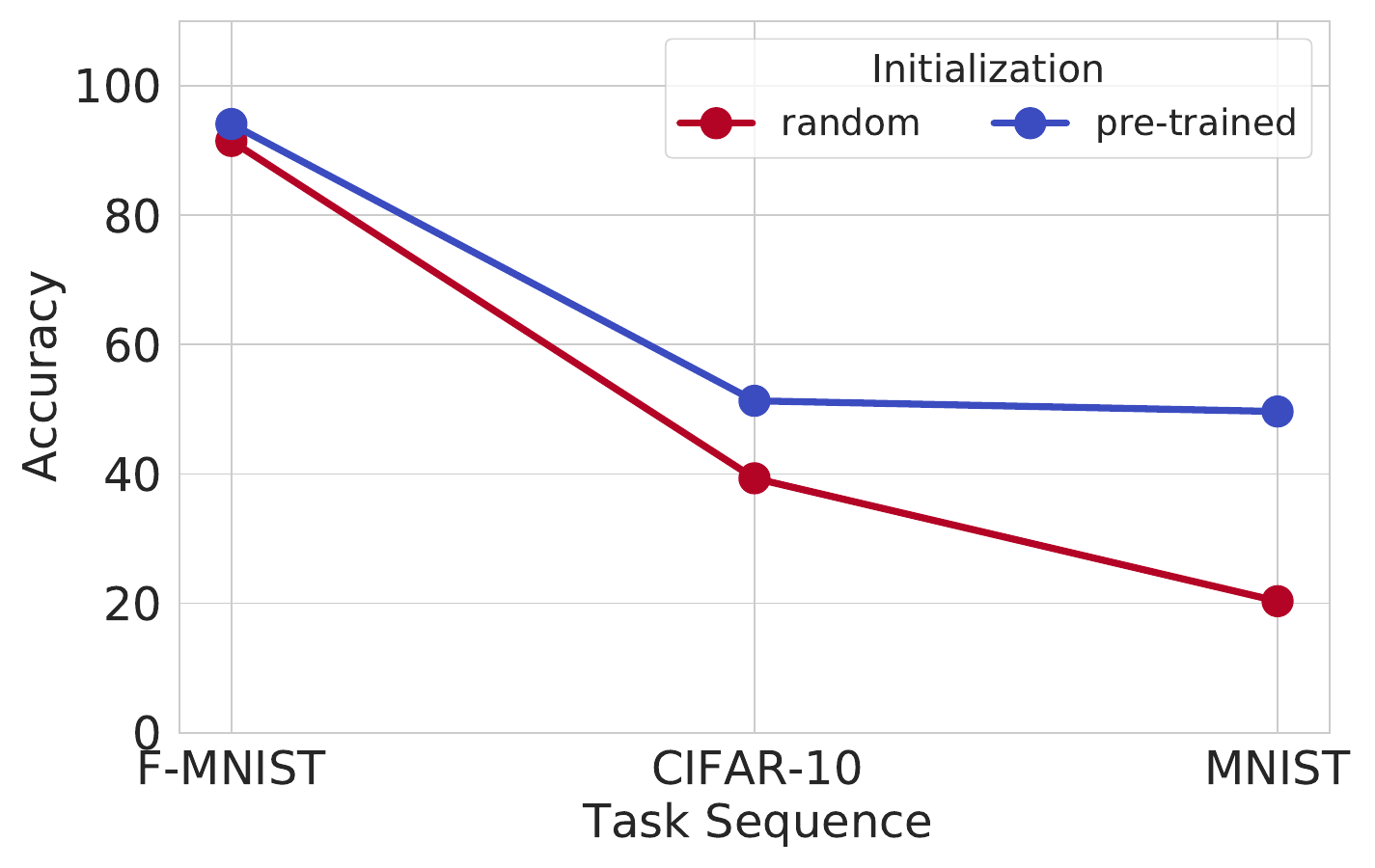}
      \caption{Fashion-MNIST (Seq1)}
      \label{fig:5data_seq1_task3}
    \end{subfigure}
    \bigskip
    \begin{subfigure}{.32\textwidth}
      \centering
      \includegraphics[width=0.8\textwidth]{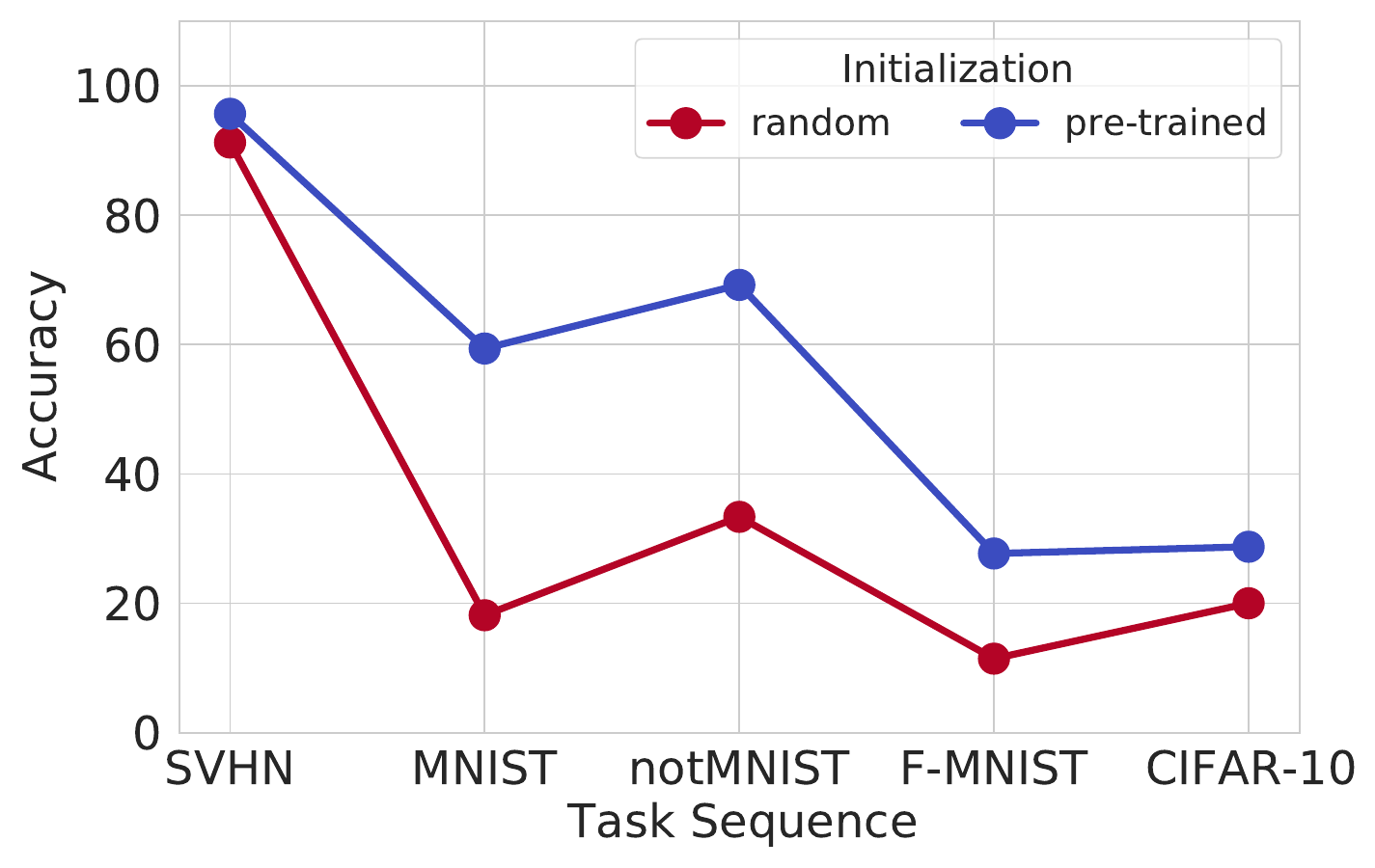}
      \caption{SVHN (Seq2)}
      \label{fig:5data_seq2_task1}
    \end{subfigure}\hspace{\fill}%
    \begin{subfigure}{.32\textwidth}
      \centering
      \includegraphics[width=0.8\textwidth]{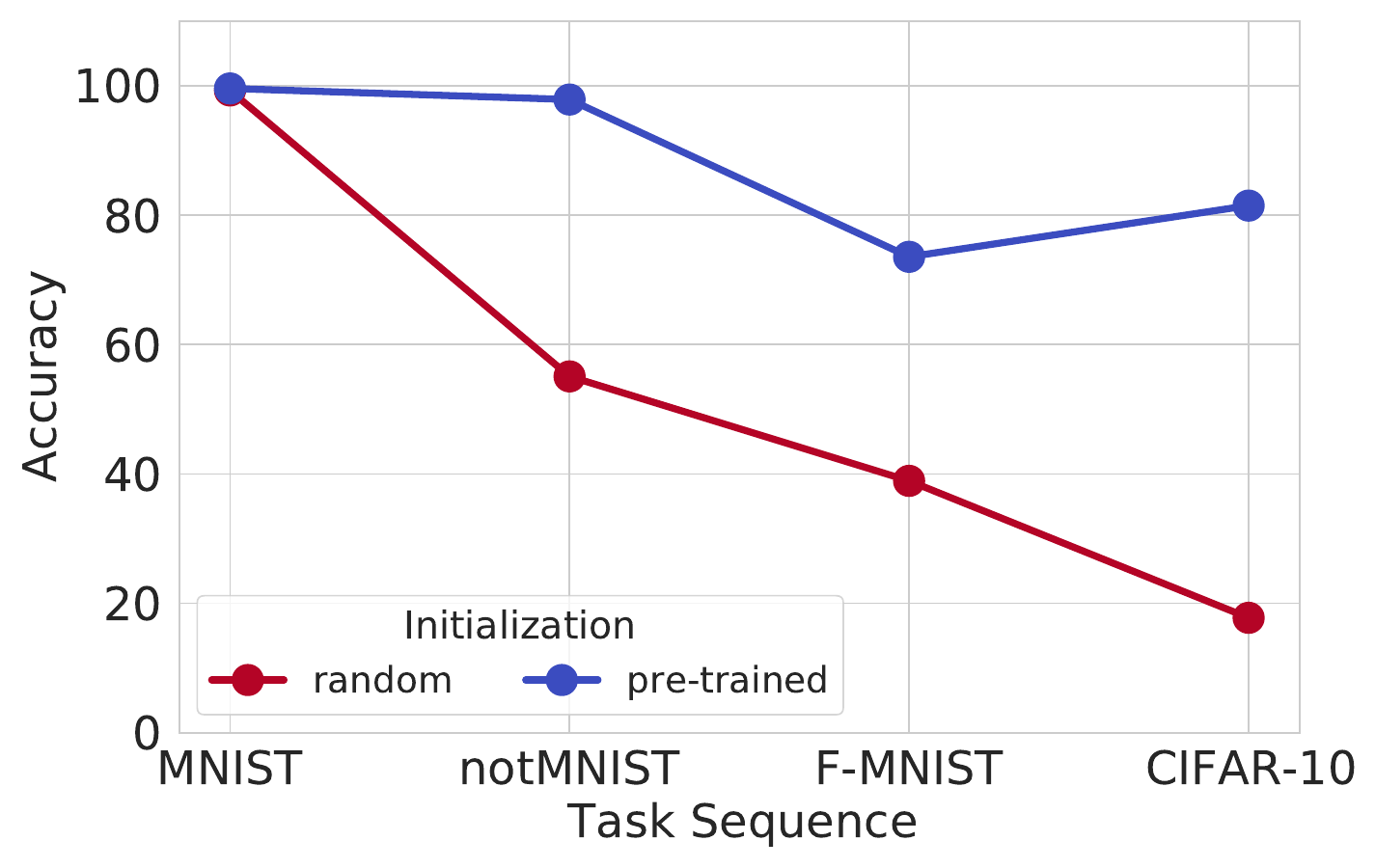}
      \caption{MNIST (Seq2)}
      \label{fig:5data_seq2_task2}
    \end{subfigure}\hspace{\fill}%
    \begin{subfigure}{.32\textwidth}
      \centering
      \includegraphics[width=0.8\textwidth]{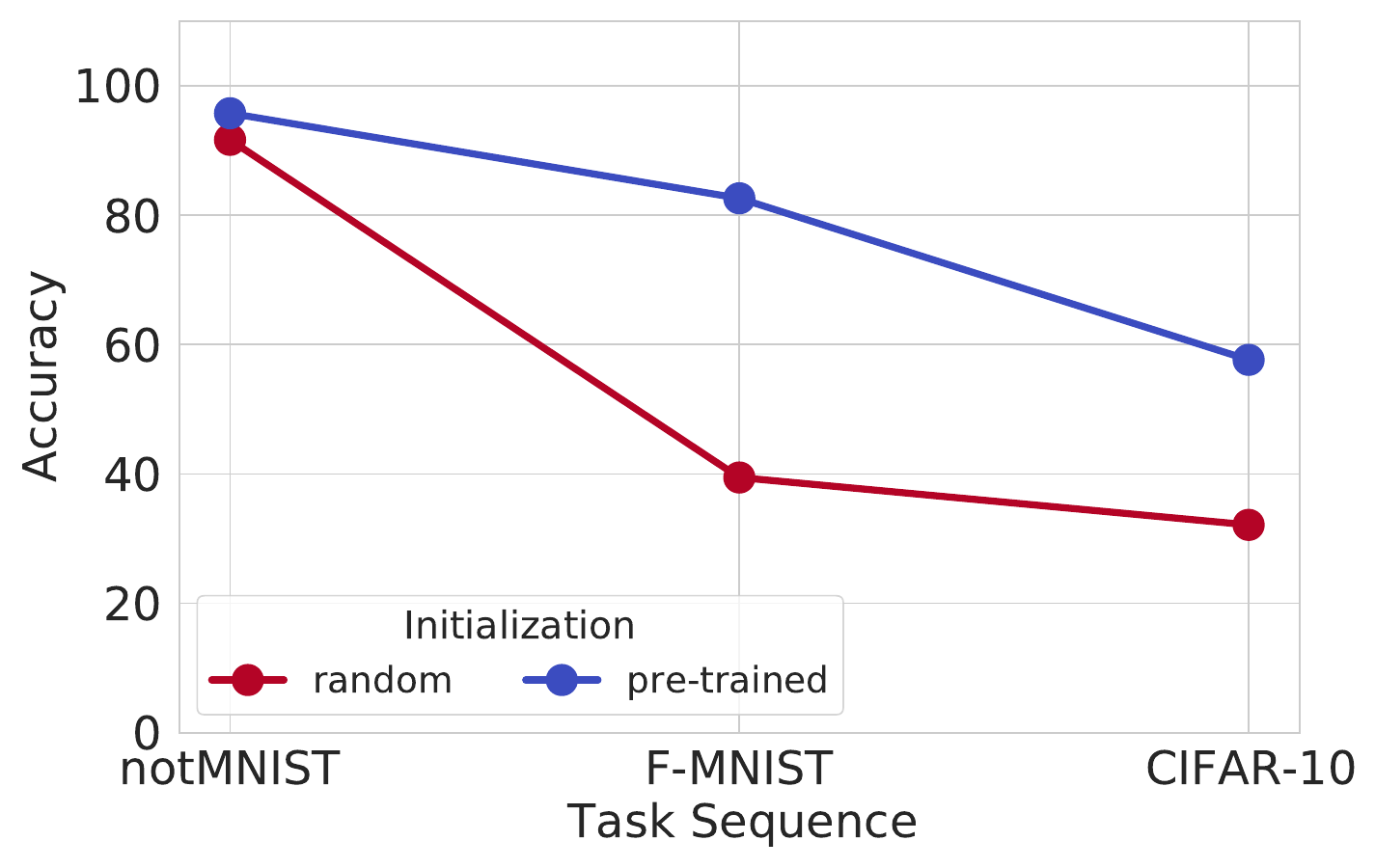}
      \caption{notMNIST (Seq2)}
      \label{fig:5data_seq2_task3}
    \end{subfigure}\hspace{\fill}%
    \bigskip
    \begin{subfigure}{.32\textwidth}
      \centering
      \includegraphics[width=0.8\textwidth]{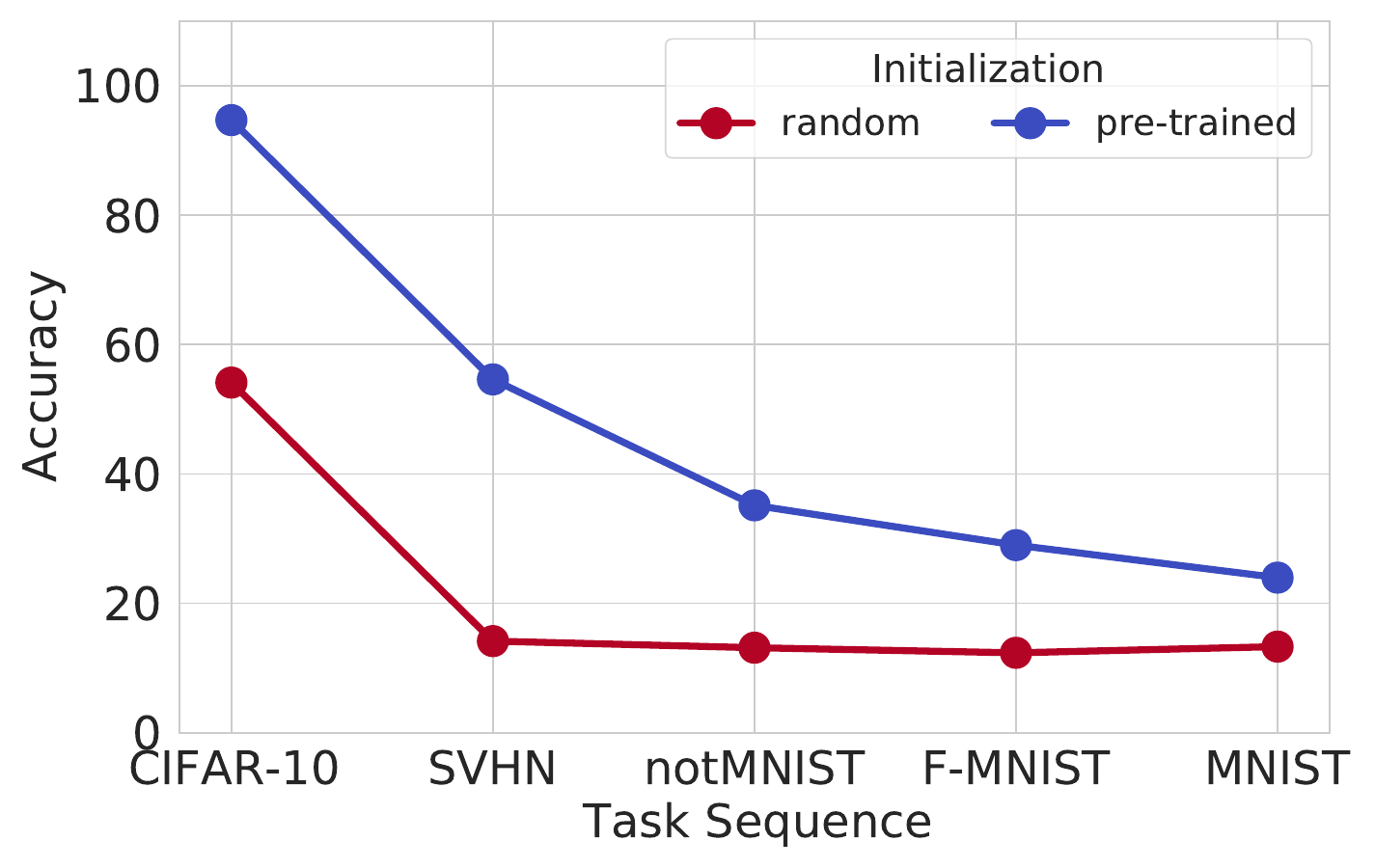}
      \caption{CIFAR-10 (Seq3)}
      \label{fig:5data_seq3_task1}
    \end{subfigure}\hspace{\fill}%
    \begin{subfigure}{.32\textwidth}
      \centering
      \includegraphics[width=0.8\textwidth]{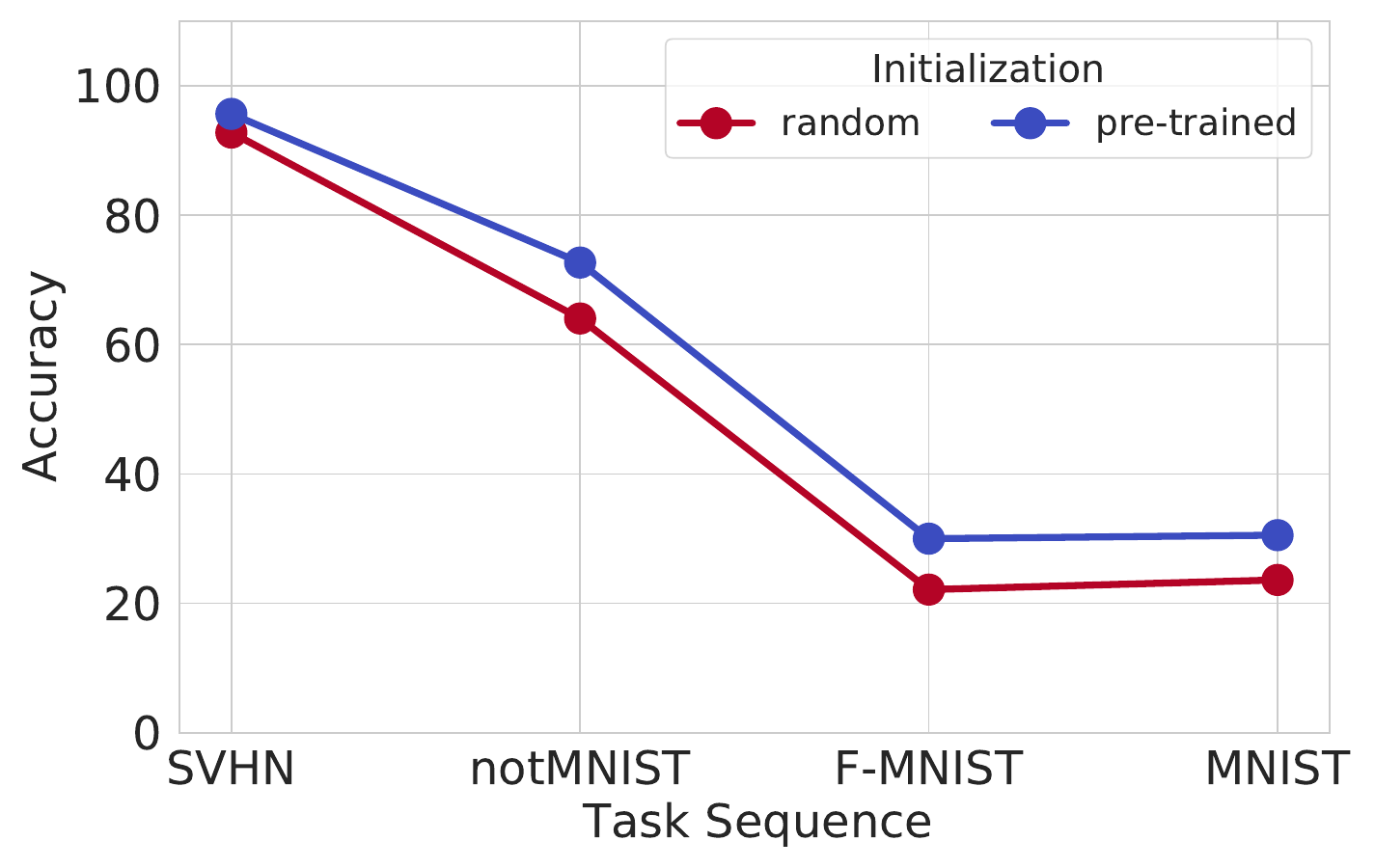}
      \caption{SVHN (Seq3)}
      \label{fig:5data_seq3_task2}
    \end{subfigure}\hspace{\fill}%
    \begin{subfigure}{.32\textwidth}
      \centering
      \includegraphics[width=0.8\textwidth]{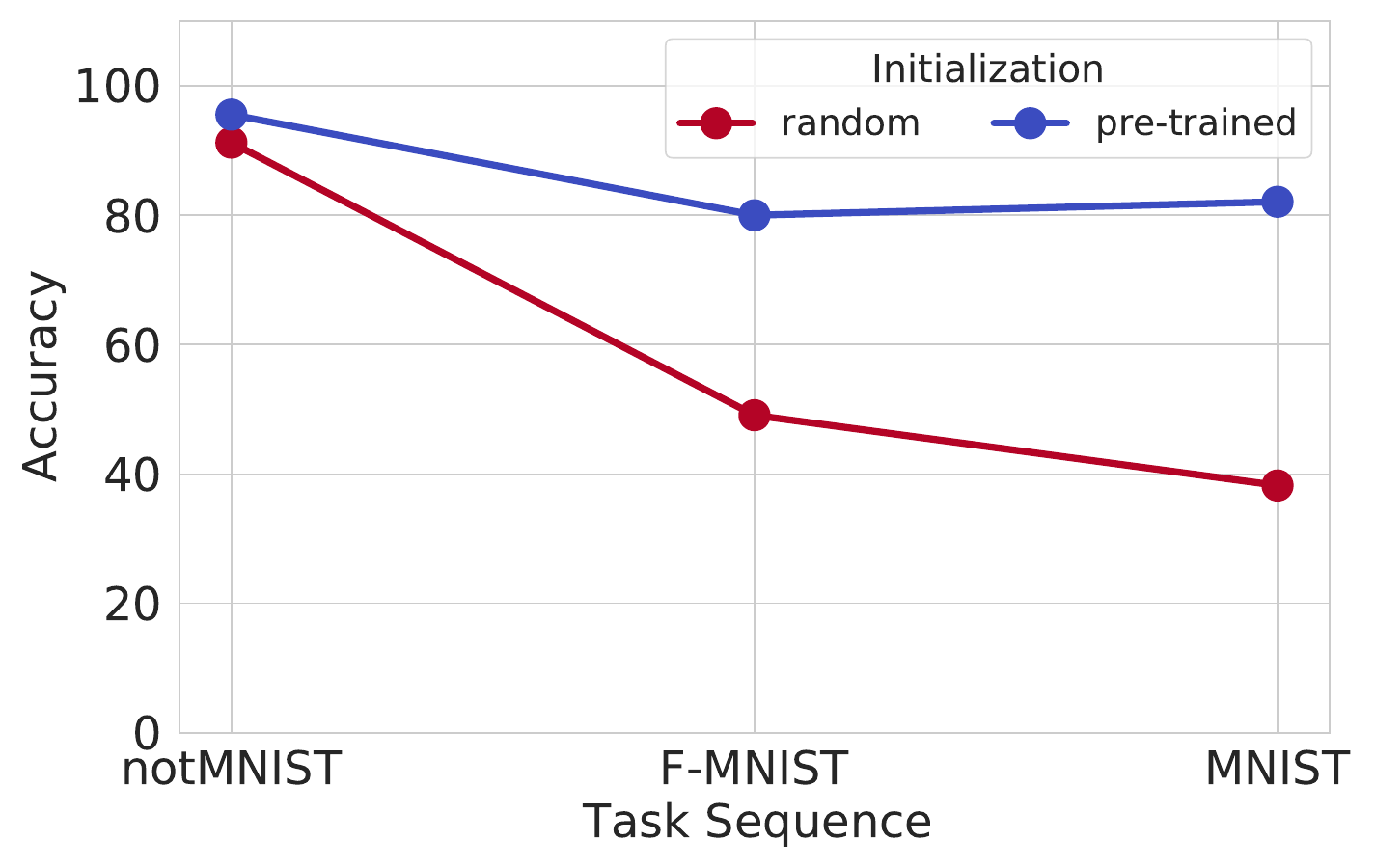}
      \caption{notMNIST (Seq3)}
      \label{fig:5data_seq3_task3}
    \end{subfigure}\hspace{\fill}%
    \bigskip
    \begin{subfigure}{.32\textwidth}
      \centering
      \includegraphics[width=0.8\textwidth]{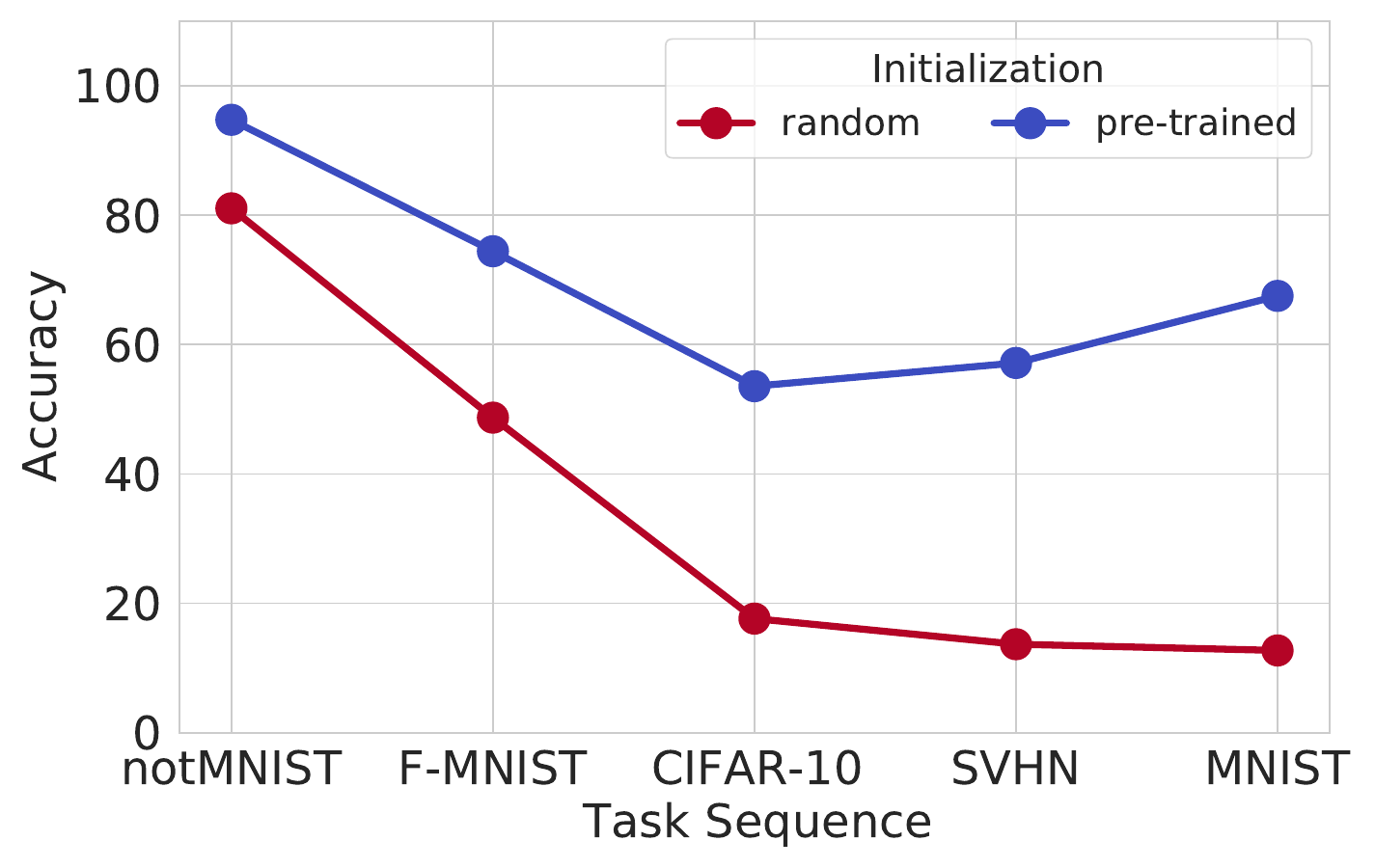}
      \caption{notMNIST (Seq4)}
      \label{fig:5data_seq4_task1}
    \end{subfigure}\hspace{\fill}%
    \begin{subfigure}{.32\textwidth}
      \centering
      \includegraphics[width=0.8\textwidth]{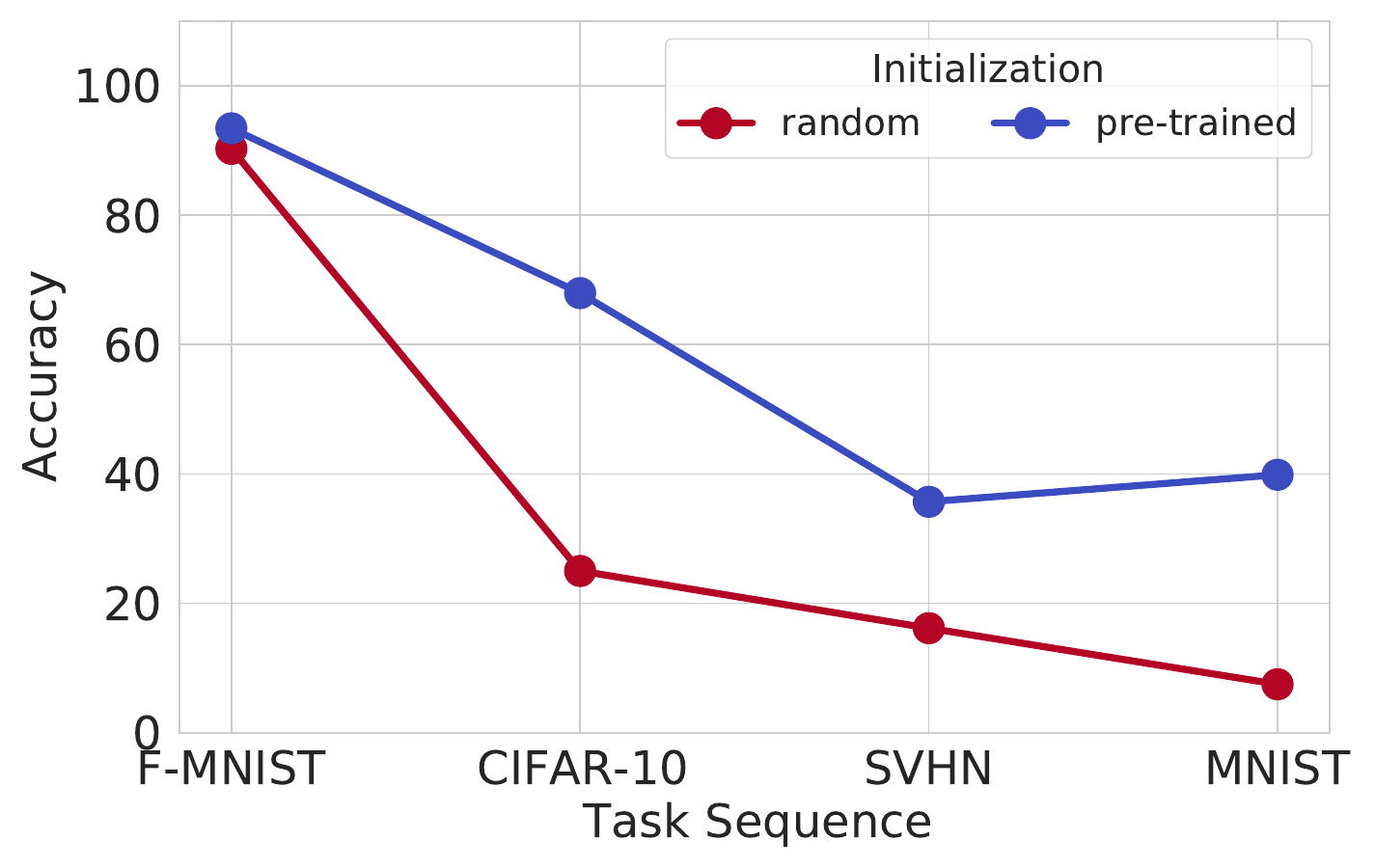}
      \caption{Fashion-MNIST (Seq4)}
      \label{fig:5data_seq4_task2}
    \end{subfigure}\hspace{\fill}%
    \begin{subfigure}{.32\textwidth}
      \centering
      \includegraphics[width=0.8\textwidth]{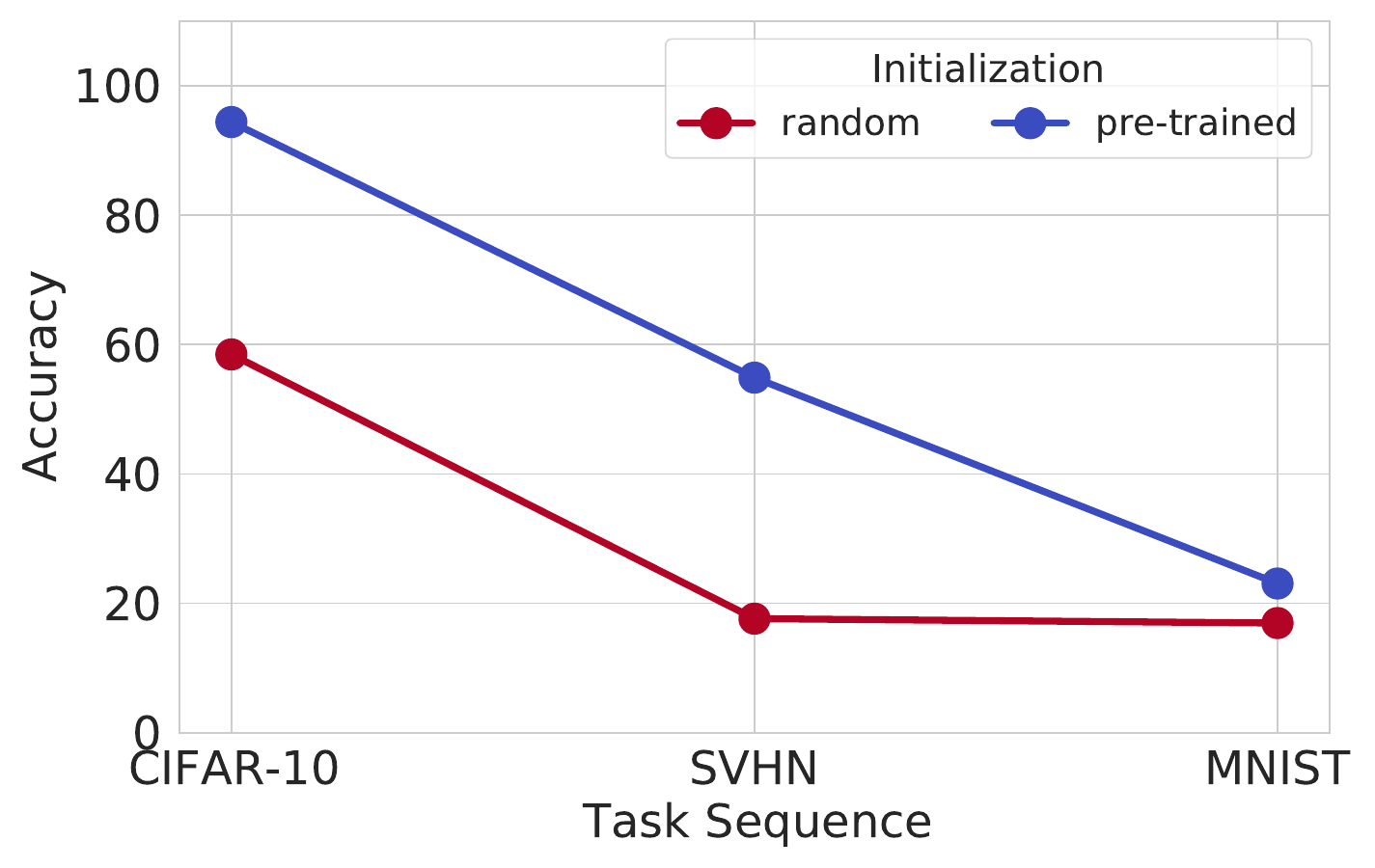}
      \caption{CIFAR-10 (Seq4)}
      \label{fig:5data_seq4_task3}
    \end{subfigure}\hspace{\fill}%
    \bigskip
    \begin{subfigure}{.32\textwidth}
      \centering
      \includegraphics[width=0.8\textwidth]{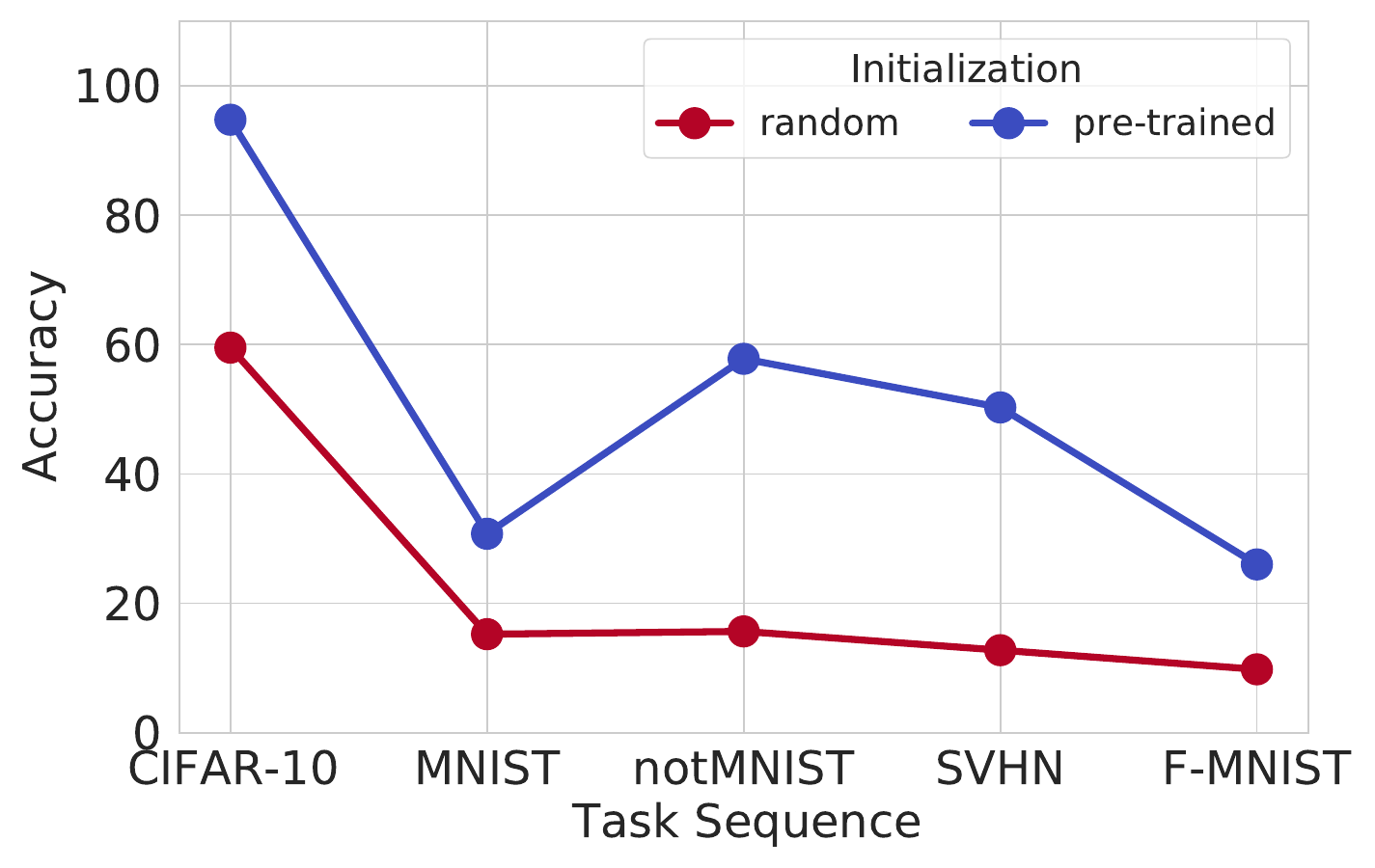}
      \caption{CIFAR-10 (Seq5)}
      \label{fig:5data_seq5_task1}
    \end{subfigure}\hspace{\fill}%
    \begin{subfigure}{.32\textwidth}
      \centering
      \includegraphics[width=0.8\textwidth]{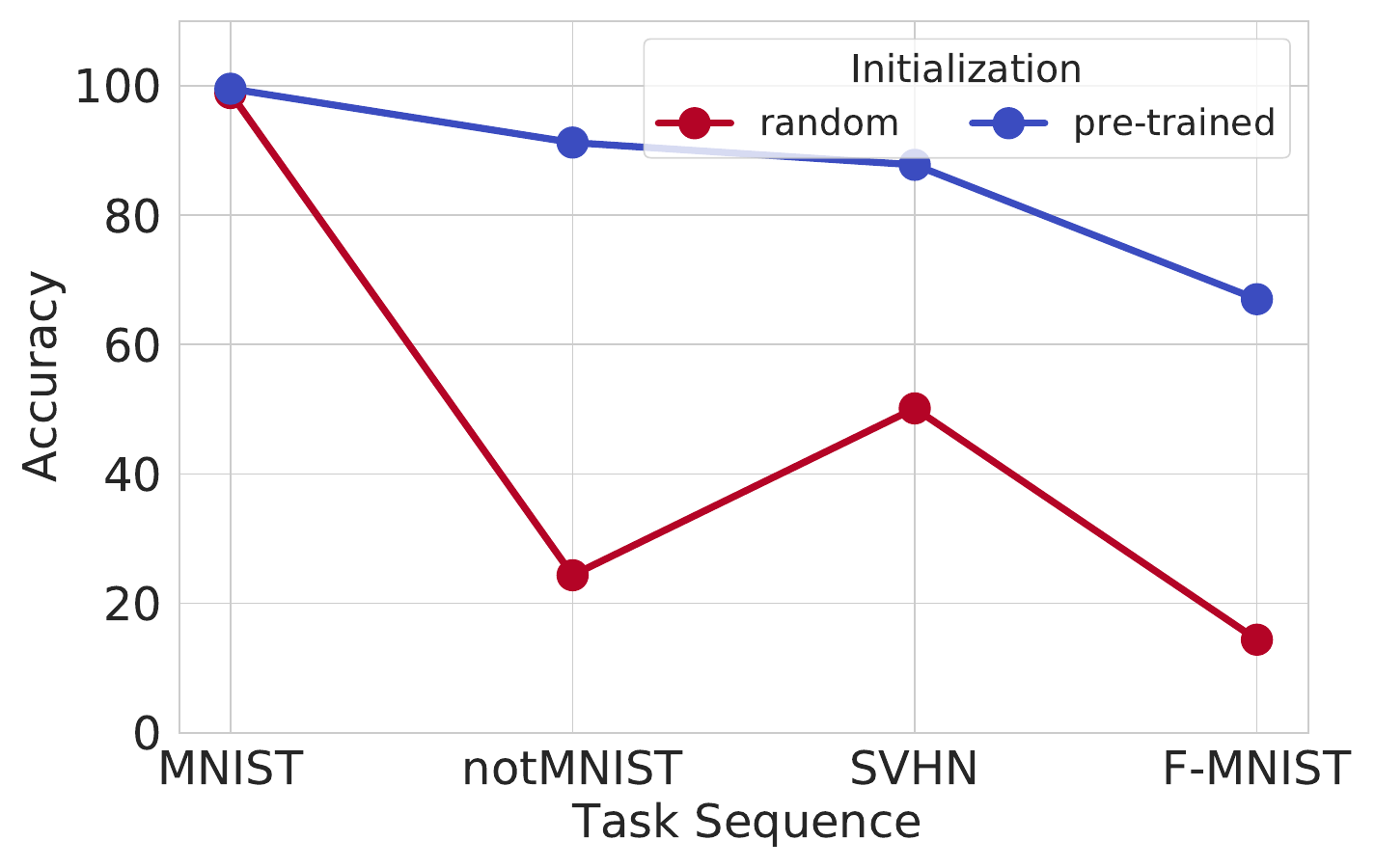}
      \caption{MNIST (Seq5)}
      \label{fig:5data_seq5_task2}
    \end{subfigure}\hspace{\fill}%
    \begin{subfigure}{.32\textwidth}
      \centering
      \includegraphics[width=0.8\textwidth]{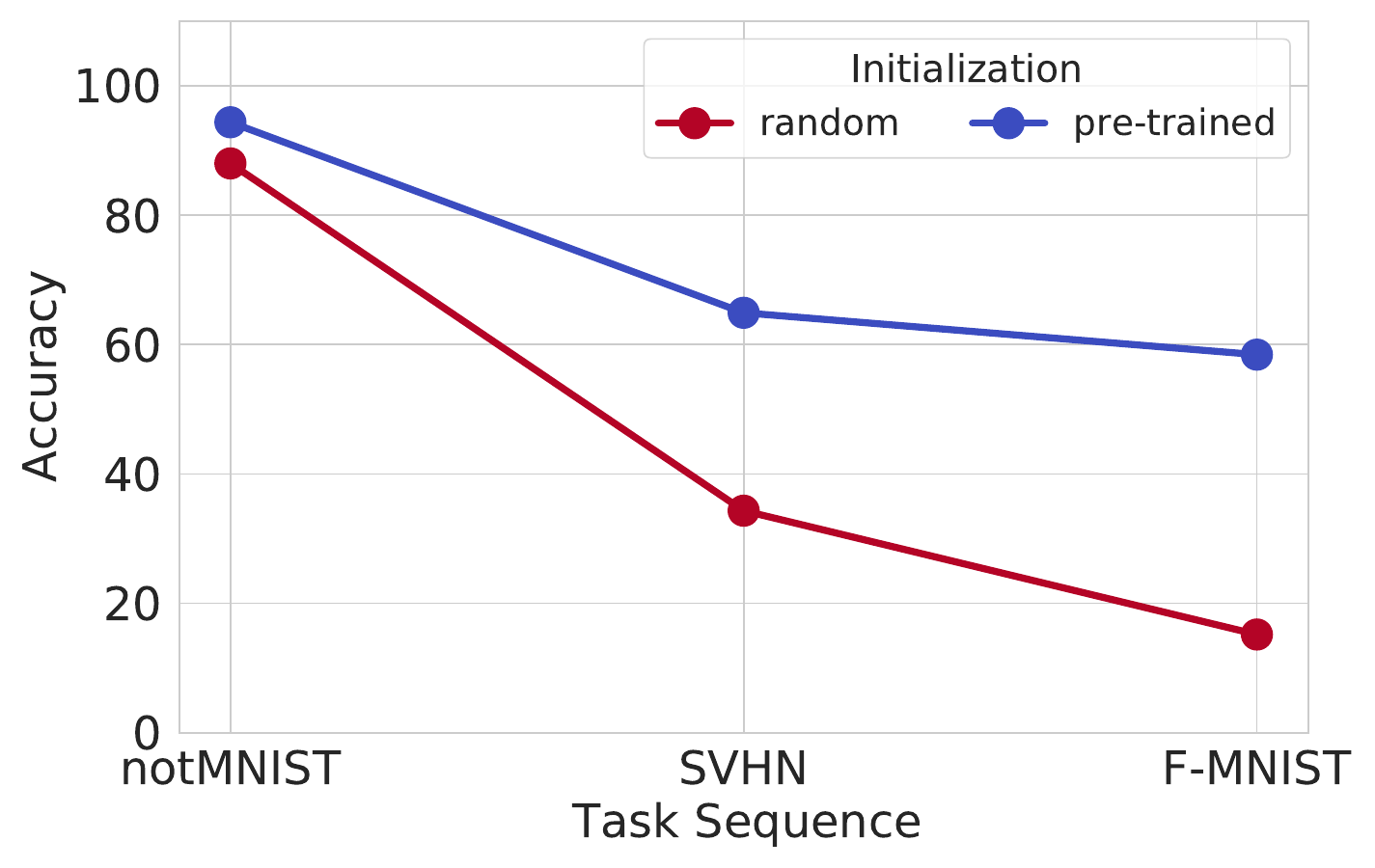}
      \caption{notMNIST (Seq5)}
      \label{fig:5data_seq5_task3}
    \end{subfigure}\hspace{\fill}%
    \caption{Evolution of task accuracy during sequential training on 5-dataset-CV. We compare the performance of pre-trained and randomly initialized models, for the first three tasks in a sequence, across five different random task orderings (Seq1, Seq2, Seq3, Seq4, Seq5). We see that both models start with approximately equal task accuracy (except for CIFAR-10), but pre-trained initialized models undergo lesser forgetting than randomly initialized models.}
    \label{fig:taskwiseplots_5data}
\end{figure}

In order to understand the evolution of task-specific performance during continuous training, we visualize the task-specific results in Figures~\ref{fig:taskwiseplots_5datanlp} and \ref{fig:taskwiseplots_5data}. Specifically, we compare the performance of pre-trained and randomly initialized ResNet-18/ DistilBERT, for the first three tasks in a sequence, across five random task ordering, when evaluated on 5-dataset-CV/5-dataset-NLP (diverse tasks). In general, we see that both models start with approximately equal task accuracy (except for CIFAR-10), but pre-trained initialization leads to lesser forgetting than randomly initialized models (consistent with our observation in Figure \ref{fig:motivation} for Split YahooQA). Moreover, given the heterogeneous nature of the downstream tasks, we see that performance gains (in terms of forgetting) from pre-trained initialization vary across different tasks. 

\subsection{5-dataset-NLP} 
For example, in the case of DBPedia (Figures~\ref{fig:5datanlp_seq1_task3}, \ref{fig:5datanlp_seq2_task1}, \ref{fig:5datanlp_seq5_task3}) and AGNews (Figures~\ref{fig:5datanlp_seq1_task2}, \ref{fig:5datanlp_seq2_task3}, \ref{fig:5datanlp_seq4_task1}) data sets, we see pre-trained DistilBERT undergoes little to almost no forgetting. One plausible explanation for these results is that both data sets are for the article classification tasks, DBPedia is Wikipedia article classification (14 classes) and AGNews is news article classification (4 classes), and share similar domains with the pre-training corpora (Wikipedia and Books). On the other hand, we see a significant forgetting in the case of Yelp (Figures~\ref{fig:5datanlp_seq1_task1}, \ref{fig:5datanlp_seq3_task1}, \ref{fig:5datanlp_seq4_task2}) and Amazon data sets (Figure~\ref{fig:5datanlp_seq4_task3}). Both of these data sets review sentiment classification tasks (5 classes). We know that the reviews domain (noisy text from Yelp.com and Amazon.com) is less similar to the pre-training corpora (clean text from Wikipedia and Books), and might be one of the reasons behind the drop in performance. Further, note that as we train on the sequence of tasks, we expect to see positive/ negative transfers from related/ unrelated tasks. For example, we see that the performance on Yelp improves significantly after training on Amazon (Figures~\ref{fig:5datanlp_seq1_task1}, \ref{fig:5datanlp_seq3_task1}, \ref{fig:5datanlp_seq5_task2}), demonstrating an example of positive transfer from the related task. 

\subsection{5-dataset-CV} Here, we report that the forgetting is more severe for SVHN (Figures~\ref{fig:5data_seq1_task1}, \ref{fig:5data_seq2_task1}, \ref{fig:5data_seq3_task2}) and CIFAR-10 (Figures~\ref{fig:5data_seq3_task1}, \ref{fig:5data_seq4_task3}, \ref{fig:5data_seq5_task1}) as compared to MNIST (Figures~\ref{fig:5data_seq2_task2}, \ref{fig:5data_seq5_task2}), notMNIST (Figures~\ref{fig:5data_seq1_task2}, \ref{fig:5data_seq2_task3}, \ref{fig:5data_seq3_task3}, \ref{fig:5data_seq5_task3}). Although SVHN and MNIST both are digit recognition tasks, we believe that the realistic nature (house numbers in Google Street View images) of SVHN images makes them more susceptible to forgetting, even in the case of pre-trained ResNet-18 models.

\section{Loss Landscape}
\label{sec:loss_landscape_appendix}

\subsection{Loss Contours}
In this section, we present loss contours for task 1/ task 2 for all task sequences (refer to Section \ref{sec:task_sequences} for task sequences) for 5-dataset-NLP, Split YahooQA, Split CIFAR-50, and 5-dataset-CV. 
% For \textbf{Split CIFAR-50} and \textbf{5-dataset}, we also present the loss contours for 5-epoch training. 
In line with our observation from the sharpness and linear model interpolation analyses, pre-trained initialized models lead to flatter task minima for subsequent tasks as well.

% \subsubsection{Loss Contours: 5-dataset-NLP} 
\begin{figure}[h]
    \centering
    \begin{subfigure}{.19\textwidth}
      \centering
      \includegraphics[width=0.9\textwidth]{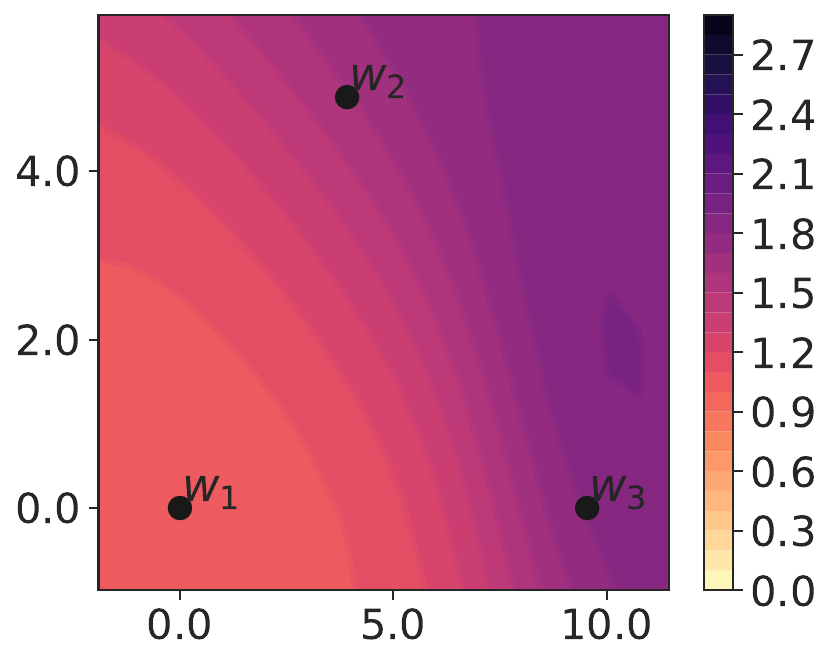}
      \caption{Seq 1 (R)}
      \label{fig:5datanlp_seq1_no_pt_contour_w1}
    \end{subfigure}\hspace{\fill}%
    \begin{subfigure}{.19\textwidth}
      \centering
      \includegraphics[width=0.9\textwidth]{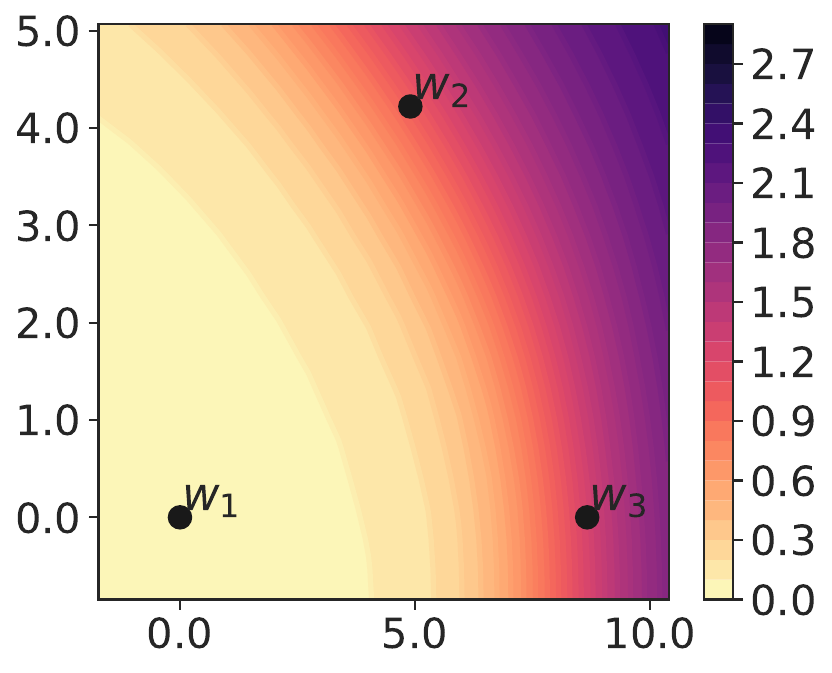}
      \caption{Seq 2 (R)}
      \label{fig:5datanlp_seq2_no_pt_contour_w1}
    \end{subfigure}\hspace{\fill}%
    \begin{subfigure}{.19\textwidth}
      \centering
      \includegraphics[width=0.9\textwidth]{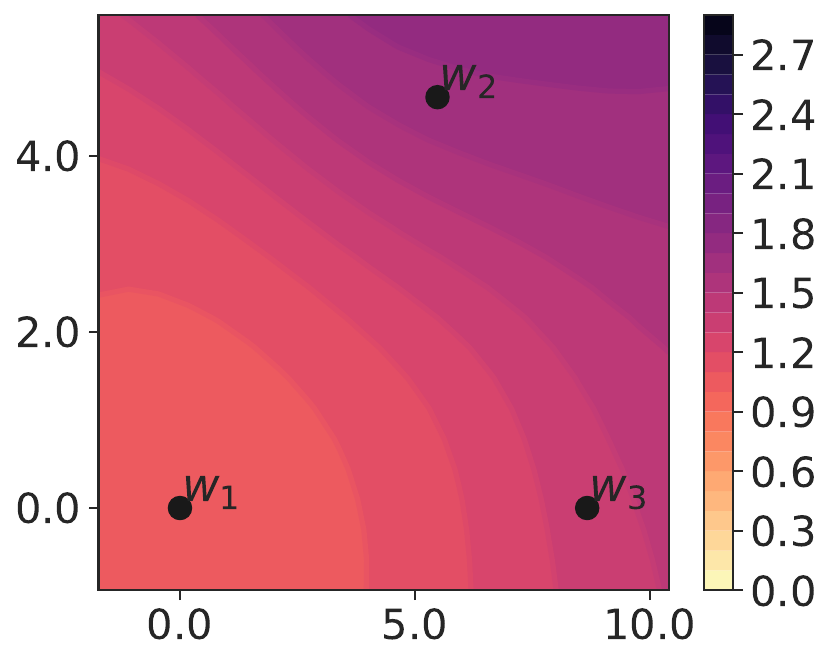}
      \caption{Seq 3 (R)}
      \label{fig:5datanlp_seq3_no_pt_contour_w1}
    \end{subfigure}\hspace{\fill}%
    \begin{subfigure}{.19\textwidth}
      \centering
      \includegraphics[width=0.9\textwidth]{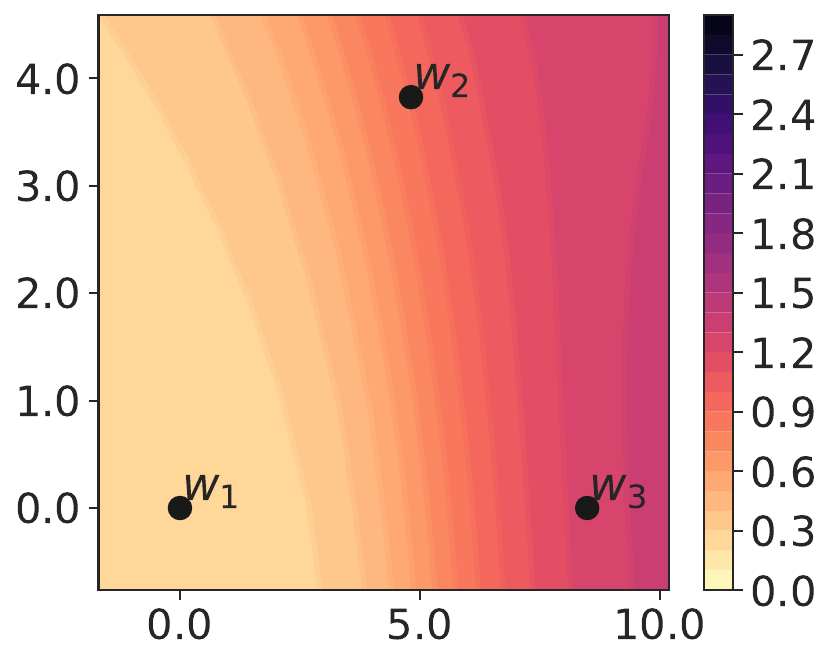}
      \caption{Seq 4 (R)}
      \label{fig:5datanlp_seq4_no_pt_contour_w1}
    \end{subfigure}
    \begin{subfigure}{.19\textwidth}
      \centering
      \includegraphics[width=0.9\textwidth]{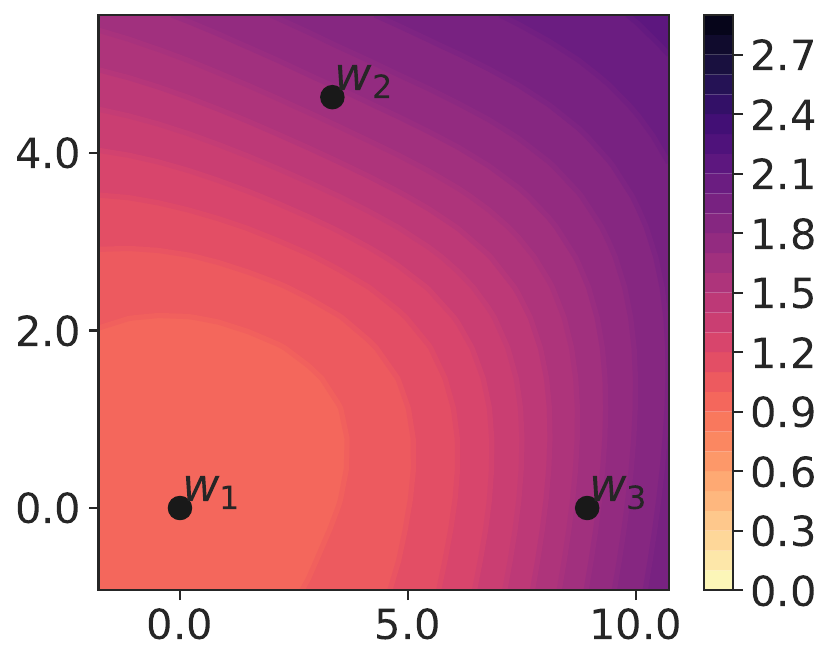}
      \caption{Seq 5 (R)}
      \label{fig:5datanlp_seq5_no_pt_contour_w1}
    \end{subfigure}
    \hspace{\fill}%
    \bigskip
    \begin{subfigure}{.19\textwidth}
      \centering
      \includegraphics[width=0.9\textwidth]{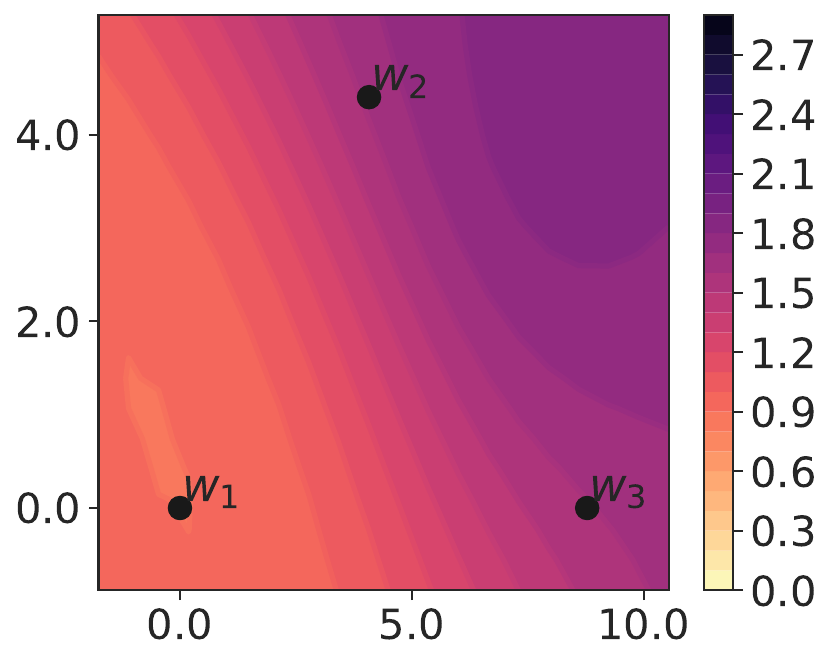}
      \caption{Seq 1 (PT)}
      \label{fig:5datanlp_seq1_pt_contour_w1}
    \end{subfigure}\hspace{\fill}%
    \begin{subfigure}{.19\textwidth}
      \centering
      \includegraphics[width=0.9\textwidth]{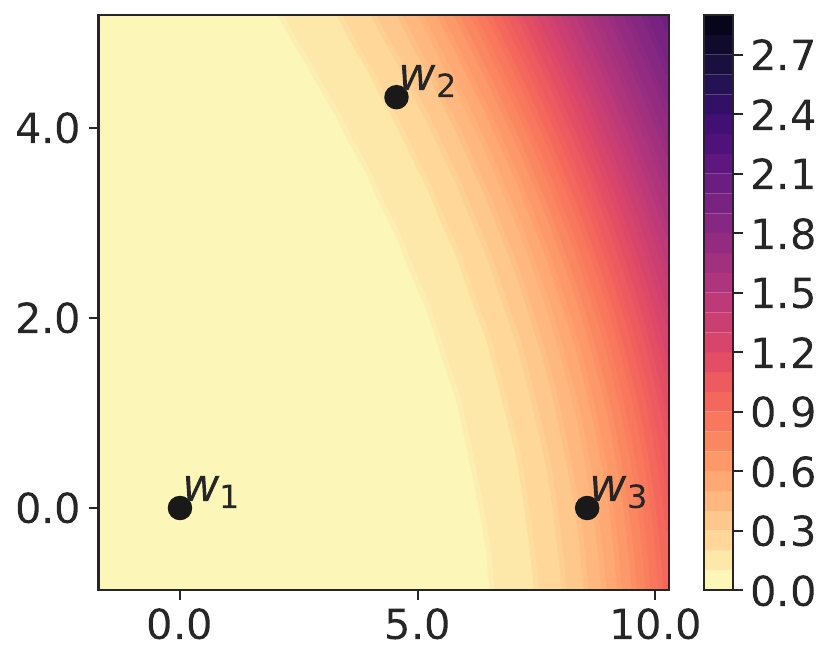}
      \caption{Seq 2 (PT)}
      \label{fig:5datanlp_seq2_pt_contour_w1}
    \end{subfigure}\hspace{\fill}%
    \begin{subfigure}{.19\textwidth}
      \centering
      \includegraphics[width=0.9\textwidth]{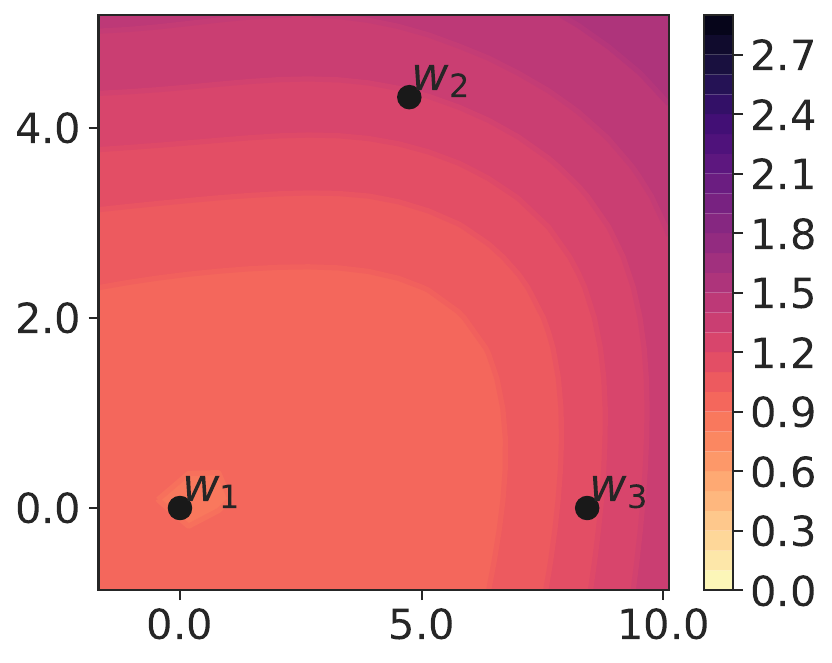}
      \caption{Seq 3 (PT)}
      \label{fig:5datanlp_seq3_pt_contour_w1}
    \end{subfigure}\hspace{\fill}%
    \begin{subfigure}{.19\textwidth}
      \centering
      \includegraphics[width=0.9\textwidth]{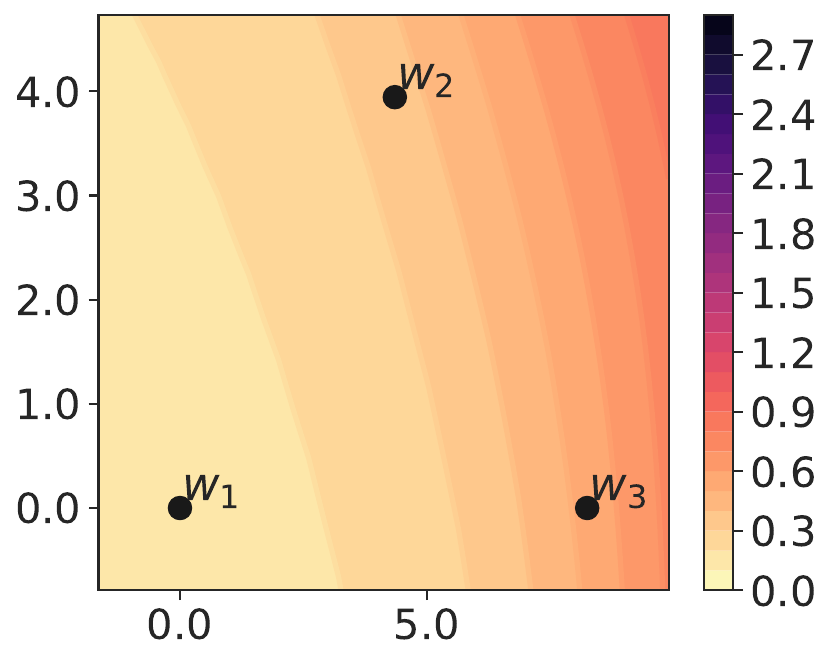}
      \caption{Seq 4 (PT)}
      \label{fig:5datanlp_seq4_pt_contour_w1}
    \end{subfigure}
    \begin{subfigure}{.19\textwidth}
      \centering
      \includegraphics[width=0.9\textwidth]{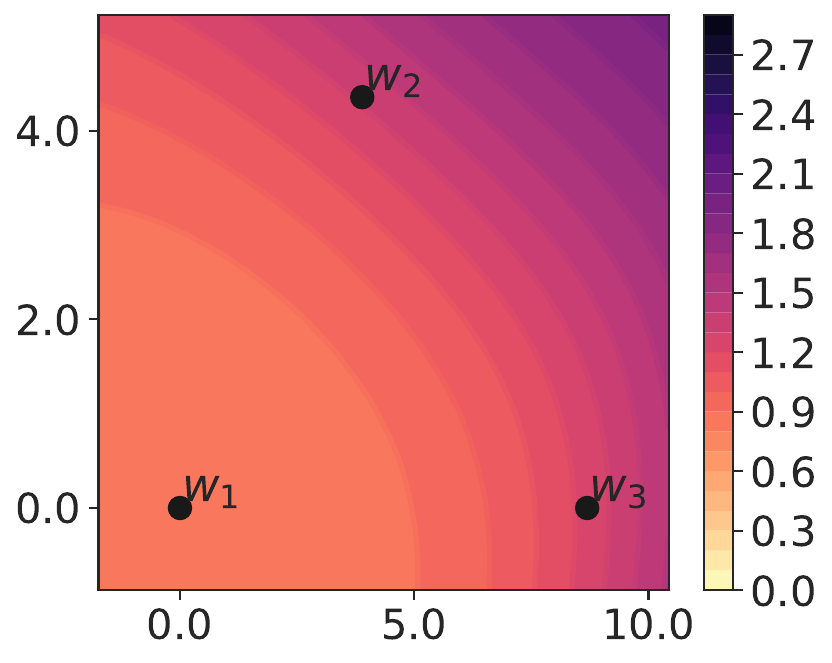}
      \caption{Seq 5 (PT)}
      \label{fig:5datanlp_seq5_pt_contour_w1}
    \end{subfigure}
    \hspace{\fill}%
    \caption{Loss contours for Task 1 on 5 task sequences of 5-dataset-NLP. Each contour shows the location of the model parameters after training sequentially on Task 1 (w$_1$), Task 2 (w$_2$), and Task 3 (w$_3$). The top row shows contours for randomly initialized models (R) and the bottom row shows contours for pre-trained initialized models (PT).}
    \label{fig:contours_w1}
\end{figure}

\begin{figure}[H]
    \centering
    \begin{subfigure}{.19\textwidth}
      \centering
      \includegraphics[width=0.88\textwidth]{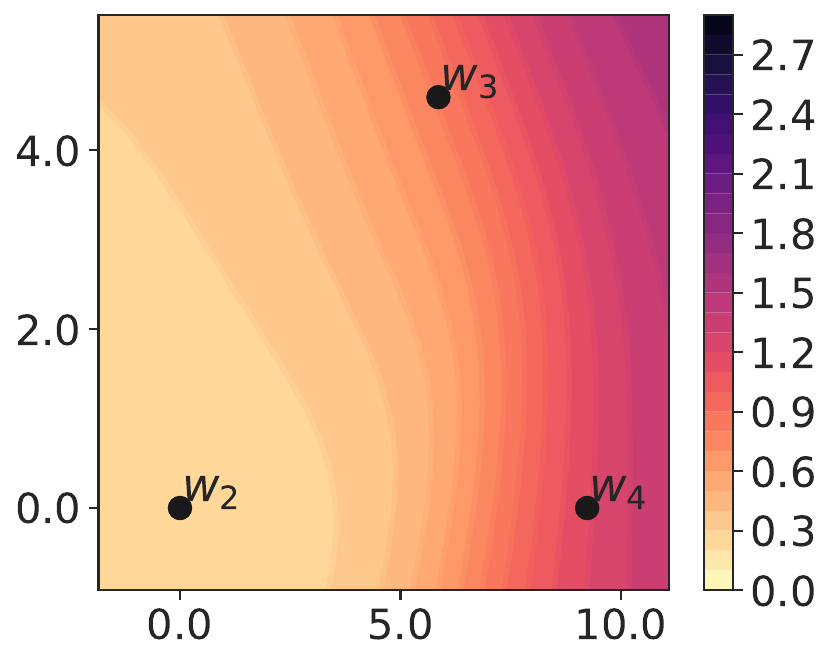}
      \caption{Seq 1 (R)}
      \label{fig:5datanlp_seq1_no_pt_contour_w2}
    \end{subfigure}\hspace{\fill}%
    \begin{subfigure}{.19\textwidth}
      \centering
      \includegraphics[width=0.88\textwidth]{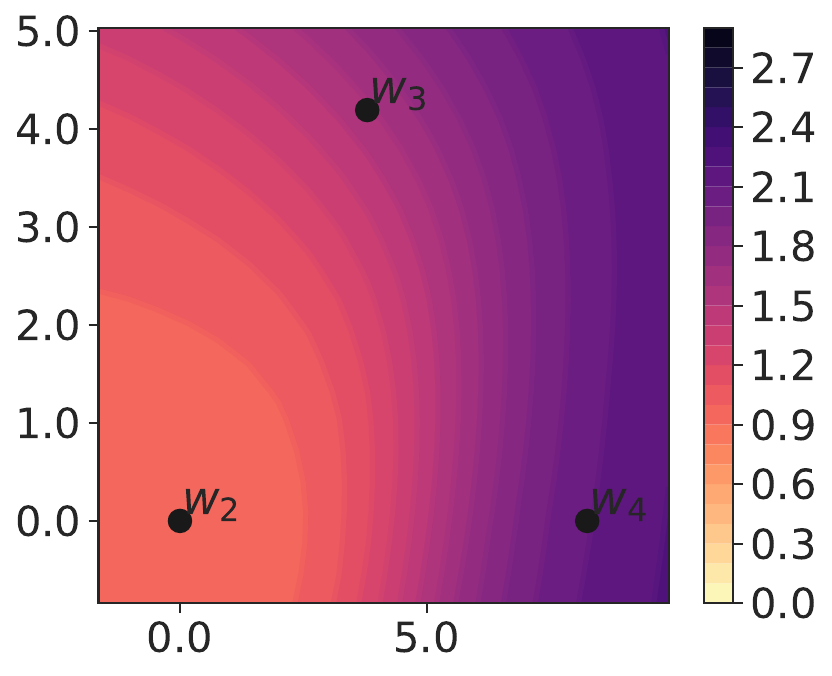}
      \caption{Seq 2 (R)}
      \label{fig:5datanlp_seq2_no_pt_contour_w2}
    \end{subfigure}\hspace{\fill}%
    \begin{subfigure}{.19\textwidth}
      \centering
      \includegraphics[width=0.88\textwidth]{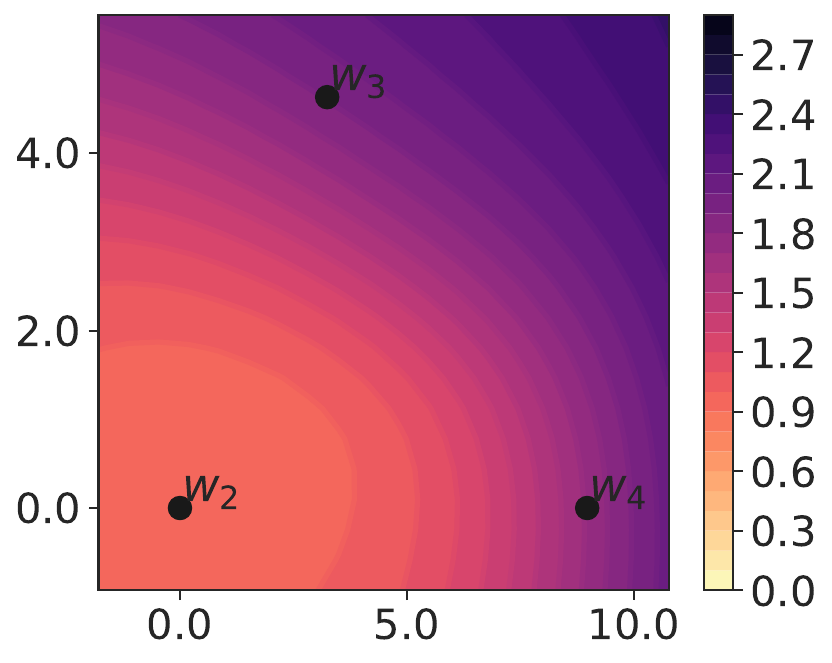}
      \caption{Seq 3 (R)}
      \label{fig:5datanlp_seq3_no_pt_contour_w2}
    \end{subfigure}\hspace{\fill}%
    \begin{subfigure}{.19\textwidth}
      \centering
      \includegraphics[width=0.88\textwidth]{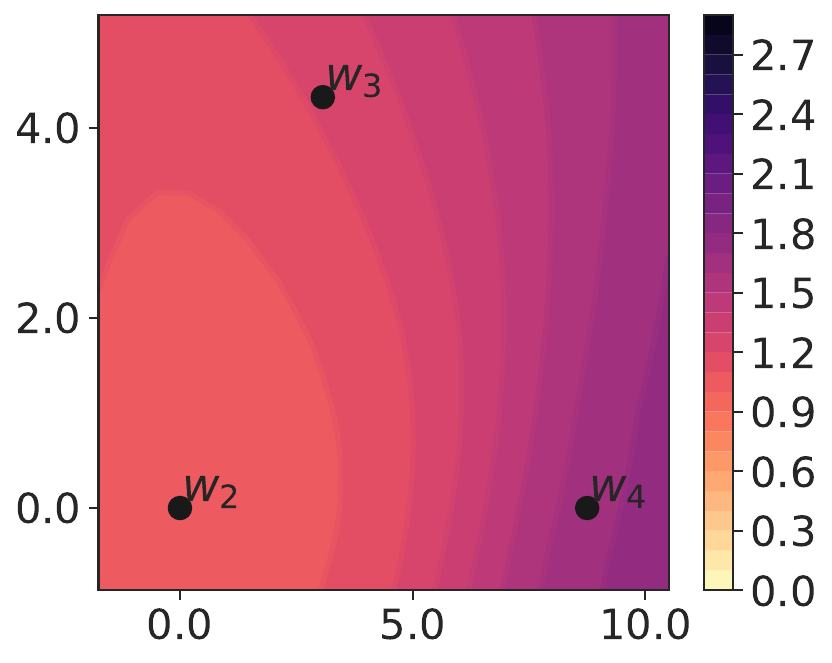}
      \caption{Seq 4 (R)}
      \label{fig:5datanlp_seq4_no_pt_contour_w2}
    \end{subfigure}
    \begin{subfigure}{.19\textwidth}
      \centering
      \includegraphics[width=0.88\textwidth]{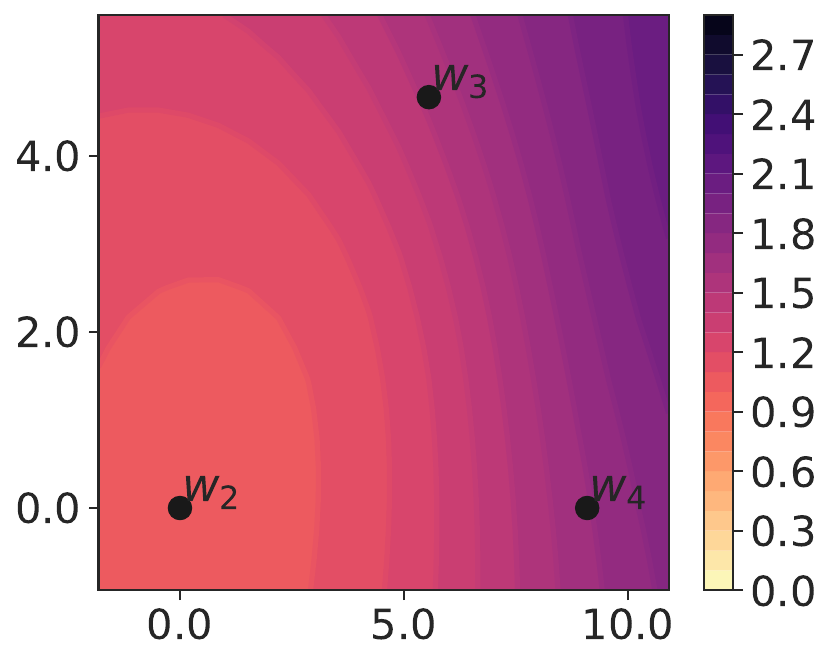}
      \caption{Seq 5 (R)}
      \label{fig:5datanlp_seq5_no_pt_contour_w2}
    \end{subfigure}
    \hspace{\fill}%
    \bigskip
    \begin{subfigure}{.19\textwidth}
      \centering
      \includegraphics[width=0.88\textwidth]{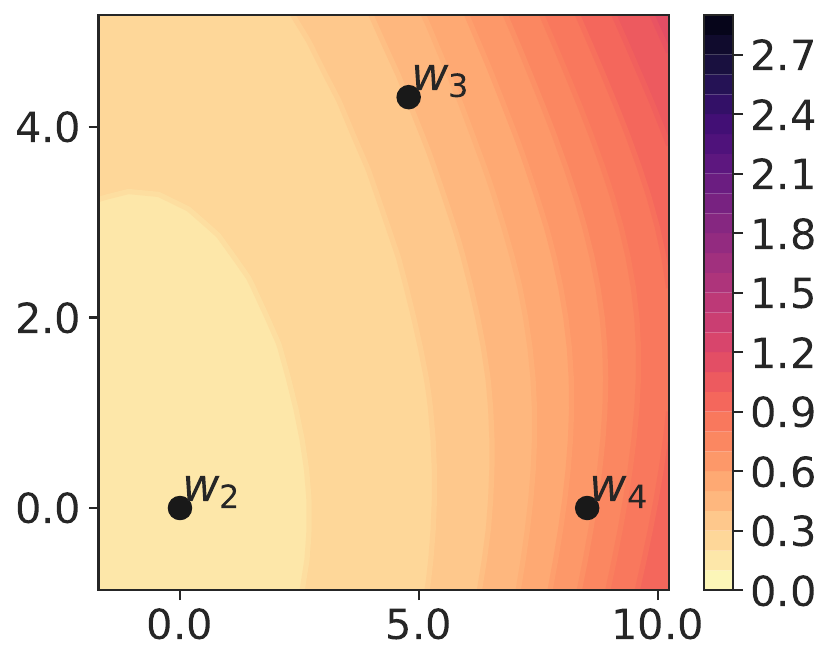}
      \caption{Seq 1 (PT)}
      \label{fig:5datanlp_seq1_pt_contour_w2}
    \end{subfigure}\hspace{\fill}%
    \begin{subfigure}{.19\textwidth}
      \centering
      \includegraphics[width=0.88\textwidth]{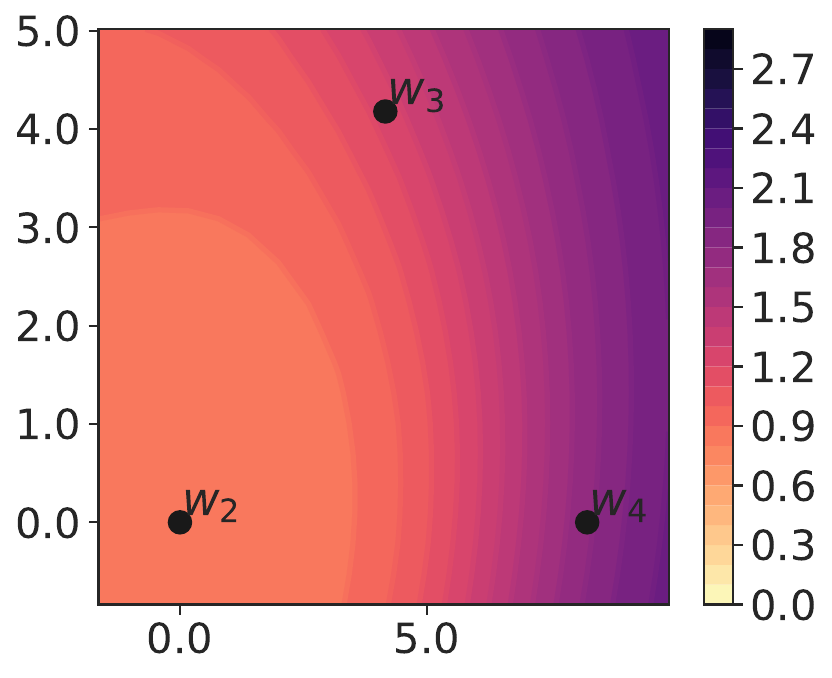}
      \caption{Seq 2 (PT)}
      \label{fig:5datanlp_seq2_pt_contour_w2}
    \end{subfigure}\hspace{\fill}%
    \begin{subfigure}{.19\textwidth}
      \centering
      \includegraphics[width=0.88\textwidth]{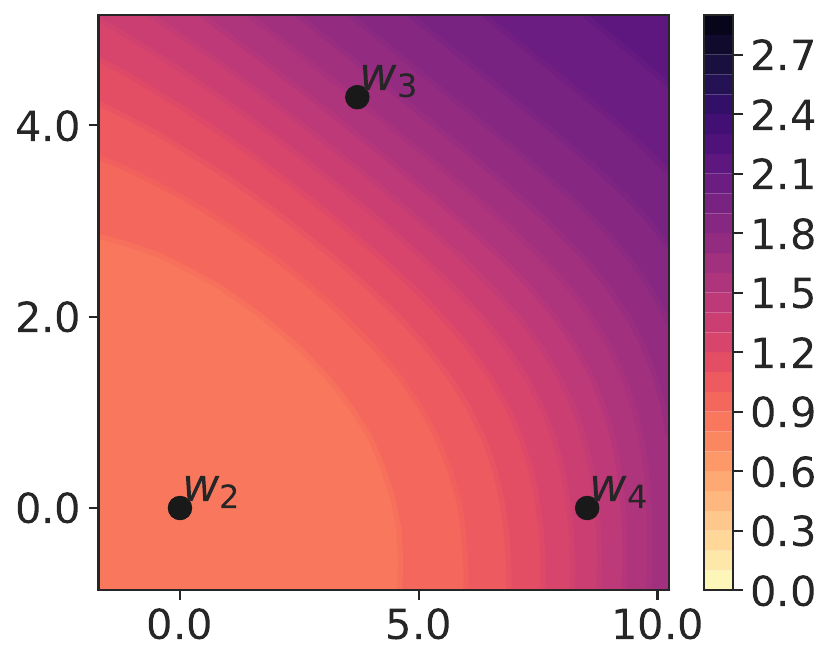}
      \caption{Seq 3 (PT)}
      \label{fig:5datanlp_seq3_pt_contour_w2}
    \end{subfigure}\hspace{\fill}%
    \begin{subfigure}{.19\textwidth}
      \centering
      \includegraphics[width=0.88\textwidth]{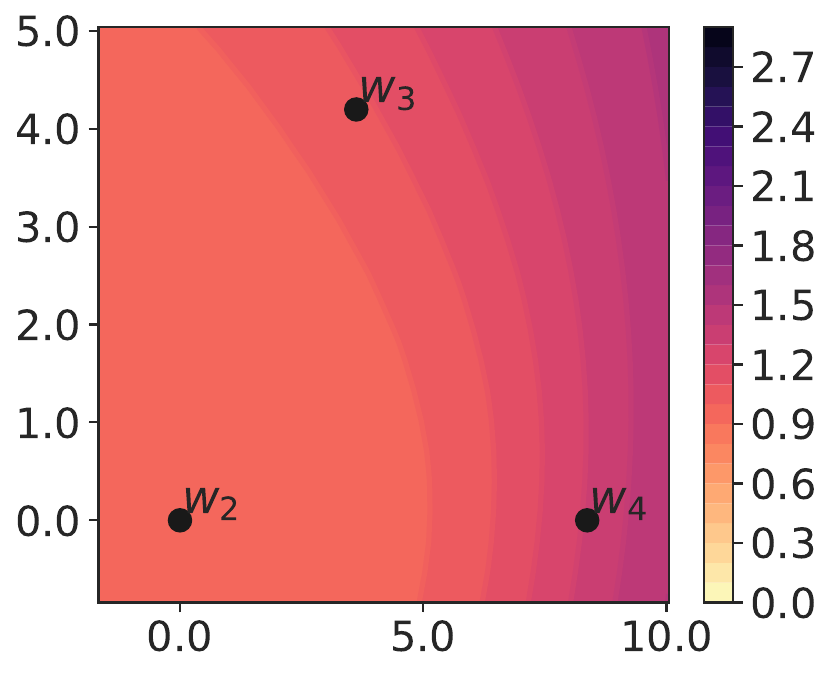}
      \caption{Seq 4 (PT)}
      \label{fig:5datanlp_seq4_pt_contour_w2}
    \end{subfigure}
    \begin{subfigure}{.19\textwidth}
      \centering
      \includegraphics[width=0.88\textwidth]{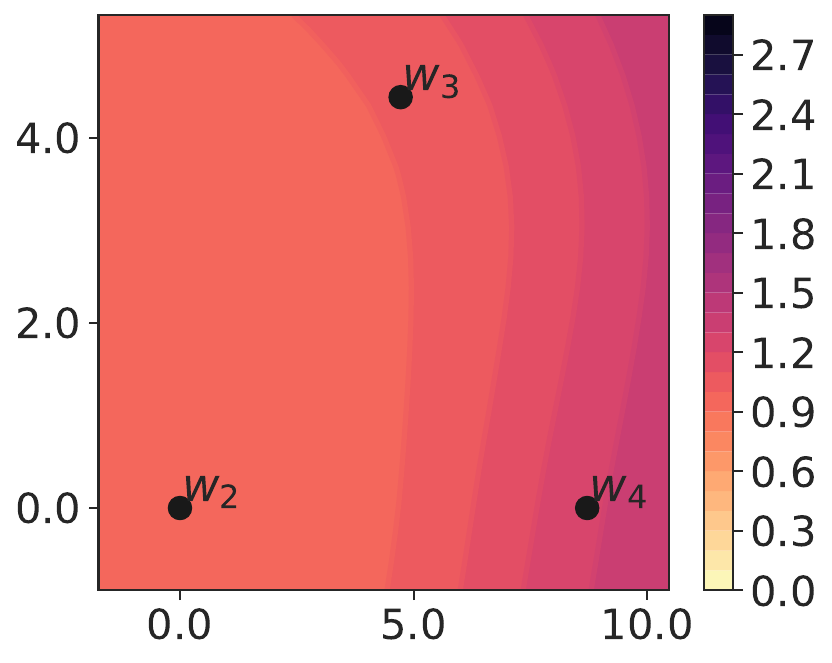}
      \caption{Seq 5 (PT)}
      \label{fig:5datanlp_seq5_pt_contour_w2}
    \end{subfigure}
    \hspace{\fill}%
    \caption{Loss contours for Task 2 on 5 task sequences of 5-dataset-NLP. 
    % Each contour shows the location of the model parameters after training sequentially on \textbf{Task 2 (w$_2$), Task 3 (w$_3$), Task 4 (w$_4$)}. The top row shows contours for randomly initialized models (w/o PT) and the bottom row shows contours for pre-trained initialized models (w/ PT).
    }
    \label{fig:contours_w2}
\end{figure}

% \subsubsection{Loss Contours: Split YahooQA}
\begin{figure}[H]
    \centering
    \begin{subfigure}{.19\textwidth}
      \centering
      \includegraphics[width=0.88\textwidth]{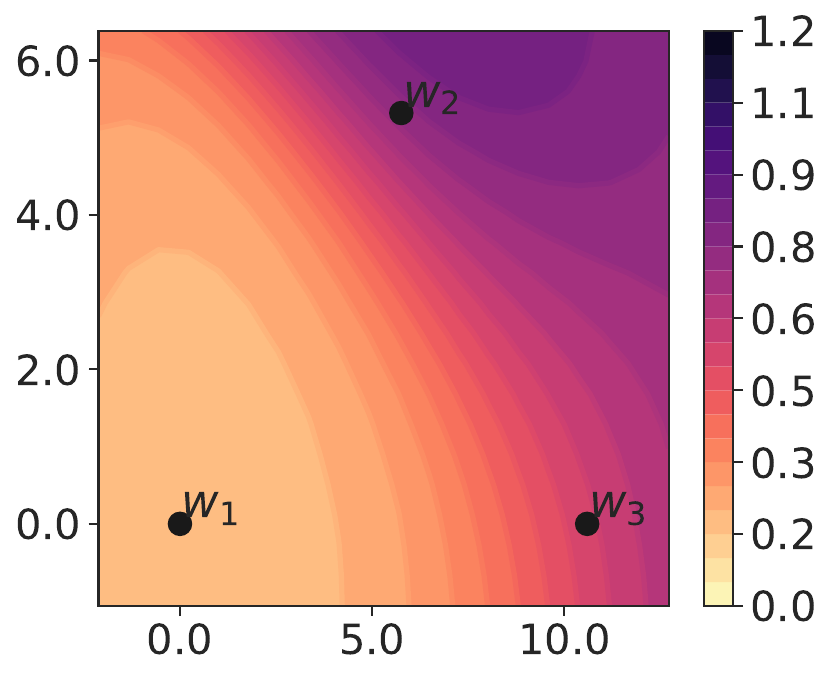}
      \caption{Seq 1 (R)}
      \label{fig:yqa_seq1_no_pt_contour_w1}
    \end{subfigure}\hspace{\fill}%
    \begin{subfigure}{.19\textwidth}
      \centering
      \includegraphics[width=0.88\textwidth]{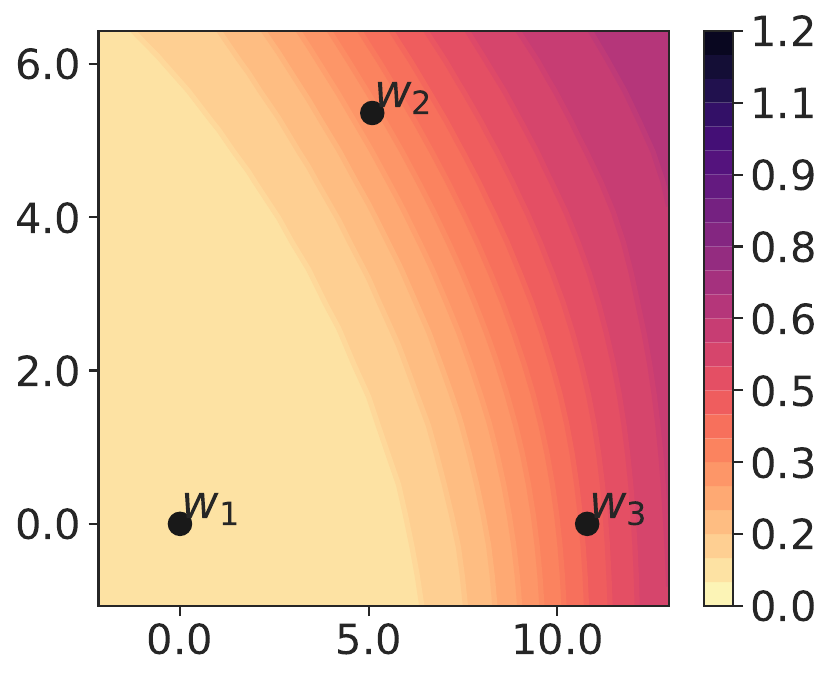}
      \caption{Seq 2 (R)}
      \label{fig:yqa_seq2_no_pt_contour_w1}
    \end{subfigure}\hspace{\fill}%
    \begin{subfigure}{.19\textwidth}
      \centering
      \includegraphics[width=0.88\textwidth]{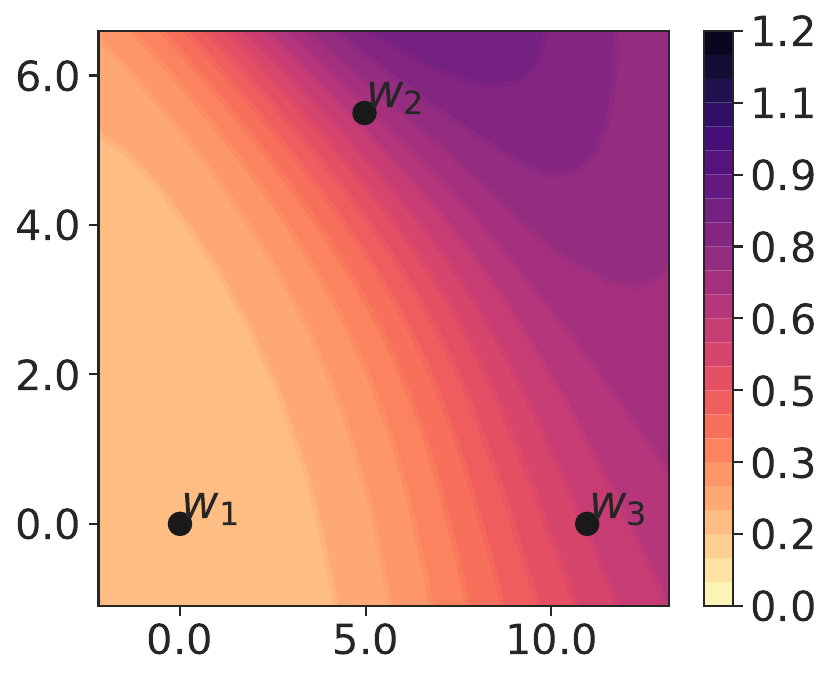}
      \caption{Seq 3 (R)}
      \label{fig:yqa_seq3_no_pt_contour_w1}
    \end{subfigure}\hspace{\fill}%
    \begin{subfigure}{.19\textwidth}
      \centering
      \includegraphics[width=0.88\textwidth]{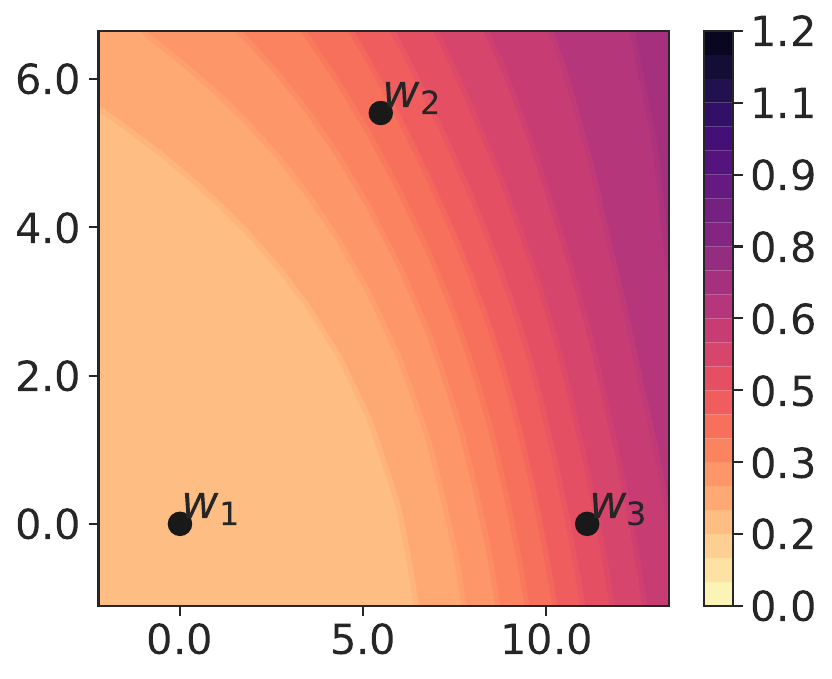}
      \caption{Seq 4 (R)}
      \label{fig:yqa_seq4_no_pt_contour_w1}
    \end{subfigure}
    \begin{subfigure}{.19\textwidth}
      \centering
      \includegraphics[width=0.88\textwidth]{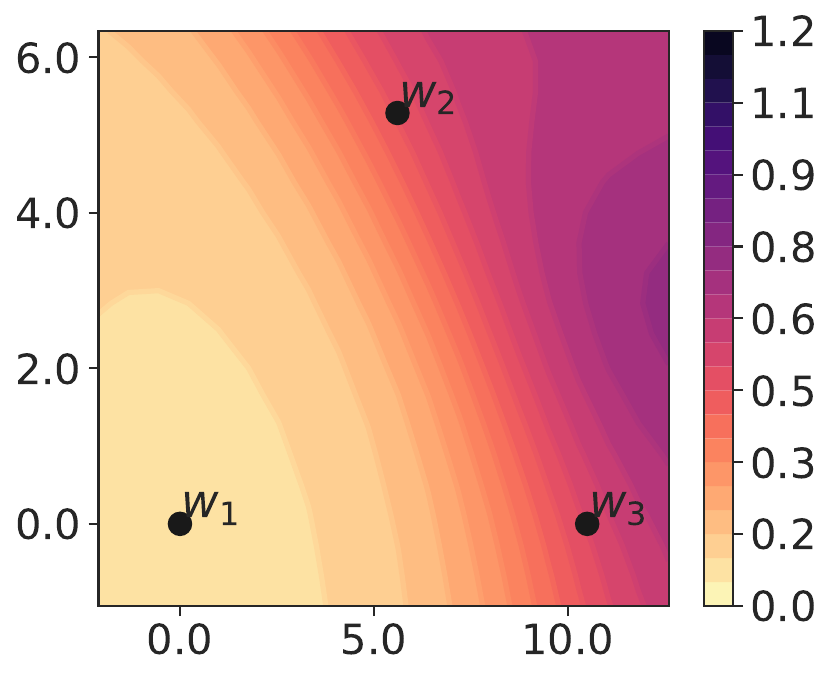}
      \caption{Seq 5 (R)}
      \label{fig:yqa_seq5_no_pt_contour_w1}
    \end{subfigure}
    \hspace{\fill}%
    \bigskip
    \begin{subfigure}{.19\textwidth}
      \centering
      \includegraphics[width=0.88\textwidth]{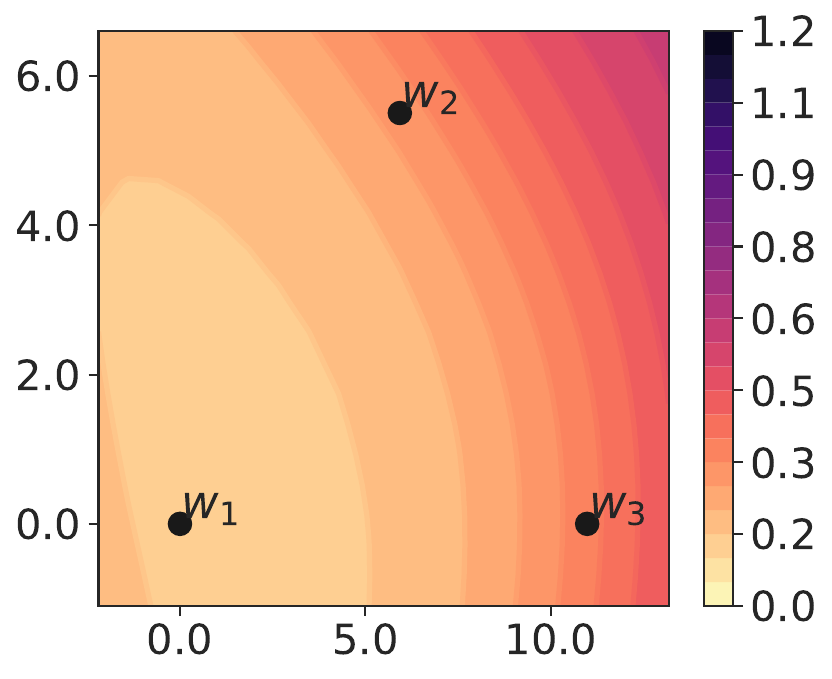}
      \caption{Seq 1 (PT)}
      \label{fig:yqa_seq1_pt_contour_w1}
    \end{subfigure}\hspace{\fill}%
    \begin{subfigure}{.19\textwidth}
      \centering
      \includegraphics[width=0.88\textwidth]{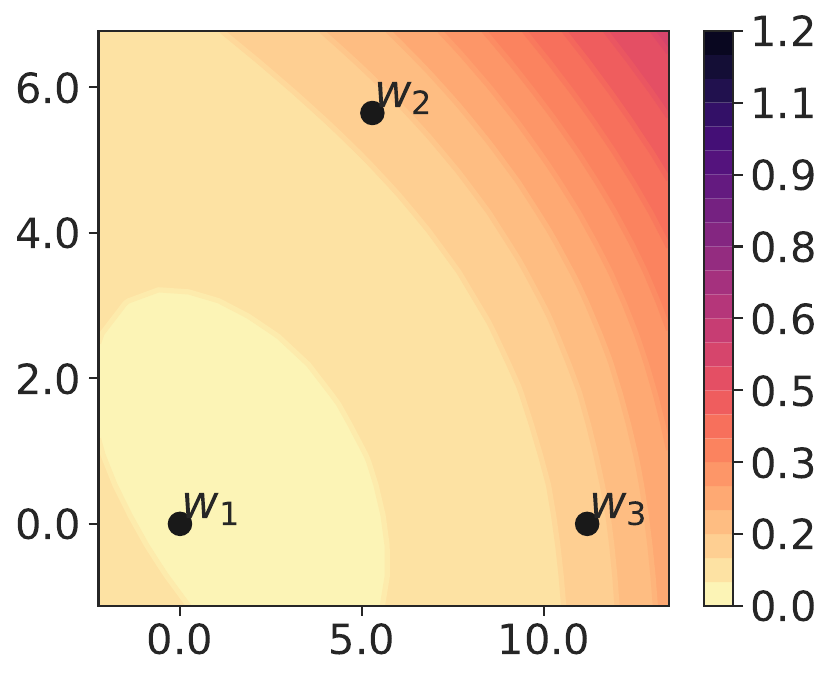}
      \caption{Seq 2 (PT)}
      \label{fig:yqa_seq2_pt_contour_w1}
    \end{subfigure}\hspace{\fill}%
    \begin{subfigure}{.19\textwidth}
      \centering
      \includegraphics[width=0.88\textwidth]{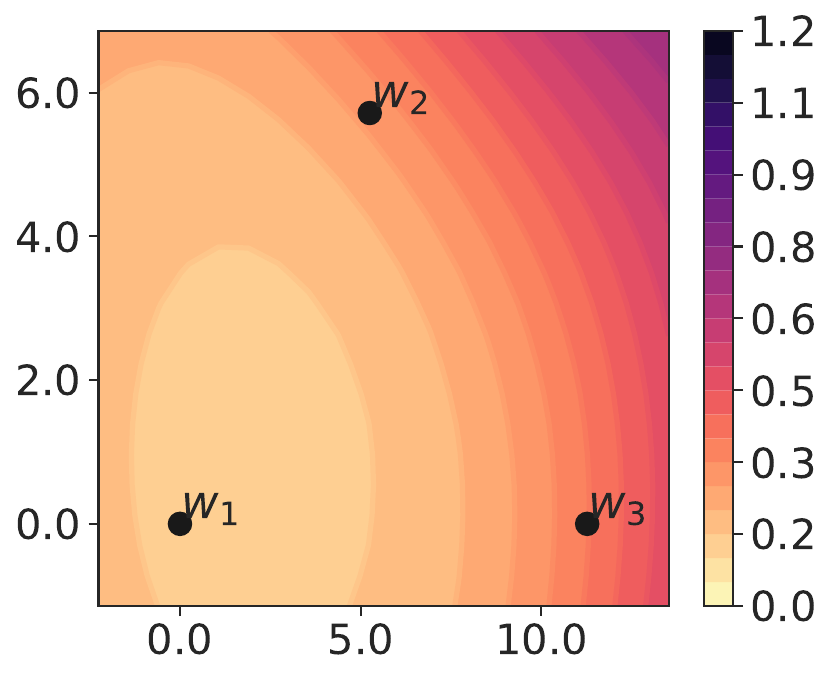}
      \caption{Seq 3 (PT)}
      \label{fig:yqa_seq3_pt_contour_w1}
    \end{subfigure}\hspace{\fill}%
    \begin{subfigure}{.19\textwidth}
      \centering
      \includegraphics[width=0.88\textwidth]{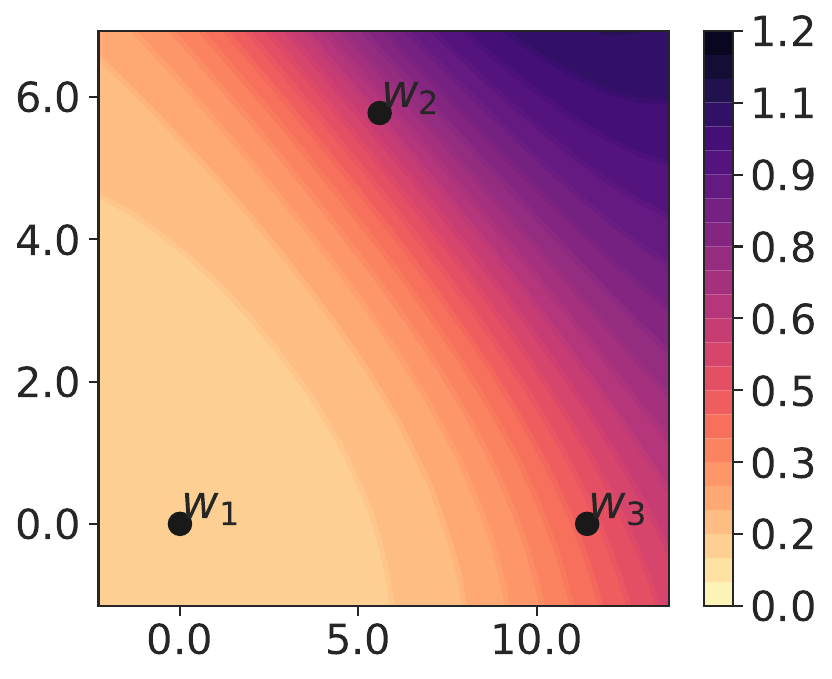}
      \caption{Seq 4 (PT)}
      \label{fig:yqa_seq4_pt_contour_w1}
    \end{subfigure}
    \begin{subfigure}{.19\textwidth}
      \centering
      \includegraphics[width=0.88\textwidth]{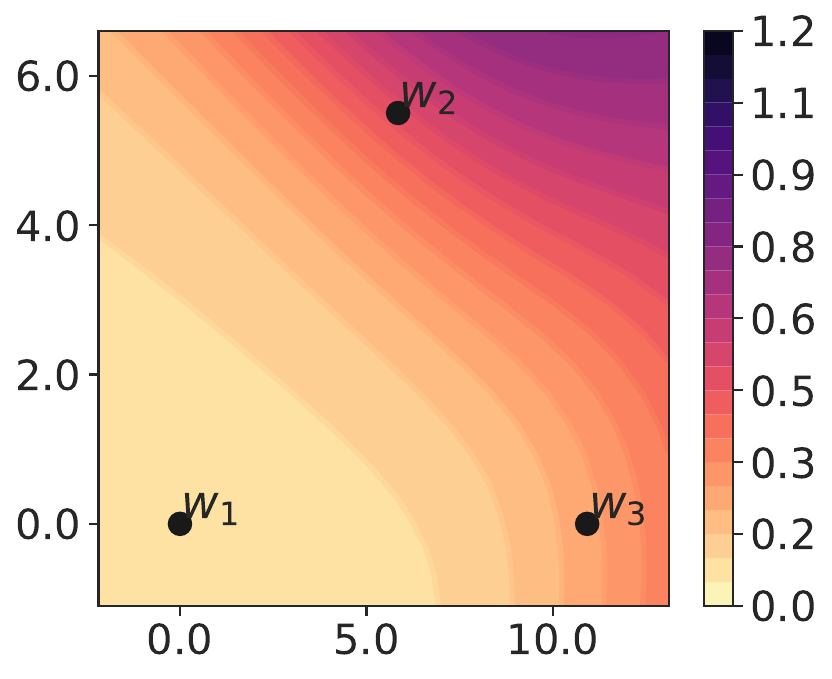}
      \caption{Seq 5 (PT)}
      \label{fig:yqa_seq5_pt_contour_w1}
    \end{subfigure}
    \hspace{\fill}%
    \caption{Loss contours for Task 1 on 5 task sequences of Split YahooQA.}
    \label{fig:yqa_contours_w1}
\end{figure}

\begin{figure}[H]
    \centering
    \begin{subfigure}{.19\textwidth}
      \centering
      \includegraphics[width=0.85\textwidth]{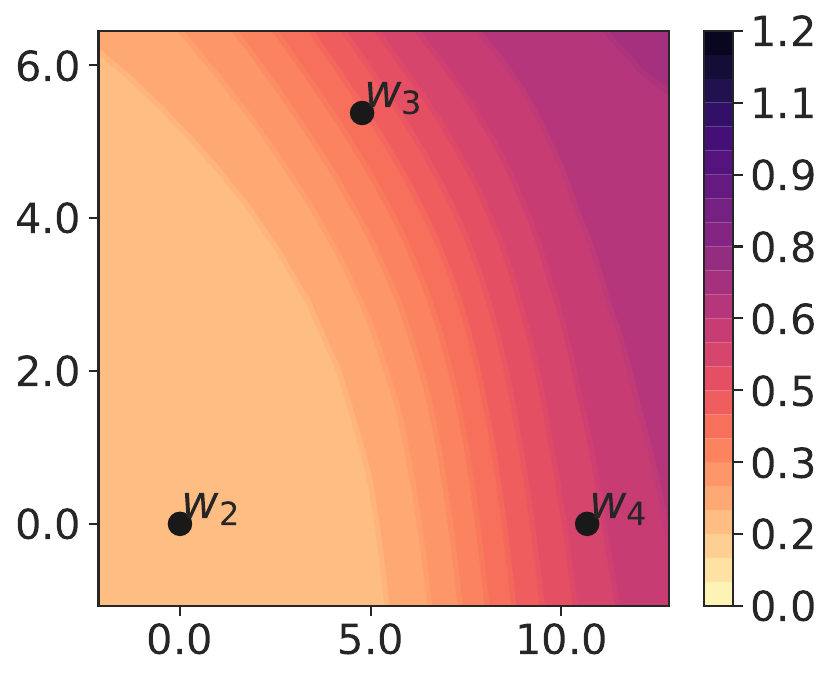}
      \caption{Seq 1 (R)}
      \label{fig:yqa_seq1_no_pt_contour_w2}
    \end{subfigure}\hspace{\fill}%
    \begin{subfigure}{.19\textwidth}
      \centering
      \includegraphics[width=0.85\textwidth]{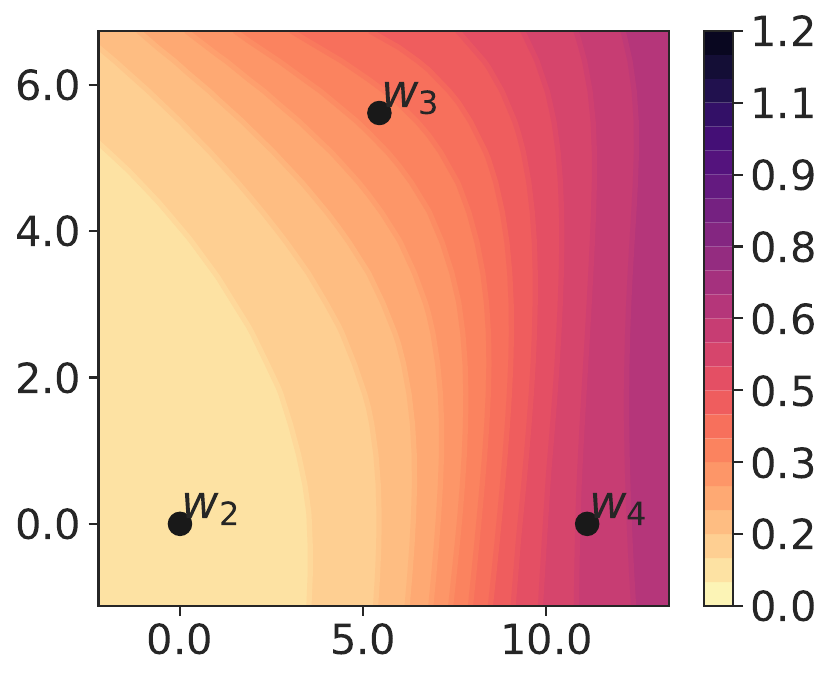}
      \caption{Seq 2 (R)}
      \label{fig:yqa_seq2_no_pt_contour_w2}
    \end{subfigure}\hspace{\fill}%
    \begin{subfigure}{.19\textwidth}
      \centering
      \includegraphics[width=0.85\textwidth]{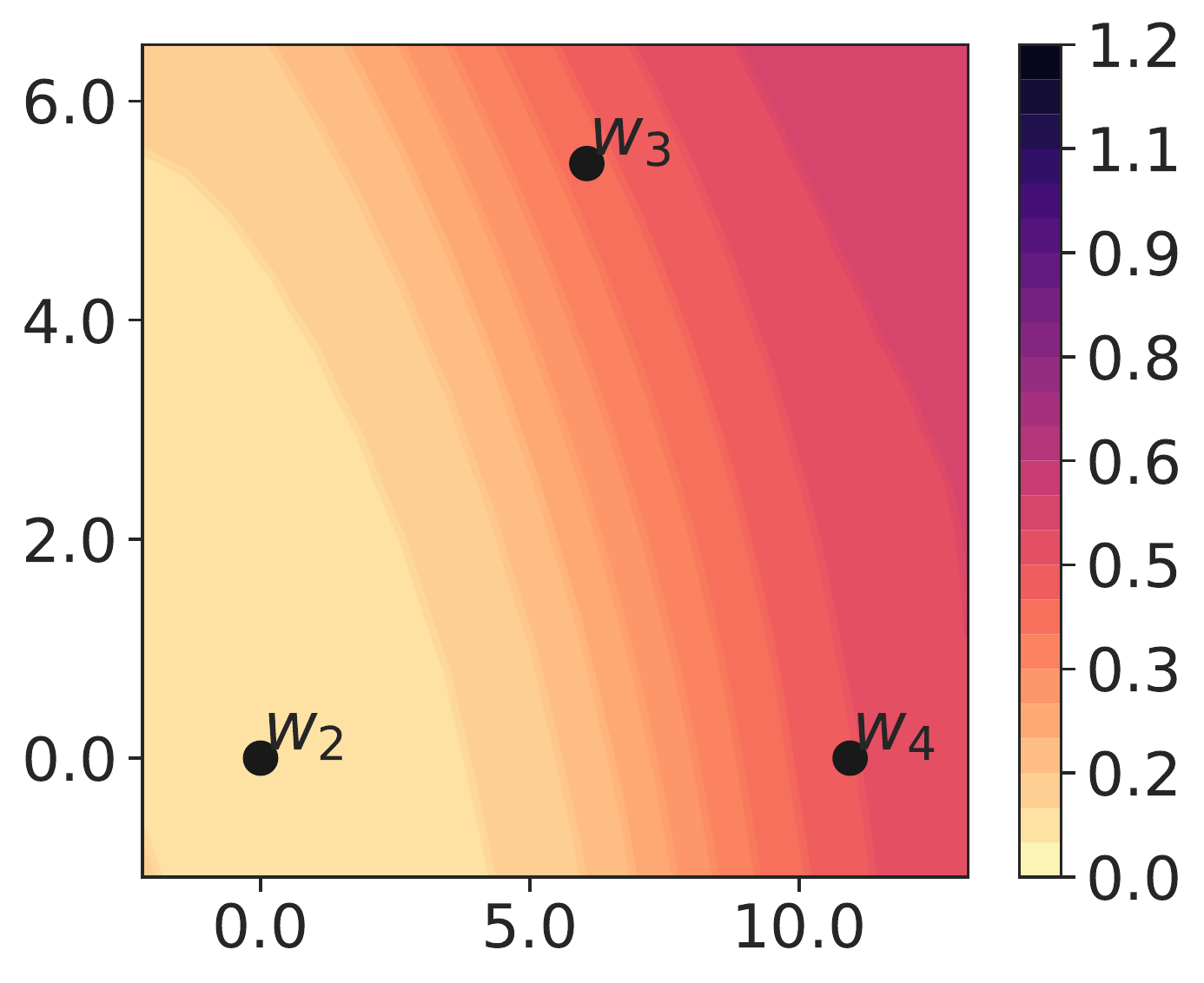}
      \caption{Seq 3 (R)}
      \label{fig:yqa_seq3_no_pt_contour_w2}
    \end{subfigure}\hspace{\fill}%
    \begin{subfigure}{.19\textwidth}
      \centering
      \includegraphics[width=0.85\textwidth]{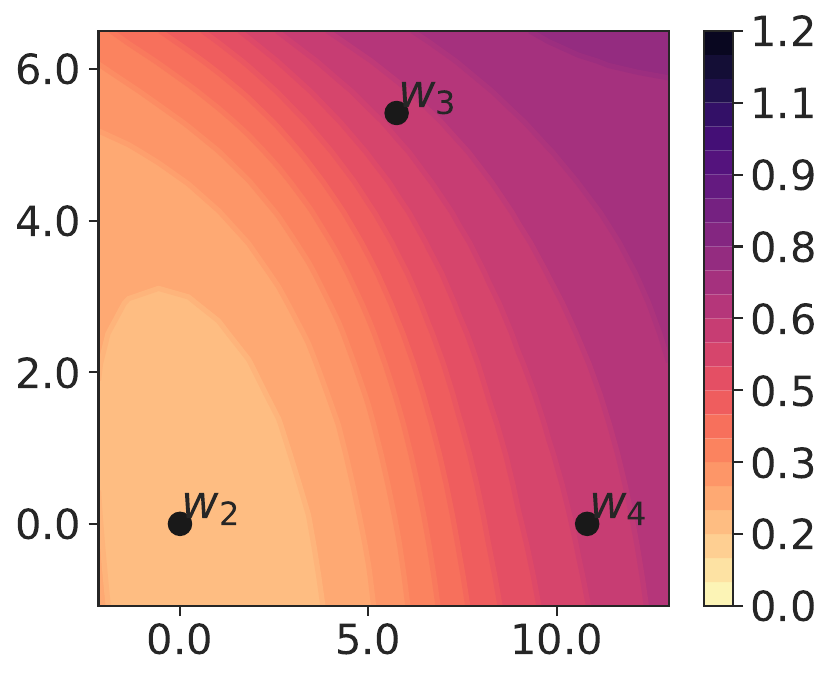}
      \caption{Seq 4 (R)}
      \label{fig:yqa_seq4_no_pt_contour_w2}
    \end{subfigure}
    \begin{subfigure}{.19\textwidth}
      \centering
      \includegraphics[width=0.85\textwidth]{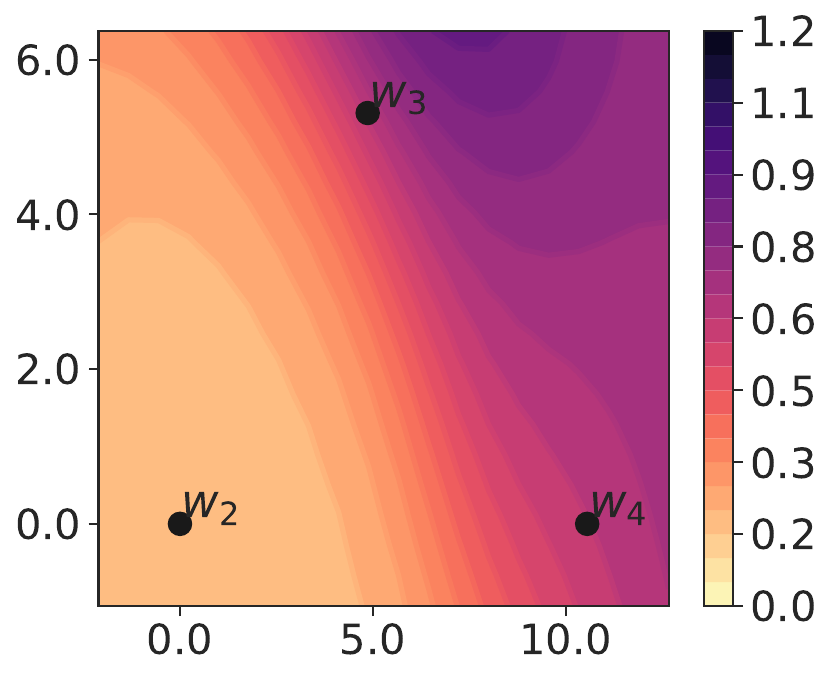}
      \caption{Seq 5 (R)}
      \label{fig:yqa_seq5_no_pt_contour_w2}
    \end{subfigure}
    \hspace{\fill}%
    \bigskip
    \begin{subfigure}{.19\textwidth}
      \centering
      \includegraphics[width=0.85\textwidth]{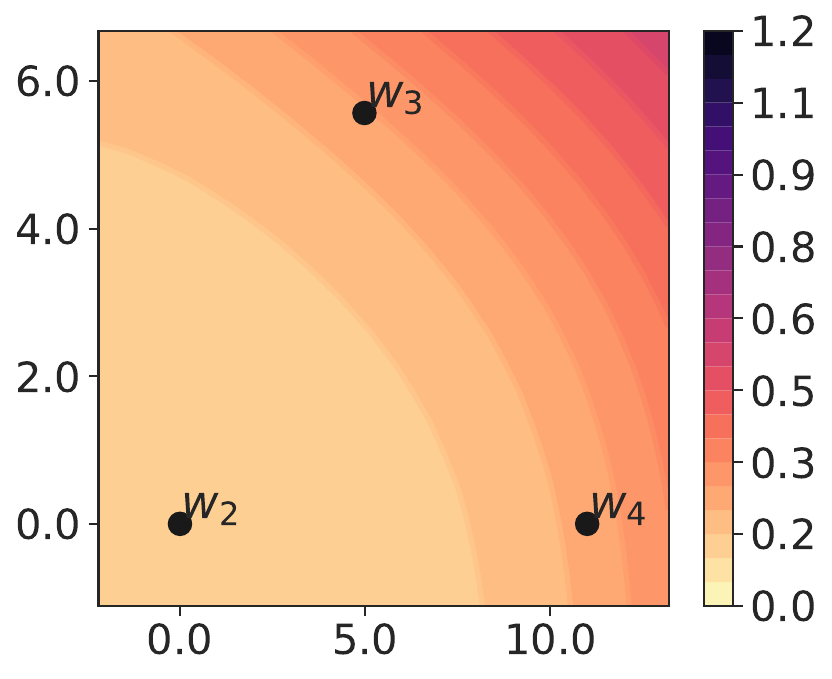}
      \caption{Seq 1 (PT)}
      \label{fig:yqa_seq1_pt_contour_w2}
    \end{subfigure}\hspace{\fill}%
    \begin{subfigure}{.19\textwidth}
      \centering
      \includegraphics[width=0.85\textwidth]{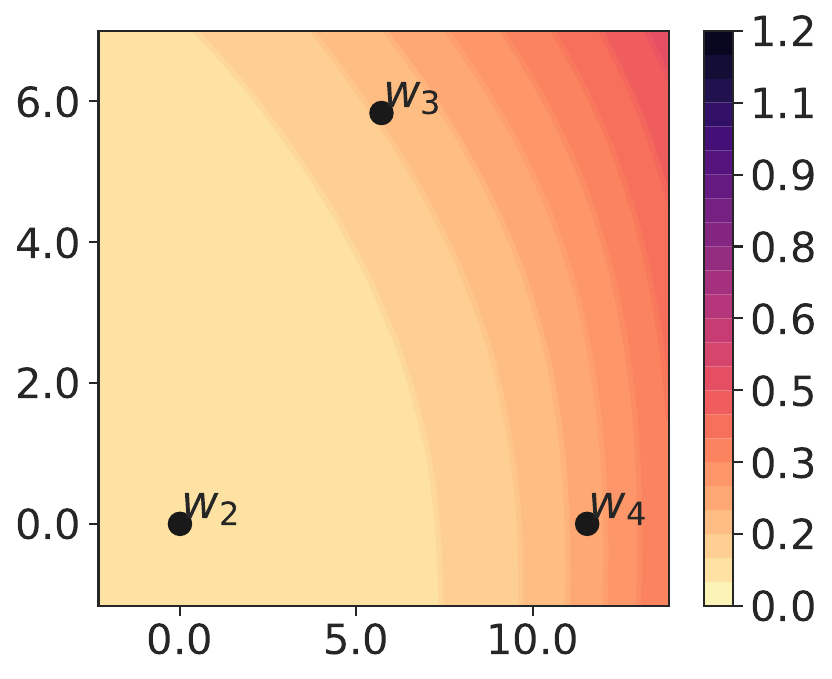}
      \caption{Seq 2 (PT)}
      \label{fig:yqa_seq2_pt_contour_w2}
    \end{subfigure}\hspace{\fill}%
    \begin{subfigure}{.19\textwidth}
      \centering
      \includegraphics[width=0.85\textwidth]{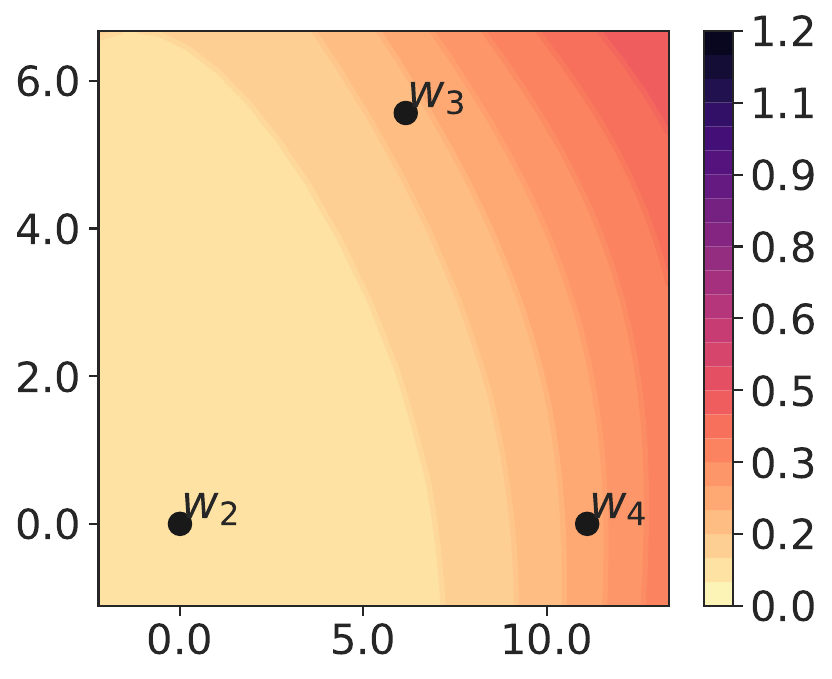}
      \caption{Seq 3 (PT)}
      \label{fig:yqa_seq3_pt_contour_w2}
    \end{subfigure}\hspace{\fill}%
    \begin{subfigure}{.19\textwidth}
      \centering
      \includegraphics[width=0.85\textwidth]{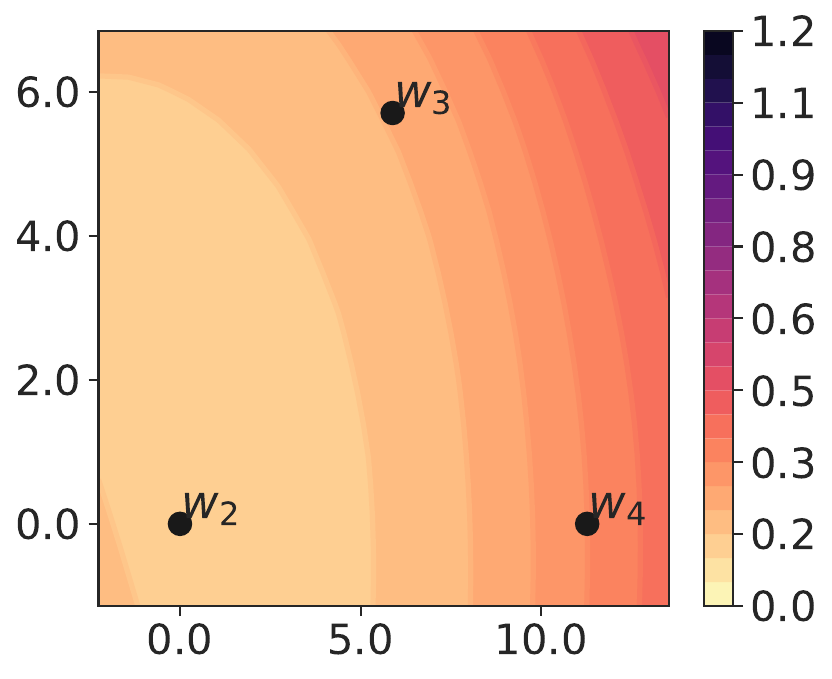}
      \caption{Seq 4 (PT)}
      \label{fig:yqa_seq4_pt_contour_w2}
    \end{subfigure}
    \begin{subfigure}{.19\textwidth}
      \centering
      \includegraphics[width=0.85\textwidth]{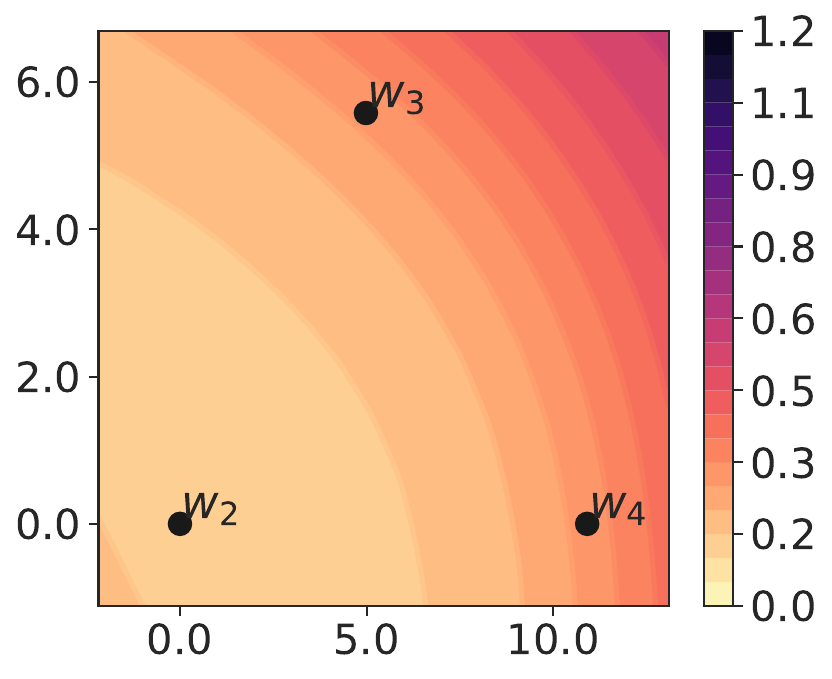}
      \caption{Seq 5 (PT)}
      \label{fig:yqa_seq5_pt_contour_w2}
    \end{subfigure}
    \hspace{\fill}%
    \caption{Loss contours for Task 2 on 5 task sequences of Split YahooQA.}
    \label{fig:yqa_contours_w2}
\end{figure}

% \subsubsection{Loss Contours: Split CIFAR-50}
\begin{figure}[H]
    \centering
    \begin{subfigure}{.19\textwidth}
      \centering
      \includegraphics[width=0.9\textwidth]{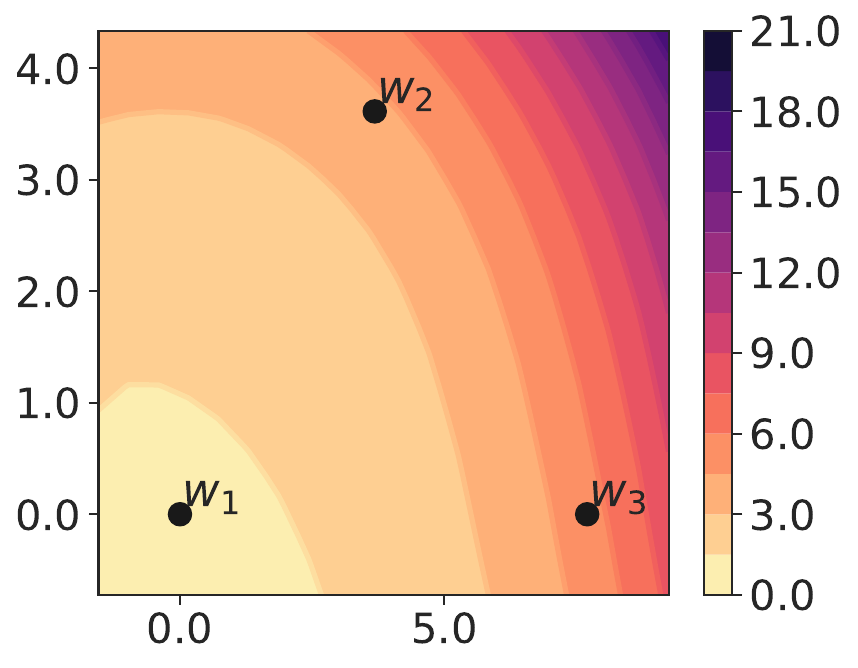}
      \caption{Seq 1 (R)}
      \label{fig:c50_5epoch_seq_1_no_contour_w1}
    \end{subfigure}\hspace{\fill}%
    \begin{subfigure}{.19\textwidth}
      \centering
      \includegraphics[width=0.9\textwidth]{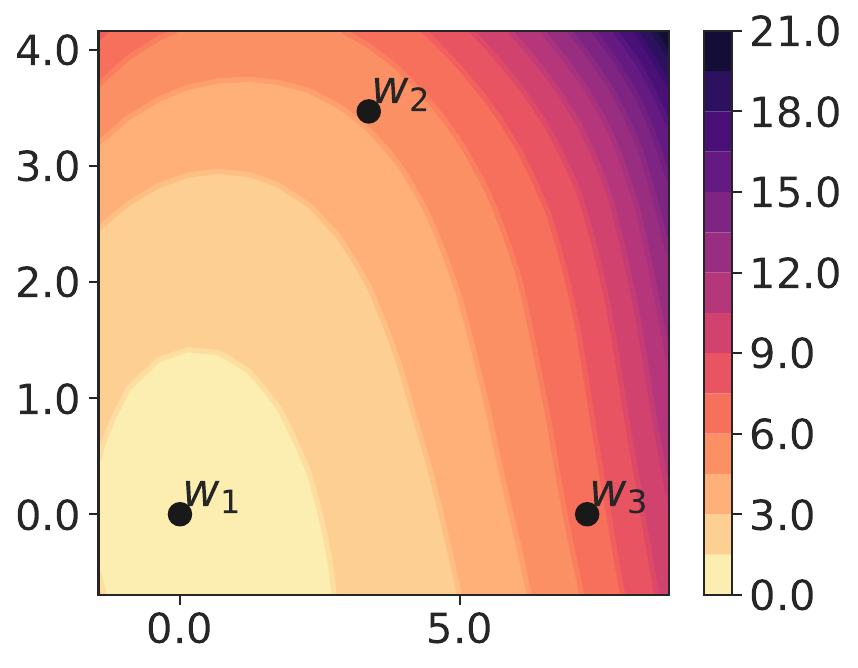}
      \caption{Seq 2 (R)}
      \label{fig:c50_5epoch_seq_2_no_contour_w1}
    \end{subfigure}\hspace{\fill}%
    \begin{subfigure}{.19\textwidth}
      \centering
      \includegraphics[width=0.9\textwidth]{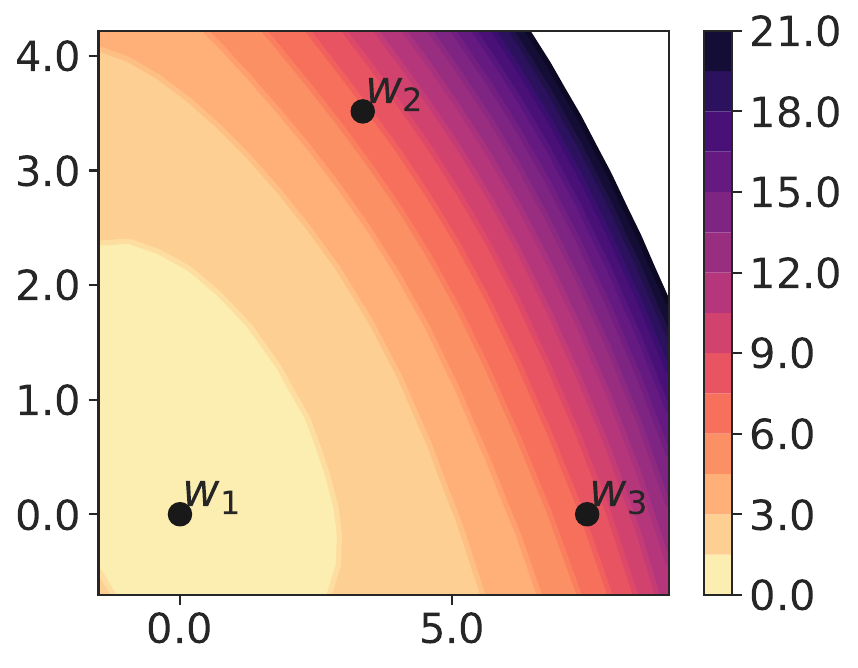}
      \caption{Seq 3 (R)}
      \label{fig:c50_5epoch_seq_3_no_contour_w1}
    \end{subfigure}\hspace{\fill}%
    \begin{subfigure}{.19\textwidth}
      \centering
      \includegraphics[width=0.9\textwidth]{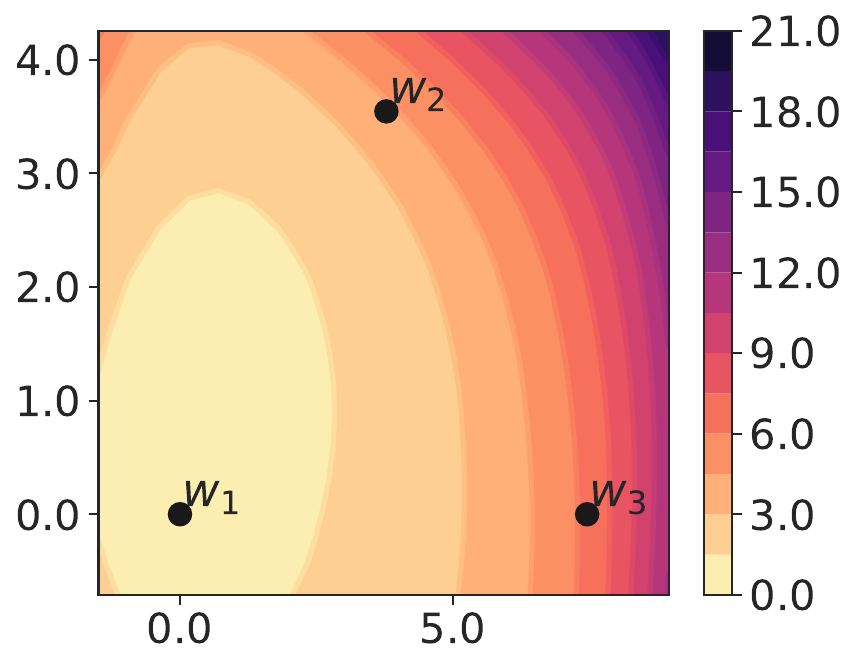}
      \caption{Seq 4 (R)}
      \label{fig:c50_5epoch_seq_4_no_contour_w1}
    \end{subfigure}
    \begin{subfigure}{.19\textwidth}
      \centering
      \includegraphics[width=0.9\textwidth]{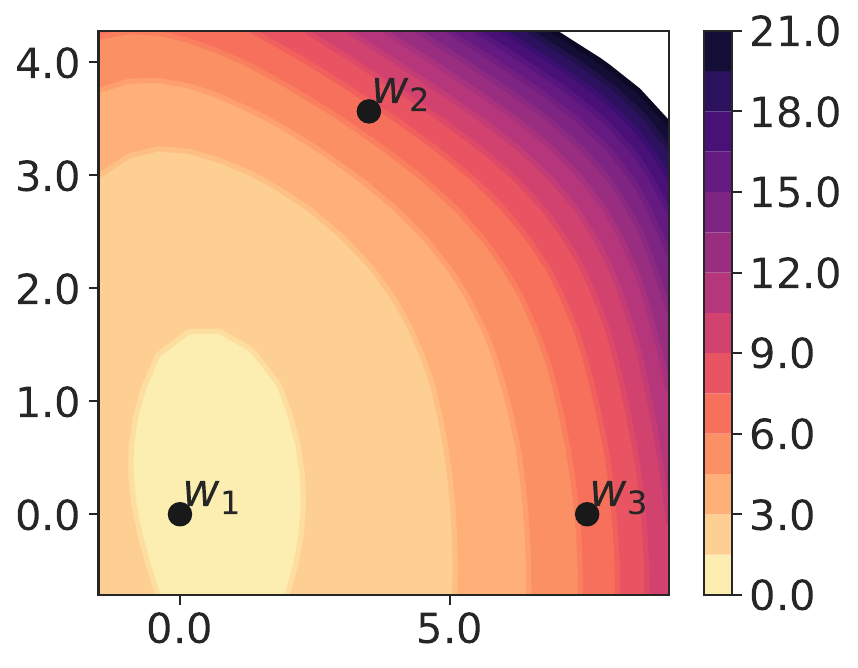}
      \caption{Seq 5 (R)}
      \label{fig:c50_5epoch_seq_5_no_contour_w1}
    \end{subfigure}
    \hspace{\fill}%
    \bigskip
    \begin{subfigure}{.19\textwidth}
      \centering
      \includegraphics[width=0.9\textwidth]{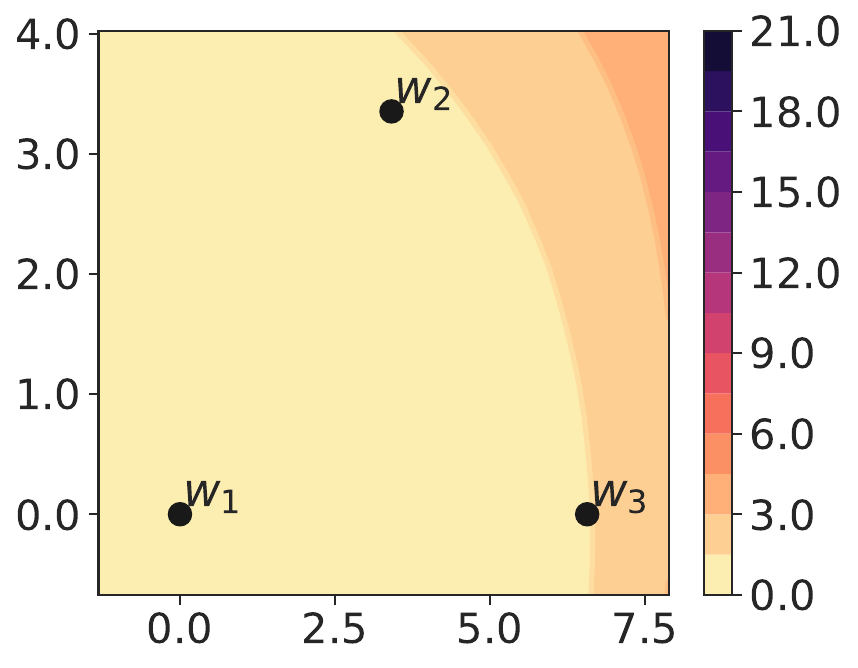}
      \caption{Seq 1 (PT)}
      \label{fig:c50_5epoch_seq_1_pt_contour_w1}
    \end{subfigure}\hspace{\fill}%
    \begin{subfigure}{.19\textwidth}
      \centering
      \includegraphics[width=0.9\textwidth]{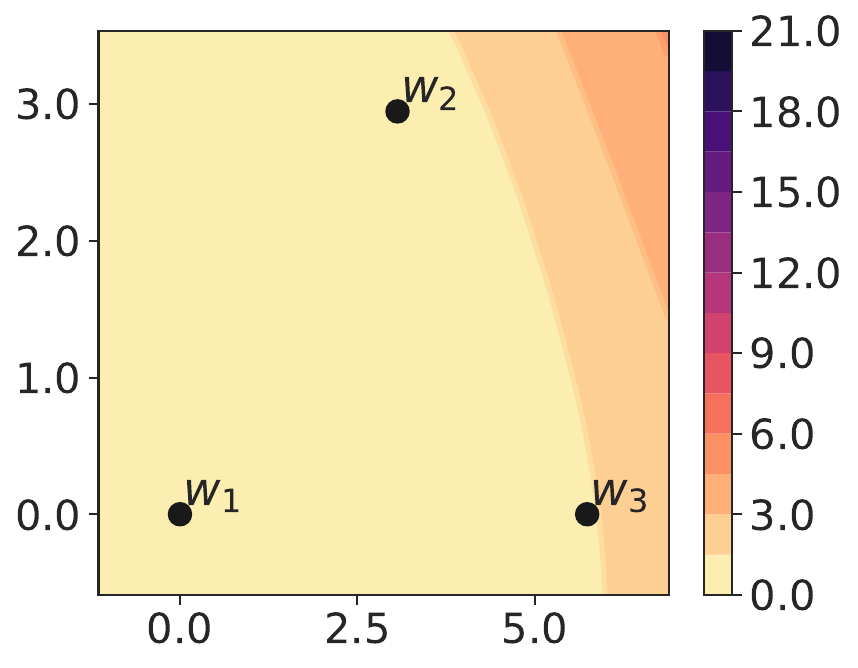}
      \caption{Seq 2 (PT)}
      \label{fig:c50_5epoch_seq_2_pt_contour_w1}
    \end{subfigure}\hspace{\fill}%
    \begin{subfigure}{.19\textwidth}
      \centering
      \includegraphics[width=0.9\textwidth]{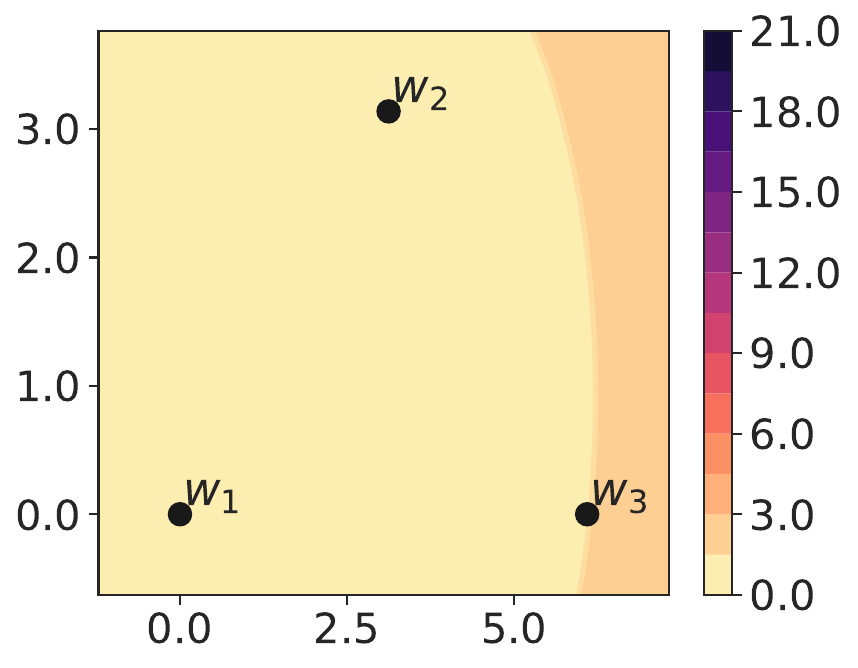}
      \caption{Seq 3 (PT)}
      \label{fig:c50_5epoch_seq_3_pt_contour_w1}
    \end{subfigure}\hspace{\fill}%
    \begin{subfigure}{.19\textwidth}
      \centering
      \includegraphics[width=0.9\textwidth]{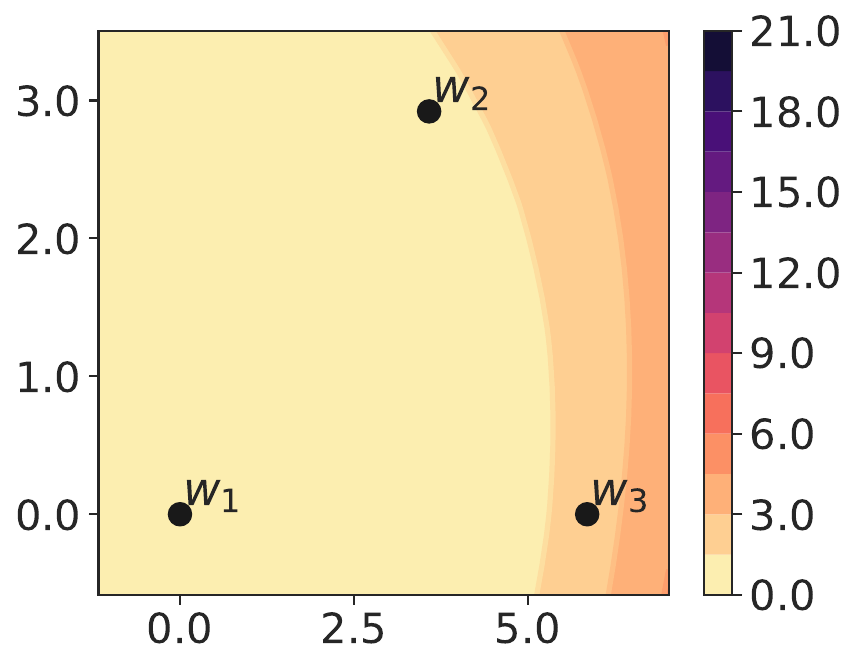}
      \caption{Seq 4 (PT)}
      \label{fig:c50_5epoch_seq_4_pt_contour_w1}
    \end{subfigure}
    \begin{subfigure}{.19\textwidth}
      \centering
      \includegraphics[width=0.9\textwidth]{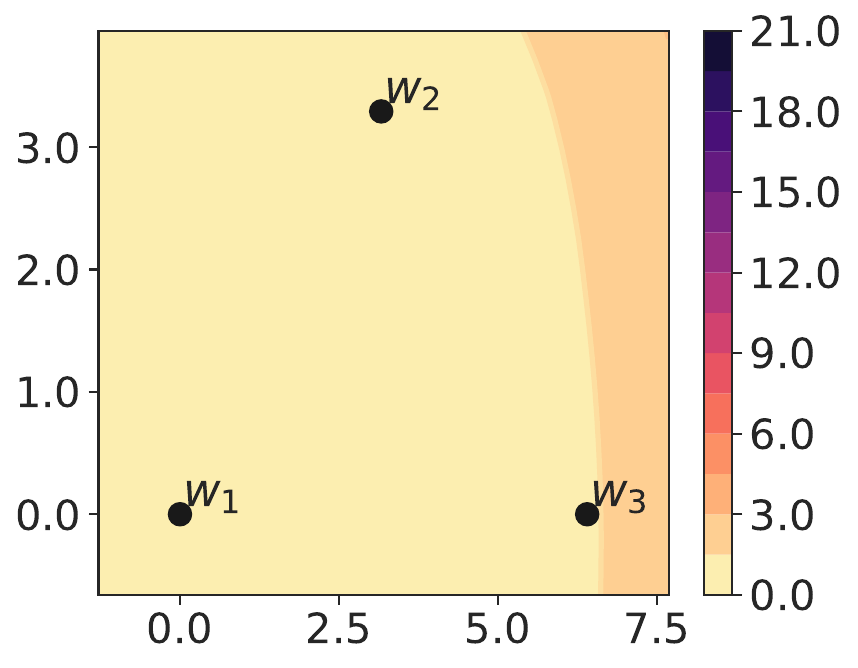}
      \caption{Seq 5 (PT)}
      \label{fig:c50_5epoch_seq_5_pt_contour_w1}
    \end{subfigure}
    \hspace{\fill}%
    \caption{Loss contours for Task 1 on 5 task sequences of Split CIFAR-50.}
    \label{fig:c50_5epoch_contours_w1}
\end{figure}

\begin{figure}[H]
    \centering
    \begin{subfigure}{.19\textwidth}
      \centering
      \includegraphics[width=0.9\textwidth]{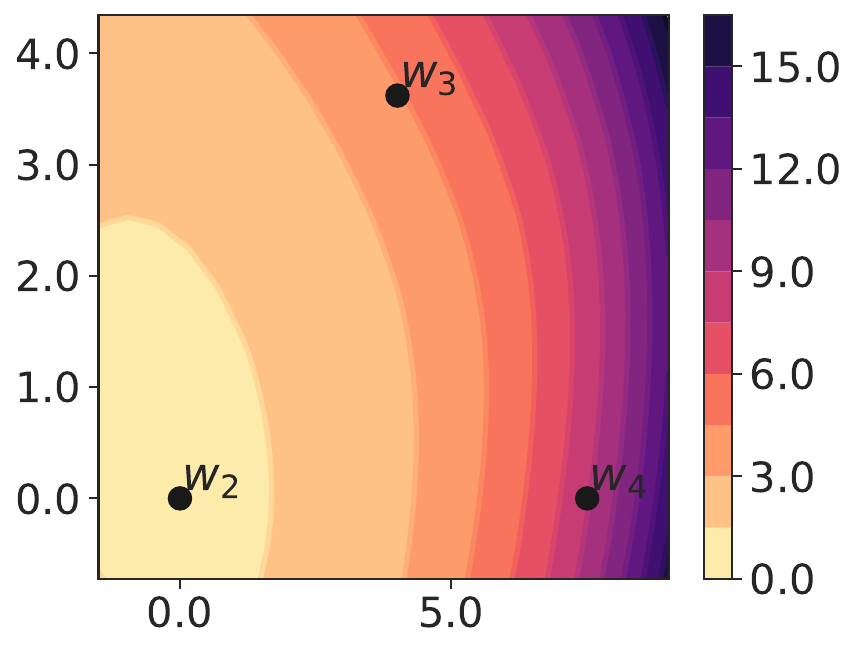}
      \caption{Seq 1 (R)}
      \label{fig:c50_5epoch_seq_1_no_contour_w2}
    \end{subfigure}\hspace{\fill}%
    \begin{subfigure}{.19\textwidth}
      \centering
      \includegraphics[width=0.9\textwidth]{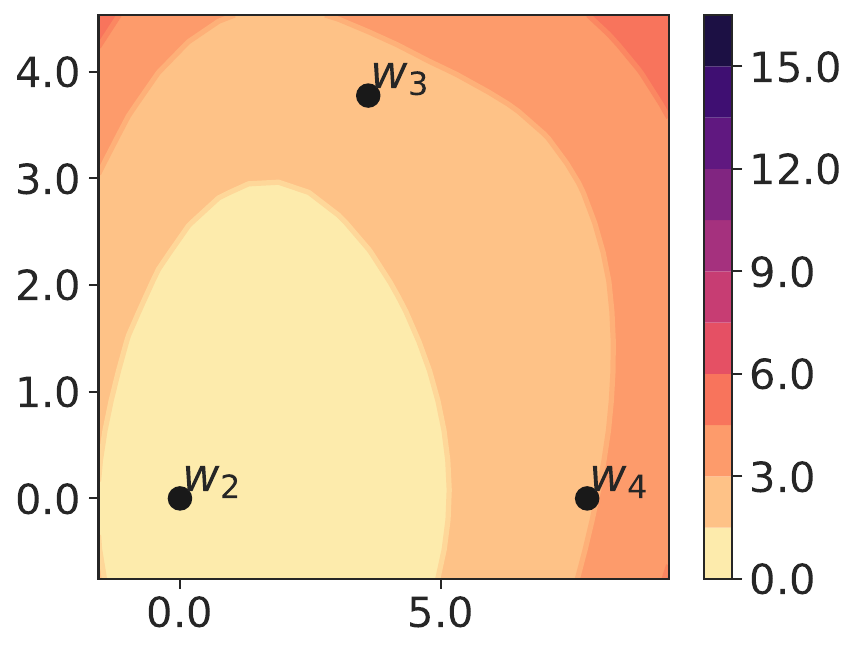}
      \caption{Seq 2 (R)}
      \label{fig:c50_5epoch_seq_2_no_contour_w2}
    \end{subfigure}\hspace{\fill}%
    \begin{subfigure}{.19\textwidth}
      \centering
      \includegraphics[width=0.9\textwidth]{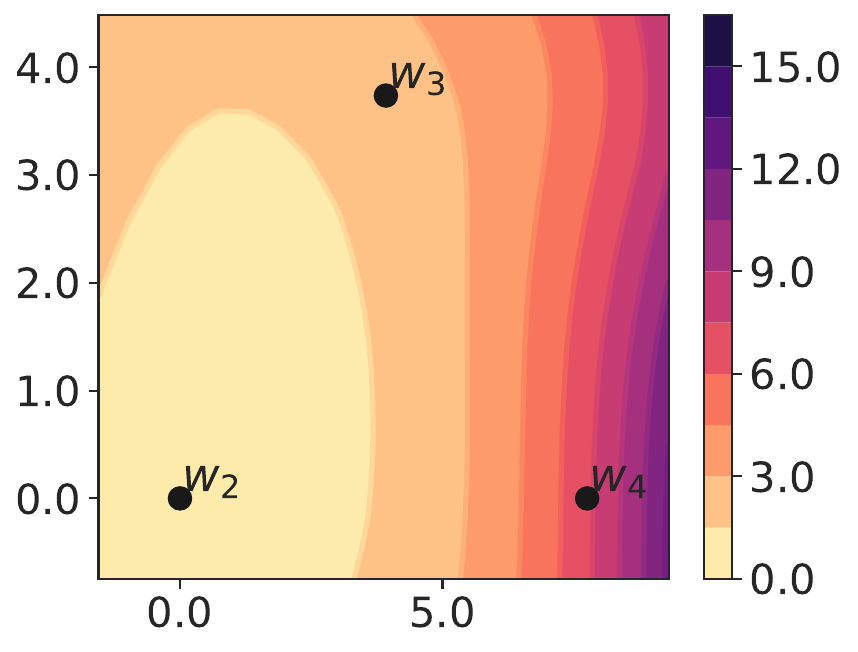}
      \caption{Seq 3 (R)}
      \label{fig:c50_5epoch_seq_3_no_contour_w2}
    \end{subfigure}\hspace{\fill}%
    \begin{subfigure}{.19\textwidth}
      \centering
      \includegraphics[width=0.9\textwidth]{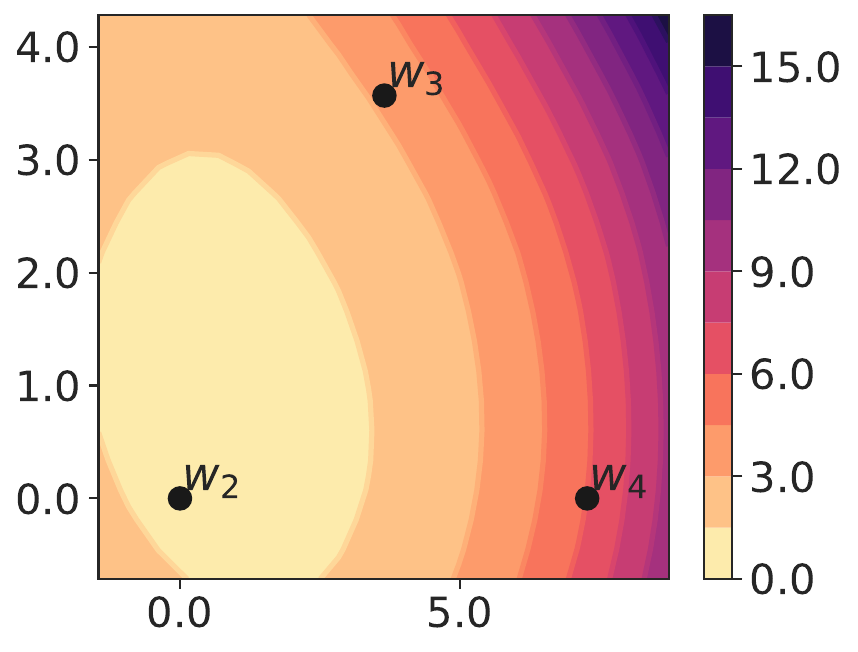}
      \caption{Seq 4 (R)}
      \label{fig:c50_5epoch_seq_4_no_contour_w2}
    \end{subfigure}
    \begin{subfigure}{.19\textwidth}
      \centering
      \includegraphics[width=0.9\textwidth]{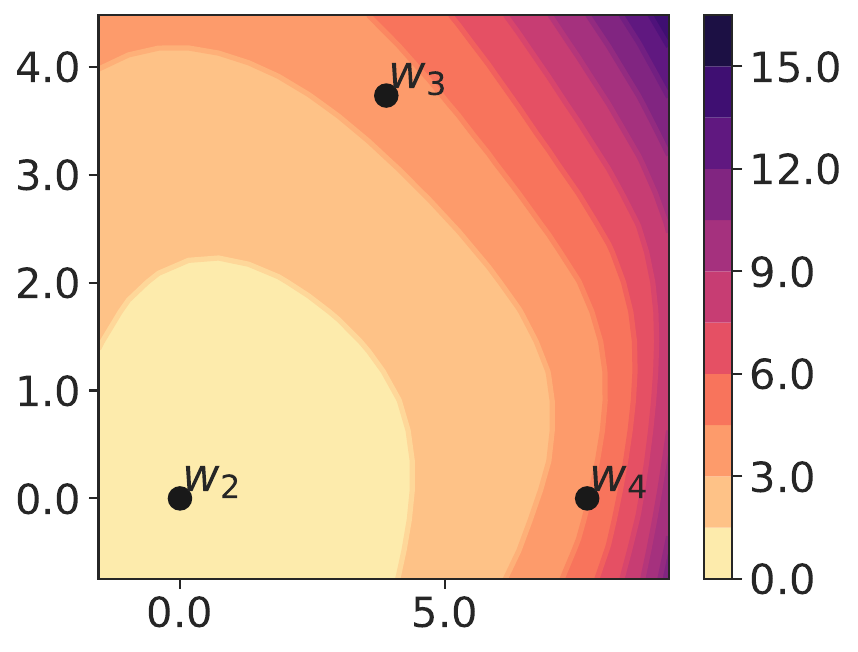}
      \caption{Seq 5 (R)}
      \label{fig:c50_5epoch_seq_5_no_contour_w2}
    \end{subfigure}
    \hspace{\fill}%
    \bigskip
    \begin{subfigure}{.19\textwidth}
      \centering
      \includegraphics[width=0.9\textwidth]{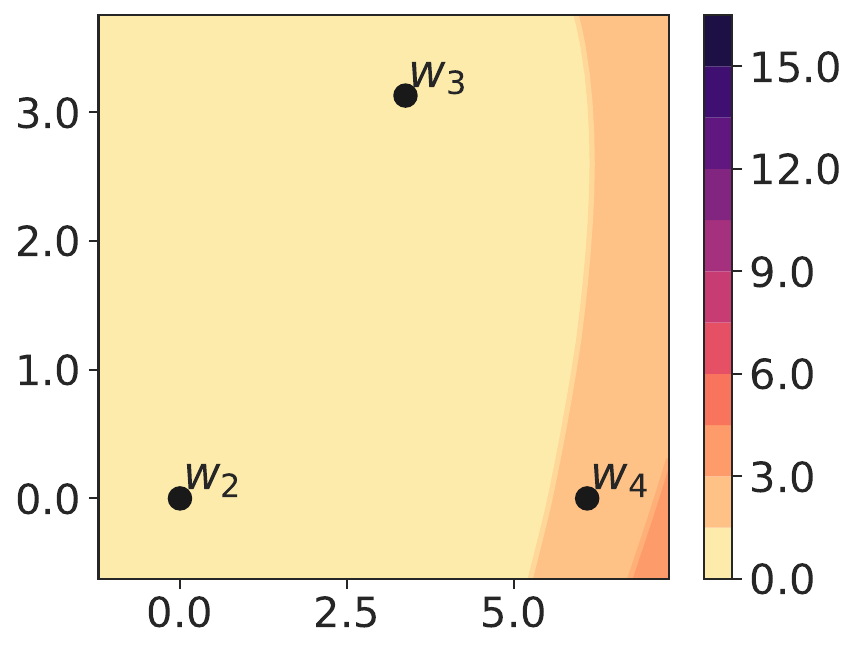}
      \caption{Seq 1 (PT)}
      \label{fig:c50_5epoch_seq_1_pt_contour_w2}
    \end{subfigure}\hspace{\fill}%
    \begin{subfigure}{.19\textwidth}
      \centering
      \includegraphics[width=0.9\textwidth]{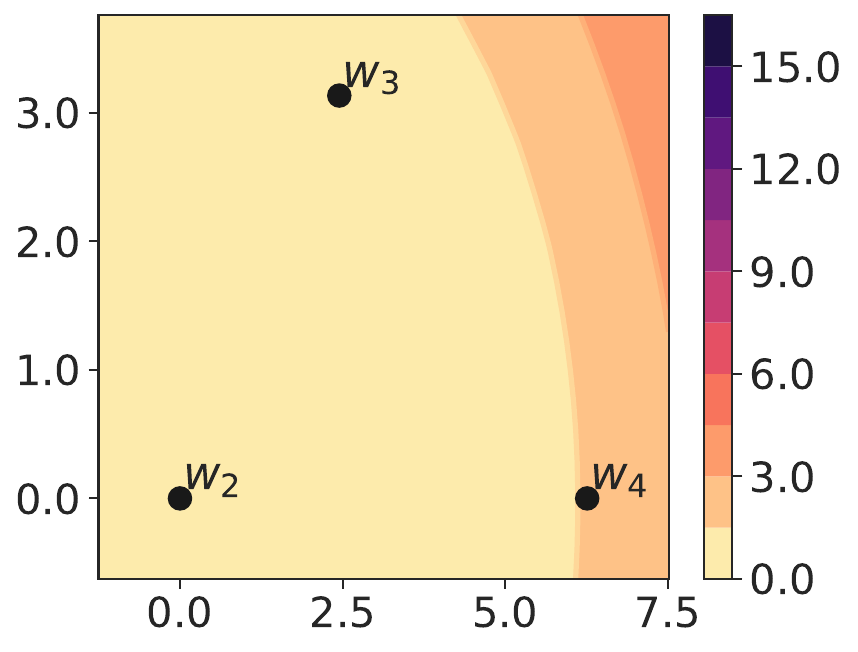}
      \caption{Seq 2 (PT)}
      \label{fig:c50_5epoch_seq_2_pt_contour_w2}
    \end{subfigure}\hspace{\fill}%
    \begin{subfigure}{.19\textwidth}
      \centering
      \includegraphics[width=0.9\textwidth]{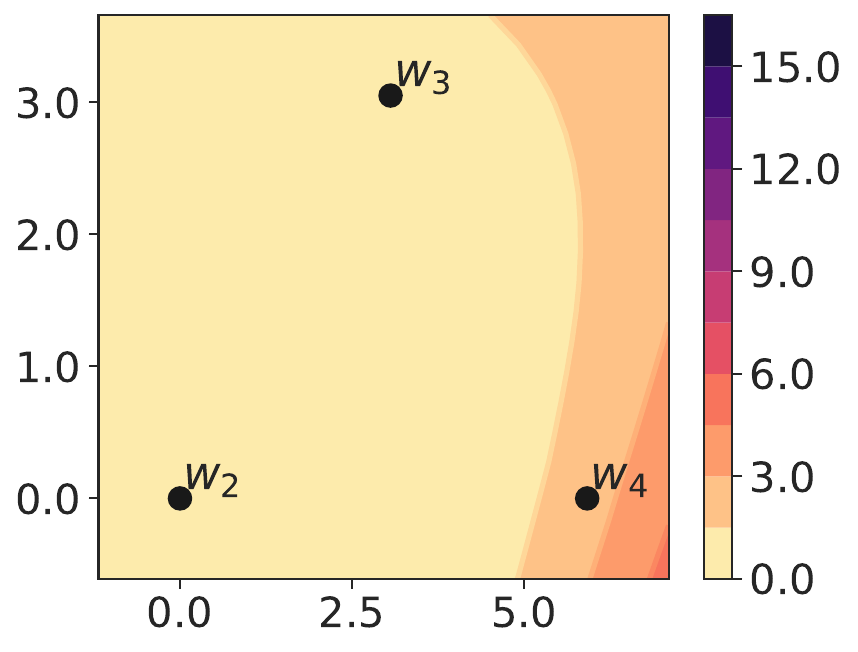}
      \caption{Seq 3 (PT)}
      \label{fig:c50_5epoch_seq_3_pt_contour_w2}
    \end{subfigure}\hspace{\fill}%
    \begin{subfigure}{.19\textwidth}
      \centering
      \includegraphics[width=0.9\textwidth]{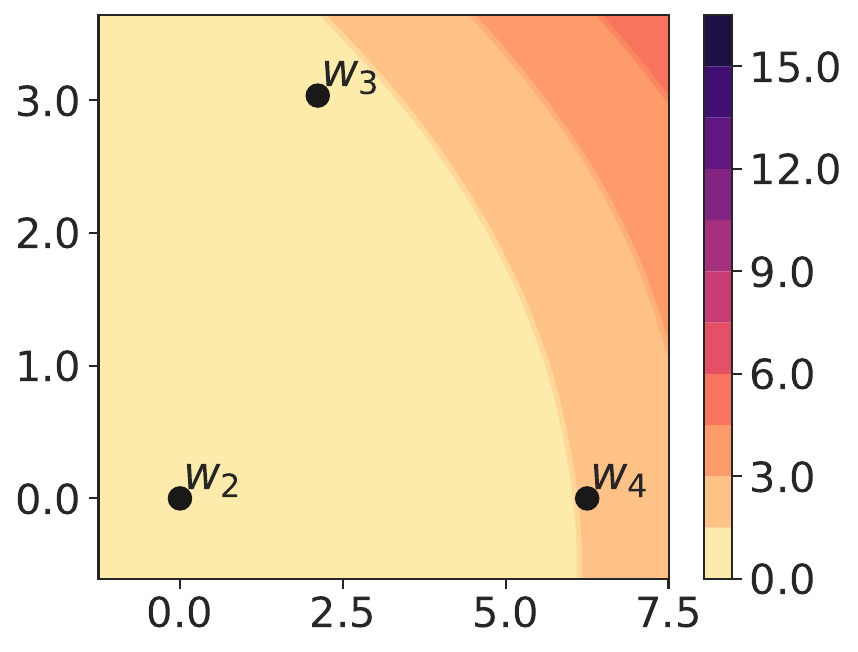}
      \caption{Seq 4 (PT)}
      \label{fig:c50_5epoch_seq_4_pt_contour_w2}
    \end{subfigure}
    \begin{subfigure}{.19\textwidth}
      \centering
      \includegraphics[width=0.9\textwidth]{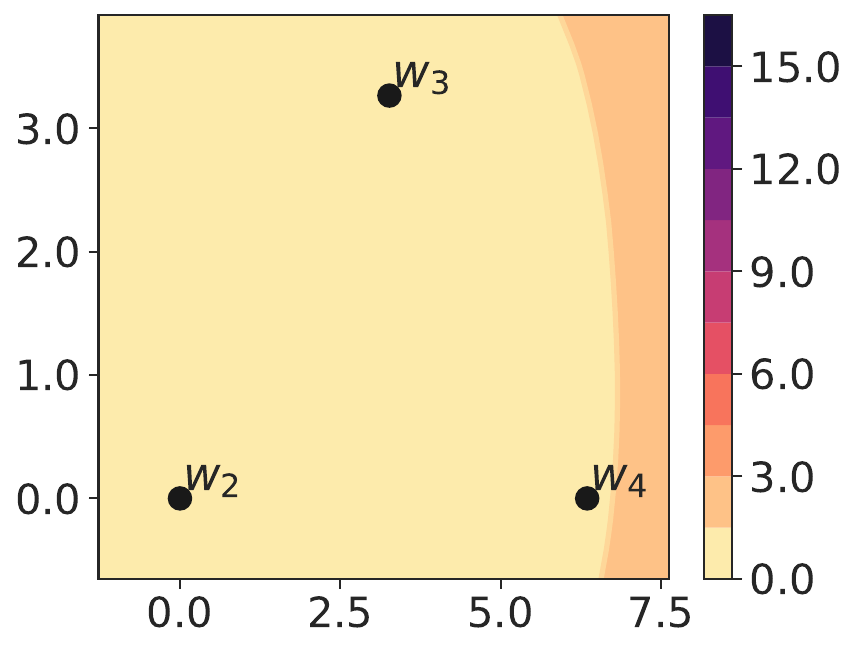}
      \caption{Seq 5 (PT)}
      \label{fig:c50_5epoch_seq_5_pt_contour_w2}
    \end{subfigure}
    \hspace{\fill}%
    \caption{Loss contours for Task 2 on 5 task sequences of Split CIFAR-50.}
    \label{fig:c50_5epoch_contours_w2}
\end{figure}

% \subsubsection{Loss Contours: 5-dataset-CV}
\begin{figure}[H]
    \centering
    \begin{subfigure}{.19\textwidth}
      \centering
      \includegraphics[width=0.9\textwidth]{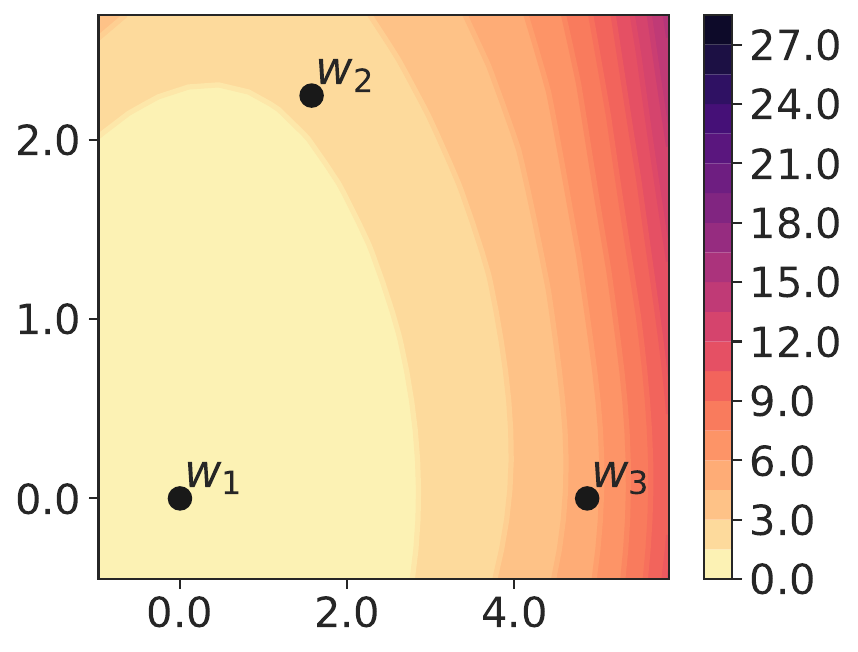}
      \caption{Seq 1 (R)}
      \label{fig:5data_5epoch_seq_1_no_contour_w1}
    \end{subfigure}\hspace{\fill}%
    \begin{subfigure}{.19\textwidth}
      \centering
      \includegraphics[width=0.9\textwidth]{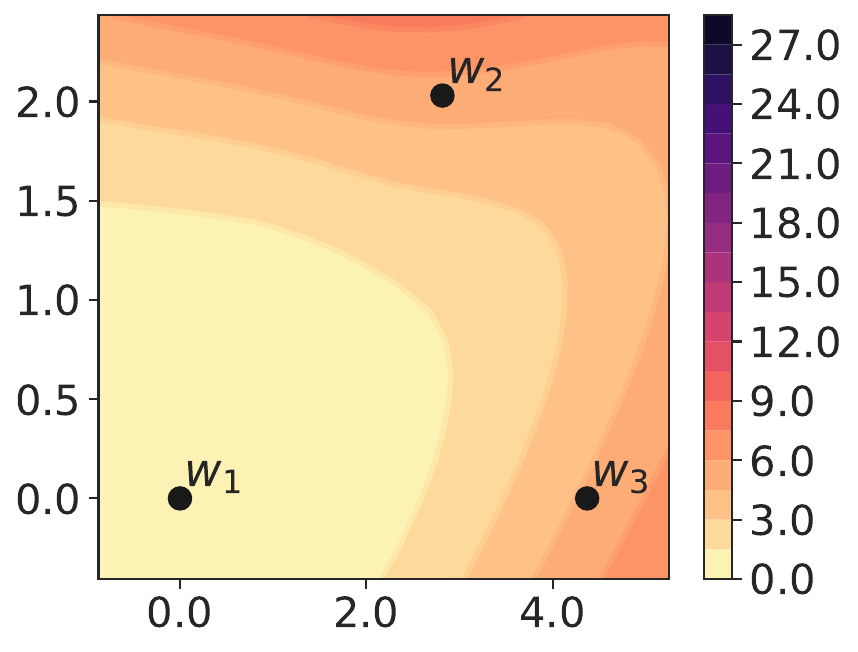}
      \caption{Seq 2 (R)}
      \label{fig:5data_5epoch_seq_2_no_contour_w1}
    \end{subfigure}\hspace{\fill}%
    \begin{subfigure}{.19\textwidth}
      \centering
      \includegraphics[width=0.9\textwidth]{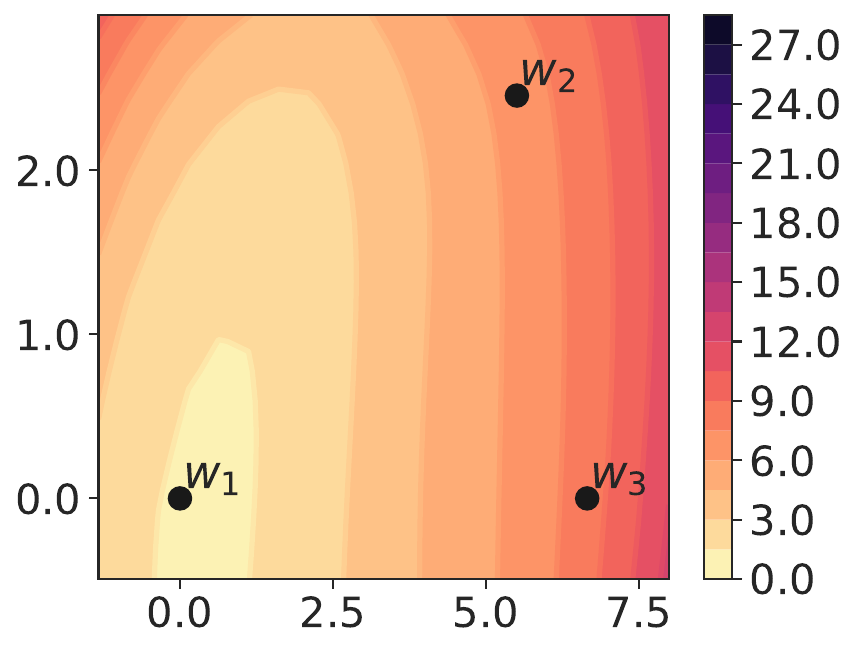}
      \caption{Seq 3 (R)}
      \label{fig:5data_5epoch_seq_3_no_contour_w1}
    \end{subfigure}\hspace{\fill}%
    \begin{subfigure}{.19\textwidth}
      \centering
      \includegraphics[width=0.9\textwidth]{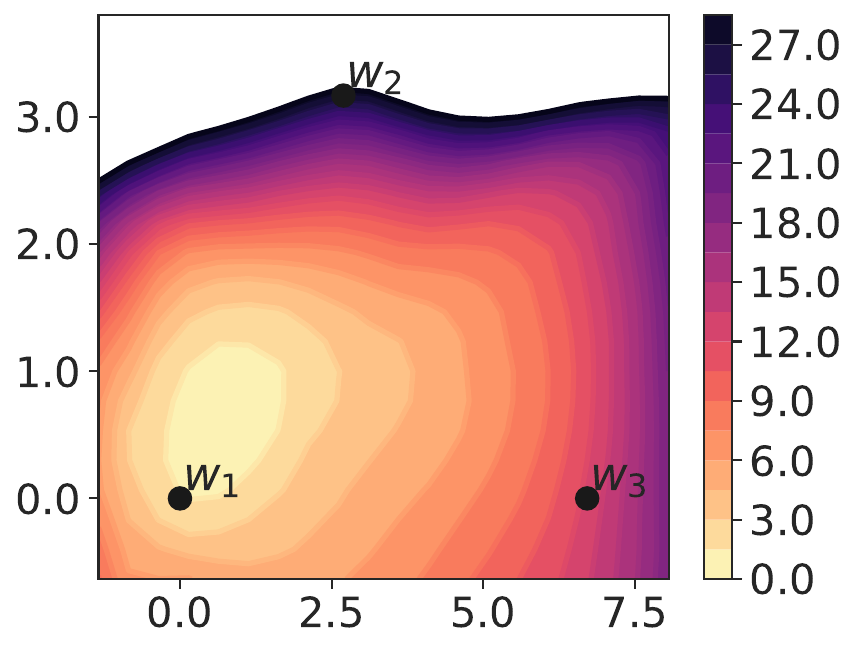}
      \caption{Seq 4 (R)}
      \label{fig:5data_5epoch_seq_4_no_contour_w1}
    \end{subfigure}
    \begin{subfigure}{.19\textwidth}
      \centering
      \includegraphics[width=0.9\textwidth]{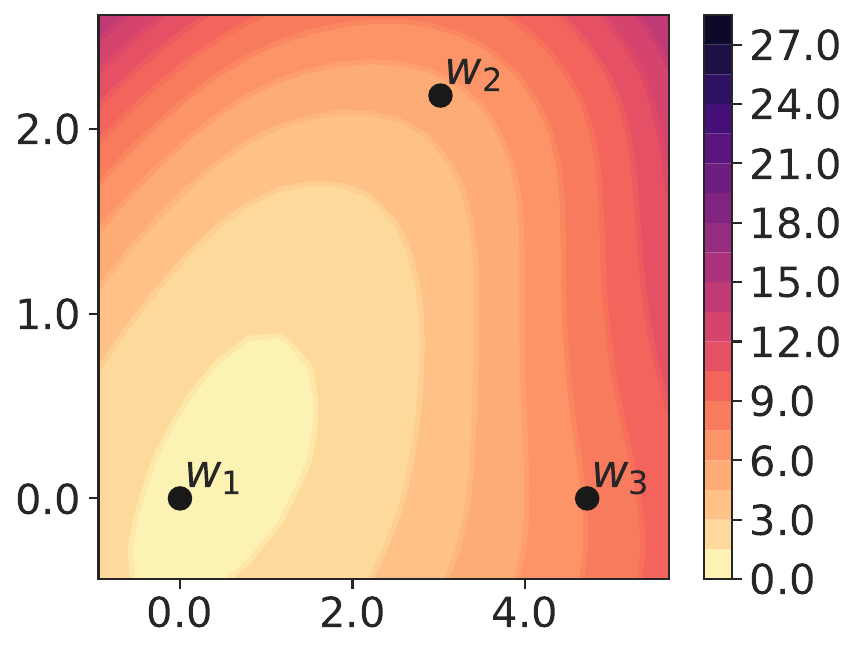}
      \caption{Seq 5 (R)}
      \label{fig:5data_5epoch_seq_5_no_contour_w1}
    \end{subfigure}
    \hspace{\fill}%
    \bigskip
    \begin{subfigure}{.19\textwidth}
      \centering
      \includegraphics[width=0.9\textwidth]{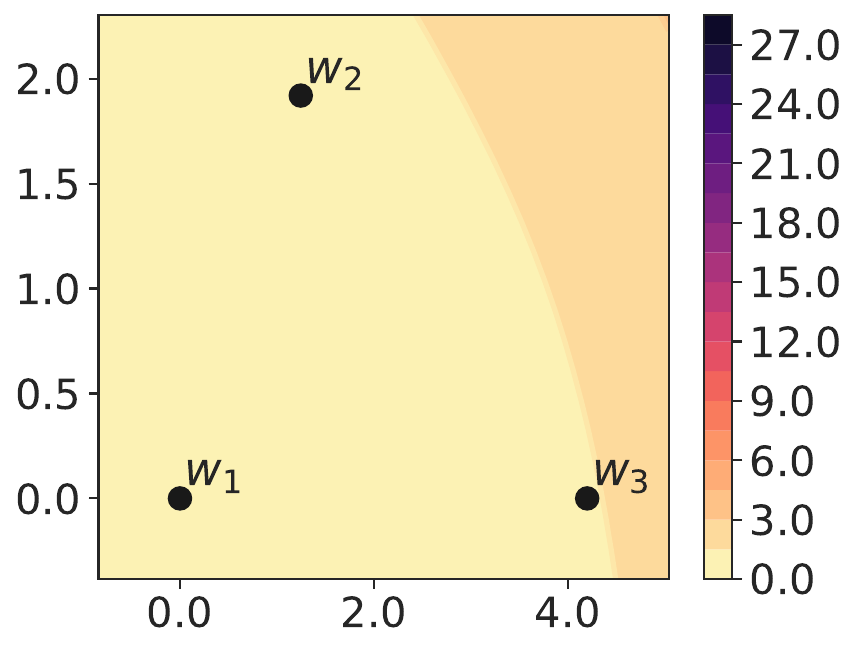}
      \caption{Seq 1 (PT)}
      \label{fig:5data_5epoch_seq_1_pt_contour_w1}
    \end{subfigure}\hspace{\fill}%
    \begin{subfigure}{.19\textwidth}
      \centering
      \includegraphics[width=0.9\textwidth]{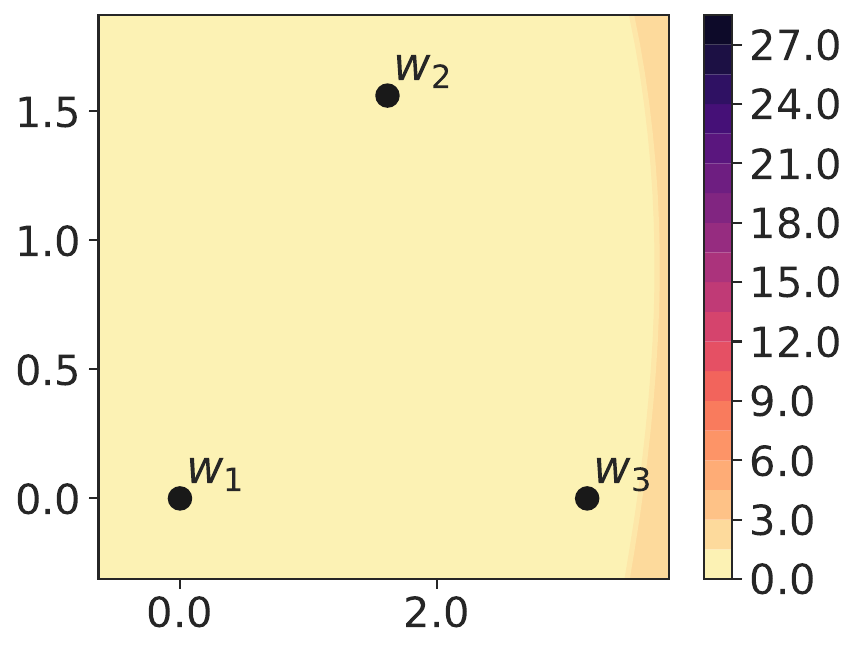}
      \caption{Seq 2 (PT)}
      \label{fig:5data_5epoch_seq_2_pt_contour_w1}
    \end{subfigure}\hspace{\fill}%
    \begin{subfigure}{.19\textwidth}
      \centering
      \includegraphics[width=0.9\textwidth]{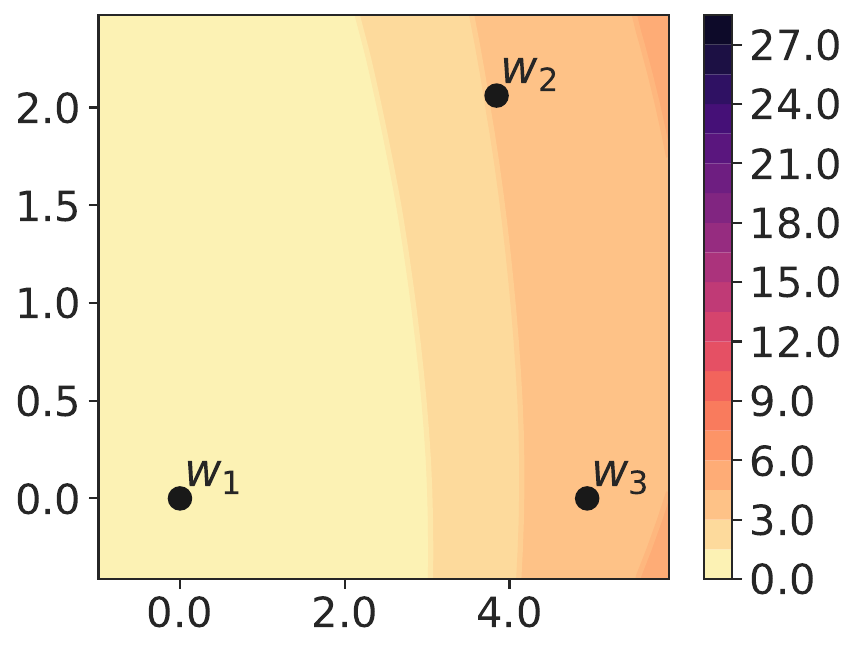}
      \caption{Seq 3 (PT)}
      \label{fig:5data_5epoch_seq_3_pt_contour_w1}
    \end{subfigure}\hspace{\fill}%
    \begin{subfigure}{.19\textwidth}
      \centering
      \includegraphics[width=0.9\textwidth]{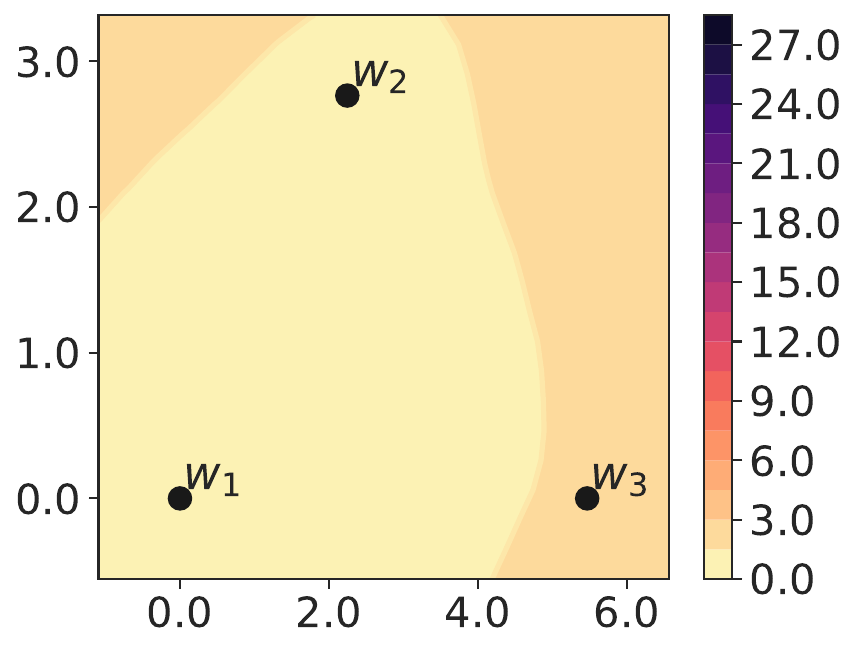}
      \caption{Seq 4 (PT)}
      \label{fig:5data_5epoch_seq_4_pt_contour_w1}
    \end{subfigure}
    \begin{subfigure}{.19\textwidth}
      \centering
      \includegraphics[width=0.9\textwidth]{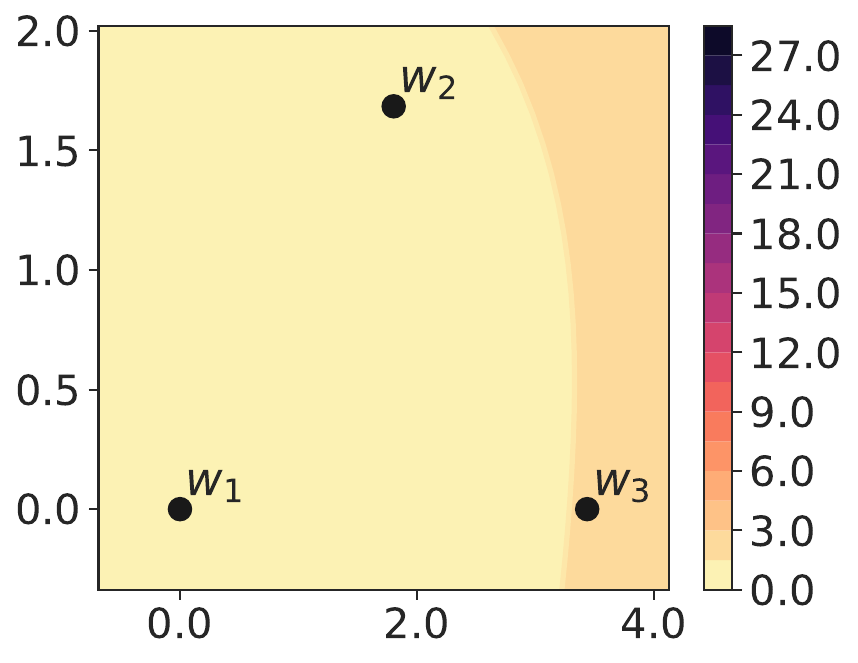}
      \caption{Seq 5 (PT)}
      \label{fig:5data_5epoch_seq_5_pt_contour_w1}
    \end{subfigure}
    \hspace{\fill}%
    \caption{Loss contours for Task 1 on 5 task sequences of 5-dataset-CV.}
    \label{fig:5data_5epoch_contours_w1}
\end{figure}

\begin{figure}[H]
    \centering
    \begin{subfigure}{.19\textwidth}
      \centering
      \includegraphics[width=0.9\textwidth]{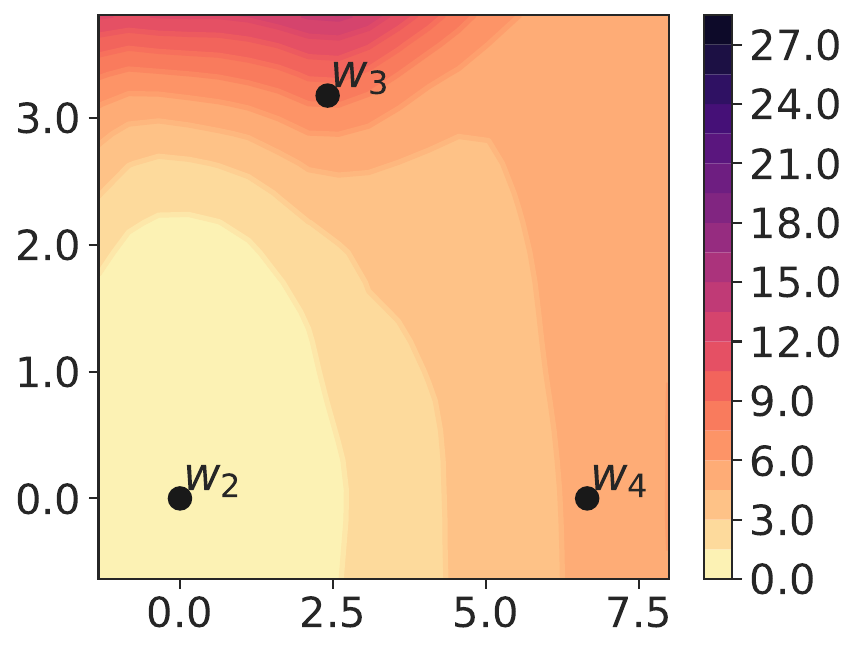}
      \caption{Seq 1 (R)}
      \label{fig:5data_5epoch_seq_1_no_contour_w2}
    \end{subfigure}\hspace{\fill}%
    \begin{subfigure}{.19\textwidth}
      \centering
      \includegraphics[width=0.9\textwidth]{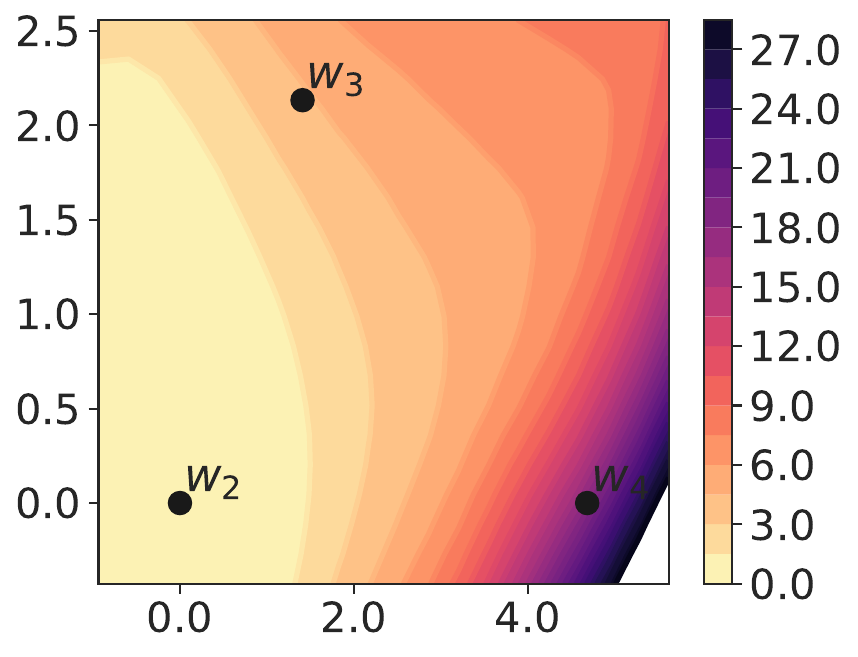}
      \caption{Seq 2 (R)}
      \label{fig:5data_5epoch_seq_2_no_contour_w2}
    \end{subfigure}\hspace{\fill}%
    \begin{subfigure}{.19\textwidth}
      \centering
      \includegraphics[width=0.9\textwidth]{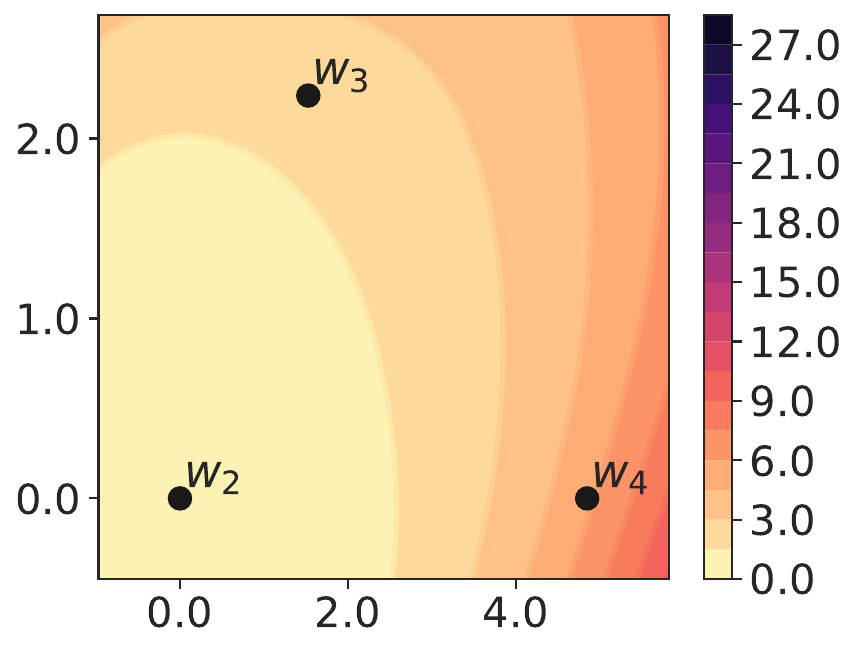}
      \caption{Seq 3 (R)}
      \label{fig:5data_5epoch_seq_3_no_contour_w2}
    \end{subfigure}\hspace{\fill}%
    \begin{subfigure}{.19\textwidth}
      \centering
      \includegraphics[width=0.9\textwidth]{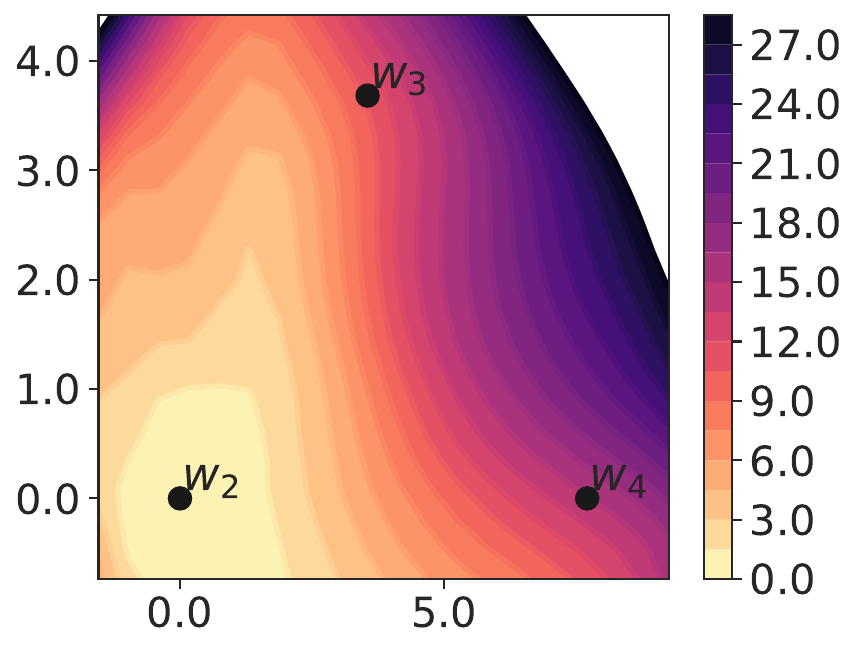}
      \caption{Seq 4 (R)}
      \label{fig:5data_5epoch_seq_4_no_contour_w2}
    \end{subfigure}
    \begin{subfigure}{.19\textwidth}
      \centering
      \includegraphics[width=0.9\textwidth]{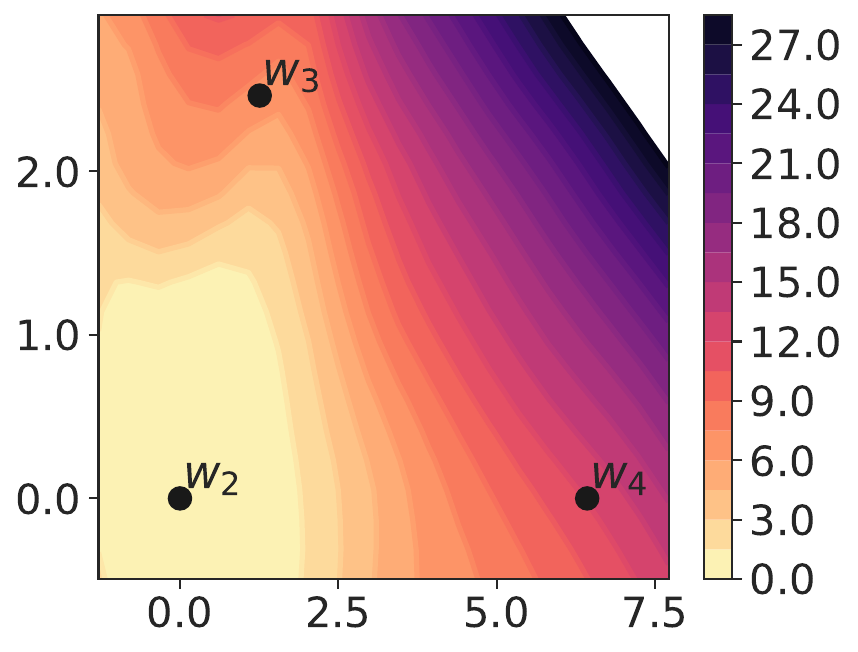}
      \caption{Seq 5 (R)}
      \label{fig:5data_5epoch_seq_5_no_contour_w2}
    \end{subfigure}
    \hspace{\fill}%
    \bigskip
    \begin{subfigure}{.19\textwidth}
      \centering
      \includegraphics[width=0.9\textwidth]{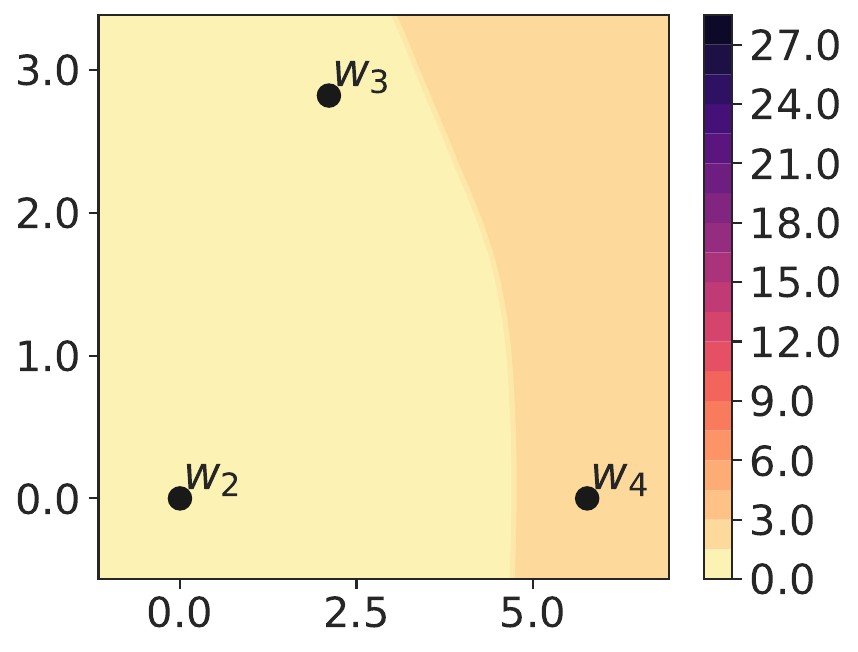}
      \caption{Seq 1 (PT)}
      \label{fig:5data_5epoch_seq_1_pt_contour_w2}
    \end{subfigure}\hspace{\fill}%
    \begin{subfigure}{.19\textwidth}
      \centering
      \includegraphics[width=0.9\textwidth]{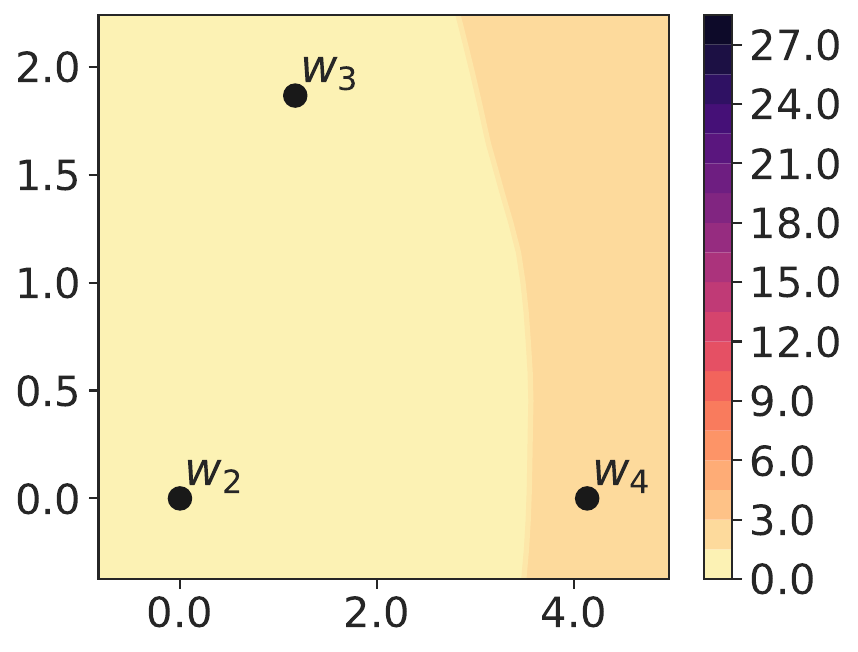}
      \caption{Seq 2 (PT)}
      \label{fig:5data_5epoch_seq_2_pt_contour_w2}
    \end{subfigure}\hspace{\fill}%
    \begin{subfigure}{.19\textwidth}
      \centering
      \includegraphics[width=0.9\textwidth]{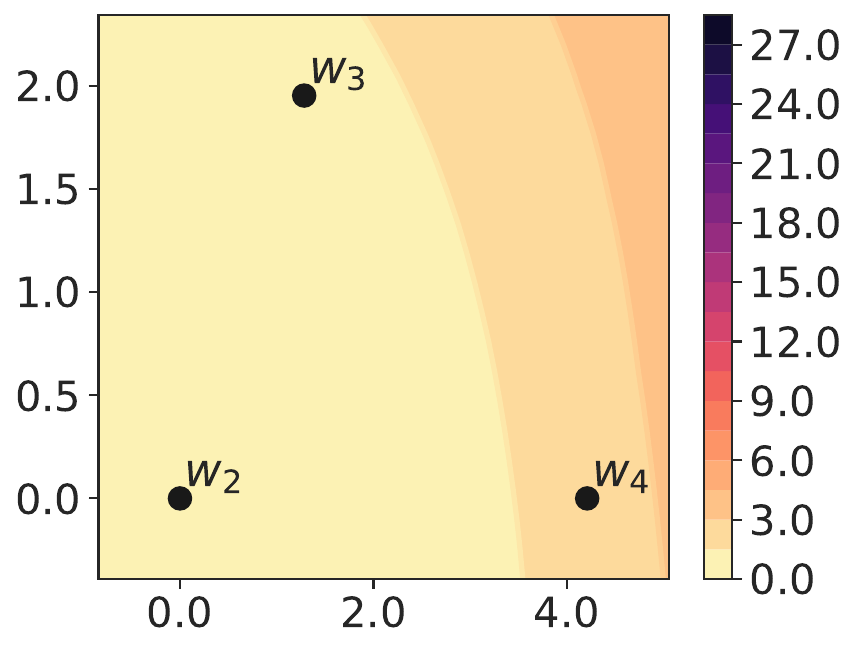}
      \caption{Seq 3 (PT)}
      \label{fig:5data_5epoch_seq_3_pt_contour_w2}
    \end{subfigure}\hspace{\fill}%
    \begin{subfigure}{.19\textwidth}
      \centering
      \includegraphics[width=0.9\textwidth]{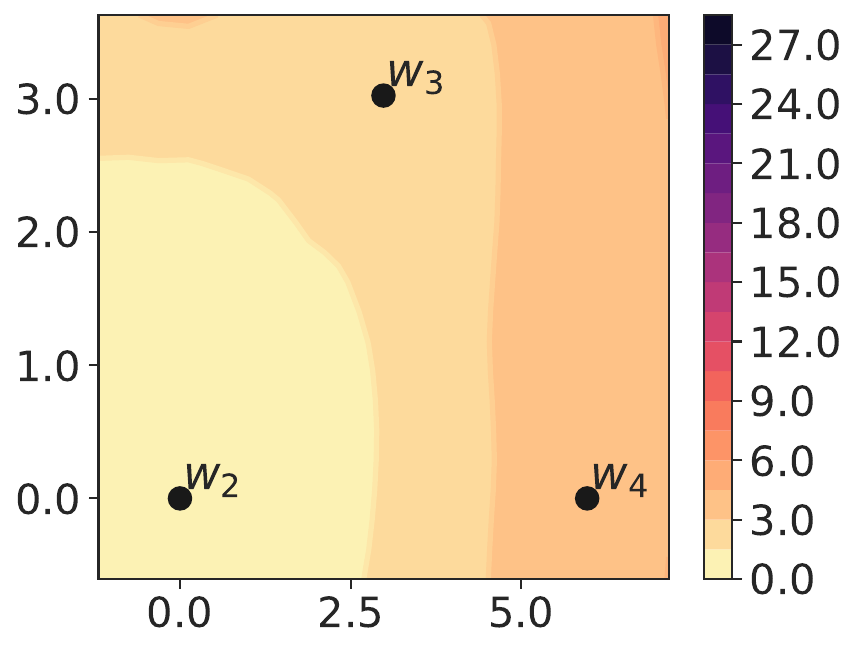}
      \caption{Seq 4 (PT)}
      \label{fig:5data_5epoch_seq_4_pt_contour_w2}
    \end{subfigure}
    \begin{subfigure}{.19\textwidth}
      \centering
      \includegraphics[width=0.9\textwidth]{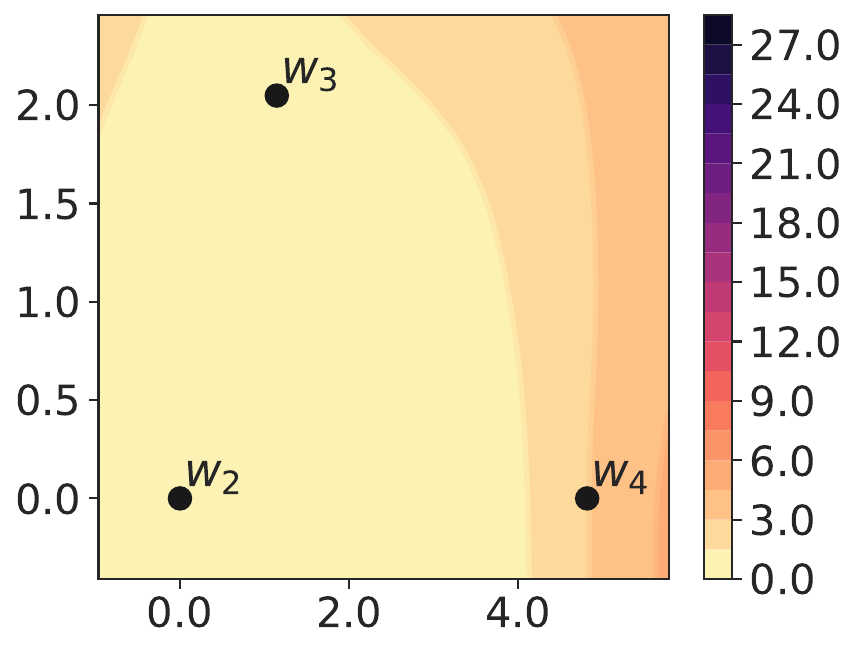}
      \caption{Seq 5 (PT)}
      \label{fig:5data_5epoch_seq_5_pt_contour_w2}
    \end{subfigure}
    \hspace{\fill}%
    \caption{Loss contours for Task 2 on 5 task sequences of 5-dataset-CV.}
    \label{fig:5data_5epoch_contours_w2}
\end{figure}